\documentclass{bmvc2kModif}

\title{AUPIMO: Redefining Anomaly Localization Benchmarks with High Speed and Low Tolerance}

\addauthor{Joao P. C. Bertoldo}{jpcbertoldo@minesparis.psl.eu}{1}
\addauthor{Dick Ameln}{dick.ameln@intel.com}{2}
\addauthor{Ashwin Vaidya}{ashwin.vaidya@intel.com}{2}
\addauthor{Samet Akçay}{samet.akcay@intel.com}{2}

\addinstitution{
   Mines Paris, PSL University,\\
   Centre for mathematical\\
   morphology (CMM),\\
   77300 Fontainebleau, France
}
\addinstitution{Intel}

\runninghead{Bertoldo, Ameln, Vaidya, Akçay}{AUPIMO}

\usepackage{graphicx}
\usepackage{booktabs}
\usepackage{dirtytalk}
\usepackage{caption}
\usepackage{subcaption}
\usepackage{setspace}
\usepackage[export]{adjustbox}
\usepackage{multicol}
\usepackage{siunitx}
\usepackage{mathtools}
\usepackage{xargs}
\usepackage{tabularx}
\usepackage{booktabs}
\usepackage{amsmath}
\usepackage{array}
\usepackage{multirow}
\usepackage{ragged2e}
\usepackage{amssymb}

\usepackage{enumitem}

\usepackage{xcolor}
\usepackage[super]{nth}
\usepackage{listings}

\usepackage[capitalize]{cleveref}
\crefname{section}{Sec.}{Secs.}
\Crefname{section}{Section}{Sections}
\crefname{table}{Tab.}{Tabs.}
\Crefname{table}{Table}{Tables}

\usepackage{hyperref}
\usepackage{glossaries}
\glsdisablehyper

\def\eg{\emph{e.g}\bmvaOneDot}

\def\ie{\emph{i.e}\bmvaOneDot}
\def\cf{\emph{c.f}\bmvaOneDot}
\def\etc{\emph{etc}\bmvaOneDot}

\renewcommand{\paragraph}[1]{\vspace{7pt}\noindent\textbf{#1}\hspace{3pt}}

\newcommand{\exampleDataset}{MVTec AD / Zipper}

\newcommand{\githubRepo}{\href{https://github.com/jpcbertoldo/aupimo}{\nolinkurl{github.com/jpcbertoldo/aupimo}}}
\newcommand{\githubAnomalib}{\href{https://github.com/openvinotoolkit/anomalib}{\nolinkurl{github.com/openvinotoolkit/anomalib}}}

\newglossaryentry{ad}{
    name={AD},
    first={Anomaly Detection (AD)},
    description={}
}

\newglossaryentry{aloc}{
    name={anomaly localization},
    description={}
}

\newglossaryentry{vad}{
    name={Visual Anomaly Detection},
    description={}
}

\newglossaryentry{sota}{
    name={SOTA},
    first={State-of-the-Art (SOTA)},
    description={}
}

\newglossaryentry{mvtecad}{
    name={MVTec AD},
    first={MVTec Anomaly Detection (MVTec AD)},
    description={},
}

\newglossaryentry{mvtecloco}{
    name={MVTec LOCO},
    first={MVTec Logical Constraints (MVTec LOCO)},
    description={},
}

\newglossaryentry{visa}{
    name={VisA},
    first={Visual Anomaly (VisA)},
    description={},
}

\newglossaryentry{ksdd}{
    name={KSDD2},
    description={},
}

\newglossaryentry{coffeead}{
    name={Coffee leaf diseases AD},
    description={},
}

\newglossaryentry{xrayad}{
    name={X-ray Fibrus Product AD},
    description={},
}

\newglossaryentry{iou}{
    name={IoU},
    first={Intersection over Union (IoU)},
    description={},
}

\newglossaryentry{auc}{
    name={AUC},
    first={Area Under the Curve (AUC)},
    description={},
}

\newglossaryentry{fp}{
    name={FP},
    first={False Positive (FP)},
    plural={FPs},
    firstplural={False Positives (FPs)},
    description={},
}

\newglossaryentry{fpty}{
    name={false positivity},
    description={},
}

\newglossaryentry{fpr}{
    name={FPR},
    first={False Positive Rate (FPR)},
    description={},
}

\newglossaryentry{shfpr}{
    name={Shared FPR},
    first={\emph{Shared} FPR},
    description={},
}

\newglossaryentry{imfpr}{
    name={Image FPR},
    description={},
}

\newglossaryentry{setfpr}{
    name={Set FPR},
    description={},
}

\newglossaryentry{numfpreg}{
    name={NumFPReg},
    first={Number of False Positive Regions (NumFPReg)},
    description={},
}

\newglossaryentry{imnumfpreg}{
    name={ImNumFPReg},
    first={Image Number of False Positive Regions (ImNumFPReg)},
    description={},
}

\newglossaryentry{tp}{
    name={TP},
    first={True Positive (TP)},
    description={},
}

\newglossaryentry{tpty}{
    name={true positivity},
    description={},
}

\newglossaryentry{tpr}{
    name={TPR},
    first={True Positive Rate (TPR)},
    plural={TPs},
    firstplural={True Positives (TPs)},
    description={},
}

\newglossaryentry{regtpr}{
    name={Region TPR},
    description={},
}

\newglossaryentry{imtpr}{
    name={Image TPR},
    description={},
}

\newglossaryentry{settpr}{
    name={Set TPR},
    description={},
}

\newglossaryentry{sem}{
    name={SEM},
    first={standard error of the mean (SEM)},
    description={},
}

\newglossaryentry{pr}{
    name={PR},
    first={Precision-Recall (PR)},
    description={}
}

\newglossaryentry{aupr}{
    name={AUPR},
    first={Area Under the Precision-Recall (AUPR)},
    description={}
}

\newglossaryentry{iap}{
    name={IAP},
    first={Instance Average Precision (IAP)},
    description={}
}

\newglossaryentry{roc}{
    name={ROC},
    first={Receiver Operating Characteristic (ROC)},
    description={}
}

\newglossaryentry{auroc}{
    name={AUROC},
    description={}
}

\newglossaryentry{pro}{
    name={PRO},
    first={Per-Region Overlap (PRO)},
    description={}
}

\newglossaryentry{spro}{
    name={sPRO},
    first={Saturated Per-Region Overlap (sPRO)},
    description={}
}

\newglossaryentry{aupro}{
    name={AUPRO},
    description={}
}

\newglossaryentry{auprofive}{
    name={AUPRO$_{5\%}$},
    description={}
}

\newglossaryentry{pimo}{
    name={PIMO},
    first={Per-Image Overlap (PIMO)},
    description={}
}

\newglossaryentry{aupimo}{
    name={AUPIMO},
    description={}
}

\newglossaryentry{fone}{
    name={F$_1$},
    description={}
}

\newglossaryentry{fonemax}{
    name={F$_1$-max},
    description={}
}

\newglossaryentry{padim}{
    name={PaDiM},
    description={},
} 

\newglossaryentry{padimRoe}{
    name={PaDiM R18},
    description={},
}  

\newglossaryentry{padimWrfz}{
    name={PaDiM WR50},
    description={},
}  

\newglossaryentry{patchcore}{
    name={PatchCore},
    description={},
}

\newglossaryentry{patchcoreWrozo}{
    name={PatchCore WR101},
    description={},
}  

\newglossaryentry{patchcoreWrfz}{
    name={PatchCore WR50},
    description={},
}  

\newglossaryentry{fastflow}{
    name={FastFlow},
    description={},
}

\newglossaryentry{fastflowCait}{
    name={FastFlow CAIT},
    description={},
}  

\newglossaryentry{fastflowWrfz}{
    name={FastFlow WR50},
    description={},
}

\newglossaryentry{efficientad}{
    name={EfficientAD},
    description={},
}

\newglossaryentry{efficientadM}{
    name={EfficientAD M},
    description={},
}  

\newglossaryentry{efficientadS}{
    name={EfficientAD S},
    description={},
}  

\newglossaryentry{simplenet}{
    name={SimpleNet},
    description={},
}

\newglossaryentry{simplenetWrfz}{
    name={SimpleNet WR50},
    description={},
}

\newglossaryentry{pyramidflow}{
    name={PyramidFlow},
    description={},
}

\newglossaryentry{pyramidFnf}{
    name={PyramidFlow FNF},
    description={},
}

\newglossaryentry{pyramidRoe}{
    name={PyramidFlow R18},
    description={},
}

\newglossaryentry{uflow}{
    name={UFlow},
    description={},
}

\newglossaryentry{reversedistppFullname}{
    name={Reverse Distilation ++},
    description={},
}
\newglossaryentry{reversedistpp}{
    name={RevDist++},
    description={},
}

\newglossaryentry{reversedistppWrfz}{
    name={RevDist++ WR50},
    description={},
}

\newglossaryentry{bbRoe}{
    name={ResNet18},
    description={},
}

\newglossaryentry{bbWrfz}{
    name={WideResNet50},
    description={},
}

\newglossaryentry{bbWrozo}{
    name={WideResNet101},
    description={},
}

\newglossaryentry{bbCait}{
    name={Cait M48},
    description={},
}

\newcommand{\Real}{\mathbb{R}}
\newcommand{\Rep}{\Real_+}

\newcommand{\zoset}{\{0, 1\}}

\newcommand{\card}[1]{\left| #1 \right|}

\newcommand{\perc}[1]{\mathrm{P}_{\text{#1}}}

\newcommand{\thr}{t}

\newcommand{\imres}{M}

\newcommand{\gt}{\mathbf{y}}
\newcommand{\gtdom}{\zoset^\imres}

\newcommand{\gtset}{\mathcal{Y}}

\newcommand{\scoredom}{\Rep}

\newcommand{\amap}{\mathbf{a}}
\newcommand{\amapdom}{\scoredom^\imres}

\newcommand{\amapset}{\mathcal{A}}

\newcommand{\pidx}{j}

\newcommand{\regmask}{\mathbf{r}}
\newcommand{\regmaskset}{\mathcal{R}}

\newcommand{\imfpr}{\mathrm{F}_{\text{i}}}
\newcommand{\imfprExp}{
        \card{(\amap \ge \thr) \land (\neg\gt)}
        /
        \card{\neg\gt}
}

\newcommand{\shfpr}{\mathrm{F}_{\text{sh}}}
\newcommand{\shfprinv}{\shfpr^{-1}}
\newcommand{\shfprExp}[1][\gtset^0]{
    \frac{1}{\card{#1}}
    \sum_{\gt \in #1}^{} \, 
    \imfpr^{\gt}(\thr)
}

\newcommand{\setfpr}{\mathrm{F}_{\text{s}}}
\newcommand{\setfprinv}{\setfpr^{-1}}
\newcommand{\setfprExp}{
    \frac{
        \sum_{\gt \in \gtset}^{} \, 
        \card{(\amap \ge \thr) \land (\neg\gt)}
    }{
        \sum_{\gt \in \gtset}^{} \, 
        \card{\neg\gt}
    } 
}

\newcommand{\regtpr}{\mathrm{T}_{\text{r}}}
\newcommand{\regtprExp}{
        \card{(\amap \ge \thr) \land \regmask}
        /
        \card{\regmask}
}
\newcommand{\avgregtpr}{\overline{\regtpr}}
\newcommand{\avgregtprExp}{
    \frac{1}{\card{\regmaskset}} 
    \sum_{\regmask \in \regmaskset}^{} \,
    \regtpr^{\regmask}(\thr)
}

\newcommand{\imtpr}{\mathrm{T}_{\text{i}}}
\newcommand{\imtprExp}{
        \card{(\amap \ge \thr) \land \gt}
        /
        \card{\gt}
}

\newcommand{\settpr}{\mathrm{T}_{\text{s}}}
\newcommand{\settprExp}{
    \frac{
        \sum_{\gt \in \gtset}^{} \, 
        \card{(\amap \ge \thr) \land \gt}
    }{
        \sum_{\gt \in \gtset}^{} \, 
        \card{\gt}
    } 
}

\newcommand{\fprlbound}{L}
\newcommand{\fprubound}{U}

\newcommand{\roc}{\mathrm{ROC}}
\newcommand{\rocExp}{\left( \setfpr(\thr) \, , \, \settpr(\thr) \right)}

\newcommand{\auroc}{\mathrm{AUROC}}
\newcommand{\aurocExp}{
    \int_{0}^{1} 
    \settpr\left( \setfprinv( z ) \right) \, 
    \mathrm{d}z
}

\newcommand{\pro}{\mathrm{PRO}}
\newcommand{\proExp}{\left( \setfpr(\thr) \, , \, \avgregtpr(\thr) \right)}

\newcommand{\aupro}{\mathrm{AUPRO}}
\newcommand{\auproExp}{
    \frac{1}{\fprubound} 
    \int_{0}^{\fprubound} 
    \avgregtpr\left( \setfprinv( z ) \right) \, 
    \mathrm{d}z  
}

\newcommand{\pimo}[1][\gt]{\mathrm{PIMO}^{#1}}
\newcommand{\pimoExp}{\left( \log \left( \shfpr(\thr) \right) \, , \, \imtpr(\thr) \right)}

\newcommand{\aupimo}[1][\gt]{\mathrm{AUPIMO}^{#1}}
\newcommand{\aupimoExpDlogz}{
    \int_{\log(\fprlbound)}^{\log(\fprubound)} 
    \frac{\imtpr\left( \shfprinv( z ) \right)}{\log\left(\fprubound / \fprlbound\right)}  
    \, 
    \mathrm{d}\log(z)   
}

\begin{document}
\maketitle
\begin{abstract}
Recent advances in anomaly localization research have seen AUROC and AUPRO scores on public benchmark datasets like MVTec and VisA converge towards perfect recall.
However, high AUROC and AUPRO scores do not always reflect qualitative performance, which limits the validity of these metrics.
We argue that the lack of an adequate and domain-specific metric restrains progression of the field, and we revisit the evaluation procedure in anomaly localization.
In response, we propose the Area Under the Per-IMage Overlap (AUPIMO) as a recall metric that introduces two major distinctions. 
First, it employs a validation scheme based solely on normal images, which avoids biasing the evaluation towards known anomalies.
Second, recall scores are assigned \emph{per image}, which is fast to compute and enables more comprehensive analyses (\eg cross-image performance variance and statistical tests).  
Our experiments (27 datasets, 8 models) show that the stricter task imposed by AUPIMO redefines anomaly localization benchmarks: current algorithms are not suitable for all datasets, problem-specific model choice is advisable, and MVTec AD and VisA have \emph{not} been near-solved.
Available on GitHub\footnote{Official implementation: \githubRepo{}. Integrated in anomalib \githubAnomalib{}. This research was conducted during Google Summer of Code 2023 (GSoC 2023) with the anomalib team from Intel's OpenVINO Toolkit.}.
\end{abstract}
\section{Introduction}\label{sec:intro}
\gls{ad} is a machine learning task based on \emph{normal} patterns, meaning they are not of special interest at inference time.
As such, the model must identify deviations from the patterns observed in the training set, \ie \emph{anomalies}.
Within this domain, \gls{vad} focuses on image or video-related applications, including both the detection of anomalies in images (answering the question, \say{Does this image contain an anomalous structure?}) and the more precise task of anomaly localization or segmentation, where the goal is to determine if specific pixels belong to an anomaly.
Our emphasis is on anomaly localization in image applications (other modalities are out of the scope of this paper, but extensions of our work are possible and briefly discussed in \cref{sec:conc}).

\begin{figure}
    \centering
    \includegraphics[height=.24\linewidth]{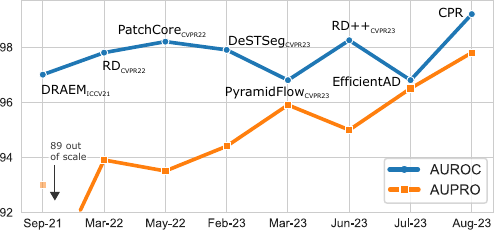}
    \hspace{5mm}
    \includegraphics[height=.24\linewidth]{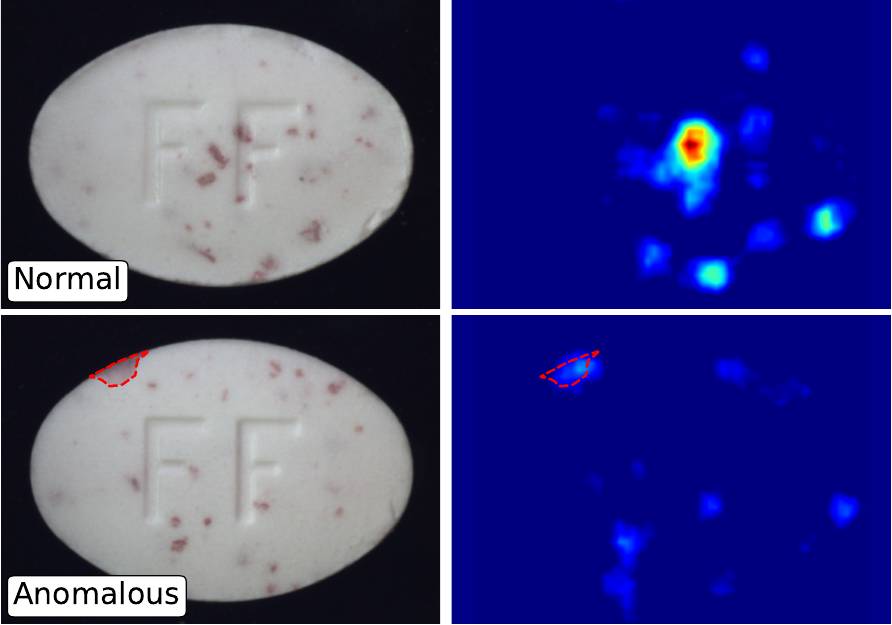}
    \caption{
        Left: performance on MVTec AD over time, approaching a near 100\% performance plateau.
        Right: images from the dataset Pill (left column) and their inferred anomaly maps (right column; higher values mean anomalous; JET colormap) from the best performing model in this dataset (EfficientAD; see \cref{app:benchmark}), with 98.7\% AUROC and 96.7\% AUPRO.
        The normal image (top) has higher anomaly scores than the anomaly (bottom).
    }
    \label{fig:paper-stats}
\end{figure}


Anomaly localization research has achieved significant progress, partly thanks to the increased availability of suitable datasets \cite{bergmann_mvtec_2019,zou_spot--difference_2022,mishra_vt-adl_2021,bozic_mixed_2021,krohling_bracol_2019}.
In particular, \gls{mvtecad} \cite{bergmann_mvtec_2019} and \gls{visa} \cite{zou_spot--difference_2022} comprise (together) 27 datasets (22 object and 5 texture-oriented) with high-resolution images and pixel-level annotations.

\gls{auroc} \cite{fawcett_introduction_2006} and \gls{aupro} \cite{bergmann_mvtec_2021} -- respectively, the \gls{auc} of the \gls{roc} and \gls{pro} curves (see \cref{sec:metrics-precursors}) -- have been used to evaluate anomaly localization, but it has been observed that the extreme class imbalance at pixel level inflates the scores produced by these metrics\footnotemark{} \cite{rafiei_pixel-level_2023,saito_precision-recall_2015}.
As a result, the performance numbers on \gls{mvtecad} and \gls{visa} reported in the literature are converging towards $100\%$ (\cref{fig:paper-stats}, left), giving the impression that these datasets have been solved.
Meanwhile, even the top performing models often fail to localize anomalous regions in some of the more challenging samples from these datasets while raising many \glspl{fp} (\ie a normal pattern wrongly flagged as anomalous in \cref{fig:paper-stats}, right).

\footnotetext{The term \say{metric} is used as a synonym for \say{performance measure} in this paper. It does \emph{not} refer to the mathematical concept of distance in a metric space.}

We argue that the anomaly localization literature urges a metric well-suited to its unique characteristic: the positive (anomalous) class is unknown beforehand and may have an unlimited number of modes.
While anomalous samples (even of different types) are available in public datasets, the goal of an \gls{ad} model is to detect \emph{any} type of anomaly.
Our work emphasizes on this unsupervised nature of the problem to build a performance metric that does \emph{not} depend on anomalies available at hand to avoid a bias towards known anomalies.

In response, we present the \textbf{Area Under the Per-Image Overlap (AUPIMO)} curve (\cref{sec:metrics-pimo}).
It relies on a clear separation of normal and anomalous images for, respectively, validation and evaluation of models -- thus avoiding class imbalance-related issues.
Its strict validation requirement sets a more challenging task in-line with the latest advances in the field.
Our work provides means to comprehensively compare models with image-specific evaluation scores and, along with the standard procedure proposed in \cref{sec:exp}, tackles cross-paper comparison issues.
In summary, our work presents the following contributions:
\begin{enumerate}[itemsep=3pt, topsep=3pt, parsep=0pt]
    \item A validation-evaluation framework based on strict low tolerance for \glspl{fp} on normal images only, which avoids conditioning the model behavior on known anomalies, thus providing a recall measure consistent with \gls{ad}'s unsupervised nature (\cref{sec:aupimo-props});
    \item Per-image recall scoring, enabling the analysis of cross-image performance variance and high-speed execution at high resolution both on CPU and GPU (\cref{sec:res}).
    \item Empirical evidence suggesting that \gls{mvtecad} and \gls{visa} datasets have \emph{not} been near-solved and that problem-specific model choice is advisable (\cref{sec:res}).
\end{enumerate}


\section{Related Work}\label{sec:related-work}
\gls{auroc} is a threshold-independent metric for binary classifiers \cite{fawcett_introduction_2006}, and it is widely used to assess anomaly localization, treating it as a pixel-level binary classification.
However, it has recently been argued that, in real-world applications, full or partial localization of anomalous regions is more relevant than pixel accuracy  \cite{Zhang_2023_CVPR,bergmann_mvtec_2019}.
Furthermore, it has been shown that \gls{auroc} is not suitable for anomaly localization datasets due to the extreme class imbalance \cite{saito_precision-recall_2015, rafiei_pixel-level_2023}, prompting the exploration of other evaluation metrics in the field \cite{rafiei_pixel-level_2023, Zhang_2023_CVPR, bergmann_mvtec_2019}.

Bergmann et al. \cite{bergmann_mvtec_2019} proposed a \gls{roc}-inspired curve called \glsfirst{pro}.
At each binarization threshold, it measures the region-scoped recall averaged across all anomalous regions available in the test set.
Notably, \gls{aupro} excludes thresholds yielding \gls{fpr} values above $30\%$ in the computation of the area under the \gls{pro} curve to force the metric to operate over a range of meaningful thresholds.

\begin{figure}[t]
    \centering
    \begin{subfigure}[b]{.49\linewidth}
        \centering
        \includegraphics[height=.385\linewidth]{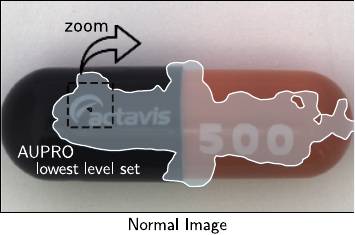}
        \includegraphics[height=.385\linewidth]{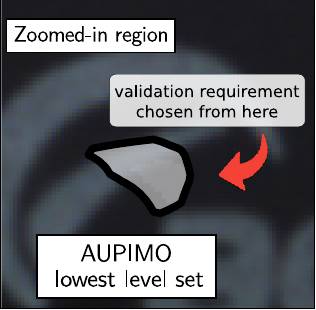}
        \caption{
            AUPIMO's integration bound is chosen so false positive regions in normal images are small.
            Zoomed-in region: the lowest (\ie largest) level set seen by AUPIMO in a normal image is insignificant compared to the structure of the image (more examples in \cref{app:fp}).
            AUPRO's equivalent is larger as it is chosen to yield recall-achievable results (\ie based on the anomalies).
        }
        \label{fig:asmaps-normal}
    \end{subfigure}
    \hfill
    \begin{subfigure}[b]{.49\linewidth}
        \centering
        \includegraphics[width=.49\linewidth]{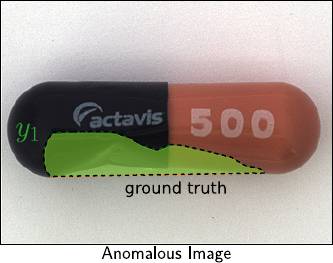}
        \hfill
        \includegraphics[width=.49\linewidth]{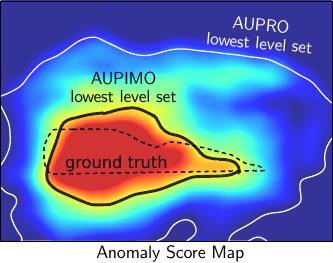}
        \caption{
            Left: anomalous image and its ground truth annotation mask (green region means anomalous).
            Right: anomaly map (JET colormap; blue/red means lower/higher anomaly score).
            The upper bound level sets are the lowest level sets seen by each metric.
            Their areas under the curve (AUCs) correspond to the average recall of the level sets above them (\ie inside these contours).
        }
        \label{fig:asmaps-anomalous}
    \end{subfigure}
    \\
    \caption{
        AUPRO and AUPIMO's upper bounds visualized as level sets from the anomaly score maps.
        Solid contours are level sets at thresholds yielding the maximum FPR in AUPRO (white) and AUPIMO (black).
        Images from the dataset MVTec AD/ Capsule.
    }
    \label{fig:asmaps}
\end{figure}


Recent studies have proposed metrics that index the thresholds based on recall instead of \gls{fpr}.
Rafiei et al. \cite{rafiei_pixel-level_2023} observed that the high pixel-level class imbalance in \gls{mvtecad} and similar anomaly localization datasets challenges the effectiveness of \gls{auroc} and \gls{aupro} for model comparison.
They concluded that the area under the \gls{pr} curve is a more suitable metric for \gls{ad} as it is conditioned on the positive class (anomalous).
Alternatively, other authors \cite{zou_spot--difference_2022,jeong_winclip_2023} have used the \gls{fonemax} score, which is the best achievable \gls{fone} (harmonic mean of recall and precision), implying an anomaly score threshold choice.
Zhang et al. \cite{Zhang_2023_CVPR} proposed the \gls{iap}, a modified version of the \gls{pr} curve where recall is defined at the region-level, counting a region as detected if at least half of its pixels are correctly detected.
This alternative recall metric is further used as a validation requirement (threshold choice) and the pixel-level precision is used to compare models (precision-at-$k\%$-recall).

\gls{aupimo} uses a validation criterium based only on normal images to avoid a bias towards detectable anomalies.
As detailed in \cref{sec:metrics}, we advocate in favor of normal-only validation to build an evaluation score in line with \gls{ad}'s unsupervised nature, while using recall only to rate models.
Finally, \gls{aupimo} uses image-scoped metrics, preserving the structured information from the images and making its computation significantly faster (\cref{fig:exectime}).

\section{Metrics}\label{sec:metrics}
We define a framework to compare \gls{auroc} and \gls{aupro} (\cref{sec:metrics-precursors}), introduce our new metric (\cref{sec:metrics-pimo}), and discuss its properties (\cref{sec:aupimo-props}). Key notation is listed in \cref{tab:notation}.

Our goal is to compare a model's output $\amap$ (an anomaly score map; higher means more likely to be anomalous) with its ground truth mask $\gt$ ($0$ and $1$ labels indicate \say{normal} and \say{anomalous}  respectively), illustrated in \cref{fig:asmaps-anomalous}.
We define $\regmask$ as a region in $\gt$ such that instances do not overlap (maximally connected components).
All metrics are \emph{pixel-wise} (one score/annotation per pixel), not \emph{image-wise} (one score/annotation per image) since our focus is to measure whether a model can detect anomalous structures \emph{within an image}.
We define the \glsfirst{fpr} and \glsfirst{tpr}, \ie recall, across three scopes: \textbf{set} (all pixels in all images confounded; subscript $\text{s}$), \textbf{per-image} (all pixels in an image; subscript $\text{i}$), and \textbf{per-region} (pixels in a single anomalous region; subscript $\text{r}$):
\begin{align}
&\setfpr: \thr \mapsto \setfprExp
&
&\settpr: \thr \mapsto \settprExp \label{eq:def-set-fpr-tpr} \\[2ex]
&\imfpr: \thr \mapsto \imfprExp
&
&\imtpr: \thr \mapsto \imtprExp \label{eq:def-im-fpr-tpr} \\[2ex]
& &
&\regtpr: \thr \mapsto \regtprExp \label{eq:def-regtpr}
\quad .
\end{align}
\noindent
Instances at each scope ($\regmask$, $\gt$, and $\amap$) are ommited in the notation for brevity.

\begin{table}[t]
    \scriptsize
    \centering
    \caption{
        Notation.
    }
    \label{tab:notation}
    \renewcommand*{\arraystretch}{1.15}
    \begin{subtable}[t]{.53\linewidth}
        \centering
        \begin{tabular}{p{0.15\linewidth}p{0.85\linewidth}}
            \toprule
            Symbol & Description \\
            \midrule
            $\imres \text{, } \pidx$ & Number and index of pixels in an image \\
            $\neg, \land$ & Pointwise logical negation/AND \\
            $\card{ \cdot }$ & Cardinality of a set or number of $1$s in a mask \\
            $\amap \geq \thr$ & Binarization of $\amap$ by \thr \\
            $\fprlbound \text{, } \fprubound$ & Integration lower/upper bounds \\
            \bottomrule
        \end{tabular}
    \end{subtable}
    \hfill  
    \begin{subtable}[t]{.43\linewidth}
        \centering
        \begin{tabular}{p{0.3\linewidth}p{0.67\linewidth}}
            \toprule
            Symbol & Description \\
            \midrule
            $\amap \in \amapdom$ & Anomaly score map\\
            $\gt \in \gtdom$ & Ground truth (GT) mask \\
            $\regmask \in \gtdom$ & Region mask \\
            $\thr \in \scoredom$ & Threshold \\
            $\amapset$, $\gtset$, $\regmaskset$ & Sets of $\amap$, $\gt$, and $\regmask$ \\
            \bottomrule
        \end{tabular}
    \end{subtable}
\end{table}

\subsection{Precursors: AUROC and AUPRO}\label{sec:metrics-precursors}
\begin{figure}[t]
    \begin{subfigure}[b]{0.325\linewidth}
        \centering
        \includegraphics[height=.72\linewidth]{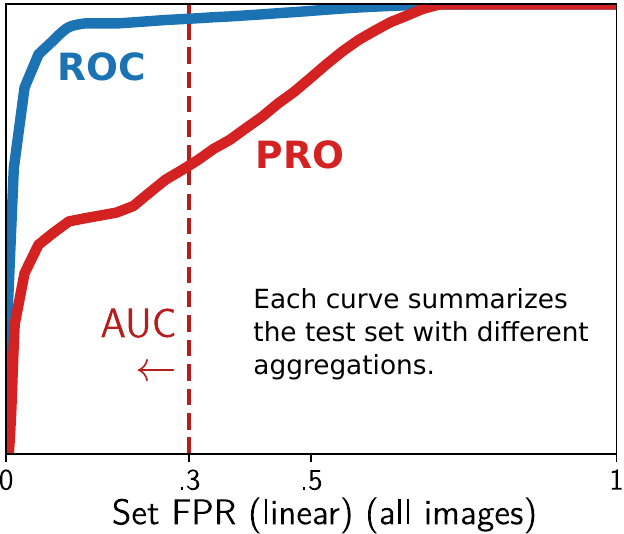}
        \caption{ROC and PRO curves}
        \label{fig:curves-bench-rocpro}
    \end{subfigure}
    \begin{subfigure}[b]{0.325\linewidth}
        \centering
        \includegraphics[height=.72\linewidth]{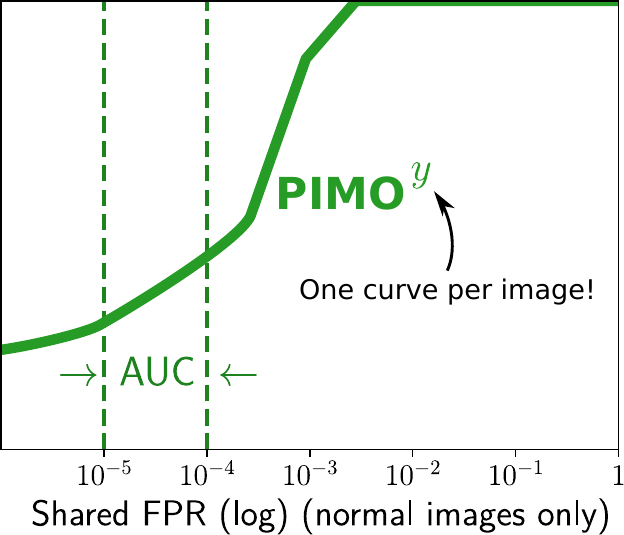}
        \caption{PIMO curve}
        \label{fig:curves-bench-pimo}
    \end{subfigure}
    \begin{subfigure}[b]{0.325\linewidth}
        \centering
        \includegraphics[height=.72\linewidth]{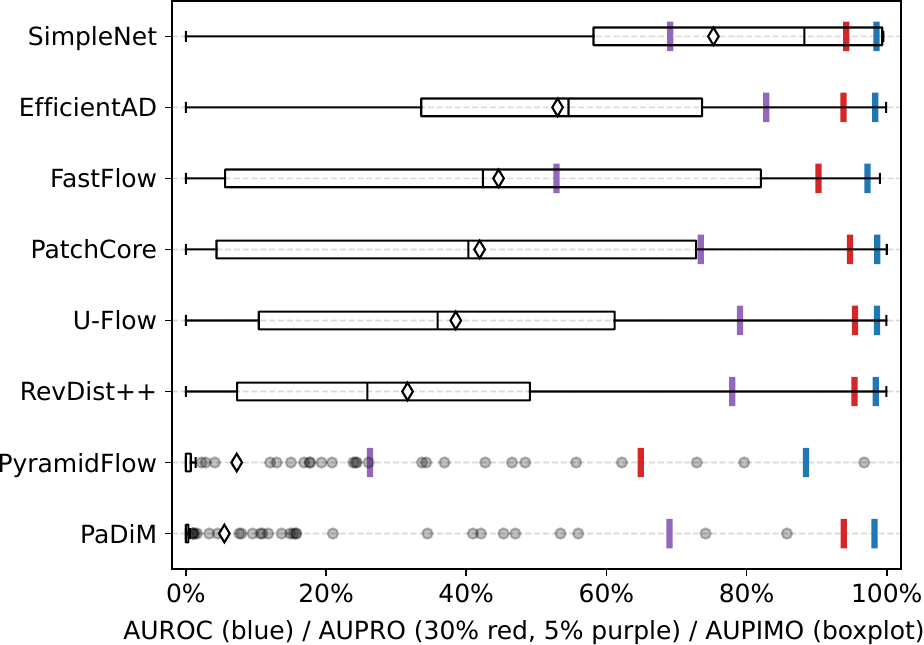}
        \caption{MVTec AD / Zipper}
        \label{fig:curves-bench-bench}
    \end{subfigure}
    \caption{
        (\subref{fig:curves-bench-rocpro}, \subref{fig:curves-bench-pimo}) ROC, PRO, and PIMO curves.
        The y-axes are TPR metrics: ROC uses the set TPR (all anomalous pixels from all images confounded); PRO uses the region-scoped TPR averaged across all regions from all images; PIMO uses the image-scoped TPR keeping one curve per anomalous image (no cross-instance averaging).
        The x-axes are FPR metrics shared by all instances (\ie anom. regions for PRO and anom. images for PIMO), which indexes the binarization thresholds.
        ROC and PRO use the set FPR (all normal pixels from all images confounded) in linear scale.
        PIMO uses the image-scoped FPR averaged accross normal images only in log scale.
        The curves are summarized by their (normalized) area under the curve (AUC), with different integration ranges: AUROC in $[0,1]$, AUPRO in $[0, 0.3]$\textsuperscript{\ref{footauprofive}}, and AUPIMO in $[10^{-5}, 10^{-4}]$.
        (\subref{fig:curves-bench-bench}) Benchmark on dataset \exampleDataset{} shows how their AUCs differ.
    }
    \label{fig:curves-bench}
\end{figure}

The \gls{roc} and \gls{pro} curves (\cref{fig:curves-bench-rocpro}) can be defined as
\begin{equation}
    \roc: \thr \mapsto \rocExp
    \quad \text{and } \quad
    \pro: \thr \mapsto \proExp
    \quad ,
\end{equation}
\noindent
where $\avgregtpr: \thr \mapsto \avgregtprExp$ is the average \gls{regtpr}; $\regtpr^{\regmask}$ refers to the $\regtpr$ applied to the instance $\regmask$ and $\regmaskset$ is the set of all $\regmask$ from all $\gt \in \gtset$.
Both curves trace the trade-off between \glsfirstplural{fp} and \glsfirstplural{tp} across all potential binarization thresholds.
Both use the \gls{setfpr} as the x-axis, but different recall measures as the y-axis, reflecting distinct \gls{regtpr} aggregation strategies.
\gls{pro} calculates the arithmetic average (equal weight to each region).
\gls{roc} uses the \gls{settpr}, which is equivalent to averaging the \glspl{regtpr} with region size weighting.
Their respective \glspl{auc}, \gls{auroc} and \gls{aupro}, summarize the curves into a single score:
\begin{equation}
    \label{eq:def-auroc-aupro}
    \auroc = \aurocExp
    \; \; \;
    \text{and}
    \; \; \;
    \aupro = \auproExp
    \; ,
\end{equation}
\noindent
where $\setfprinv$ is the inverse of $\setfpr$.
In practice, they are computed using the trapezoidal rule with discrete curves given by a sequence of anomaly score thresholds.

\gls{aupro} is restricted to thresholds such that $\setfpr(\thr) \in [0, \fprubound]$ (\ie to the left of the vertical line in \cref{fig:curves-bench-rocpro}), where $\fprubound$ is the upper bound \gls{fpr}.
This means that \gls{aupro} only accounts for recall values obtained from level sets higher than (\ie inside) the white level set in the anomaly score map in \cref{fig:asmaps}.
The default value of $\fprubound = 30\%$\footnotemark{} is based on the intuition that at such \gls{fpr} levels the segmentation contours of the anomalies are no longer meaningful \cite{bergmann_mvtec_2021}, so that should be the \say{worst case}.
From this perspective, the \gls{fpr} restriction in \gls{aupro} acts as a model validation -- an implicit requirement since a partial threshold choice is imposed.

\footnotetext{
    We also considered a \gls{aupro} with $\fprubound = 5\%$ (noted \gls{auprofive}) in our experiments for the sake of making the metric more challenging.
    \label{footauprofive}
}

\subsection{Our Approach: AUPIMO}\label{sec:metrics-pimo}

\gls{pro} measures region-scoped recall at each binarization threshold, which are indexed by an \gls{fpr} metric (the x-axis) shared by all region instances.
We generalize this idea and employ the term \gls{shfpr} ($\shfpr$) to refer to \say{\emph{any} \gls{fp} measure shared by all anomalous instances.}
In our approach, the \gls{setfpr} used as x-axis by \gls{roc} and \gls{pro} is replaced by the average \gls{imfpr} on normal images only: $\shfpr: \thr \mapsto \shfprExp$, where $\gtset^0 \subset \gtset$ contains only and all normal images in $\gtset$, and $\imfpr^{\gt}$ refers to $\imfpr$ computed on instance $\gt$.
This design choice is a major counterpoint with previous approaches, and its implications are discussed in \cref{sec:aupimo-props}.
The \textbf{\glsfirst{pimo}} curve (\cref{fig:curves-bench-pimo}) and its \gls{auc} are defined as
\begin{equation}
    \pimo: \thr \mapsto \pimoExp
    \; \; \;
    \text{and}
    \; \; \;
    \aupimo = \aupimoExpDlogz
    \quad ,
    \label{eq:def-pimo-aupimo}
\end{equation}
\noindent
where the integration bounds have default values $\fprlbound = 10^{-5}$ and $\fprubound = 10^{-4}$.
To have a better resolution at low \gls{fpr} levels, the x-axis is in log-scale, and the term $1/\log(\fprubound / \fprlbound)$ normalizes the integral's score to $[0, 1]$.
Contrasting with \gls{auroc} and \gls{aupro}, which define a single score for the entire test set, we keep one score per image (superscript $^\gt$).

\subsection{AUPIMO's properties}\label{sec:aupimo-props}
\gls{aupimo} significantly diverges from its predecessors by: (1) considering only normal instances for validation and using a stricter requirement (integration range in the x-axis), (2) evaluating metrics at the image scope, and (3) calculating individual scores for each image.
This section discusses the implications and advantages of these design choices.

\paragraph{Bias-free validation}
\gls{auroc} is a threshold-independent metric, which limits its usage in real-world applications that require threshold selection for inference.
\gls{aupro} addresses this by imposing an \gls{fpr} restriction, which selects a range of valid thresholds, thus carrying an implicit model validation based on the \gls{setfpr}.
\gls{aupimo} uses a similar strategy, but -- to produce a bias-free score -- we propose that the validation metric (x-axis of the curve) should only use normal images, while anomalous images are only used for evaluation.

\gls{ad} is often viewed as a binary classification problem, yet this simplification is misleading.
While the normal class is well-defined by the training set, the anomalous class is, by definition, unknown, unbounded, thus inherently multi-modal.
Public datasets (\eg \gls{mvtecad} and \gls{visa}) provide various types of anomalies, but the objective in \gls{ad} is to detect \emph{any} type of anomaly.
As the positive class in \gls{ad} can have an unlimited number of modes, we argue that an evaluation metric in benchmarks should avoid conditioning the model behavior (\ie creating a bias, \eg selecting a threshold range) based on \emph{known} anomalies.

The x-axis in \gls{aupimo} ($\shfpr$) is built only from normal images, which can be reasonably assumed from the same distribution as the training set.
In this framework, the variance of the normal class coming from acquisition conditions, sensor noise, \etc is accounted for in the validation metric ($\shfpr$).
By ensuring that these variations are not falsely detected, the model's capacity to detect anomalies is isolated from the normal class's variability.
This essential change avoids biasing the evaluation metric towards available anomalies, which is consistent with the unsupervised nature of \gls{ad}.
Note that an alternative \gls{aupro} could be defined in the same way, but \gls{aupimo} carries additional advantages discussed below.

\paragraph{Anomaly-dependent metrics}
The \gls{aupr} and its variant \glsfirst{iap} \cite{Zhang_2023_CVPR} use recall measures on the x-axis and precision on the y-axis.
Similar to the \glspl{auc} defined in \cref{sec:metrics-precursors} and \cref{sec:metrics-pimo}, they express the average of the y-axis over a range of thresholds, which are indexed by the x-axis.
Using the recall as x-axis biases the metric in favor of detectable anomalies, making the metric sensitive to the distribution of known anomalies.
The threshold at the integration lower bound is the maximum full-recall threshold, making them sensitive to hard anomalies\footnote{Reminder: lower threshold means higher recall, so the anomalies with lowest anomaly score are the hardest.} -- while not revealing them.
Conversely, easy anomalies can be over-represented because low-recall thresholds are coverered  -- \ie unnecessarily high thresholds are accounted for.

The \gls{fonemax} score and \gls{iap} further choose, respectively, optimal and minimum thresholds based on the recall.
Similarly, \gls{aupro} validates models using anomalous images as well because it restricts the \gls{setfpr} (\cref{eq:def-set-fpr-tpr}), which encompasses all test images (thus the normal-annotated pixels in anomalous images).
While such threshold choices are useful for practical applications, we argue that benchmarks should prefer bias-free metrics so that model comparison is more consistent across different datasets and applications.

Finally, \gls{aupimo}'s validation is insentive to imprecisions in the anomaly annotations -- \ie when only loose bounding box annotations are available.
Other model conditioning criteria -- as in \gls{fonemax} and \gls{iap} in particular -- carry pixel-level imprecision but \gls{aupimo} is not affected because normal images are only annotated at the image level. 

\paragraph{Low tolerance}
From an application perspective, anomalies are expected to contain information deserving the user's attention.
A high \gls{fpr} can lead to user frustration and diminish trust in the model.
To tighten evaluation, we restrict the \gls{fpr} range in \gls{aupimo} to be between $10^{-5}$ and $10^{-4}$ for datasets like \gls{mvtecad} and \gls{visa}.
At such levels, the \gls{fp} regions in normal images are small compared to the structures seen in the images (see \cref{fig:asmaps} and \cref{app:fp}).
An \gls{aupimo} score can be interpreted as the \say{\emph{average segmentation recall in an anomalous image given that the model (nearly) does not yield \gls{fp} regions in normal images}}.
These default values were chosen to establish a challenging task in-line with recent advances in research, but they can be adapted to application-specific needs.

\paragraph{AUPRO vs. AUPIMO}
\cref{fig:asmaps} shows a visual comparison between \gls{aupro} and \gls{aupimo}.
The upper bound in \gls{aupro} is chosen from a precionsion-inspired criterion (\say{beyond that point the anomaly segmentations are no longer useful}), so the \gls{fp} regions on normal images can be large.
In contrast, \gls{aupimo} chooses a more conservative upper bound.
The model conditioning in \gls{aupimo} ensures that \gls{fp} regions in normal images are insignificant.
As a result, its recall on the anomalous region (on the right in \cref{fig:asmaps-anomalous}) is lower than \gls{aupro}'s -- which is expected.

\paragraph{Image-scoped metrics}
Note that the set-scoped metrics in \gls{auroc} and \gls{aupro} are ill-suited for images because information within each image is disregarded (all pixels are confounded).
\gls{aupimo} avoids this problem by only using image-scoped metrics (\ie ratios of pixels within each image).
Image-scoped measures account for image structure, are fast to compute (\cref{fig:exectime}), and are robust to noisy annotations (see \cref{fig:robustness-tiny-blobs}).

\paragraph{Image-specific scores}
Since each curve/score refers to an image file, it is easy to index scores to instances\footnotemark.
Achieving the same with region-based scores would require more metadata, and finding connected regions is implementation-sensitive.
For instance, Anomalib's \cite{akcay_anomalib_2022} CPU and GPU-based implementations are from \texttt{opencv-python} \cite{bradski_opencv_2000} and \texttt{kornia} \cite{riba_kornia_2020}, and the \gls{aupro} scores slightly differ.
Per-image scores enable fine-grained analyses otherwise impossible with \gls{auroc} and \gls{aupro}.
Score distributions (\eg \cref{fig:curves-bench-bench}) -- instead of single-valued scores -- provide insight into performance variance, which we exploit to select representative samples for qualitative analysis in \cref{app:benchmark}.
Finally, it also enables the use of statistical tests, which we showcase in an ablation study in \cref{app:ablation}.

\footnotetext{A standard format is proposed in \cref{app:benchmark} and implemented in our repository.\label{footnotescorepublicationformat}}

\section{Experimental Setup}\label{sec:exp}
We benchmark the datasets from \gls{mvtecad} and \gls{visa} with \gls{sota} models to compare the performances reported in terms of \gls{auroc}, \gls{aupro}, and \gls{aupimo}.
We also report \gls{aupro} with $\fprubound = 5\%$ (\gls{auprofive}) for the sake of comparing with a more challenging alternative of that metric.

We reproduce a selection of models: \gls{padim} \cite{defard_padim_2021} from ICPR 2021, \gls{patchcore} \cite{roth_towards_2022} from CVPR 2022, \gls{simplenet} \cite{liu_simplenet_2023}, \gls{pyramidflow}\footnotemark \cite{lei_pyramidflow_2023}, and \gls{reversedistpp} \cite{tien_revisiting_2023} from CVPR 2023, along with the recently published models \gls{uflow} \cite{tailanian_u-flow_2023}, \gls{fastflow} \cite{yu_fastflow_2021}, and \gls{efficientad} \cite{batzner_efficientad_2023}.
Our aim is to ensure a comprehensive evaluation with a set of different algorithm families.
This selection includes methods based on memory bank (\gls{patchcore}), reconstruction (\gls{simplenet}), student-teacher framework (\gls{reversedistpp}, \gls{efficientad}), probability density modelling (\gls{padim}), and normalizing flows (\gls{fastflow}, \gls{pyramidflow}, \gls{uflow}).

\footnotetext{
Our \gls{aupro} results significantly differ from \gls{pyramidflow}'s paper.
Their implementation has higher scores because it does not apply the maximum \gls{fpr} ($30\%$) as proposed by \cite{bergmann_mvtec_2021}
\url{https://github.com/gasharper/PyramidFlow} (commit \texttt{6977d5a}), see function \texttt{compute\_pro\_score\_fast} in the file \texttt{util.py}.
\label{footpyramidflow}
}

All models were trained with $256 \! \times \! 256$ images (downsampled with bilinear interpolation, no center crop), and with the hyperparameters reported in the original papers.
We used the official implementations or Anomalib \cite{akcay_anomalib_2022}.
The implementations of \gls{auroc} and \gls{aupro} are from Anomalib \cite{akcay_anomalib_2022}.
Details provided in \cref{app:benchmark}.

Cross-paper comparisons in the anomaly localization literature often have conflicting evaluation procedures.
We aim to tackle this issue by proposing our evaluation guidelines as a standard:
(1) compute test set metrics at the annotations' full resolution with bilinear interpolation for resizing the anomaly score maps if necessary;
(2) do \emph{not} apply crop to the input images;
(3) publish per-image scores\textsuperscript{\ref{footnotescorepublicationformat}};
(4) (ideally) report the score distribution (\eg boxplots as in \cref{fig:curves-bench-bench}).
Details in \cref{app:benchmark}.

\begin{figure*}[t]
    \centering
    \includegraphics[width=\linewidth]{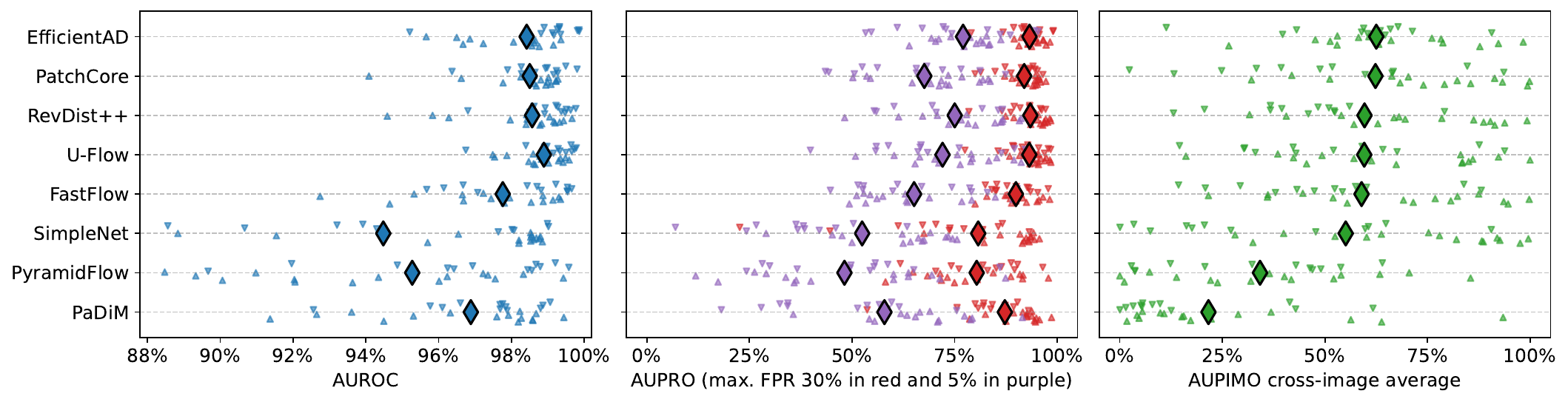}
    \caption{
        Dataset-wise comparison.
        Each triangle is a set-scoped score (AUROC, AUPRO, and AUPRO$_{5\%}$) or a cross-image statistic (average AUPIMO) from a dataset in MVTec AD ($\vartriangle$) or VisA ($\triangledown$).
        Diamonds are cross-dataset averages (all confounded).
        Plots have different x-axis scales.
        AUPIMO reveals that all models have a large cross-problem variance, meaning that none of the models is robust to all problems.
    }
    \label{fig:datasetwise-maintext}
  \end{figure*}


\section{Results}\label{sec:res}

In this section we comment on the results of a single dataset (\cref{fig:curves-bench-bench}), present a summary across all datasets (\cref{fig:datasetwise-maintext}), and compare \gls{auroc}, \gls{aupro}, and \gls{aupimo} in terms of the execution time and robustness to noisy annotation.
Due to the space constraints, additional results are available in \cref{app:add-res} and the benchmarks from all datasets in \gls{mvtecad} and \gls{visa} are documented in \cref{app:benchmark}.


\begin{figure}[t]
\begin{subfigure}[b]{.495\linewidth}
    \centering
    \includegraphics[width=.9\linewidth]{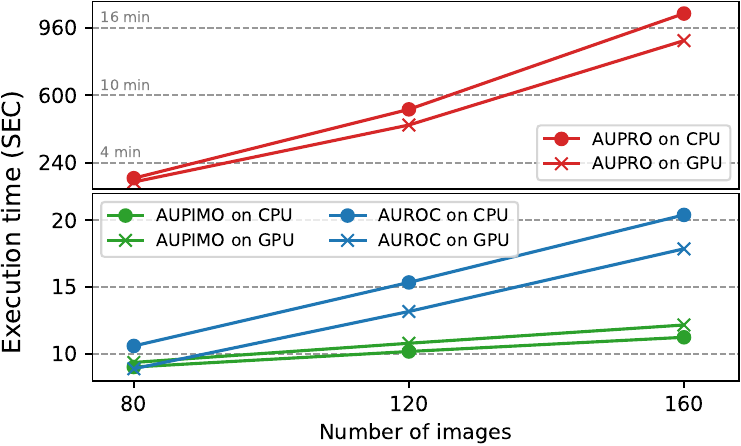}
    \caption{Execution time}
    \label{fig:exectime}
\end{subfigure}
\begin{subfigure}[b]{.495\linewidth}
    \centering
    \includegraphics[width=\linewidth]{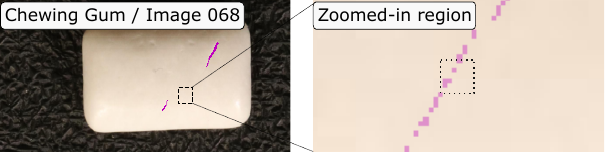}
    \includegraphics[width=\linewidth]{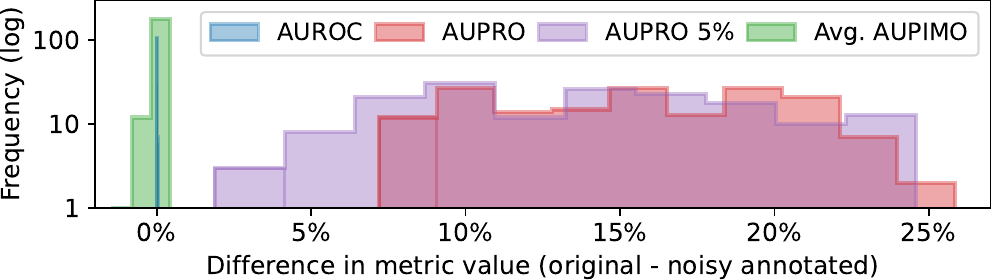}
    \caption{Robustness}
    \label{fig:robustness-tiny-blobs}
\end{subfigure}
\caption{
    (a) Execution time of metrics on MVTec AD / Screw dataset (image resolution of $1024 \times 1024$; average times over 3 runs).
    (b, top) An anomalous sample from the dataset VisA / Chewing Gum superimposed with its annotation (pink) shows meaningless, tiny (even 1-pixel) regions (the mask has \textit{not} been downsampled).
    (b, bottom) Robustness to noisy annotation.
    Histograms show the distribution of the difference between the scores without and with the synthetic mistakes (closer to zero is better).
}
\end{figure}


\paragraph{Benchmark on \exampleDataset{}}
\cref{fig:curves-bench-bench} illustrates two common observations in our benchmarks.
First, it shows how \gls{auroc} and \gls{aupro} fail to reveal differences between models (\eg differences of $0.1\%$ and $0.4\%$ between the two best models).
While \gls{auprofive} amplifies the differences, \gls{aupimo}'s strict validation causes the best model to stand out more clearly.
Note that \gls{auprofive} and \gls{aupimo} show different rankings, which might be attributed to how they weight small anomalies differently.
Second, image-specific performance often has large variance and the best models have left-skewed \gls{aupimo} distributions -- \cf the best models per dataset in \cref{app:permodel}.
In \cref{fig:curves-bench-bench} for example, several models have worst and best-case samples at 0\% and 100\% \gls{aupimo} respectively.
Fortunately, \gls{aupimo} provides the means to investigate this by programatically identifying specific instances or anomaly types not well-detected by a model.

\paragraph{Cross-dataset analysis}
\cref{fig:datasetwise-maintext} reveals two key insights regarding the \gls{sota} in anomaly localization.
First, the benchmark datasets from \gls{mvtecad} and \gls{visa} still have room for improvement.
While \gls{auprofive}'s (purple) stricter validation is more challenging, \gls{aupimo} (green) reveals that even the best models have failure cases when constrained to low \gls{fp} tolerance.
We argue that setting such a challenging standard will push the next generation of models to achieve a more trustworthy task: high anomaly recall with near-zero false positives.
Second, none of the models consistently achieves reasonable performance across all datasets.
For example, despite \gls{patchcore}'s high performance in many problems, it performs poorly on \gls{visa} / Macaroni 2 (details in \cref{app:permodel}).
Meanwhile, \gls{efficientad} has a reasonable performance on this dataset, thus the dataset is not unsolvable with the current models.
This provides a useful insight for practitioners: problem-specific model choice is highly advised because a model's failure in one dataset does not imply failure in another one.

\paragraph{Execution time}
Having computationally efficient metrics is essential to enable fast iterations and not create computational bottlenecks in research and development.
\cref{fig:exectime} shows that AUROC and AUPIMO have comparable execution time, but AUPRO is significantly slower both on CPU and GPU.
The main reason is that \gls{aupro} requires connected component analysis, while \gls{auroc} and \gls{aupimo} do not.
\gls{aupimo}'s implementation relies on simple operations, enabling the use of \texttt{numba} \cite{lam_numba_2015} to further accelerate the computation (reported execution times include the just-in-time compilation).
The GPU used was an NVIDIA GeForce RTX 3090 and the CPU was an Intel Core i9-10980XE.
Note that the chosen model does not influence the execution time because the anomaly score maps are precomputed.

\paragraph{Robustness}
In real-world use-cases, high-quality annotation is hard to acquire or even to define.
\cref{fig:robustness-tiny-blobs} shows an example of a ground truth mask where noisy regions can be seen.
We found this issue to be prevalent in \gls{visa} (more examples in \cref{app:tiny-blobs}).
In the \gls{pro} curve, these tiny regions have the same weight as the actual anomalous regions.
In contrast, \gls{aupimo} is more robust to this issue due to their limited contribution to the overall image score.
\cref{fig:robustness-tiny-blobs} demonstrates this in an experiment with artificially added noise.
Random mistakes mimicking statistics from VisA are added to the datasets in MVTec AD.
We generate one noisy mask for each anomalous mask by adding randomly shaped anomalous regions to it. 
The number and size of the noisy regions are randomly sampled with probabilities matching the statistics of the VisA dataset (average frequencies from \cref{tab:tiny-blobs} in \cref{app:tiny-blobs}).


\section{Conclusion}\label{sec:conc}

We introduced \glsfirst{aupimo}: a novel recall metric tailored for anomaly localization addressing the limitations of its predecessors (\gls{auroc} and \gls{aupro}) and formalizing a validation-evaluation framework.
As a guiding principle, it was proposed that the validation step should only depend on normal images to avoid biasing the model behaviour towards known anomalies, thus making the metric consistent with the unsupervised nature of \gls{ad}.
Finally, a stringent false positive restriction is proposed to establish a more challenging task on contemporary benchmark datasets and expose differences between models.

\gls{aupimo} is built with image-scoped metrics and enables simple assignment of image-specific scores.
As demonstrated, these design choices offer advantages in terms of computational efficiency (see \cref{fig:exectime}), fine-grained performance analysis (see \cref{fig:curves-bench-bench} and \cref{app:ablation}), and resilience against noisy annotation (see \cref{fig:robustness-tiny-blobs}).

Evaluating eight recent models on 27 datasets with \gls{aupimo} revealed a significant insights about the \gls{sota} in anomaly localization.
We show evidence that problem-specific model selection is highly advised, raising further questions for future research.
Namely, can one identify dataset traits causing a model to succeed or fail?
Or conversely, which model features should one look for to succeed on a specific problem?

\paragraph{Limitations}
In this paper we focused on (2D) image anomaly localization, but \gls{aupimo} can be easily adapted to 3D imaging (\eg X-ray tomography), 3D point clouds (\eg LiDAR), and video-based applications (a proof of concept is shown in \cref{app:video}).
Other domains like times series would require more careful adaptation, which is left for future work.
As a recall metric, the notion of segmentation quality is not covered by \gls{aupimo}, but \cref{app:prec-vs-iou} briefly discusses alternatives based on the same validation-evaluation principle.

\section{Acknowledgements}
This research was conducted during Google Summer of Code 2023 (GSoC 2023) with the anomalib team from Intel's OpenVINO Toolkit.
We would like to thank the OpenVINO team for their support and feedback during the project.
We would like to thank Mat\'ias Tailanian for having collaborated by training the \gls{uflow} models and providing the evaluation results for the benchmark.
\bibliography{main}
\newpage
\appendix
\onecolumn
\clearpage
\section{False positives on normal images}\label{app:fp}

We argue that, in \glsfirst{ad}, the negative class (normal) is the only well-defined class, and that \gls{fpr} is a meaningful metric to validate models.
The positive class (anomalous) is \emph{not} a well-defined concept because it covers the entire complement of the normal class.
As such, it is impossible to cover all types and variations.
Based on this principle, we argue that it is problematic to use anomalous samples for model validation (not to be confused with model \textit{evaluation}).
For this reason, we propose the validation to depend solely on normal instances, thus based on \glspl{fp}.

For the sake of complementing the discussion, we present an alternative to the (pixel-wise, image-scoped) \gls{fpr} used in \gls{aupimo}.
Counting the number of regions falsely detected as anomalous can be used as meaningful metric to validate (\ie constrain) models.
However, such metric is not used in \gls{aupimo} because it is inconvenient to compute, so we propose the \gls{fpr} as a proxy.
Finally, we present visual examples of \gls{fp} masks at different levels of \gls{fpr} to provide an intuition of what it represents in practice.

\subsection{Rate vs. number of regions}\label{app:fp-levels-num-blobs}

In this section the relation between two (pixel-wise, image-scoped) metrics is analyzed (both measured on normal images at different binarization thresholds of anomaly score maps):
\begin{enumerate}
    \item \glsfirst{fpr}: the ratio between the number of \gls{fp} pixels and the total number of pixels;
    \item \glsfirst{numfpreg}: the number of maximally connected \gls{fp} regions.
\end{enumerate}
\noindent

To be trusted in real-world applications, an \gls{aloc} model is expected to find image structures worth the user's attention.
Raising false detections eventually diminishes the user's interest, so it should happen as rarely as possible.
One could assume, for instance, that users eventually investigate detected anomalies manually -- or even programatically.
From this perspective, we argue that the \gls{numfpreg} is an informative metric in practice because it directly relates to how often a user would investigate \gls{fp}s, so it is a good estimator of the human cost for using the model (\ie how often one's time is wasted).
A good estimate of the expected \gls{numfpreg} would allow a user to set a threshold based on its operational cost.

However, computing \gls{numfpreg} requires connected component analysis, which has two major drawbacks.
First, it is slow to compute, especially on the CPU.
Second, some implementations use an iterative process that may not converge in some cases.
For instance, the implementation in \texttt{kornia} \cite{riba_kornia_2020} (see \texttt{kornia.contrib.connected\_components}).
The \gls{fpr}, on the other hand, is fast to compute and, as we show next, can be used as a proxy for the \gls{numfpreg} at low \gls{fp} levels.

\paragraph{Experiment}
Anomaly score maps from our experiments were randomly sampled from the set of normal images, upscaled with bilinear interpolation to the same resolution as the original annotation masks, binarized with a series of thresholds, and the \gls{numfpreg} and the \gls{fpr} were computed for each binary mask.
All models and datasets were confounded on purpose because we seek to understand the relationship between \gls{fpr} and \gls{numfpreg} \emph{in general}, not for a specific model or dataset.
Thresholds were chosen such that a series of logarithmically-spaced \gls{fpr} levels from $10^{-5}$ to $10^{-3}$ are covered.
A random multiplying factor $\in \left[ 0.9 \, , \, 1.1 \right]$ was added to each target \gls{fpr} value in this range (like a jitter).
Assumptions:
\begin{enumerate}
    \item Each threshold is interpreted as an operational threshold set to automatically obtain binary masks from an \gls{ad} model;
    \item Both metrics are computed at the image scope (\ie ratio of pixels and number of regions in each image);
    \item In an real-life scenario, the expected values of these metrics would be estimated to describe a model's behavior to control its operational cost.
\end{enumerate}

\begin{figure}[t]
  \centering
  \begin{subfigure}{0.49\linewidth}
    \includegraphics[width=\linewidth]{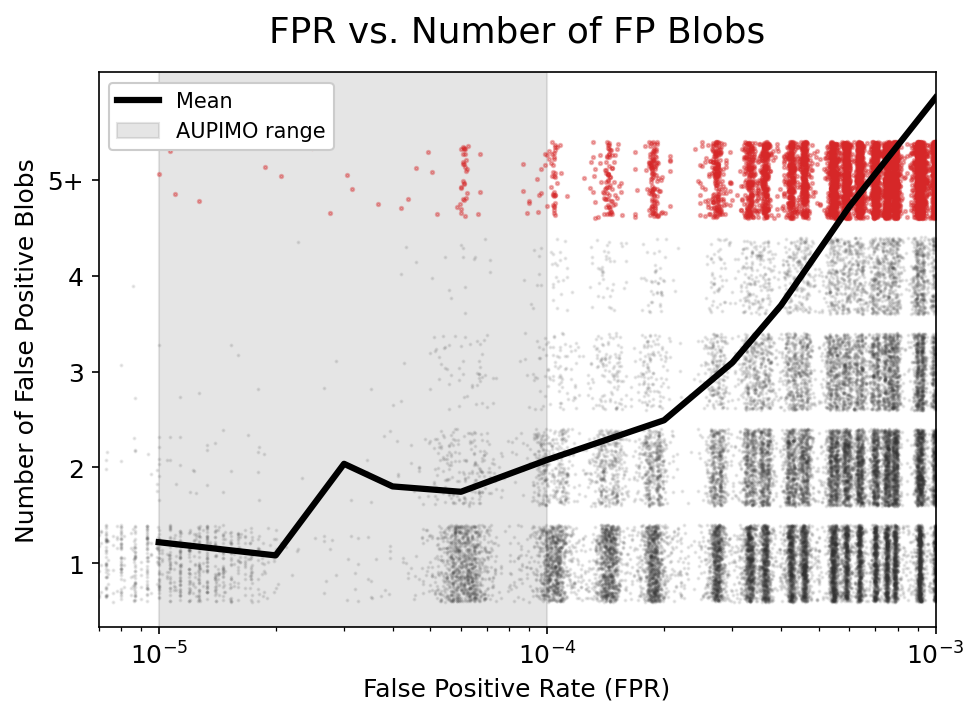}
    \caption{Scatter plot.}
    \label{fig:fp-levels-num-blobs-scatter}
  \end{subfigure}
  \hfill
  \begin{subfigure}{0.49\linewidth}
    \includegraphics[width=\linewidth]{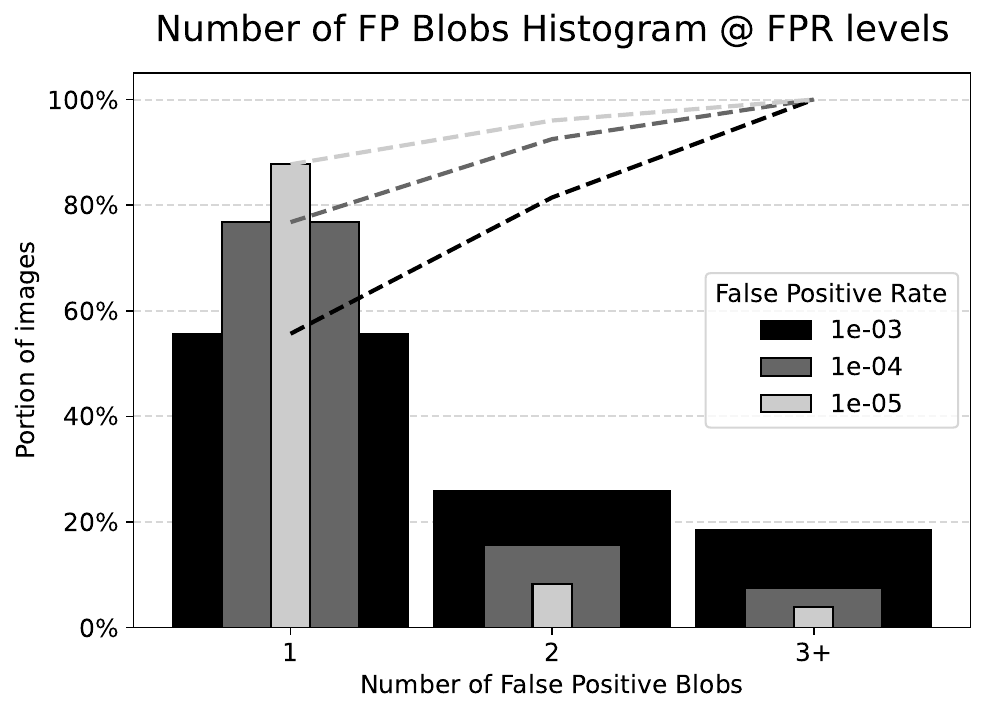}
    \caption{Histograms.}
    \label{fig:fp-levels-num-blobs-hist}
  \end{subfigure}
  \caption{
        False Positivity. 
        How Image False Positive Rate (ImFPR) relates to the Number of False Positive Regions (NumFPReg).
    }
  \label{fig:fp-levels-num-blobs}
\end{figure}

\cref{fig:fp-levels-num-blobs-scatter} shows a scatter plot of \gls{fpr} (X-axis, in logarithmic scale) vs. \gls{numfpreg} (Y-axis).
\gls{numfpreg} was clipped to the maximum value of $5$ and jitter was added to avoid overlapping points.
A mean line is displayed in black.
The Y-axis values of the mean line are computed as the average \gls{numfpreg} in the bins centered around the pre-set \gls{fpr} levels.

\cref{fig:fp-levels-num-blobs-hist} shows histograms (counts are numbers of images) of \gls{numfpreg} at three \gls{fpr} levels: $10^{-5}$, $10^{-4}$, and $10^{-3}$.
At each level $L$, all the points from the scatter in the range $[L/2, 2L]$ are accounted to have a sufficient number of samples.
The histograms are normalized to sum to 1.
The dashed lines show the sum of the bars' values from left to right.

\paragraph{Results}
\cref{fig:fp-levels-num-blobs} shows that the \gls{fpr} can effectively be used as a proxy for the number of \gls{fp} regions:
\begin{enumerate}
    \item \gls{fpr} and \gls{numfpreg} correlate positively;
    \item The majority of images have $\le 2$ regions (more than $90\%$ at \gls{fpr} $10^{-4}$ and nearly $100\%$ at \gls{fpr} $10^{-5}$);
    \item Inside \gls{aupimo}'s integration range, the average \gls{numfpreg} tends to $1$, so the \gls{fpr} generally equals the relative size of the single \gls{fp} region in the mask.
\end{enumerate}
In summary, at \gls{aupimo}'s default integration range, the \gls{fpr} tends to translate to the maximum relative size of \gls{fp} regions in normal images because they tend to have a single \gls{fp} region.

As a practical implication, \gls{aupimo}'s bounds can be leveraged to filter out model predictions.
For instance, one can ignore detected regions with areas smaller than \gls{aupimo}'s lower bound.
Notice in  that \gls{mvtecad}'s datasets do not have anomalies with relative size smaller than $10^{-5}$, and very few as small as $10^{-4}$ (see \cref{app:tiny-blobs}).

\subsection{Visual intuition of FP levels}\label{app:fp-levels}

We intend to build an intuition of what \glsfirst{imfpr} ($\imfpr$) levels visually represent on normal images.
The \gls{imfpr} on normal images is the relative area covered by an \gls{fp} mask.
As shown in the previous section, with \gls{aupimo}'s low levels of \gls{fpr}, it further tends to translate to the size of a single \gls{fp} region.

\cref{fig:fp-levels} shows examples of normal images from all the datasets in \gls{mvtecad} and \gls{visa} superposed by \gls{fp} masks.
Each dataset is in a row with three samples from the test set.
Each image is presented with a zoom on the right (the zoomed area is highlighted in the original image with a dashed rectangle).
Each color corresponds to a predicted mask at a given ImFPR level.
Color code:
\begin{multicols}{2}
\begin{enumerate}
    \item Darker blue is $\imfpr = 10^{-2}$;
    \item Lighter blue is $\imfpr = 10^{-3}$;
    \item White is $\imfpr = 10^{-4}$;
    \item Black is $\imfpr = 10^{-5}$.
\end{enumerate}
\end{multicols}
The masks are generated from the anomaly score maps produced by a randomly picked model from our benchmark.
The different masks in a single image are generated from the same anomaly score map (\ie same model), but different samples may have masks from different models.

Inside \gls{aupimo}'s integration bounds ($10^{-5} \sim 10^{-4}$, \ie between black and white in \cref{fig:fp-levels}), \gls{fp} regions become barely visible at the image scale and generally irrelevant compared to the objects' structures.

Disclaimer: the \glsfirst{shfpr} used in \gls{pimo} is the \emph{average} \gls{imfpr} across all normal images, so it is not to be confused with the \gls{imfpr} of a single image.
This visual intuition should be understood as an average behavior, not as a strict rule.


\begin{figure}[ht]
\centering

\begin{subfigure}{\linewidth}
    \centering
    \includegraphics[width=0.16\linewidth,valign=t,keepaspectratio]{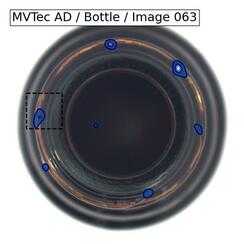}
    \includegraphics[width=0.16\linewidth,valign=t,keepaspectratio]{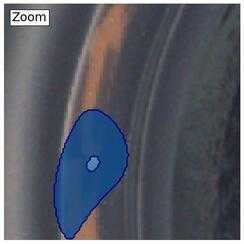}
    \includegraphics[width=0.16\linewidth,valign=t,keepaspectratio]{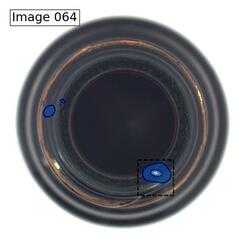}
    \includegraphics[width=0.16\linewidth,valign=t,keepaspectratio]{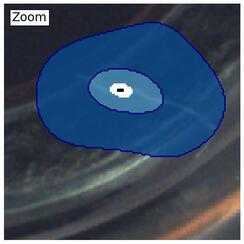}

\end{subfigure}
\\ 
\begin{subfigure}{\linewidth}
    \centering
    \includegraphics[width=0.16\linewidth,valign=t,keepaspectratio]{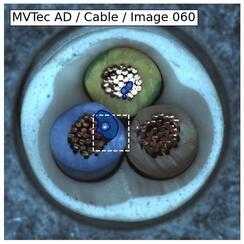}
    \includegraphics[width=0.16\linewidth,valign=t,keepaspectratio]{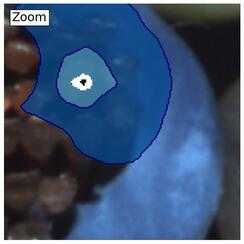}
    \includegraphics[width=0.16\linewidth,valign=t,keepaspectratio]{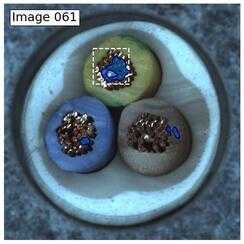}
    \includegraphics[width=0.16\linewidth,valign=t,keepaspectratio]{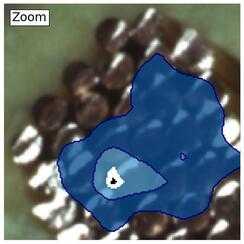}

\end{subfigure}
\\ 
\begin{subfigure}{\linewidth}
    \centering

\end{subfigure}
\\ 
\begin{subfigure}{\linewidth}
    \centering
    \includegraphics[width=0.16\linewidth,valign=t,keepaspectratio]{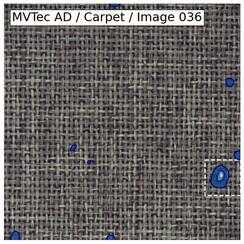}
    \includegraphics[width=0.16\linewidth,valign=t,keepaspectratio]{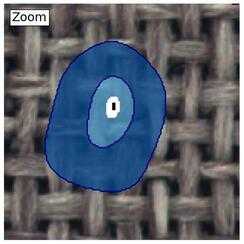}
    \includegraphics[width=0.16\linewidth,valign=t,keepaspectratio]{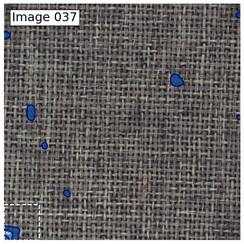}
    \includegraphics[width=0.16\linewidth,valign=t,keepaspectratio]{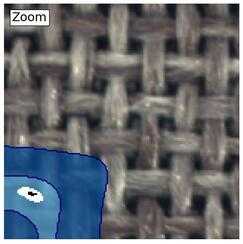}

\end{subfigure}
\\ 
\begin{subfigure}{\linewidth}
    \centering
    \includegraphics[width=0.16\linewidth,valign=t,keepaspectratio]{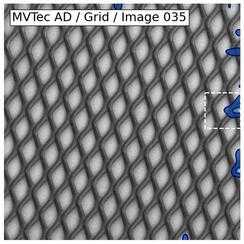}
    \includegraphics[width=0.16\linewidth,valign=t,keepaspectratio]{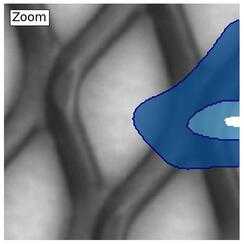}
    \includegraphics[width=0.16\linewidth,valign=t,keepaspectratio]{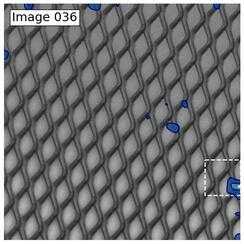}
    \includegraphics[width=0.16\linewidth,valign=t,keepaspectratio]{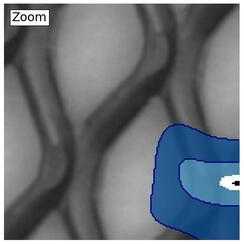}

\end{subfigure}
\\ 
\begin{subfigure}{\linewidth}
    \centering
    \includegraphics[width=0.16\linewidth,valign=t,keepaspectratio]{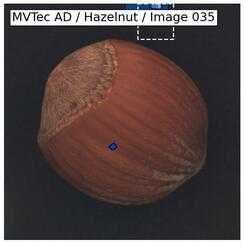}
    \includegraphics[width=0.16\linewidth,valign=t,keepaspectratio]{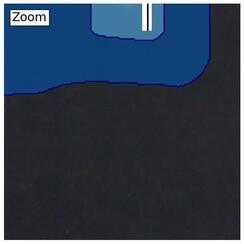}
    \includegraphics[width=0.16\linewidth,valign=t,keepaspectratio]{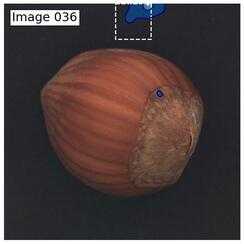}
    \includegraphics[width=0.16\linewidth,valign=t,keepaspectratio]{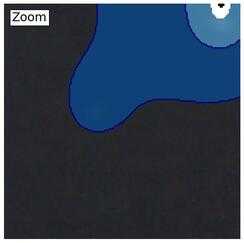}

\end{subfigure}
\\ 
\begin{subfigure}{\linewidth}
    \centering
    \includegraphics[width=0.16\linewidth,valign=t,keepaspectratio]{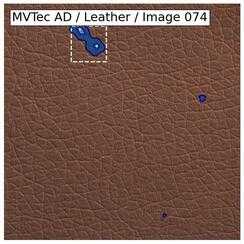}
    \includegraphics[width=0.16\linewidth,valign=t,keepaspectratio]{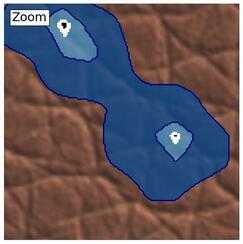}
    \includegraphics[width=0.16\linewidth,valign=t,keepaspectratio]{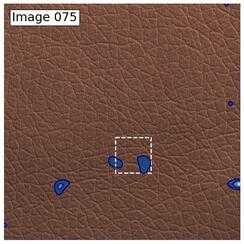}
    \includegraphics[width=0.16\linewidth,valign=t,keepaspectratio]{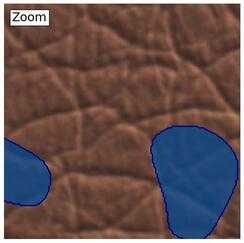}

\end{subfigure}
\\ 
\begin{subfigure}{\linewidth}
    \centering
    \includegraphics[width=0.16\linewidth,valign=t,keepaspectratio]{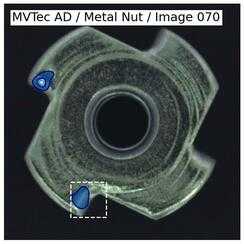}
    \includegraphics[width=0.16\linewidth,valign=t,keepaspectratio]{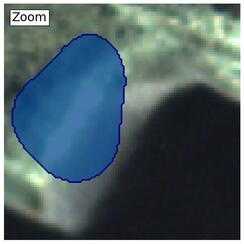}
    \includegraphics[width=0.16\linewidth,valign=t,keepaspectratio]{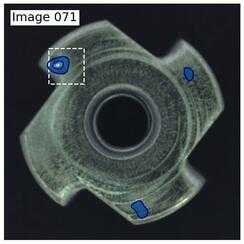}
    \includegraphics[width=0.16\linewidth,valign=t,keepaspectratio]{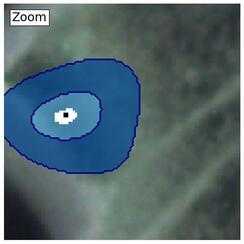}

\end{subfigure}
\\ 
\begin{subfigure}{\linewidth}
    \centering
    \includegraphics[width=0.16\linewidth,valign=t,keepaspectratio]{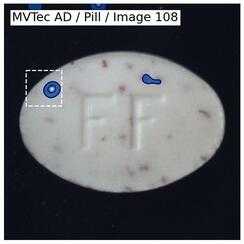}
    \includegraphics[width=0.16\linewidth,valign=t,keepaspectratio]{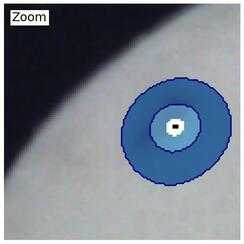}
    \includegraphics[width=0.16\linewidth,valign=t,keepaspectratio]{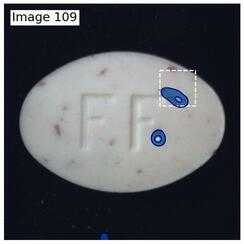}
    \includegraphics[width=0.16\linewidth,valign=t,keepaspectratio]{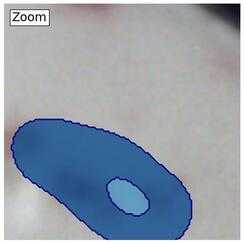}

\end{subfigure}

\phantomcaption
\end{figure}

\begin{figure}[ht]
\centering
\ContinuedFloat

\begin{subfigure}{\linewidth}
    \centering
    \includegraphics[width=0.16\linewidth,valign=t,keepaspectratio]{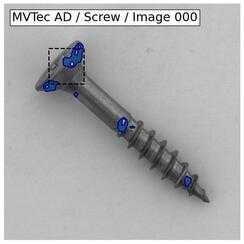}
    \includegraphics[width=0.16\linewidth,valign=t,keepaspectratio]{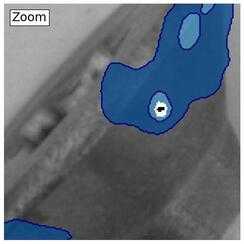}
    \includegraphics[width=0.16\linewidth,valign=t,keepaspectratio]{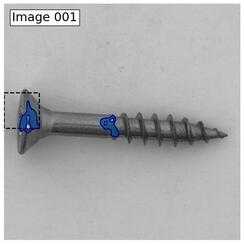}
    \includegraphics[width=0.16\linewidth,valign=t,keepaspectratio]{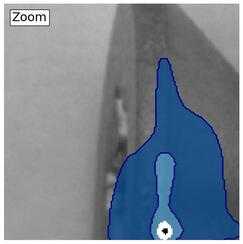}

\end{subfigure}
\\ 
\begin{subfigure}{\linewidth}
    \centering
    \includegraphics[width=0.16\linewidth,valign=t,keepaspectratio]{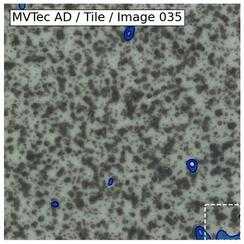}
    \includegraphics[width=0.16\linewidth,valign=t,keepaspectratio]{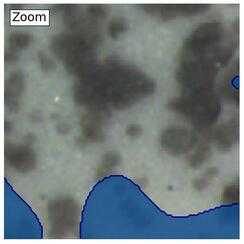}
    \includegraphics[width=0.16\linewidth,valign=t,keepaspectratio]{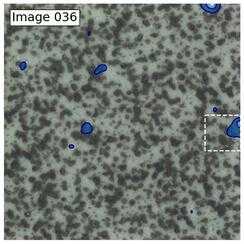}
    \includegraphics[width=0.16\linewidth,valign=t,keepaspectratio]{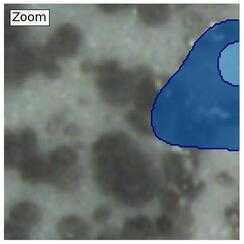}

\end{subfigure}
\\ 
\begin{subfigure}{\linewidth}
    \centering
    \includegraphics[width=0.16\linewidth,valign=t,keepaspectratio]{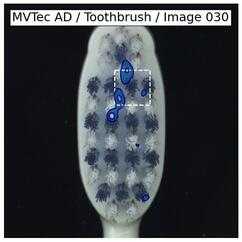}
    \includegraphics[width=0.16\linewidth,valign=t,keepaspectratio]{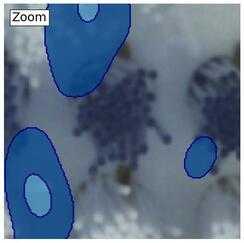}
    \includegraphics[width=0.16\linewidth,valign=t,keepaspectratio]{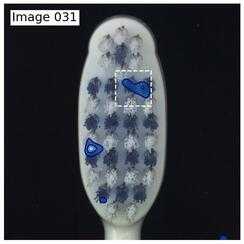}
    \includegraphics[width=0.16\linewidth,valign=t,keepaspectratio]{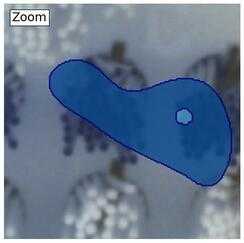}

\end{subfigure}
\\ 
\begin{subfigure}{\linewidth}
    \centering
    \includegraphics[width=0.16\linewidth,valign=t,keepaspectratio]{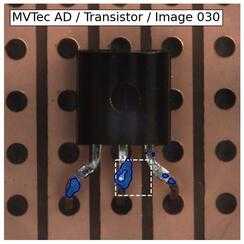}
    \includegraphics[width=0.16\linewidth,valign=t,keepaspectratio]{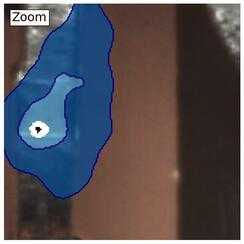}
    \includegraphics[width=0.16\linewidth,valign=t,keepaspectratio]{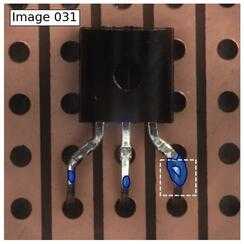}
    \includegraphics[width=0.16\linewidth,valign=t,keepaspectratio]{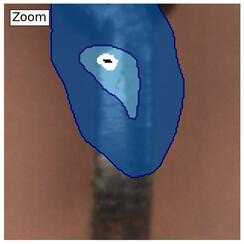}

\end{subfigure}
\\ 
\begin{subfigure}{\linewidth}
    \centering
    \includegraphics[width=0.16\linewidth,valign=t,keepaspectratio]{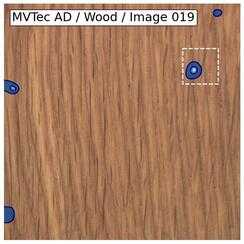}
    \includegraphics[width=0.16\linewidth,valign=t,keepaspectratio]{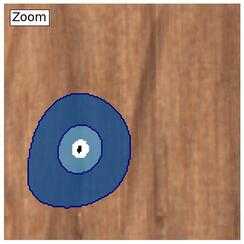}
    \includegraphics[width=0.16\linewidth,valign=t,keepaspectratio]{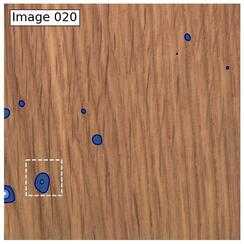}
    \includegraphics[width=0.16\linewidth,valign=t,keepaspectratio]{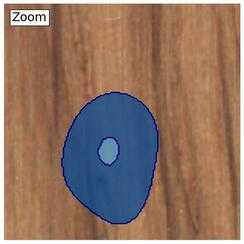}

\end{subfigure}
\\ 
\begin{subfigure}{\linewidth}
    \centering
    \includegraphics[width=0.16\linewidth,valign=t,keepaspectratio]{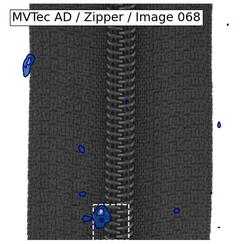}
    \includegraphics[width=0.16\linewidth,valign=t,keepaspectratio]{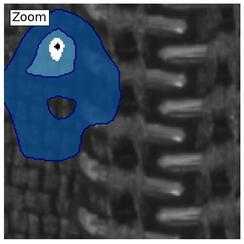}
    \includegraphics[width=0.16\linewidth,valign=t,keepaspectratio]{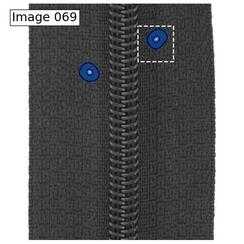}
    \includegraphics[width=0.16\linewidth,valign=t,keepaspectratio]{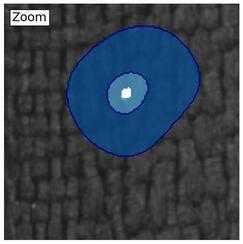}

\end{subfigure}
\\ 
\begin{subfigure}{\linewidth}
    \centering
    \includegraphics[width=0.16\linewidth,valign=t,keepaspectratio]{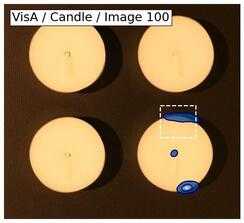}
    \includegraphics[width=0.16\linewidth,valign=t,keepaspectratio]{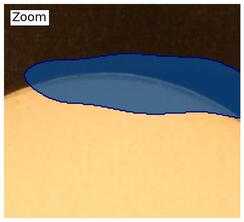}
    \includegraphics[width=0.16\linewidth,valign=t,keepaspectratio]{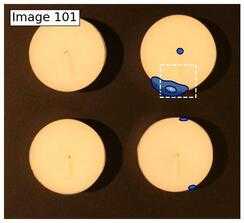}
    \includegraphics[width=0.16\linewidth,valign=t,keepaspectratio]{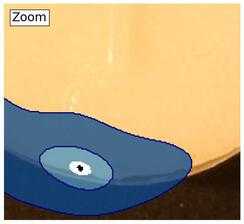}

\end{subfigure}
\\ 
\begin{subfigure}{\linewidth}
    \centering
    \includegraphics[width=0.16\linewidth,valign=t,keepaspectratio]{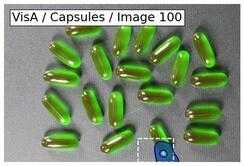}
    \includegraphics[width=0.16\linewidth,valign=t,keepaspectratio]{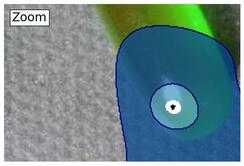}
    \includegraphics[width=0.16\linewidth,valign=t,keepaspectratio]{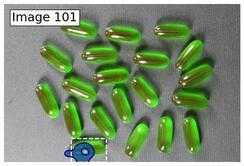}
    \includegraphics[width=0.16\linewidth,valign=t,keepaspectratio]{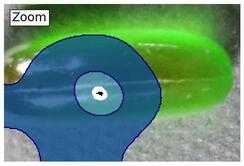}

\end{subfigure}
\\ 
\begin{subfigure}{\linewidth}
    \centering
    \includegraphics[width=0.16\linewidth,valign=t,keepaspectratio]{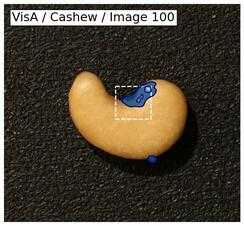}
    \includegraphics[width=0.16\linewidth,valign=t,keepaspectratio]{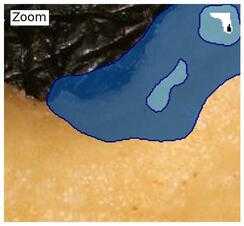}
    \includegraphics[width=0.16\linewidth,valign=t,keepaspectratio]{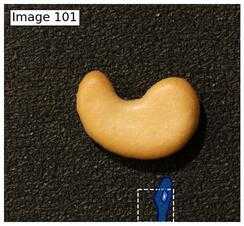}
    \includegraphics[width=0.16\linewidth,valign=t,keepaspectratio]{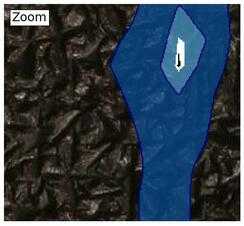}

\end{subfigure}

\phantomcaption
\end{figure}

\begin{figure}[ht]
\centering
\ContinuedFloat

\begin{subfigure}{\linewidth}
    \centering
    \includegraphics[width=0.16\linewidth,valign=t,keepaspectratio]{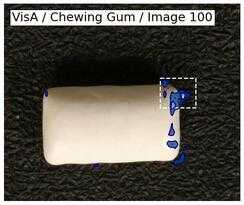}
    \includegraphics[width=0.16\linewidth,valign=t,keepaspectratio]{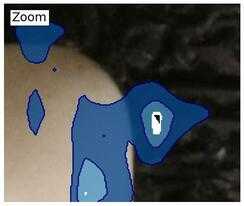}
    \includegraphics[width=0.16\linewidth,valign=t,keepaspectratio]{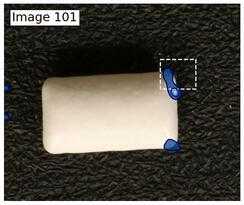}
    \includegraphics[width=0.16\linewidth,valign=t,keepaspectratio]{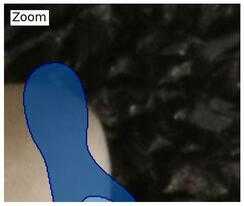}

\end{subfigure}
\\ 
\begin{subfigure}{\linewidth}
    \centering
    \includegraphics[width=0.16\linewidth,valign=t,keepaspectratio]{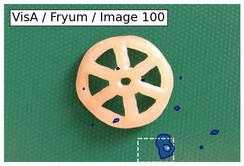}
    \includegraphics[width=0.16\linewidth,valign=t,keepaspectratio]{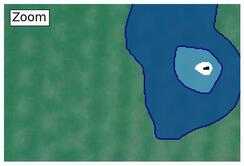}
    \includegraphics[width=0.16\linewidth,valign=t,keepaspectratio]{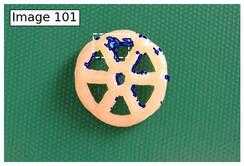}
    \includegraphics[width=0.16\linewidth,valign=t,keepaspectratio]{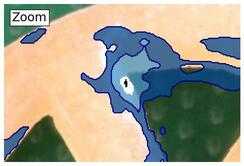}

\end{subfigure}
\\ 
\begin{subfigure}{\linewidth}
    \centering
    \includegraphics[width=0.16\linewidth,valign=t,keepaspectratio]{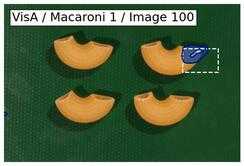}
    \includegraphics[width=0.16\linewidth,valign=t,keepaspectratio]{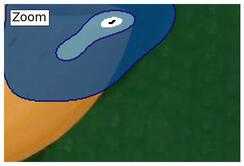}
    \includegraphics[width=0.16\linewidth,valign=t,keepaspectratio]{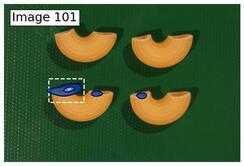}
    \includegraphics[width=0.16\linewidth,valign=t,keepaspectratio]{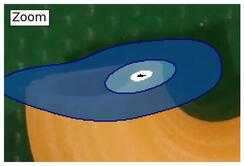}

\end{subfigure}
\\ 
\begin{subfigure}{\linewidth}
    \centering
    \includegraphics[width=0.16\linewidth,valign=t,keepaspectratio]{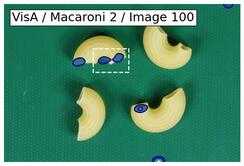}
    \includegraphics[width=0.16\linewidth,valign=t,keepaspectratio]{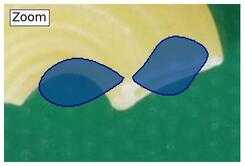}
    \includegraphics[width=0.16\linewidth,valign=t,keepaspectratio]{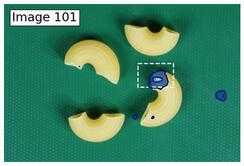}
    \includegraphics[width=0.16\linewidth,valign=t,keepaspectratio]{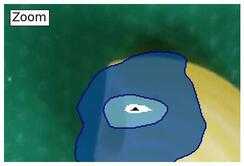}

\end{subfigure}
\\ 
\begin{subfigure}{\linewidth}
    \centering
    \includegraphics[width=0.16\linewidth,valign=t,keepaspectratio]{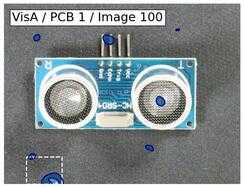}
    \includegraphics[width=0.16\linewidth,valign=t,keepaspectratio]{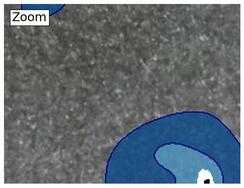}
    \includegraphics[width=0.16\linewidth,valign=t,keepaspectratio]{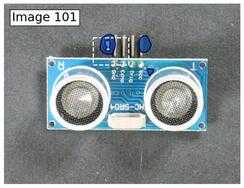}
    \includegraphics[width=0.16\linewidth,valign=t,keepaspectratio]{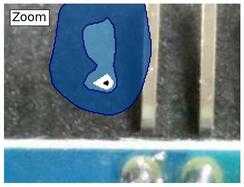}

\end{subfigure}
\\ 
\begin{subfigure}{\linewidth}
    \centering
    \includegraphics[width=0.16\linewidth,valign=t,keepaspectratio]{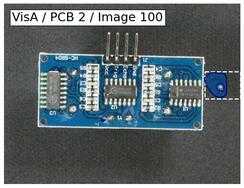}
    \includegraphics[width=0.16\linewidth,valign=t,keepaspectratio]{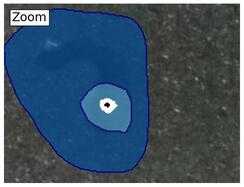}
    \includegraphics[width=0.16\linewidth,valign=t,keepaspectratio]{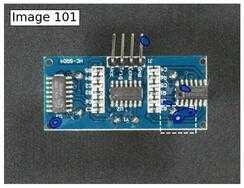}
    \includegraphics[width=0.16\linewidth,valign=t,keepaspectratio]{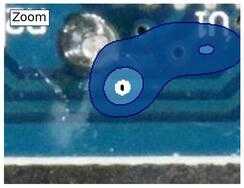}

\end{subfigure}
\\ 
\begin{subfigure}{\linewidth}
    \centering
    \includegraphics[width=0.16\linewidth,valign=t,keepaspectratio]{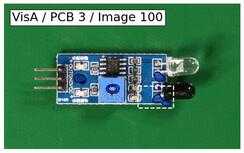}
    \includegraphics[width=0.16\linewidth,valign=t,keepaspectratio]{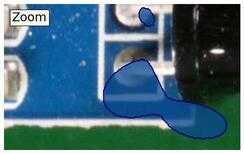}
    \includegraphics[width=0.16\linewidth,valign=t,keepaspectratio]{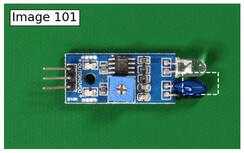}
    \includegraphics[width=0.16\linewidth,valign=t,keepaspectratio]{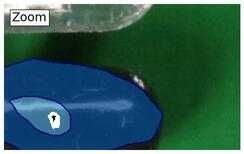}

\end{subfigure}
\\ 
\begin{subfigure}{\linewidth}
    \centering
    \includegraphics[width=0.16\linewidth,valign=t,keepaspectratio]{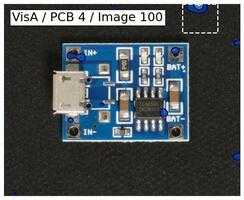}
    \includegraphics[width=0.16\linewidth,valign=t,keepaspectratio]{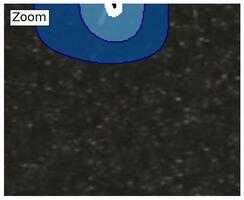}
    \includegraphics[width=0.16\linewidth,valign=t,keepaspectratio]{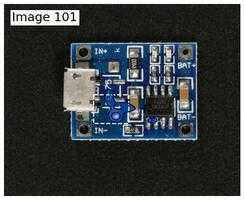}
    \includegraphics[width=0.16\linewidth,valign=t,keepaspectratio]{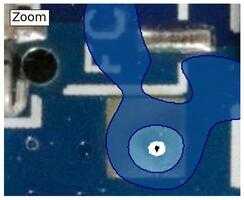}

\end{subfigure}
\\ 
\begin{subfigure}{\linewidth}
    \centering
    \includegraphics[width=0.16\linewidth,valign=t,keepaspectratio]{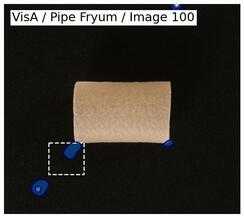}
    \includegraphics[width=0.16\linewidth,valign=t,keepaspectratio]{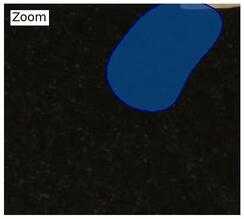}
    \includegraphics[width=0.16\linewidth,valign=t,keepaspectratio]{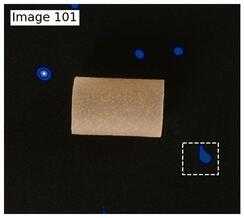}
    \includegraphics[width=0.16\linewidth,valign=t,keepaspectratio]{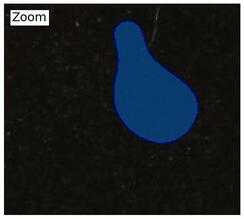}

\end{subfigure}

\caption{
    Visual intuition of Image False Positive Rate (ImFPR) levels on normal images.
    Images are normal samples from the datasets in MVTecAD and VisA.
    Each color corresponds to a predicted mask at a given ImFPR level: darker blue is $10^{-2}$, lighter blue is $10^{-3}$, white is $10^{-4}$, and black is $10^{-5}$. 
}
\label{fig:fp-levels}
    
\end{figure}


\clearpage
\section{Anomaly size}\label{app:tiny-blobs}
\begin{figure}[ht]
    \centering
    \includegraphics[width=\linewidth]{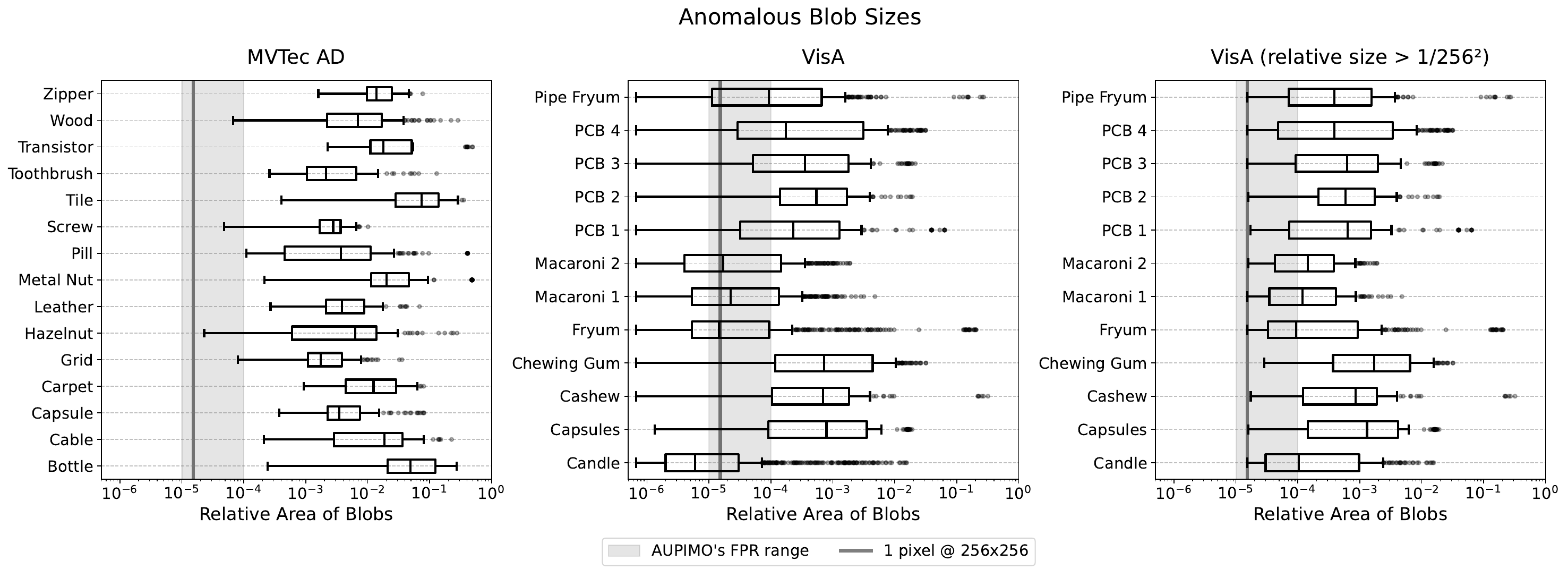}
    \caption{
        Distribuition of relative size of anomalous regions.
    }
    \label{fig:anomalous-blobs-sizes}
\end{figure}

\cref{fig:anomalous-blobs-sizes} shows the distributions of the relative region size in ground truth annotations in each dataset from \gls{mvtecad} and \gls{visa}.
Reminder: relative size is the number of pixels in a maximally connected component divided by the number of pixels in the image.
Lower and upper whiskers are set with maximum size to $1.5$ inter-quartile range (IQR), and fliers (outliers) are shown as gray dots.
The gray-shaded span is \gls{aupimo}'s integration range, and the vertical gray line represents the relative size of a single pixel at resolution $256 \times 256$ (input size seen by the models in our experiments).

\paragraph{MVTec AD}
\cref{fig:anomalous-blobs-sizes} shows that the size of the anomalies in \gls{mvtecad} are generally between $10^{-3}$ and $10^{-1}$. 
Few cases are as small $10^{-4}$.
Given this distribution, the \gls{aupimo} scores from our experiments can be interpreted as a (near) \gls{fp}-free recall.
Since (almost) none of the anomalies are as small as the \gls{fpr} integration range, any prediction with relative size above the integration range is a \gls{tp}.
Conversely, one could dismiss any prediction with relative size below the integration range.

\paragraph{VisA}
The anomalies in \gls{visa} are largely biased towards small regions of relative sizes as small as $\sim10^{-6}$ (\ie a single pixel at resolution $1000 \! \times \! 1000$).
They are so numerous that the actual anomalous regions show as outliers in \cref{fig:anomalous-blobs-sizes}.

\begin{figure*}[ht]
  \centering
  \begin{subfigure}{\linewidth}
      \centering
      \includegraphics[width=0.49\linewidth,valign=t,keepaspectratio]{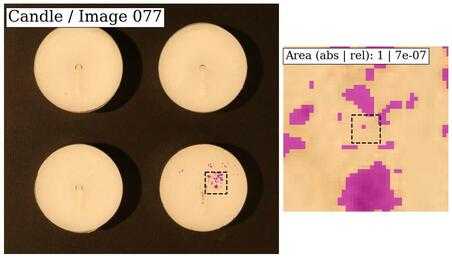}
      \hfill
      \includegraphics[width=0.49\linewidth,valign=t,keepaspectratio]{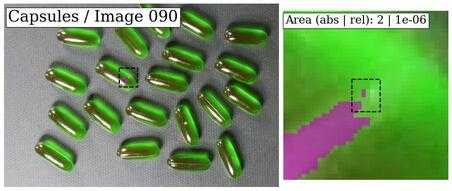}
      \includegraphics[width=0.49\linewidth,valign=t,keepaspectratio]{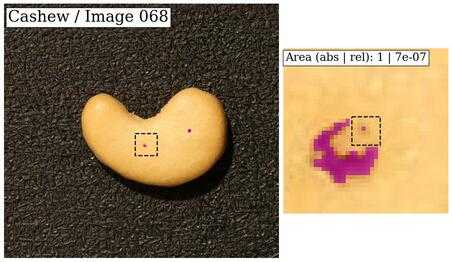}
      \hfill
      \includegraphics[width=0.49\linewidth,valign=t,keepaspectratio]{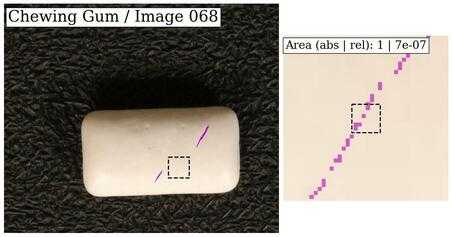}
      \includegraphics[width=0.49\linewidth,valign=t,keepaspectratio]{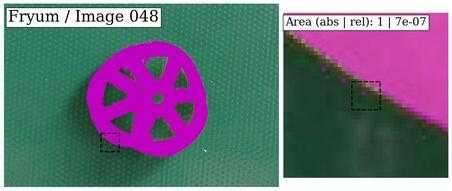}
      \hfill
      \includegraphics[width=0.49\linewidth,valign=t,keepaspectratio]{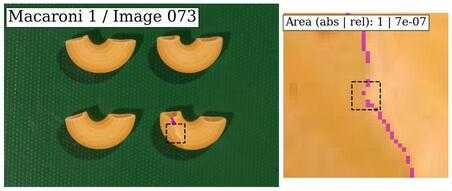}
      \includegraphics[width=0.49\linewidth,valign=t,keepaspectratio]{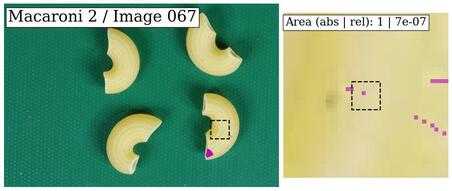}
      \hfill
      \includegraphics[width=0.49\linewidth,valign=t,keepaspectratio]{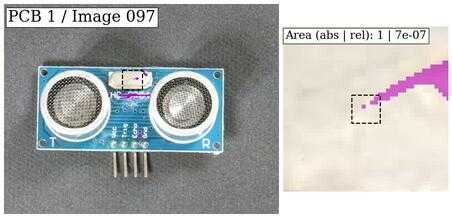}
      \includegraphics[width=0.49\linewidth,valign=t,keepaspectratio]{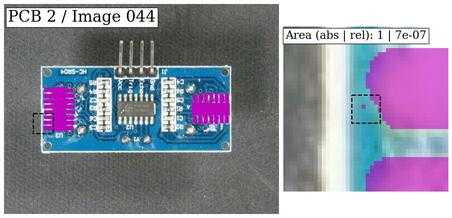}
      \hfill
      \includegraphics[width=0.49\linewidth,valign=t,keepaspectratio]{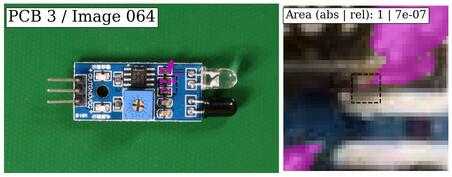}
      \includegraphics[width=0.49\linewidth,valign=t,keepaspectratio]{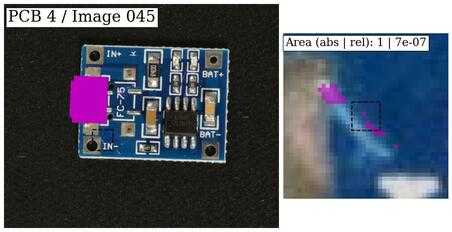}
      \hfill
      \includegraphics[width=0.49\linewidth,valign=t,keepaspectratio]{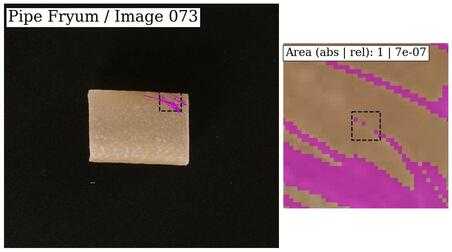}
  \end{subfigure}
  \caption{Tiny anomalous regions in VisA.}
  \label{fig:tiny-blobs-viz}
\end{figure*}

\paragraph{Tiny regions}
Let \say{tiny} refer to connected components of relative size smaller than $\frac{1}{256^2}$, which corresponds to a single pixel at resolution $256 \! \times \! 256$.
In other words, an actual anomaly this small would be seen as a single pixel by the models in our experiments or simply not seen at all.
\cref{fig:tiny-blobs-viz} displays several examples of tiny regions in \gls{visa} with zoomed-in views on the right.
These regions are meaningless: 
as \cref{fig:tiny-blobs-viz} shows, they are often 1-pixel (or \say{very few}-pixel) regions.
They are often near the surroundings of an actual anomaly (e.g. Fryum/Image 048).
Extreme cases where completely isolated regions with insignificant size also occur (e.g. Chewing Gum/Image 068 and Macaroni 2/Image 067). 

\paragraph{How often and how small are these tiny regions?}
\cref{tab:tiny-blobs} shows statistics about the absolute sizes (at original resolution) and the number of tiny regions per image in each dataset from \gls{visa}.
The right-most plot in \cref{fig:anomalous-blobs-sizes} shows \gls{visa}'s anomalous region size distribution when discarding the tiny regions.

\begin{table}[th]
    \singlespacing
    \centering
    \caption{Statistics from tiny blobs in \gls{visa} \cite{zou_spot--difference_2022}.}
    \label{tab:tiny-blobs}
    \subfloat[Sizes.]{
        \begin{tabular}{lrrr}
            \toprule
            Reg Size (abs) & 1 - 9 & 10 - 19 & 20 - 29 \\
            Category &  &  &  \\
            \midrule
            Candle & 358 & 98 & 20 \\
            Capsules & 8 & 7 & 3 \\
            Cashew & 10 & 0 & 1 \\
            Chewing Gum & 39 & 1 & 0 \\
            Fryum & 158 & 96 & 22 \\
            Macaroni 1 & 114 & 52 & 14 \\
            Macaroni 2 & 123 & 54 & 6 \\
            PCB 1 & 19 & 20 & 9 \\
            PCB 2 & 11 & 8 & 4 \\
            PCB 3 & 20 & 11 & 0 \\
            PCB 4 & 32 & 19 & 12 \\
            Pipe Fryum & 44 & 34 & 9 \\
            \bottomrule
        \end{tabular}
    }
    \hspace{5mm}
    \subfloat[Number of regions per image.]{
        \begin{tabular}{lrrr}
            \toprule
            Nb Reg/Img & 1 - 5 & 6+ & Total \\
            Category &  &  &  \\
            \midrule
            Candle & 5 & 18 & 23 \\
            Capsules & 6 & 1 & 7 \\
            Cashew & 5 & 0 & 5 \\
            Chewing Gum & 6 & 1 & 7 \\
            Fryum & 22 & 13 & 35 \\
            Macaroni 1 & 27 & 10 & 37 \\
            Macaroni 2 & 21 & 8 & 29 \\
            PCB 1 & 10 & 2 & 12 \\
            PCB 2 & 10 & 1 & 11 \\
            PCB 3 & 8 & 1 & 9 \\
            PCB 4 & 10 & 5 & 15 \\
            Pipe Fryum & 17 & 3 & 20 \\
            \bottomrule
        \end{tabular}
    }
\end{table}


\clearpage
\section{Additional results}\label{app:add-res}

\newcommand{\nulhyp}{\mathrm{H}_0}
\newcommand{\althyp}{\mathrm{H}_1}
\newcommand{\confidence}{C}
\newcommand{\modA}{A}
\newcommand{\modB}{B}

\subsection{Ablation study}\label{app:ablation}
\cref{tab:ablation_studies} showcases the use of statistical tests in an ablation study of \gls{efficientad} \cite{batzner_efficientad_2023} on the dataset \gls{mvtecad} / Capsule.
The Wilcoxon signed-rank test \cite{benavoli_should_2016,demsar_statistical_2006} is used to assess the consistency of performance gain given by different components of the model.
The null hypothesis $\nulhyp$ is that two models $\modA$ and $\modB$ are equivalent (average ranks tend to equal), and the alternative hypothesis $\althyp$ is that one of the two models (say, $\modA$) is \emph{more often} better than $\modB$.
No assumption is made about the scores distributions making it robust to outliers \cite{benavoli_should_2016,demsar_statistical_2006}.
Interpretation: high confidence ($\confidence \! = \! 1 - \text{p-value}$) to reject the null hypothesis (\ie low p-value) means that $\modA$ \emph{consistently} outperforms $\modB$.

\begin{table}[ht]
    \centering
    \singlespacing
    \small
    \caption{
        Ablation study (use-case of statistical tests).
        Layout and model configurations based on Tab. 4 in \cite{batzner_efficientad_2023}.
        At each row a component is added to the model above starting with Patch Description Network (PDN) at top and resulting in EfficientAD at the bottom.
        $\confidence$ refers to the confidence to reject the null hypothesis ($1 - $ p-value); higher means more confidence on the improvement by adding the new component.
        Each component generally shows significant improvements, but the bottom right cell is an exception.
        Pretraining penalty causes a score drop, and the low confidence on the alternative hypothesis confirms that the drop is consistent across images.
    }
    \label{tab:ablation_studies}
    \begin{tabular}{p{0.44\linewidth}p{0.20\linewidth}p{0.20\linewidth}}
        Avg. AUPIMO~(Diff. [\%]; $\confidence$ [\%]) & EfficientAD-S & EfficientAD-M                                        \\ 
    \toprule
         \begin{tabular}[c]{@{}c@{}} PDN \end{tabular} & $\sim$ 0&  $\sim$ 0\\
    $\hookrightarrow$ map normalization    & 22 (+22; 100)          & 23 (+23; 100)                                                 \\
    $\hookrightarrow$ hard feature loss    & 57 (+35; 100)          & 59 (+36; 100)                                                 \\
    $\hookrightarrow$ pretraining penalty  & 64 (+7; 100)           & 57 (-2; 0.02)                                                
    \end{tabular}
\end{table}


\subsection{Does AUPIMO correlate with AUROC and AUPRO?}\label{app:others-vs-aupimo}
\begin{figure}[ht]
  \centering

  \hfill
  \begin{subfigure}{0.49\linewidth}
    \includegraphics[width=\linewidth]{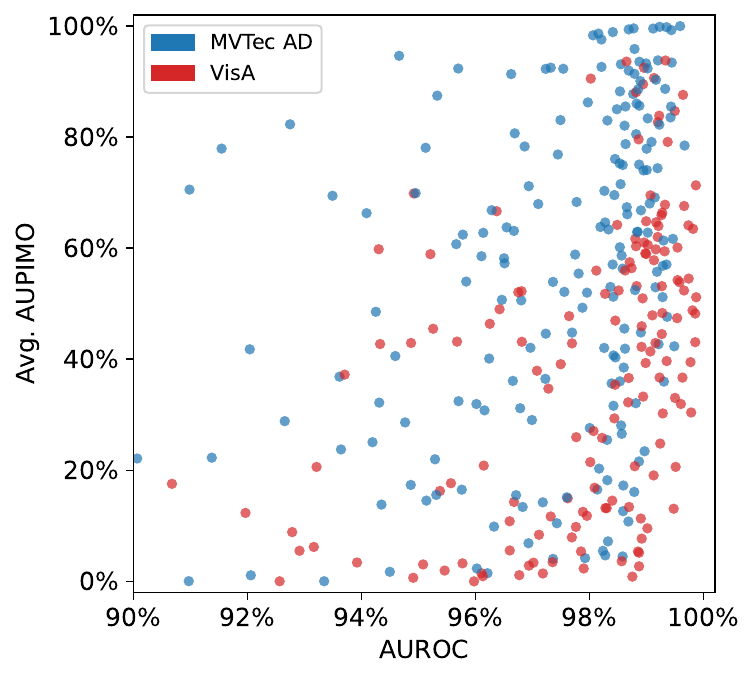}
    \caption{AUPIMO vs. AUROC}
    \label{fig:scatter-auroc}
  \end{subfigure}
  \hfill
  \begin{subfigure}{0.49\linewidth}
    \includegraphics[width=\linewidth]{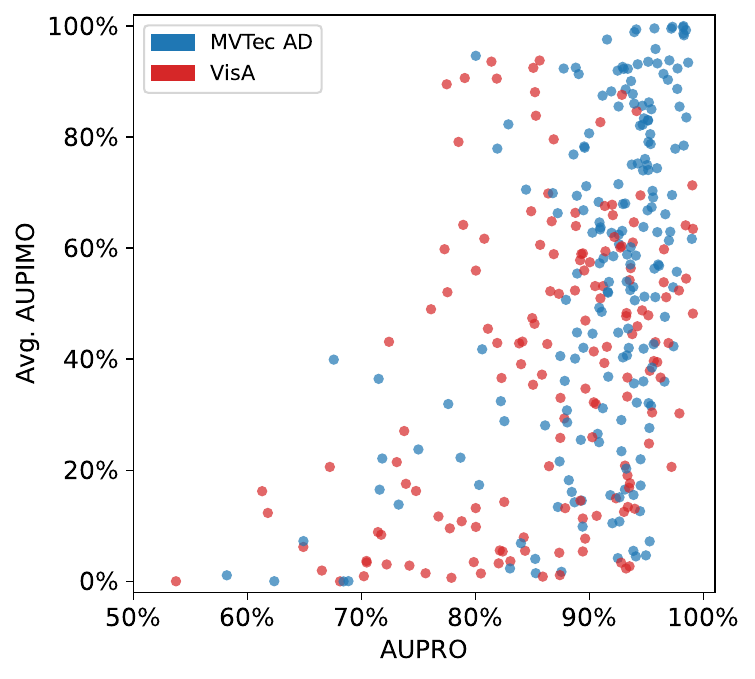}
    \caption{AUPIMO vs. AUPRO}
    \label{fig:scatter-aupro}
  \end{subfigure}
  \hfill

  \caption{Scatter plots of AUPIMO vs. \{AUROC, AUPRO\}}
  \label{fig:scatter}
\end{figure}

\cref{fig:scatter} shows scatter plots of \gls{auroc} and \gls{aupro} vs. (cross-image) average \gls{aupimo}.
All models and datasets in the benchmark confounded.
Notice that the scales of the axes are different for each metric. 
Both plots seem to show a positive correlation, but one metric is not enough to imply the other.
High levels of \gls{auroc} and \gls{aupro} do not guarantee high levels of \gls{aupimo}.
Conversely, high levels of \gls{aupimo} \emph{tend} to imply higher levels of \gls{aupro} and \gls{auroc} (notice the slightly triangular shape of the point clouds).

\subsection{Video}\label{app:video}
In this section we present how \gls{aupimo} can be used in video applications. 
It must be stressed that this is not a full-fledged video \gls{aupimo} application, but rather a proof of concept.
The UCSD Pedestrian dataset \cite{mahadevan_anomaly_2010} was used to illustrate this concept because it has been widely used and cited in the literature, but other datasets like A Day on Campus (ADOC) \cite{pranav_day_2020} and Street Scene \cite{ramachandra_street_2020} would also be relevant to this discussion.

\begin{figure}[hb]
    \centering
    \includegraphics[width=.9\linewidth]{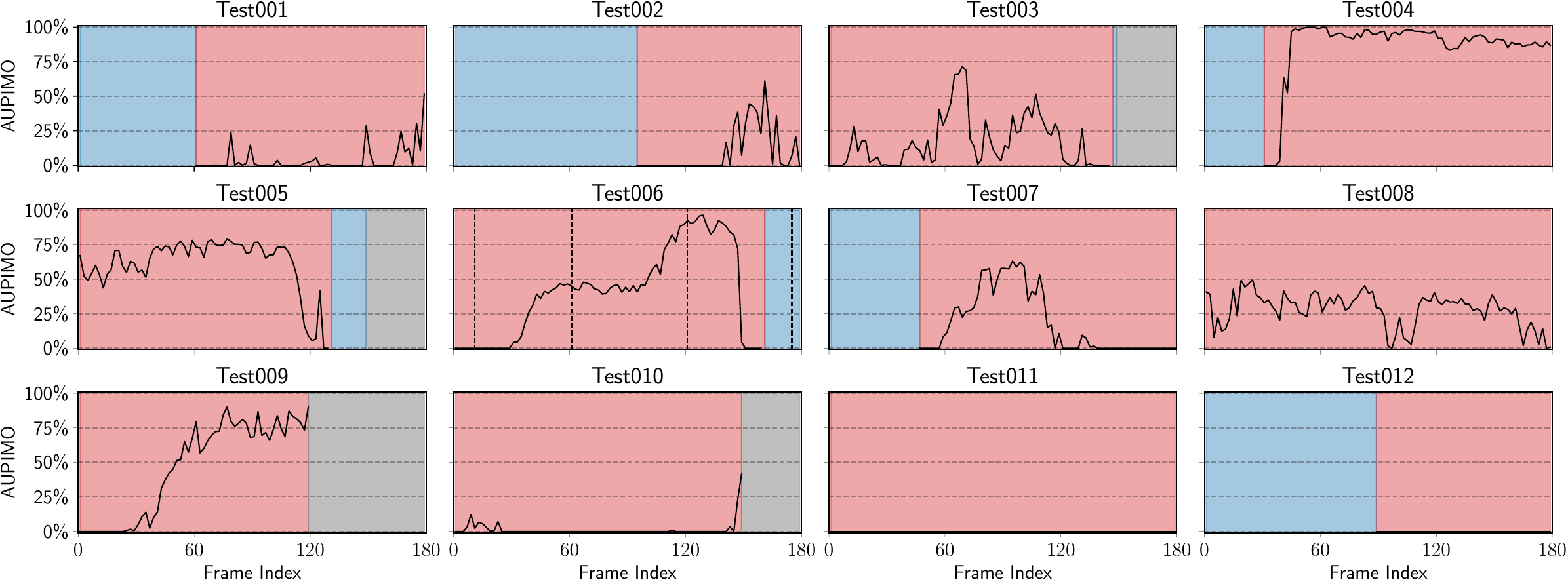}
    \caption{
        Time vs. \gls{aupimo} in test videos from the UCSD Pedestrian dataset.
        The x-axis is the frame index in each video and the y-axis is the \gls{aupimo} score at that frame.
        Blue zones indicate the frame is normal, red zones indicate the frame has an anomaly, and gray zones indicate there is no frame.
        Vertical dashed lines in "Test006" correspond to the frames shown in \cref{fig:video-frames}.
    }
    \label{fig:video}
\end{figure}

\begin{figure}[ht]
    \centering
    \includegraphics[width=.9\linewidth]{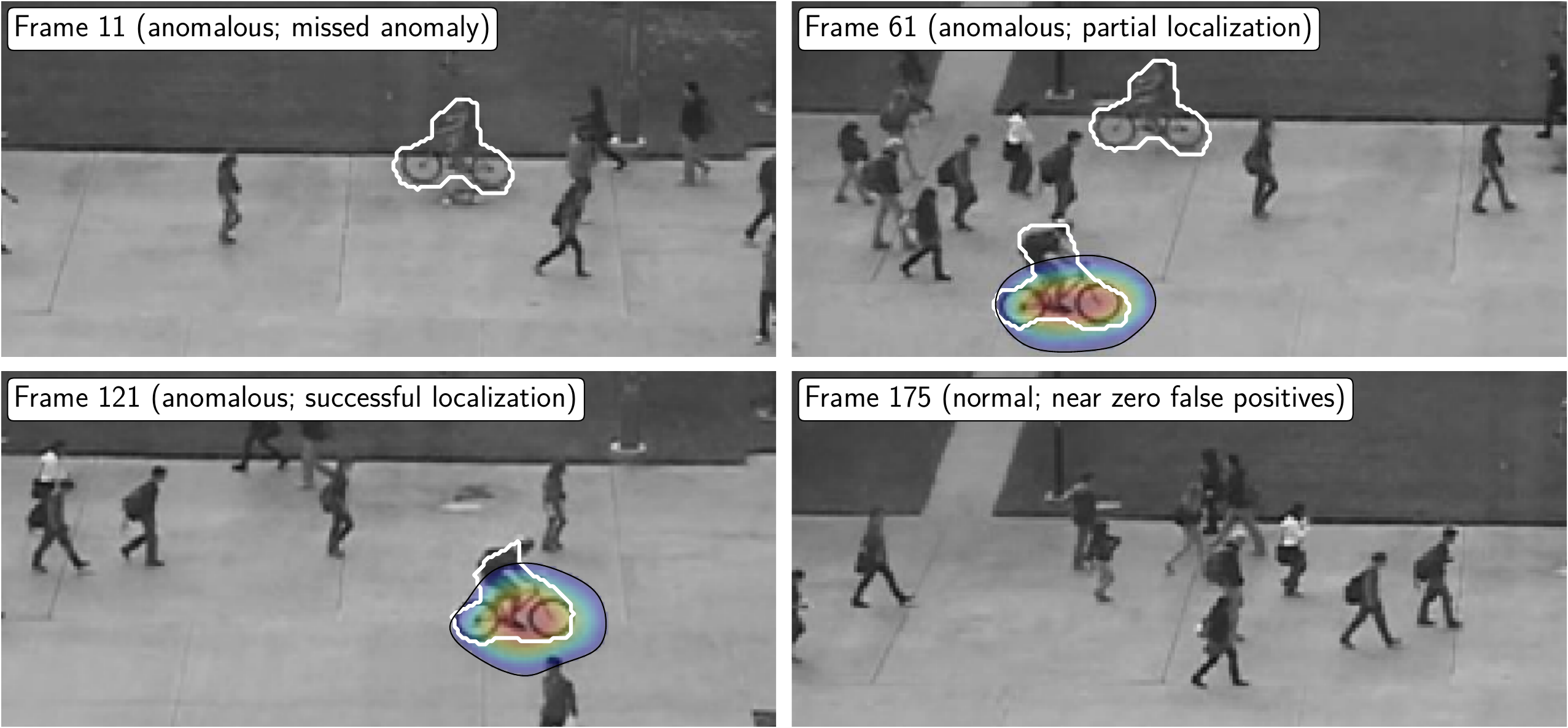}
    \caption{
        Frames from the video Test006.
        White contours indicate the ground truth anomalous regions.
        Black contours correspond to the level sets in each anomaly score map $\amap$ at $\thr_L$, where $\shfpr(\thr_L) = \fprlbound$.
        Anomaly scores above below $\thr_L$ are not shown and above are colored using the JET colormap with local maxima in red and $\thr_L$ in blue.
    }
    \label{fig:video-frames}    
\end{figure}

A \gls{patchcore} \cite{roth_towards_2022} model was trained on the normal videos from UCSD Pedestrian dataset at every 2 frames.
The model was evaluated with the same procedure than our experiments by ignoring the temporal dimension of the videos and treating all the frames from all the videos as a single dataset.
In \cref{fig:video} we show the \gls{aupimo} scores for each frame in the test videos along the time axis (referenced by the frame index).
A selection of frames from the video Test006 are shown in \cref{fig:video-frames}. 

Notice how \gls{aupimo}'s validation works in practice: the normal frame (175) does not have any visible \gls{fp} region -- \ie anomaly score values above the threshold $\thr_L$, corresponding to the lowest \gls{fpr} level $\fprlbound$ used in \gls{aupimo}.
Frame 61 shows an example case where the image-scoped has limitations: the \gls{aupimo} score is around $50\%$ because there are two indendent anomalous regions in the frame; one of them is well detected by the model, but the other is ignored..
A better modelization for this case would require a more complete annotation where each instance of anomaly is labeled separately, which is not the case in the UCSD Pedestrian dataset.


\subsection{Precision vs. Intersection over Union}\label{app:prec-vs-iou}
Since \gls{aupimo} only concerns recall, our analysis lacks a discussion about the segmentation quality. 
In this section we aim to mitigate this shortcoming by extending our validation-evaluation framework.
Two candidate metrics are considered: the image-scoped precision and the image-scoped \gls{iou}.
As detailed in the next paragraphs, the precision is not suitable for our purposes, so the \gls{iou} is chosen to build a \gls{shfpr}-based curve and an \gls{auc} score like \gls{pimo} and \gls{aupimo} respectively.
The anomaly score maps in this section are from \gls{patchcore} in the dataset \gls{mvtecad} / Metal Nut. 
We made this restricted choice to simplify the discussion, but similar results are obtained for the other datasets and models.

\begin{figure}[ht]
    \centering
    \begin{subfigure}[b]{\linewidth}
        \centering
        \includegraphics[width=.32\linewidth]{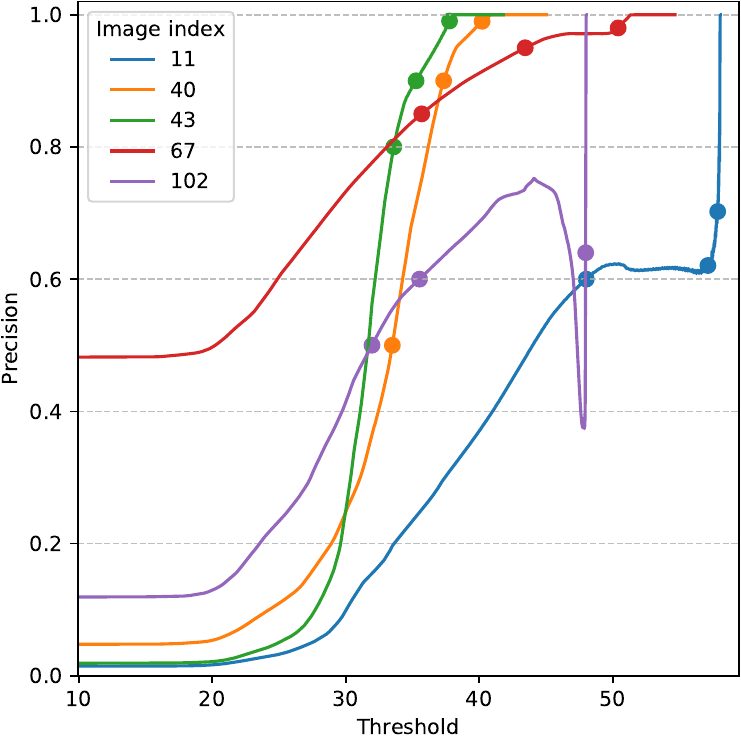}
        \includegraphics[width=.32\linewidth]{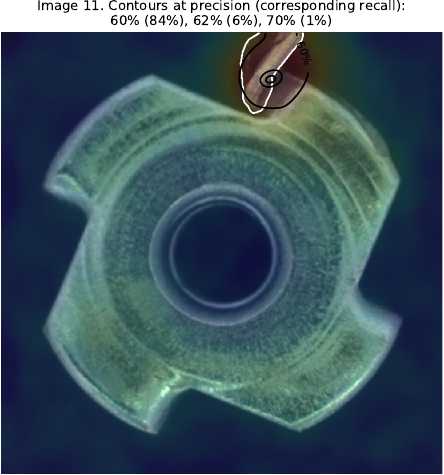}
        \includegraphics[width=.32\linewidth]{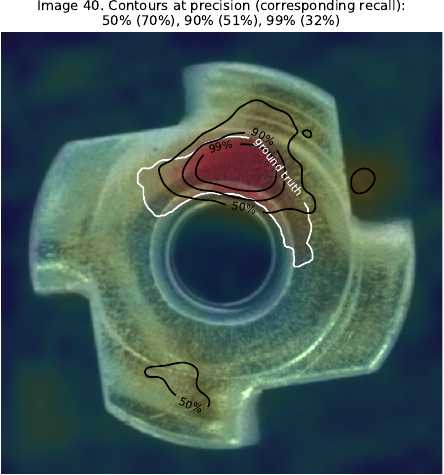}
    \end{subfigure}
        \\[.5em]
    \begin{subfigure}[b]{\linewidth}
        \centering
        \includegraphics[width=.32\linewidth]{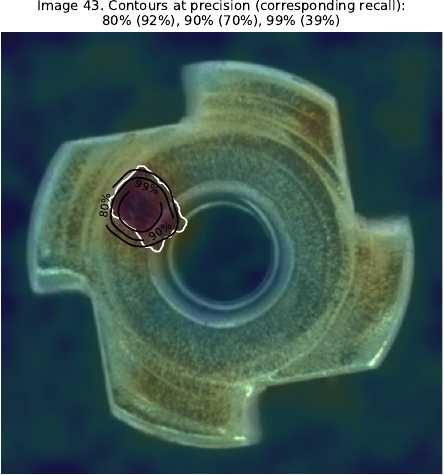}
        \includegraphics[width=.32\linewidth]{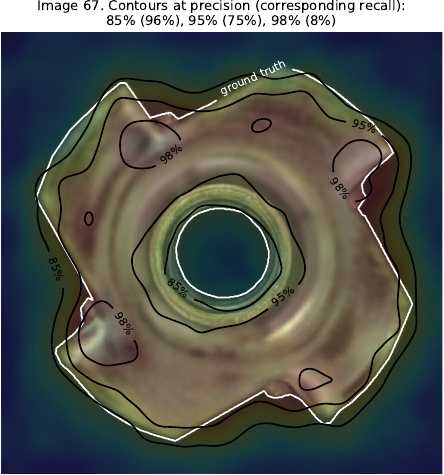}
        \includegraphics[width=.32\linewidth]{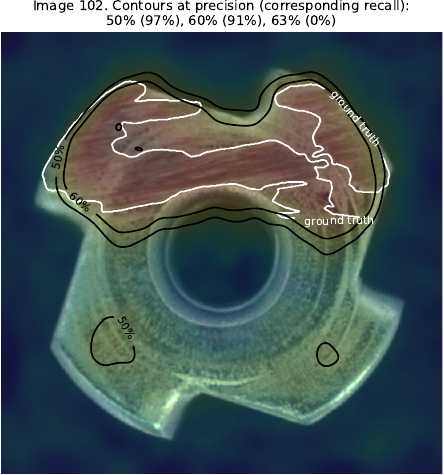}
    \end{subfigure}
    \caption{Precision curves and contours at different points along the curves.}
    \label{fig:thresh-prec-curves}
    \end{figure}

\cref{fig:thresh-prec-curves} shows the precision as a function of binarization thresholds in five images (note: not indexed by $\shfpr$ like \gls{pimo}). 
The level sets of the anomaly score maps at three thresholds along these curves are shown in black superimposed on the images, which can be compared to the contour from the ground truth annotations in white.
The precision curves are not smooth, and optimizing this metric does not correspond to improving the visual aspect of the segmentation.
It can be seen that optimizing for precision is not a viable option, as the segmentations tend to have a recall-disaster behavior as the precision increases.

The threshold-vs-precision  curves show a breakpoint phenomenon where increasing the threshold generaly increases the precision but dramatically decreases the recall at some point.
For instance, in image 11 the breakpoint is between $60\%$ and $62\%$ precision; \ie somewhere between their respective contour lines the segmentation switches from being too big to being too small (recall drops from $84\%$ to $6\%$).
In image 67, on the other hand, the breakpoint is between $95\%$ and $98\%$ precision (recall drops from $75\%$ to $8\%$ respectively).
Image 102 shows an extreme case of this, where the segmentation is reduced to a nearly invisible region as the precision increases from $60\%$ to $63\%$ (recall drops from $91\%$ to almost $0\%$).

\begin{figure}[ht]
\centering
    \begin{subfigure}[b]{\linewidth}
        \centering
        \includegraphics[width=.32\linewidth]{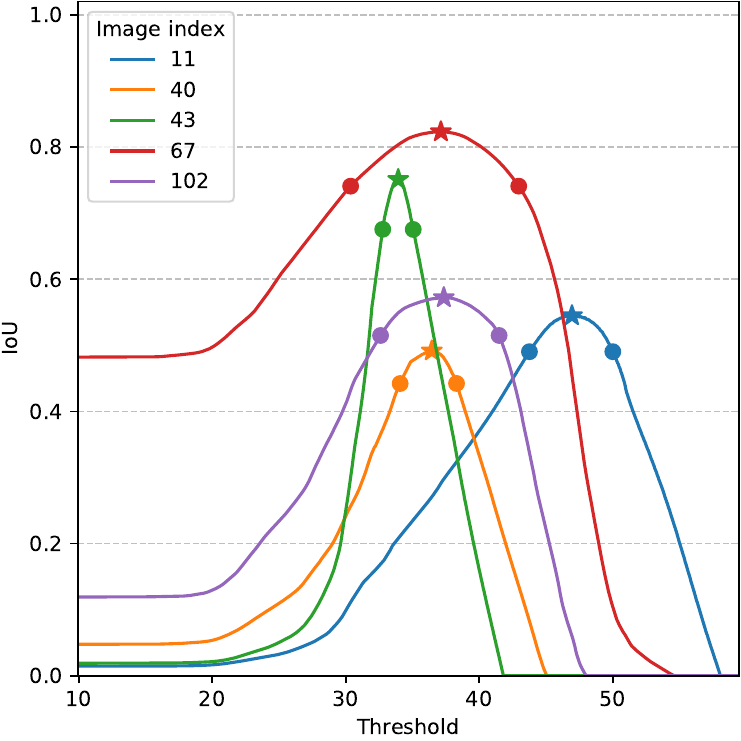}
        \includegraphics[width=.32\linewidth]{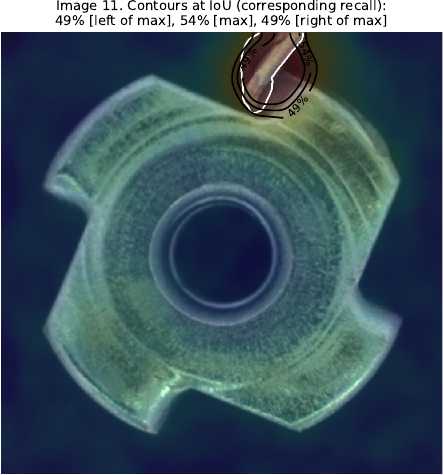}
        \includegraphics[width=.32\linewidth]{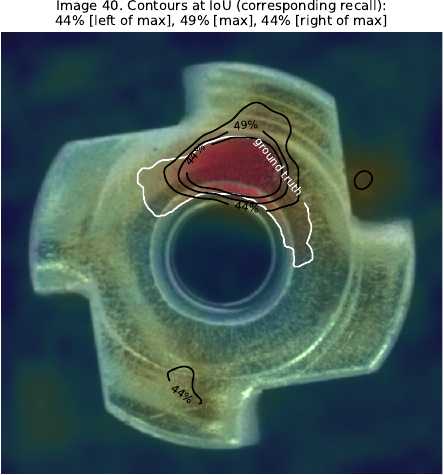}
    \end{subfigure}
    \\[.5em]
    \begin{subfigure}[b]{\linewidth}
        \centering
        \includegraphics[width=.32\linewidth]{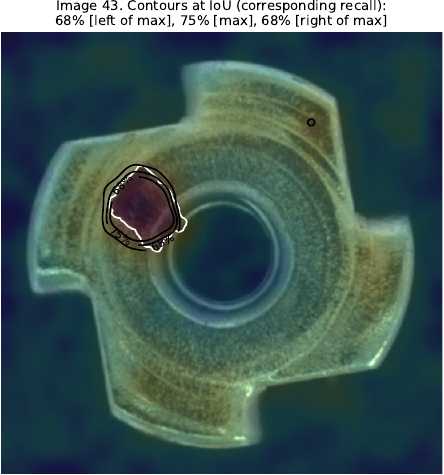}
        \includegraphics[width=.32\linewidth]{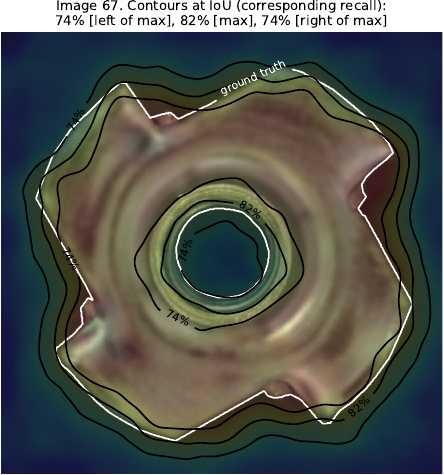}
        \includegraphics[width=.32\linewidth]{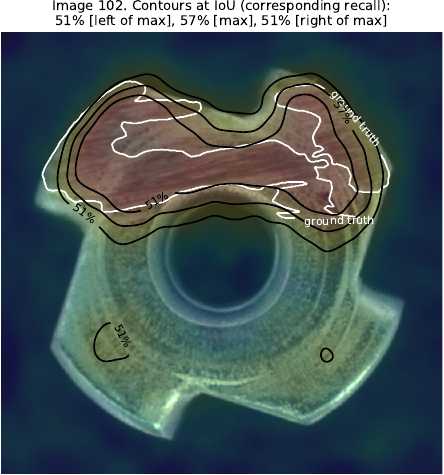}
    \end{subfigure}
\caption{Intersection over union curves and contours at different points along the curves.}
\label{fig:thresh-iou-curves}
\end{figure}

The \gls{iou} curves in \cref{fig:thresh-iou-curves} (built in the same way as \cref{fig:thresh-prec-curves} described above) are smoother, generally show a global maximum, and the level sets at near-maximum-\gls{iou} are more visually stable.
As the \gls{iou} accounts for a balance between precision and recall, it is a more suitable metric for our purposes.

\begin{figure}[hb]
    \centering
    \includegraphics[width=.8\linewidth]{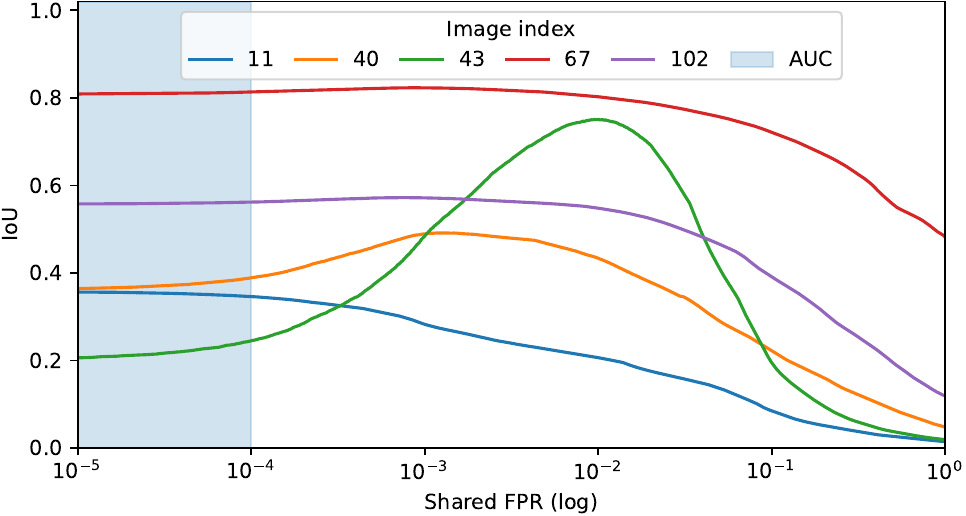}
    \caption{Shared FPR vs. IoU curve.}
    \label{fig:shfpr-vs-iou}
\end{figure}

\cref{fig:shfpr-vs-iou} shows the \gls{shfpr} vs. \gls{iou} curve, which is analogous to the \gls{pimo} curve. 
From this curve, the \gls{auc} score is computed like \gls{aupimo} using the same integration bounds (blue area in \cref{fig:shfpr-vs-iou}).

The cross-image average \gls{auc} scores were added to the results in our benchmark in \cref{app:benchmark}. 
Since the paper already contains a large number of figures, we decided to not include the distributions of the \gls{iou} scores in the paper, but this promissing metric deserves in-depth analysis in future work.


\clearpage
\section{Benchmark}\label{app:benchmark}

In this section we provide additional details about our experiments and results ommited from the main text for brevity.
The following paragraphs provide discuss and detail the evaluation guidelines in our experiments and define a standard format to publish \gls{aupimo} scores.

\paragraph{Full resolution}
Many models typically downsample input images, which conveniently reduces computational costs. 
However, for a fair and model agnostic evaluation, it is important to use the original resolution as it impacts the decision-making when choosing the most suitable model for a use-case. 
If a small anomaly is missed due to downsampling, it is desirable to penalize this, while rewarding models that can handle higher resolution.
As \cite{Zhang_2023_CVPR} pointed out, downsampling ground truth masks creates artifacts, leading to inconsistent results across papers.
While \gls{aupro}'s computational cost is high at full resolution -- especially on CPU -- \gls{aupimo} is orders of magnitude faster (see our results in the paper). 
Our recommendation is to apply bilinear interpolation to upsample anomaly score maps and evaluate at the original resolution in each image.

\paragraph{No crop}
Center crop has been used to leverage the center alignment of the objects depicted in \gls{mvtecad} and \gls{visa}.
However, this is a prior knowledge, hence we do not apply crop.

\paragraph{Sample selection}
To avoid biases from cherry-picking qualitative samples, we propose a systematic selection procedure.
Select the images whose \gls{aupimo} are closest to the statistics in a boxplot: mean, first/second/third quartiles, and low/high whiskers set with maximum size of $1.5 \, \mathrm{IQR}$ (inter-quartile range).
We applied this procedure to select the samples shown in \cref{app:benchmark}.
Note that this is applicable to any per-instance score.

\paragraph{Score publication}
We recommend to publish \gls{aupimo} scores for all images. 
A standard format is specified below.
The field \texttt{paths} is optional but recommended.
For standard datasets like \gls{mvtecad} and \gls{visa}, it is a list of paths to the images in the test set with the path truncated to the dataset root directory.
The field \texttt{num\_threshs} is the effective number of thresholds used to compute the \gls{auc}, which differs from the number of thresholds used to compute the \gls{pimo} curve because only a portion of the curve is used to compute the \gls{aupimo} score.

It is advised to report score distributions (\eg as boxplots and histograms) when possible for a more comprehensive evaluation.
All the scores from our experiments are available in this format at \githubRepo{}.

{
\small
\begin{lstlisting}[language=Python]
    {
        "shared_fpr_metric": "mean_perimage_fpr",
        "fpr_lower_bound": 0.00001, 
        "fpr_upper_bound": 0.0001,
        "num_threshs": 300,
        "thresh_lower_bound": 0.3342, 
        "thresh_upper_bound": 1.1588,
        "aupimos": [0.72107, 0.02415, 0.98991],
        "paths": [
            "MVTec/bottle/test/broken_large/000.png", 
            "MVTec/bottle/test/broken_large/001.png", 
            "MVTec/bottle/test/broken_large/002.png", 
        ]
    }
\end{lstlisting}
}

\clearpage
\subsection{Models}\label{app:benchmark-models}
\cref{app:benchmark-models} lists the models in the benchmark and provides details on the implementation sources and hyperparameters.

We trained and evaluated 13 models from 8 papers listed in \cref{tab:models}.
For some models we considered two backbones and selected the (generally) best out of the two to show in the main text of the paper (see column \cref{tab:models}).

We used the following implementations with the same hyperparameters reported in the papers:
\begin{itemize}
    \item \texttt{anomalib} \cite{akcay_anomalib_2022} (\href{https://github.com/openvinotoolkit/anomalib}{github.com/openvinotoolkit/anomalib}\footnote{Commit \texttt{09ad1d4b1e8f634b72f788314275d3aea33815dd}.}) for \gls{padim} \cite{defard_padim_2021}, \gls{patchcore} \cite{roth_towards_2022}, and \gls{fastflow} \cite{yu_fastflow_2021};
    \item \href{https://github.com/gasharper/PyramidFlow}{github.com/gasharper/PyramidFlow} for \gls{pyramidflow} \cite{lei_pyramidflow_2023};
    \item \href{https://github.com/donaldrr/simplenet}{github.com/donaldrr/simplenet} for \gls{simplenet} \cite{liu_simplenet_2023};
    \item \href{https://github.com/tientrandinh/revisiting-reverse-distillation}{github.com/tientrandinh/revisiting-reverse-distillation} for \gls{reversedistpp} \cite{tien_revisiting_2023};
    \item \href{https://github.com/mtailanian/uflow}{github.com/mtailanian/uflow} for \gls{uflow} \cite{tailanian_u-flow_2023};
    \item \href{https://github.com/nelson1425/EfficientAD}{github.com/nelson1425/EfficientAD} for \gls{efficientad} \cite{batzner_efficientad_2023}.
\end{itemize}
The non-official implementations are the ones from \texttt{anomalib} and \gls{efficientad}.

\begin{table}[ht]
    \centering
    \footnotesize
    \begin{tabular}{llllcl} 
        \textbf{Model}&\textbf{Publication}&  \textbf{Backbone}&  \textbf{Family}&  \textbf{Paper} &\textbf{Implem.}\\ 
        \gls{padim}&ICPR 21&  \gls{bbRoe}&  probability density&   \checkmark&anomalib\\ 
        \gls{padim}&ICPR 21&  \gls{bbWrfz}&  probability density&  --&anomalib\\ 
        \gls{patchcore}&CVPR 22&  \gls{bbWrfz}&  memory bank&   --&anomalib\\ 
        \gls{patchcore}&CVPR 22&  \gls{bbWrozo}&  memory bank&  \checkmark &anomalib\\ 
        \gls{simplenet}&CVPR 23&  \gls{bbWrfz}&  reconstruction&  \checkmark &official\\ 
        \gls{pyramidflow}&CVPR 23& \gls{bbRoe}& normalizing flow&  --&official\\ 
        \gls{pyramidflow}&CVPR 23& --& normalizing flow& \checkmark &official\\ 
        \gls{reversedistpp}&CVPR 23& \gls{bbWrfz}& student-teacher& \checkmark &official\\
        \gls{fastflow}& arXiv (21)& \gls{bbWrfz}& normalizing flow&  --&anomalib\\
        \gls{fastflow}& arXiv (21)& \gls{bbCait}& normalizing flow& \checkmark &anomalib\\
        \gls{efficientad}-S& arXiv (23)& \gls{bbWrozo}& student-teacher&  --& unofficial \\
        \gls{efficientad}-M& arXiv (23)& \gls{bbWrozo}& student-teacher& \checkmark & unofficial \\
        \gls{uflow}& arXiv (23)& --& normalizing flow& \checkmark &official\\ 
    \end{tabular}
    \caption{Models. Years were abbreviated to the last two digits.}
    \label{tab:models}
\end{table}

\clearpage
\subsection{Cross-dataset analysis}

\begin{figure}[ht]
  \centering
  \includegraphics[height=60mm]{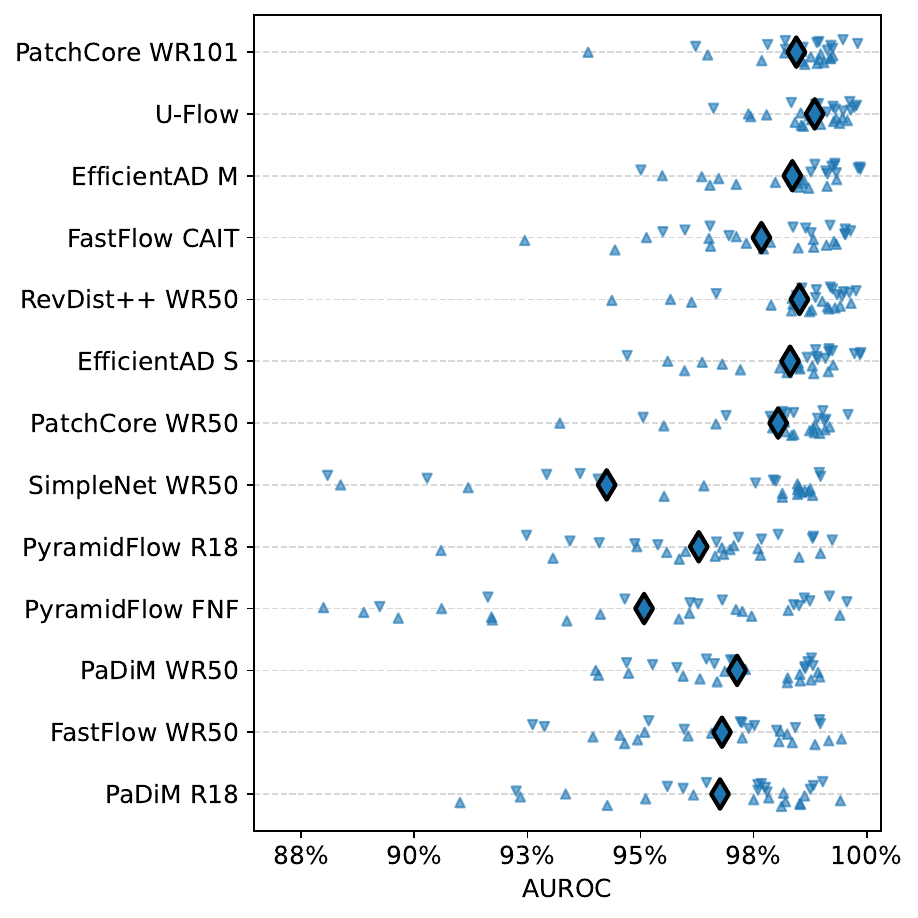}
  \includegraphics[height=60mm]{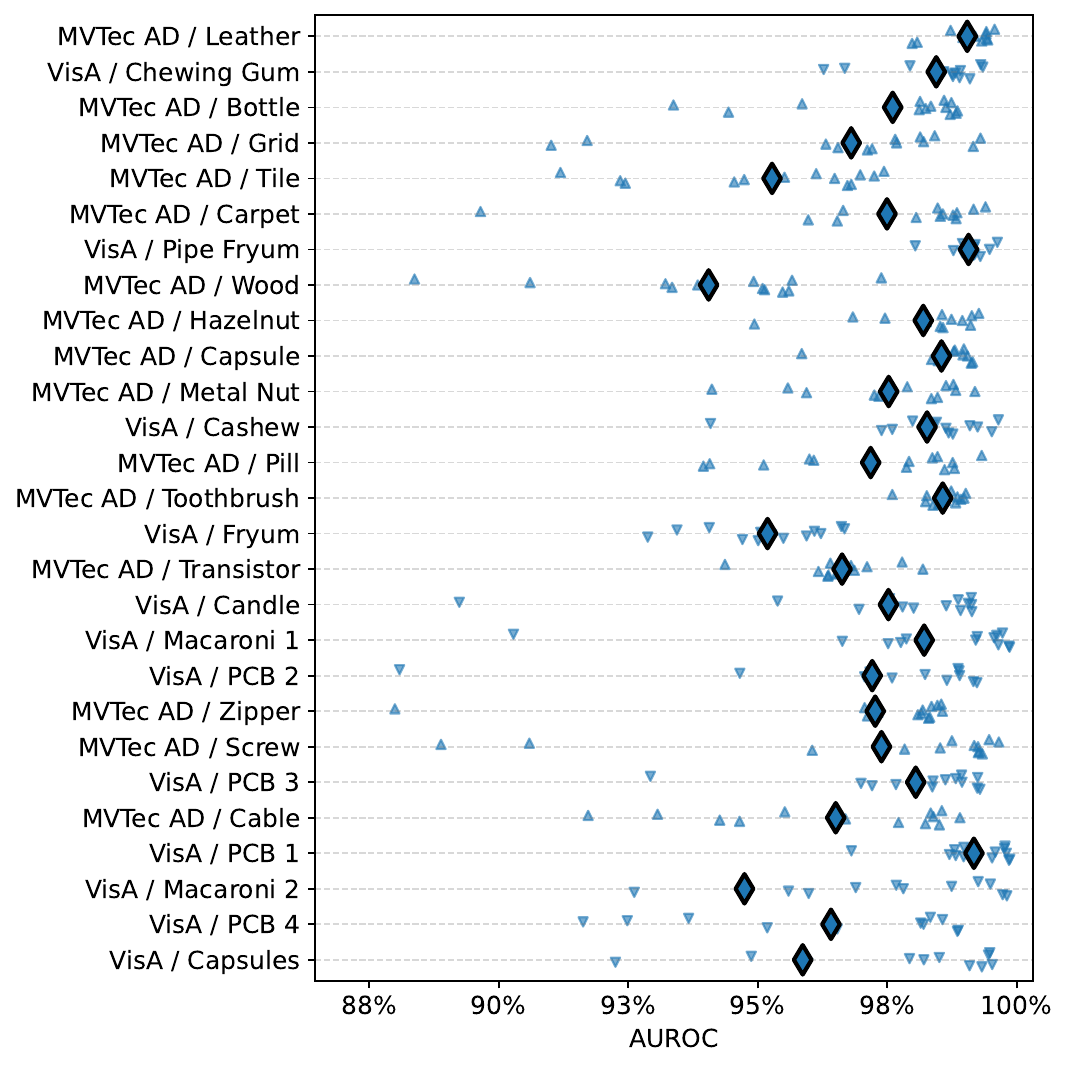}
  \caption{AUROC}
  \label{fig:datasetwise-scores-auroc}
\end{figure}

\begin{figure}[ht]
  \centering
  \includegraphics[height=60mm]{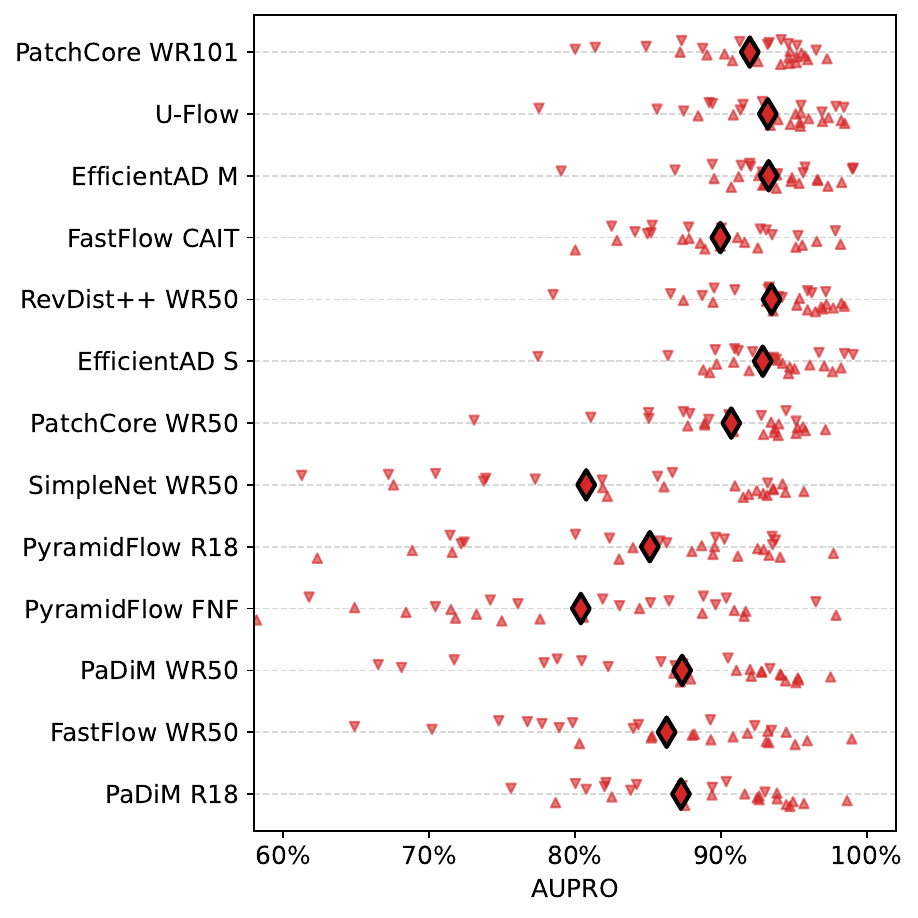}
  \includegraphics[height=60mm]{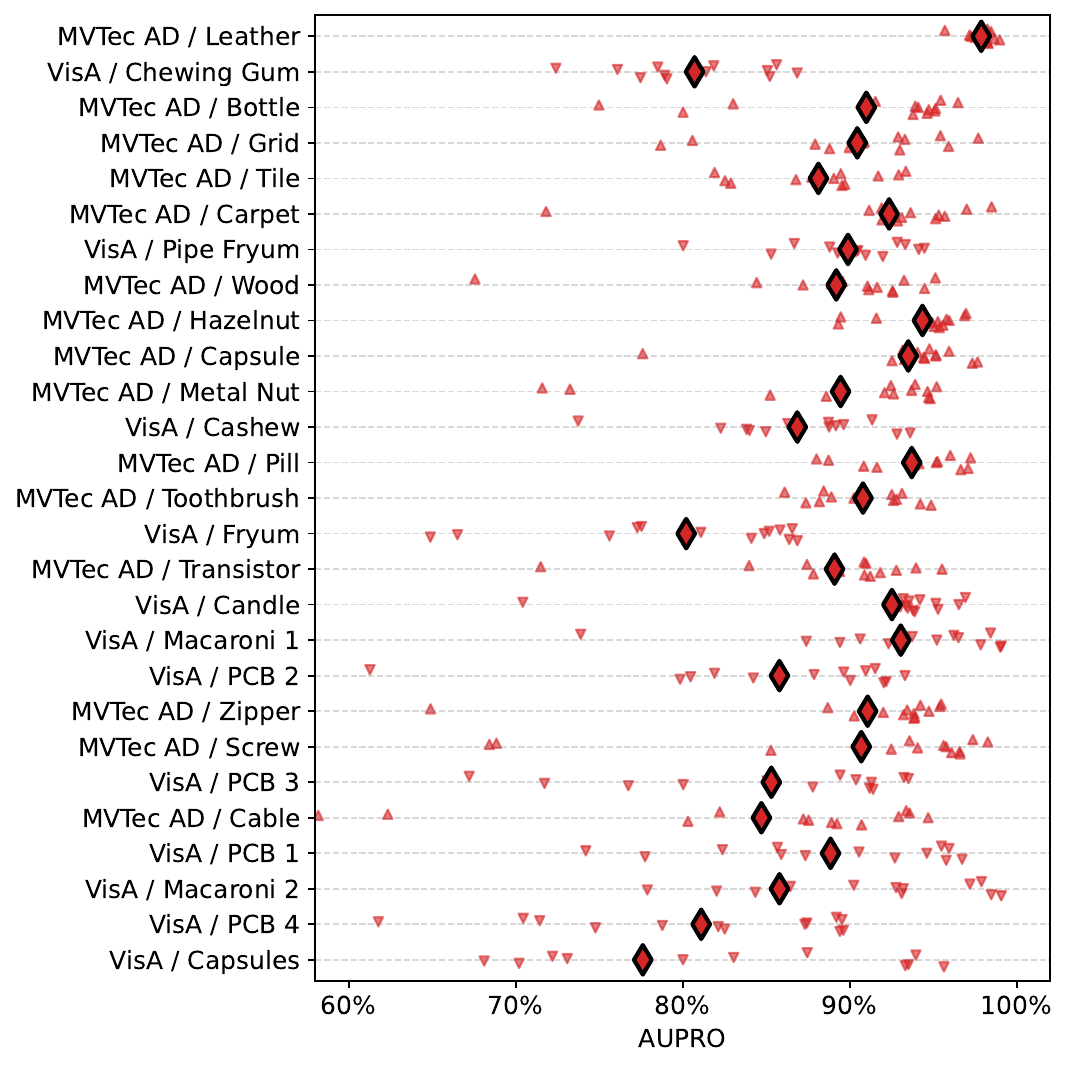}
  \caption{AUPRO}
  \label{fig:datasetwise-scores-aupro}
\end{figure}

\begin{figure}[ht]
  \centering
  \includegraphics[height=60mm]{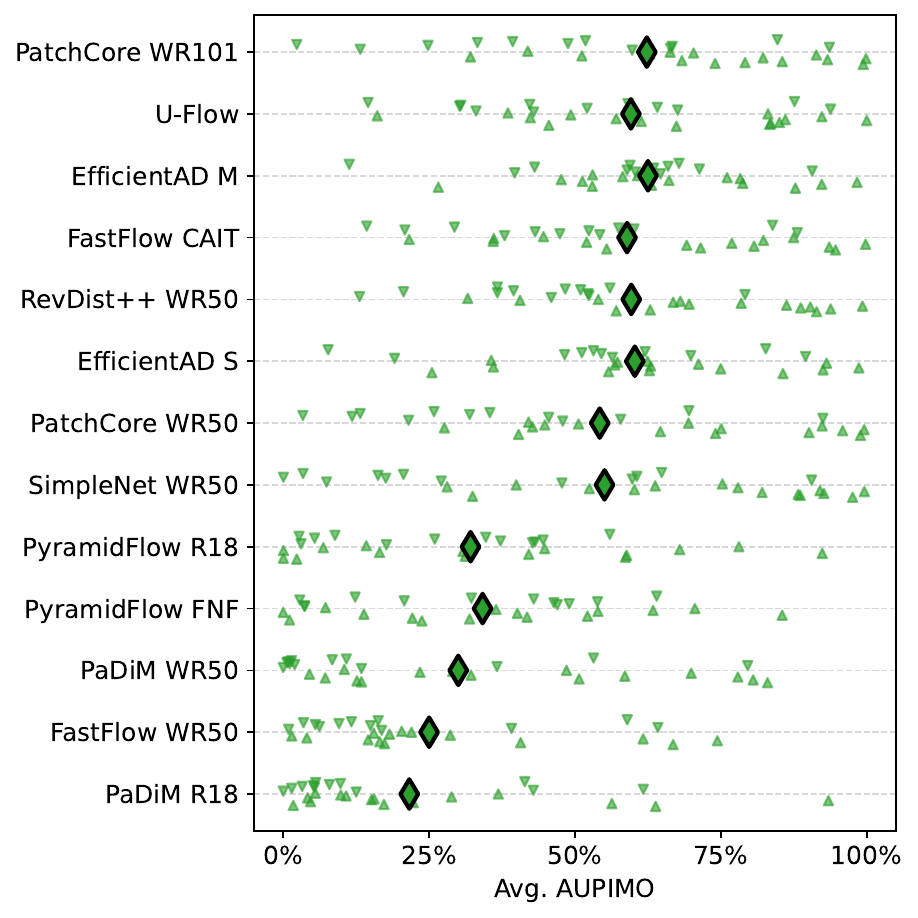}
  \includegraphics[height=60mm]{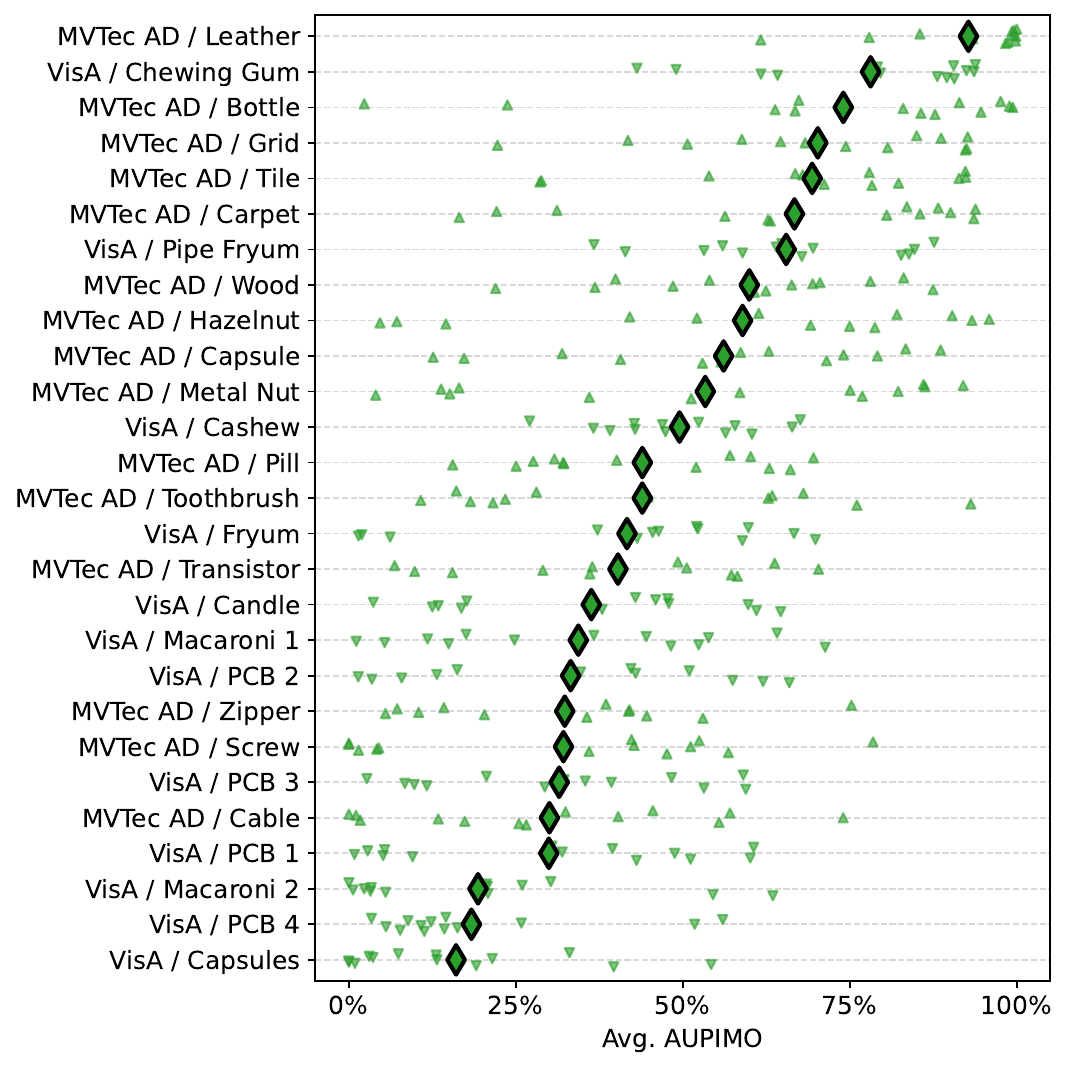}
  \caption{Average AUPIMO}
  \label{fig:datasetwise-scores-avgaupimo}
\end{figure}

\begin{figure}[ht]
  \centering
  \includegraphics[height=60mm]{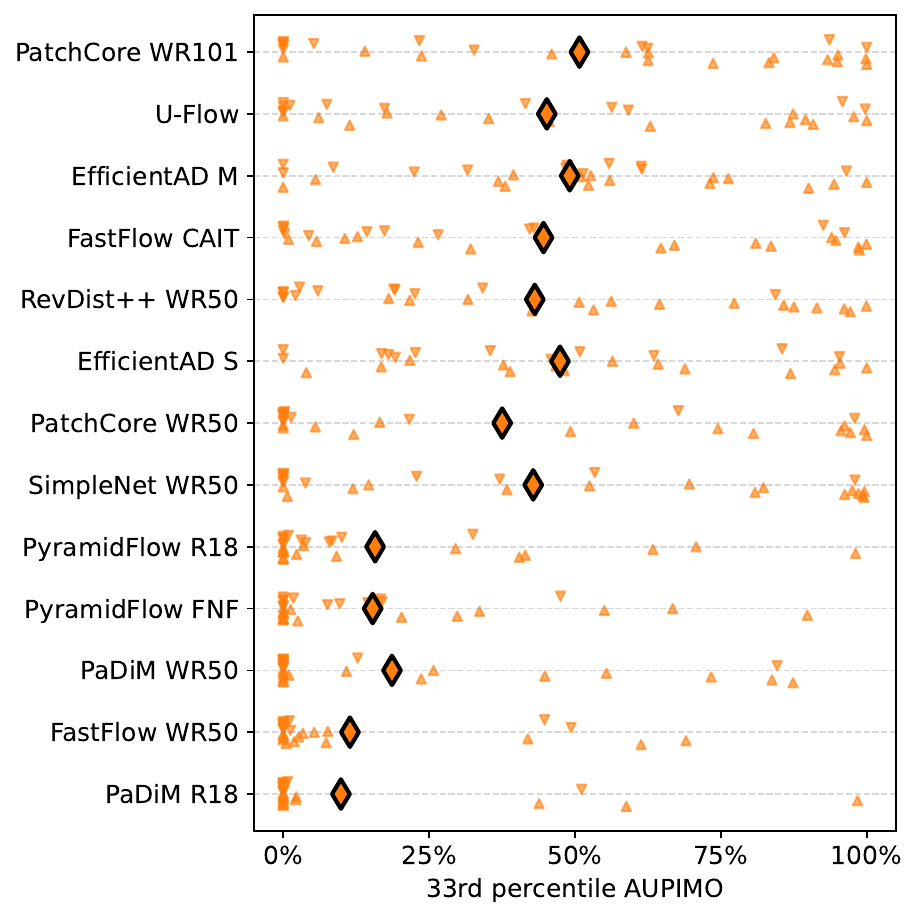}
  \includegraphics[height=60mm]{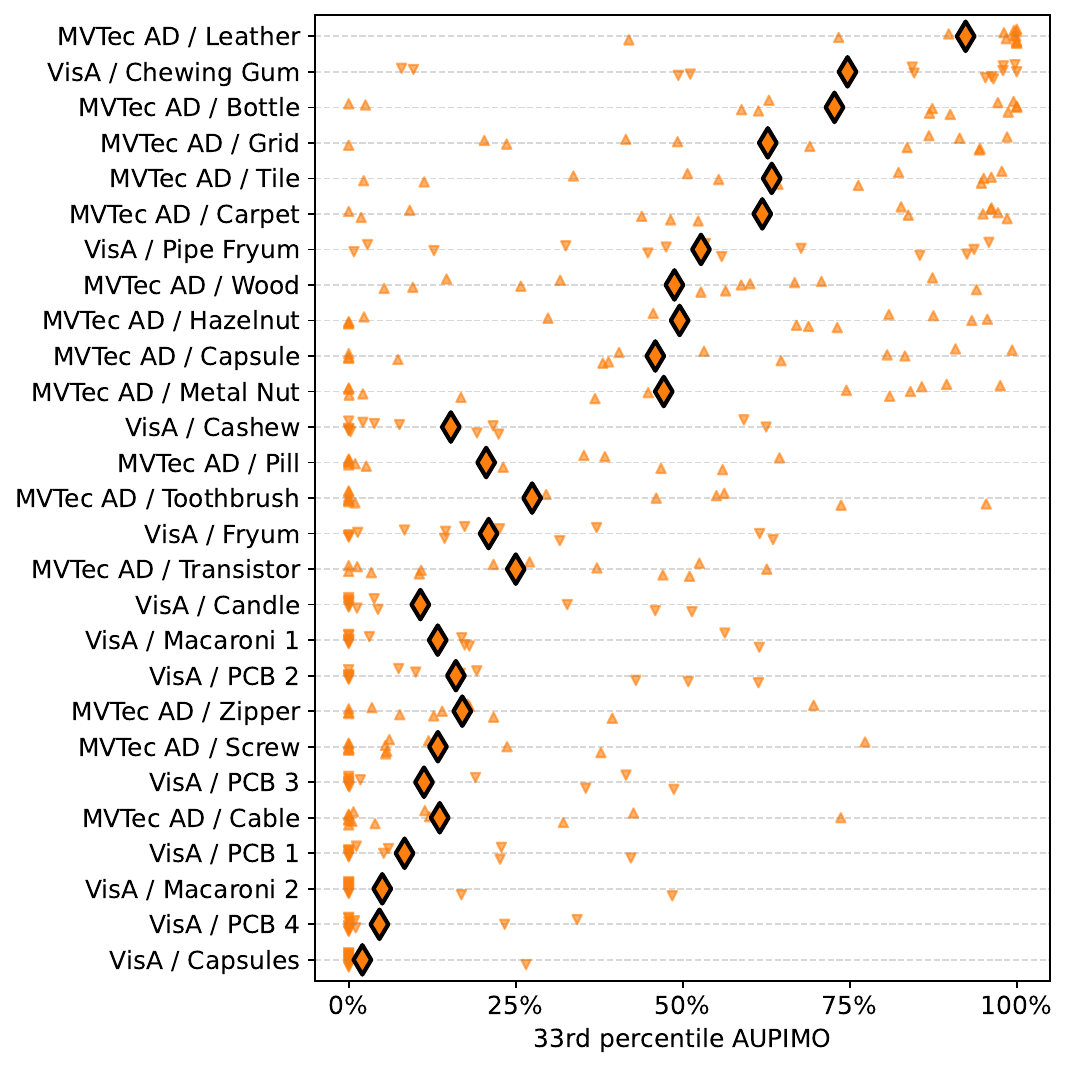}
  \caption{$\perc{33}$ AUPIMO}
  \label{fig:datasetwise-scores-pt33aupimo}
\end{figure}

\begin{figure}[ht]
  \centering
  \includegraphics[height=60mm]{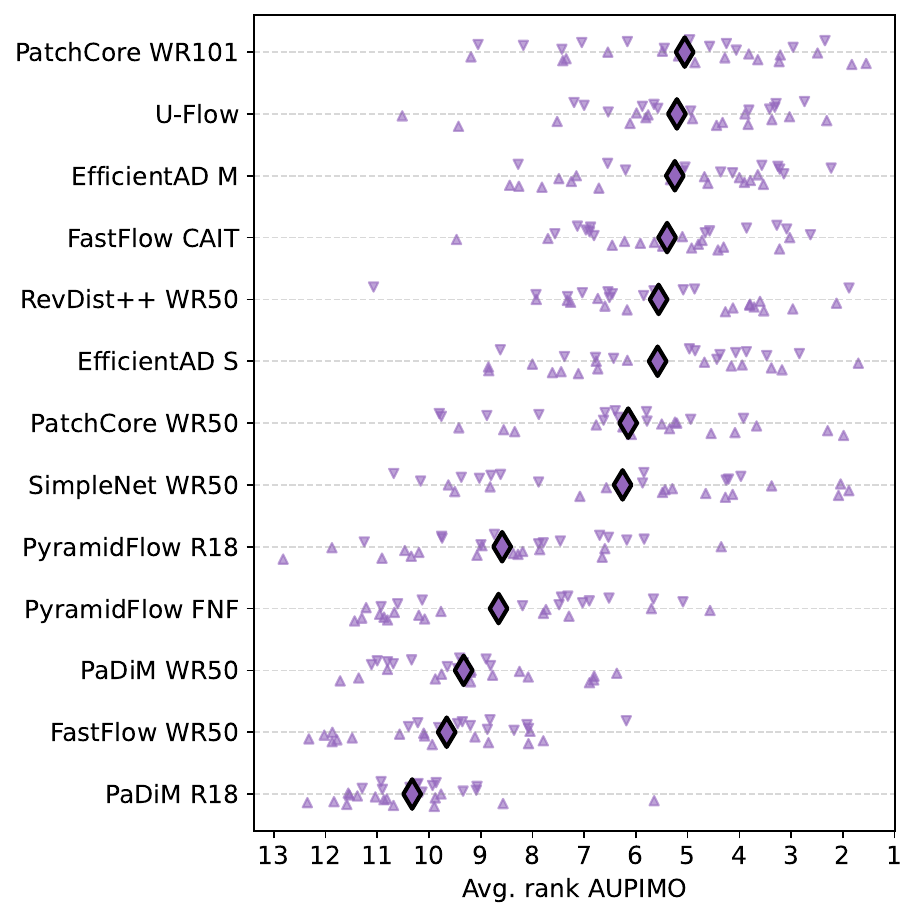}
  \caption{Average rank according to AUPIMO}
  \label{fig:datasetwise-aupimo-avgrank}
\end{figure}

In this section, the model performances are summarized across all the datasets in \gls{mvtecad} and \gls{visa} (all confounded) according to
\begin{enumerate}
    \item \gls{auroc} (\cref{fig:datasetwise-scores-auroc}), 
    \item \gls{aupro} (\cref{fig:datasetwise-scores-aupro}),
    \item average \gls{aupimo} (\cref{fig:datasetwise-scores-avgaupimo}),
    \item \nth{33} percentile \gls{aupimo} (\cref{fig:datasetwise-scores-pt33aupimo}),
    \item average image-wise rank according to \gls{aupimo} scores (\cref{fig:datasetwise-aupimo-avgrank}).
\end{enumerate}

\paragraph{Scores}
In \cref{fig:datasetwise-scores-auroc}, \cref{fig:datasetwise-scores-aupro}, \cref{fig:datasetwise-scores-avgaupimo}, and \cref{fig:datasetwise-scores-pt33aupimo}, each point represents the score in the test set or an \gls{aupimo} statistic (average and \nth{33} percentile) across the images in the test set.
Diamonds are averages across datasets (both collections confounded) or across models.
Notice the difference in the X-axis scales.

\paragraph{Percentile 33 score}
While the average \gls{aupimo} is a useful indicator, we propose the use of the \nth{33} percentile of \gls{aupimo} scores, denoted $\perc{33}$, for a more rigorous, worst-case evaluation. 
A $\perc{33}$ score of value $V$ indicates that two thirds of the images in the test set have an \gls{aupimo} score of \emph{at least} $V$.
Otherwise stated, a $\perc{33}$ score of value $V$ indicates that one third of the images in the test set have an \gls{aupimo} score of \emph{at most} $V$.

\paragraph{Average ranks}
\cref{fig:datasetwise-aupimo-avgrank} shows the average image ranks according to \gls{aupimo} as points and the average across datasets as diamonds. 
At each image from a given dataset, ranks are assigned to the models (\say{which model best detects this specific image?}), and the average is taken across all images from the same dataset.
The range of rank values is from 1 (best) to number of models (worst).

\paragraph{Summary table}
\cref{tab:model-averages} summarizes the average scores across datasets within each dataset collection (\gls{mvtecad} and \gls{visa}) and across all datasets (both collections confounded).


\begin{table}[ht]
    \centering
    \scriptsize
    \begin{tabular}{llcccccc}
\toprule
Model & Dataset Collection & AUROC & AUPRO & AUPIMO & Avg. & $\perc{33}$ & Avg. Rank \\
\midrule
PaDiM R18 & MVTec AD & 96.62 & 91.58 &   & 25.75 & 14.34 & 10.5 \\
PaDiM R18 & VisA & 97.22 & 81.87 &   & 16.42 & 4.33 & 10.1 \\
PaDiM R18 & All & 96.89 & 87.27 &   & 21.61 & 9.89 & 10.3 \\
FastFlow WR50 & MVTec AD & 97.01 & 90.87 &   & 28.49 & 14.15 & 10.3 \\
FastFlow WR50 & VisA & 96.83 & 80.54 &   & 20.65 & 8.03 & 8.9 \\
FastFlow WR50 & All & 96.93 & 86.28 &   & 25.00 & 11.43 & 9.7 \\
PaDiM WR50 & MVTec AD & 97.19 & 92.57 &   & 40.14 & 27.06 & 8.9 \\
PaDiM WR50 & VisA & 97.32 & 80.81 &   & 17.34 & 8.12 & 9.9 \\
PaDiM WR50 & All & 97.25 & 87.35 &   & 30.01 & 18.64 & 9.3 \\
PyramidFlow FNF & MVTec AD & 94.21 & 79.10 &   & 36.26 & 19.94 & 9.4 \\
PyramidFlow FNF & VisA & 96.62 & 82.03 &   & 31.55 & 9.56 & 7.8 \\
PyramidFlow FNF & All & 95.28 & 80.40 &   & 34.17 & 15.33 & 8.7 \\
PyramidFlow R18 & MVTec AD & 96.36 & 85.81 &   & 36.32 & 23.91 & 9.0 \\
PyramidFlow R18 & VisA & 96.53 & 84.27 &   & 26.84 & 5.55 & 8.1 \\
PyramidFlow R18 & All & 96.44 & 85.13 &   & 32.11 & 15.75 & 8.6 \\
SimpleNet WR50 & MVTec AD & 97.13 & 89.48 &   & 71.39 & 62.78 & 5.3 \\
SimpleNet WR50 & VisA & 91.17 & 69.88 &   & 34.66 & 17.93 & 7.4 \\
SimpleNet WR50 & All & 94.48 & 80.77 &   & 55.07 & 42.84 & 6.3 \\
PatchCore WR50 & MVTec AD & 98.01 & 93.13 &   & 67.21 & 54.95 & 5.6 \\
PatchCore WR50 & VisA & 98.26 & 87.69 &   & 38.02 & 15.74 & 6.9 \\
PatchCore WR50 & All & 98.12 & 90.72 &   & 54.24 & 37.53 & 6.1 \\
EfficientAD S & MVTec AD & 97.96 & 93.65 &   & 64.76 & 55.16 & 5.9 \\
EfficientAD S & VisA & 98.89 & 91.90 &   & 54.62 & 37.78 & 5.2 \\
EfficientAD S & All & 98.37 & 92.87 &   & 60.25 & 47.44 & 5.6 \\
RevDist++ WR50 & MVTec AD & 98.23 & 95.03 &   & 71.93 & 64.93 & 4.9 \\
RevDist++ WR50 & VisA & 99.00 & 91.53 &   & 44.30 & 15.85 & 6.3 \\
RevDist++ WR50 & All & 98.57 & 93.48 &   & 59.65 & 43.11 & 5.6 \\
FastFlow CAIT & MVTec AD & 97.37 & 90.44 &   & 66.79 & 57.83 & 5.4 \\
FastFlow CAIT & VisA & 98.25 & 89.37 &   & 49.10 & 28.09 & 5.4 \\
FastFlow CAIT & All & 97.76 & 89.96 &   & 58.93 & 44.61 & 5.4 \\
EfficientAD M & MVTec AD & 97.96 & 94.10 &   & 66.08 & 55.97 & 5.8 \\
EfficientAD M & VisA & 99.00 & 92.25 &   & 58.06 & 40.52 & 4.6 \\
EfficientAD M & All & 98.42 & 93.28 &   & 62.52 & 49.10 & 5.2 \\
U-Flow & MVTec AD & 98.74 & 94.89 &   & 66.07 & 56.07 & 5.4 \\
U-Flow & VisA & 99.09 & 91.14 &   & 51.48 & 31.54 & 4.9 \\
U-Flow & All & 98.89 & 93.22 &   & 59.58 & 45.17 & 5.2 \\
PatchCore WR101 & MVTec AD & 98.35 & 93.53 &   & 73.19 & 66.12 & 4.7 \\
PatchCore WR101 & VisA & 98.70 & 90.06 &   & 48.72 & 31.58 & 5.5 \\
PatchCore WR101 & All & 98.51 & 91.99 &   & 62.31 & 50.77 & 5.1 \\
\bottomrule
\end{tabular}

    \caption{Model averages. Scores are in percentages. Ranks range from 1 (best) to number of models (worst). }
    \label{tab:model-averages}
\end{table}

\clearpage
\subsection{Per-model analyses}\label{app:permodel}

\begin{figure}[ht]
    \centering
    \begin{subfigure}{0.49\linewidth}
        \centering
        \includegraphics[width=\linewidth]{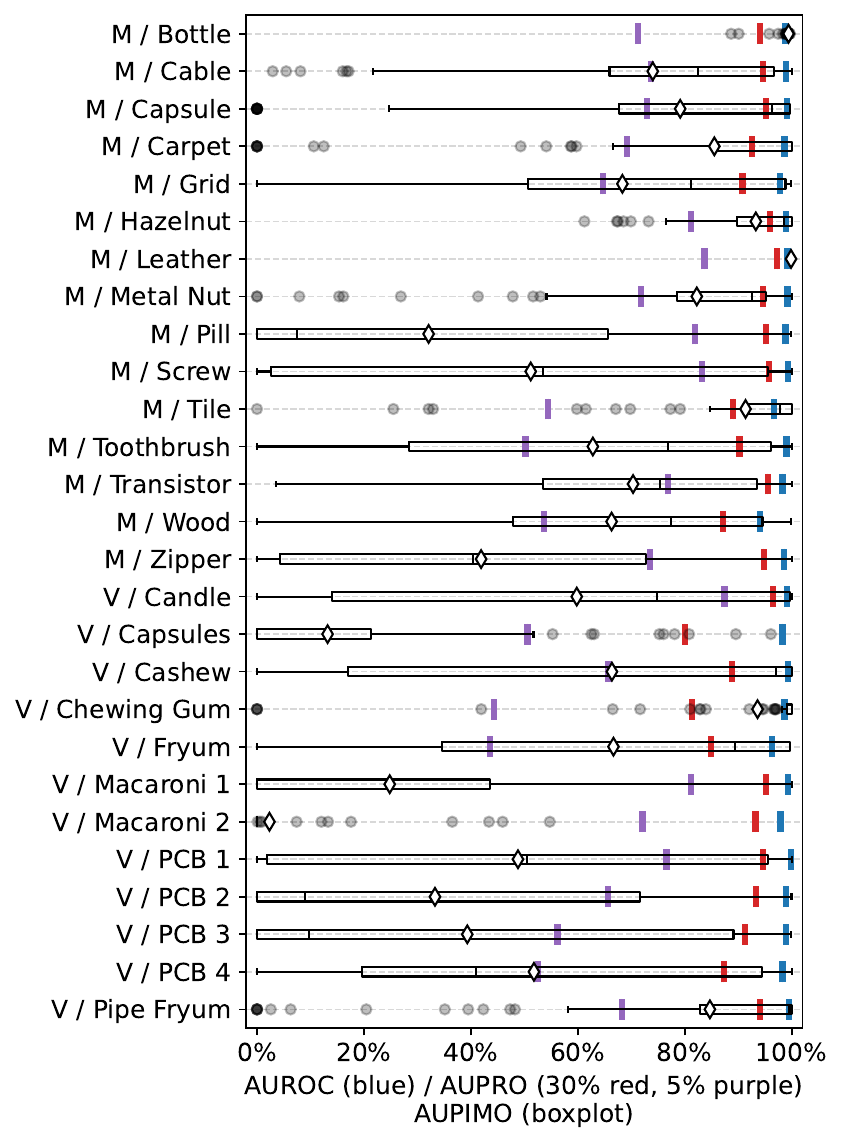}
        \caption{PatchCore-WR101}
        \label{fig:permodel-showcase-patchcore}
    \end{subfigure}
    \begin{subfigure}{0.49\linewidth}
        \centering
        \includegraphics[width=\linewidth]{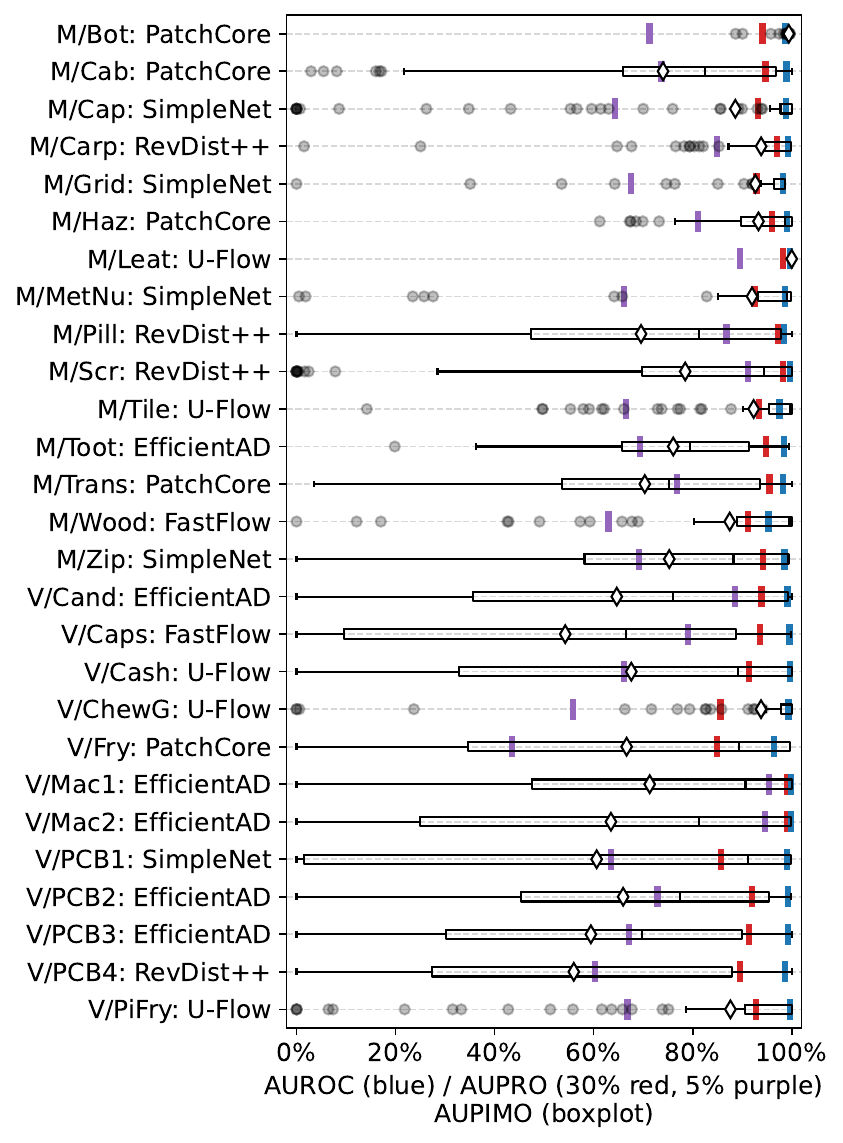}
        \caption{Per-dataset best models}
        \label{fig:permodel-showcase-best}
    \end{subfigure}
    \caption{AUPIMO distributions for PatchCore-WR101 (left) and per-dataset best models (right).}
    \label{fig:permodel-showcase}
\end{figure}

\cref{fig:permodel-showcase} shows that current \gls{aloc} models still are not capable of cracking the datasets from \gls{mvtecad} and \gls{visa}. 
\cref{fig:permodel-showcase-best} shows the \gls{aupimo} distributions of \gls{patchcoreWrozo}, the model with best cross-dataset average.
Even though it is the overall best, it still has a long tail of low \gls{aupimo} scores on several datasets like Grid and Wood, or in some cases it practically fails to detect any anomaly at all, like in Capsules and Macaroni 2.
\cref{fig:permodel-showcase-best} shows the \gls{aupimo} distributions of the best model per dataset.
Even if a user would be willing to select a model per dataset, there is no clear winner, and most datasets from \gls{visa} show challenging images that are not detected. 

\clearpage
\subsection{Per-dataset analyses}

The following figures are detailed results from the benchmark of all the datasets from \gls{mvtecad} or \gls{visa}.

\begin{figure}[ht]
    \centering
    \begin{subfigure}[b]{\linewidth}
      \includegraphics[width=\linewidth,valign=t,keepaspectratio]{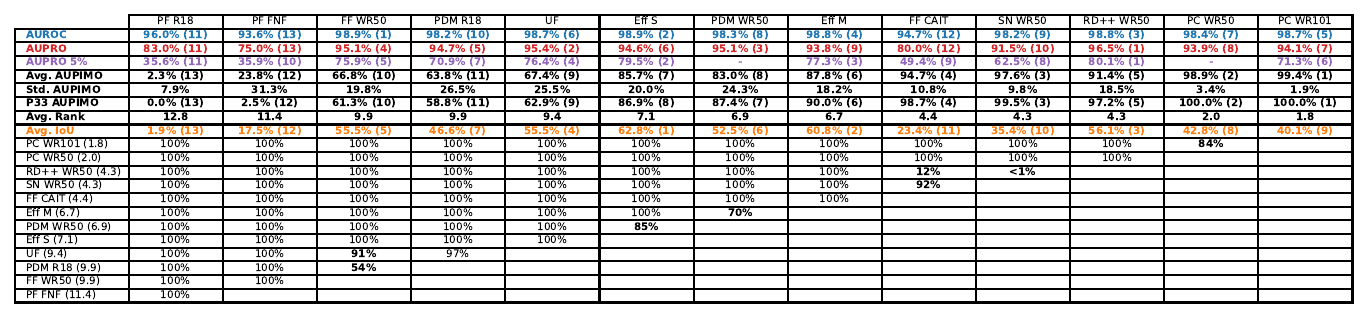}
      \caption{Statistics and pairwise statistical tests.}
      \label{fig:benchmark-000-table}
    \end{subfigure}
    \\ \vspace{2mm}
    \begin{subfigure}[b]{0.5\linewidth}
      \includegraphics[width=\linewidth,valign=t,keepaspectratio]{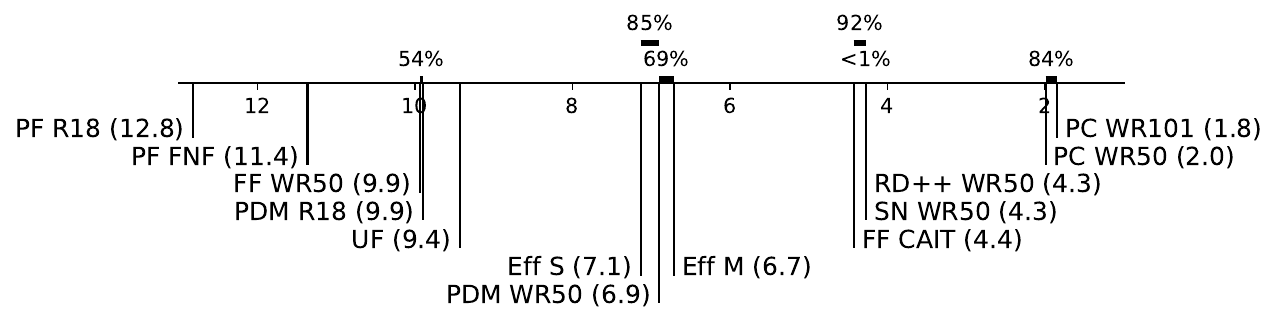}
      \caption{Average rank diagram.}
      \label{fig:benchmark-000-diagram}
    \end{subfigure}
    \\ \vspace{2mm}
    \begin{subfigure}[b]{0.45\linewidth}
      \includegraphics[width=\linewidth,valign=t,keepaspectratio]{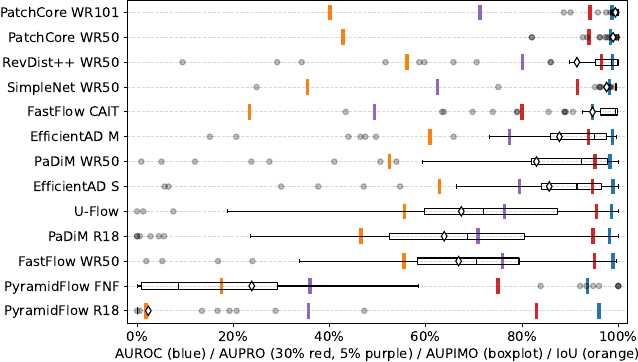}
      \caption{Score distributions.}
      \label{fig:benchmark-000-boxplot}
    \end{subfigure}
    ~
    \begin{subfigure}[b]{0.45\linewidth}
      \includegraphics[width=\linewidth,valign=t,keepaspectratio]{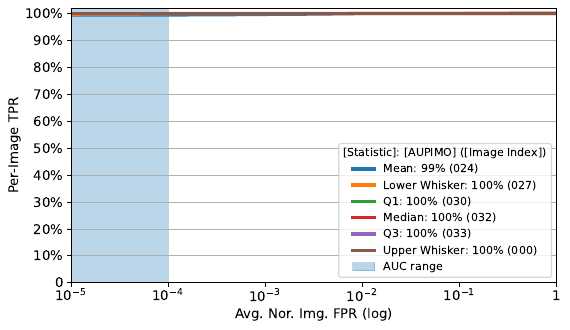}
      \caption{PIMO curves.}
      \label{fig:benchmark-000-pimo-curves}
    \end{subfigure}
    \\  \vspace{2mm}
    \begin{subfigure}[b]{\linewidth}
    
      \begin{minipage}{\linewidth}
        \centering
        \includegraphics[width=.3\linewidth,valign=t,keepaspectratio]{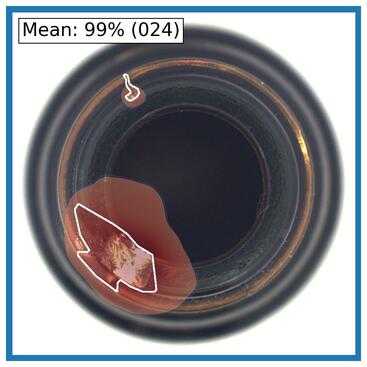}
        \includegraphics[width=.3\linewidth,valign=t,keepaspectratio]{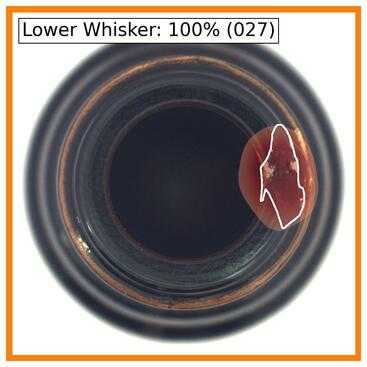}
        \includegraphics[width=.3\linewidth,valign=t,keepaspectratio]{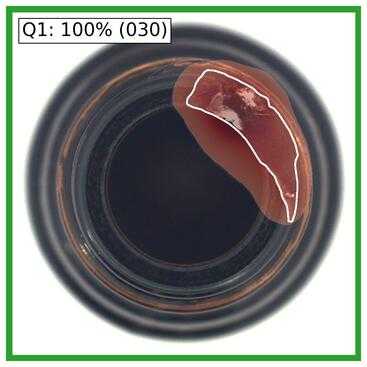}
      \end{minipage}
      \\
      \begin{minipage}{\linewidth}
        \centering
        \includegraphics[width=.3\linewidth,valign=t,keepaspectratio]{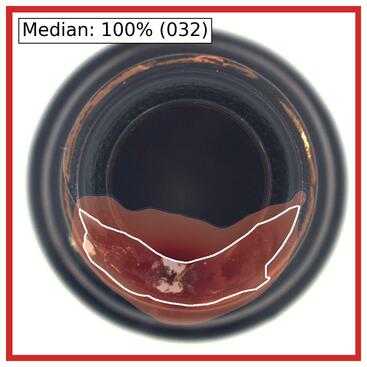}
        \includegraphics[width=.3\linewidth,valign=t,keepaspectratio]{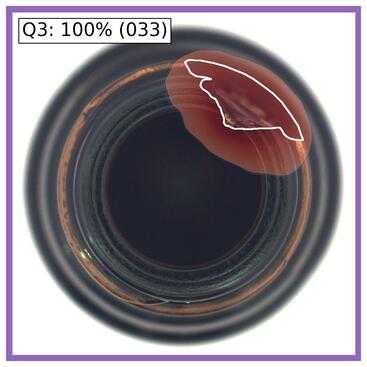}
        \includegraphics[width=.3\linewidth,valign=t,keepaspectratio]{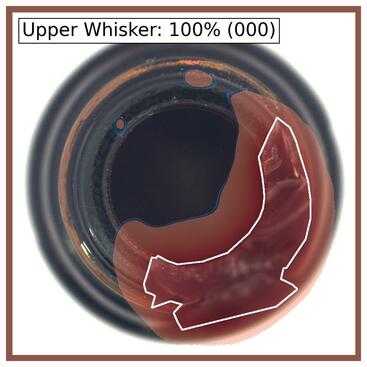}
      \end{minipage}
      \caption{
        Heatmaps.
        Images selected according to AUPIMO's statistics.
        Statistic and image index annotated on upper left corner.
      }
      \label{fig:benchmark-000-heatmap}
    \end{subfigure}
    \caption{
      Benchmark on MVTec AD / Bottle.
      PIMO curves and heatmaps are from PatchCore WR101.
      083 images (020 normal, 063 anomalous).
    }
    \label{fig:benchmark-000}
\end{figure}

\begin{figure}[ht]
    \centering
    \begin{subfigure}[b]{\linewidth}
      \includegraphics[width=\linewidth,valign=t,keepaspectratio]{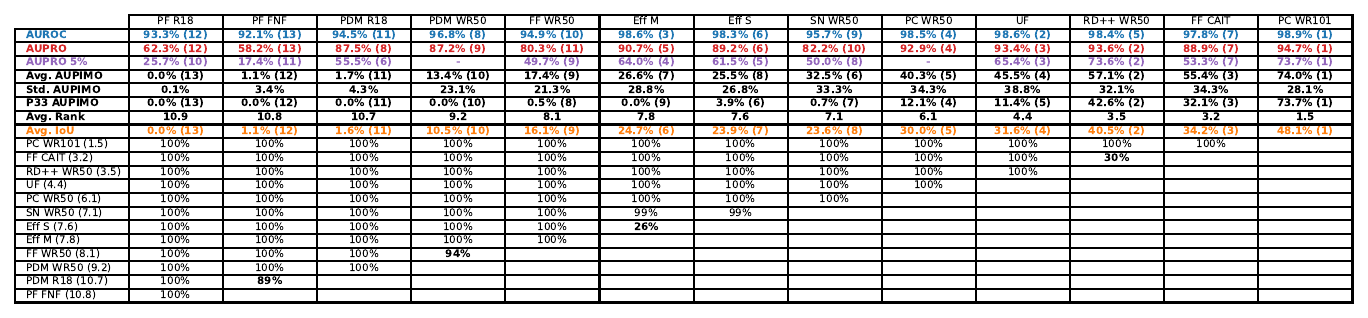}
      \caption{Statistics and pairwise statistical tests.}
      \label{fig:benchmark-001-table}
    \end{subfigure}
    \\ \vspace{2mm}
    \begin{subfigure}[b]{0.5\linewidth}
      \includegraphics[width=\linewidth,valign=t,keepaspectratio]{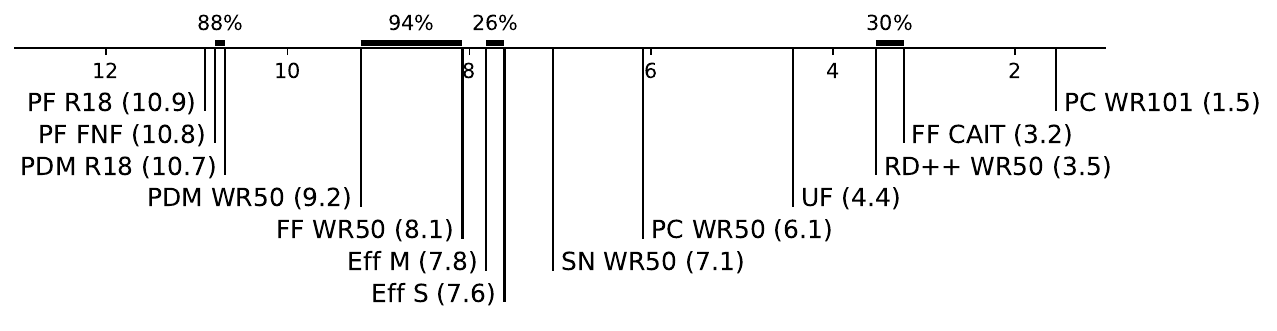}
      \caption{Average rank diagram.}
      \label{fig:benchmark-001-diagram}
    \end{subfigure}
    \\ \vspace{2mm}
    \begin{subfigure}[b]{0.45\linewidth}
      \includegraphics[width=\linewidth,valign=t,keepaspectratio]{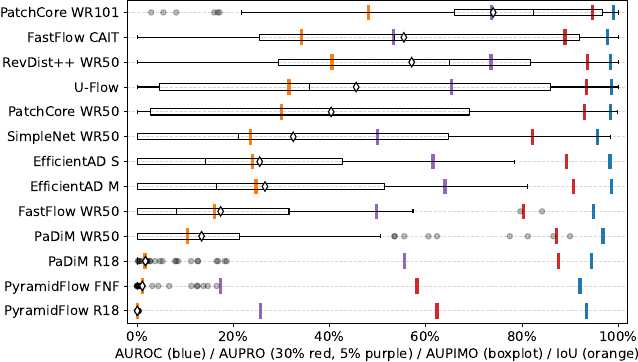}
      \caption{Score distributions.}
      \label{fig:benchmark-001-boxplot}
    \end{subfigure}
    ~
    \begin{subfigure}[b]{0.45\linewidth}
      \includegraphics[width=\linewidth,valign=t,keepaspectratio]{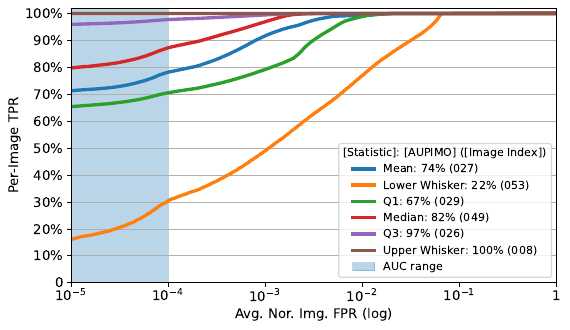}
      \caption{PIMO curves.}
      \label{fig:benchmark-001-pimo-curves}
    \end{subfigure}
    \\  \vspace{2mm}
    \begin{subfigure}[b]{\linewidth}
    
      \begin{minipage}{\linewidth}
        \centering
        \includegraphics[width=.3\linewidth,valign=t,keepaspectratio]{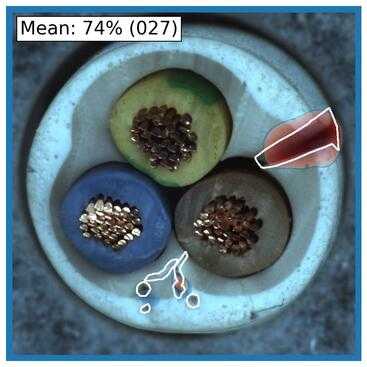}
        \includegraphics[width=.3\linewidth,valign=t,keepaspectratio]{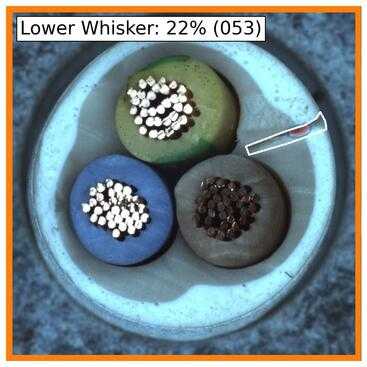}
        \includegraphics[width=.3\linewidth,valign=t,keepaspectratio]{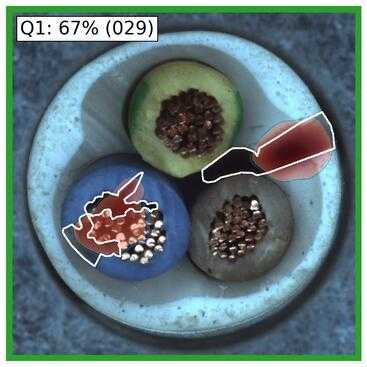}
      \end{minipage}
      \\
      \begin{minipage}{\linewidth}
        \centering
        \includegraphics[width=.3\linewidth,valign=t,keepaspectratio]{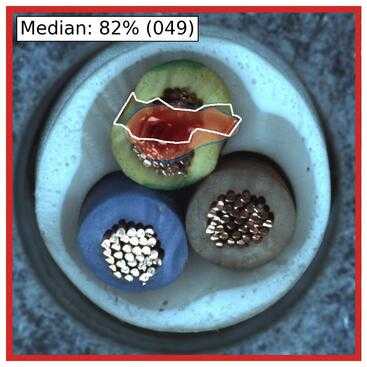}
        \includegraphics[width=.3\linewidth,valign=t,keepaspectratio]{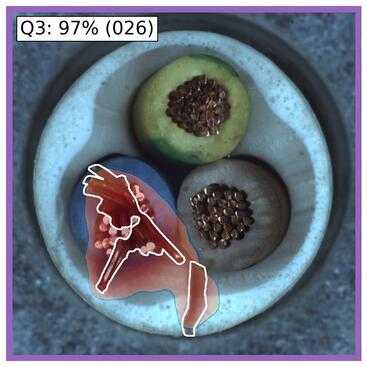}
        \includegraphics[width=.3\linewidth,valign=t,keepaspectratio]{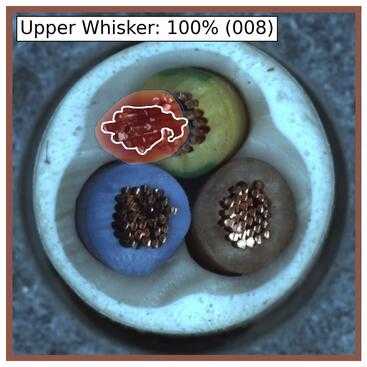}
      \end{minipage}
      \caption{
        Heatmaps.
        Images selected according to AUPIMO's statistics.
        Statistic and image index annotated on upper left corner.
        Image index annotated on upper left corner.
      }
      \label{fig:benchmark-001-heatmap}
    \end{subfigure}
    \caption{
      Benchmark on MVTec AD / Cable.
      PIMO curves and heatmaps are from PatchCore WR101.
      150 images (058 normal, 092 anomalous).
    }
    \label{fig:benchmark-001}
\end{figure}

\begin{figure}[ht]
    \centering
    \begin{subfigure}[b]{\linewidth}
      \includegraphics[width=\linewidth,valign=t,keepaspectratio]{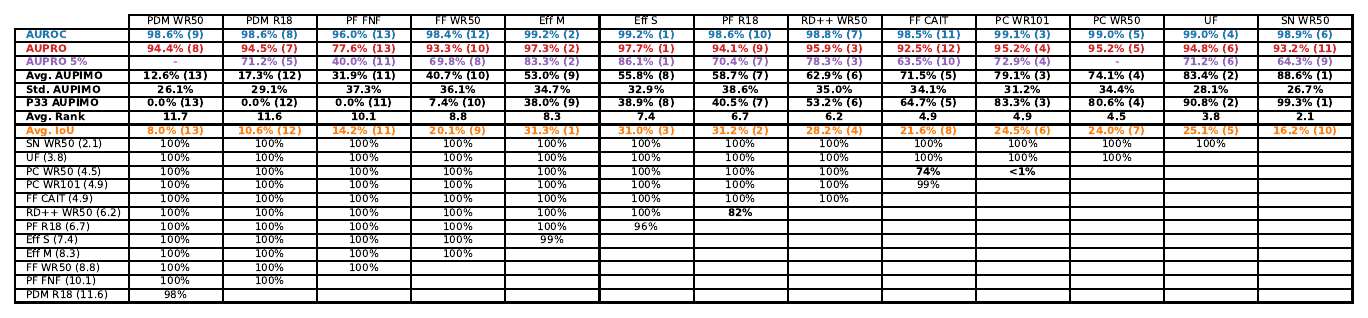}
      \caption{Statistics and pairwise statistical tests.}
      \label{fig:benchmark-002-table}
    \end{subfigure}
    \\ \vspace{2mm}
    \begin{subfigure}[b]{0.5\linewidth}
      \includegraphics[width=\linewidth,valign=t,keepaspectratio]{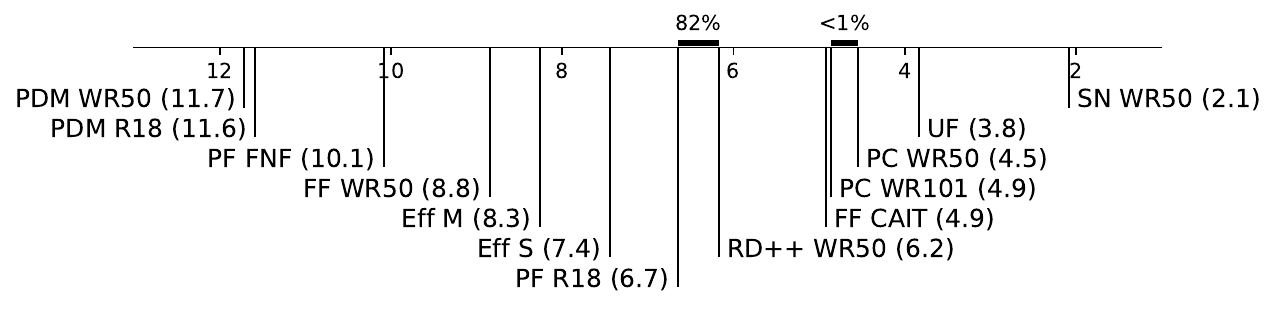}
      \caption{Average rank diagram.}
      \label{fig:benchmark-002-diagram}
    \end{subfigure}
    \\ \vspace{2mm}
    \begin{subfigure}[b]{0.45\linewidth}
      \includegraphics[width=\linewidth,valign=t,keepaspectratio]{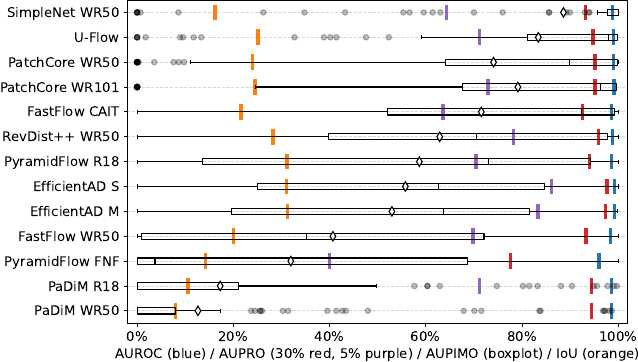}
      \caption{Score distributions.}
      \label{fig:benchmark-002-boxplot}
    \end{subfigure}
    ~
    \begin{subfigure}[b]{0.45\linewidth}
      \includegraphics[width=\linewidth,valign=t,keepaspectratio]{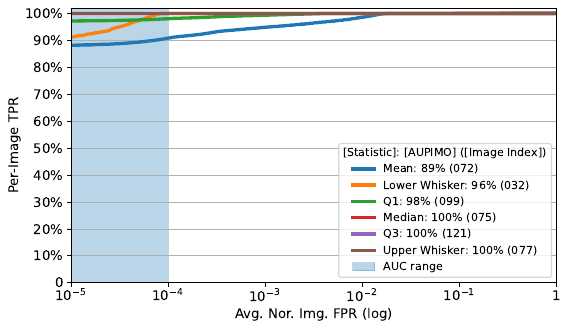}
      \caption{PIMO curves.}
      \label{fig:benchmark-002-pimo-curves}
    \end{subfigure}
    \\  \vspace{2mm}
    \begin{subfigure}[b]{\linewidth}
    
      \begin{minipage}{\linewidth}
        \centering
        \includegraphics[width=.3\linewidth,valign=t,keepaspectratio]{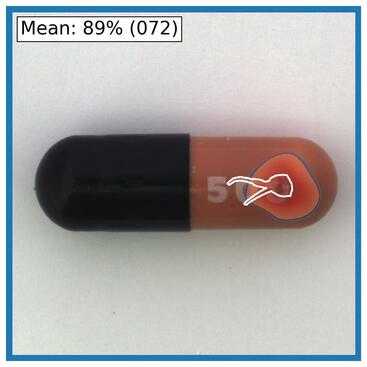}
        \includegraphics[width=.3\linewidth,valign=t,keepaspectratio]{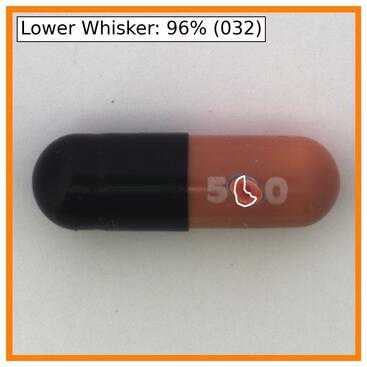}
        \includegraphics[width=.3\linewidth,valign=t,keepaspectratio]{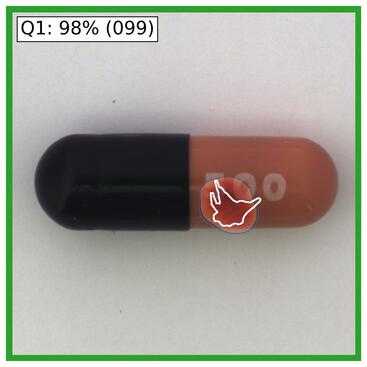}
      \end{minipage}
      \\
      \begin{minipage}{\linewidth}
        \centering
        \includegraphics[width=.3\linewidth,valign=t,keepaspectratio]{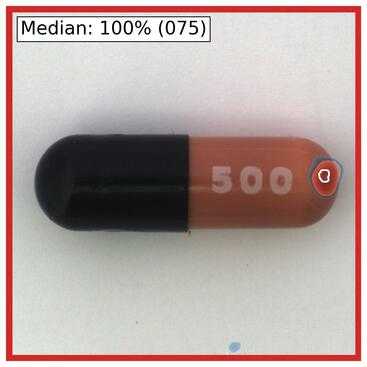}
        \includegraphics[width=.3\linewidth,valign=t,keepaspectratio]{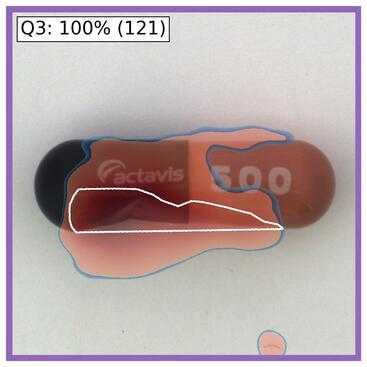}
        \includegraphics[width=.3\linewidth,valign=t,keepaspectratio]{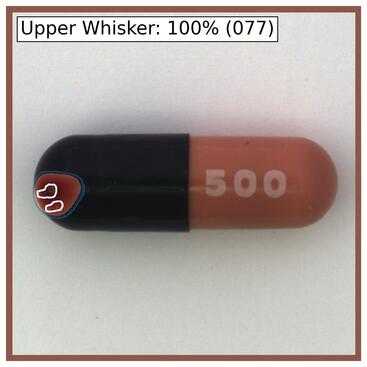}
      \end{minipage}
      \caption{
        Heatmaps.
        Images selected according to AUPIMO's statistics.
        Statistic and image index annotated on upper left corner.
      }
      \label{fig:benchmark-002-heatmap}
    \end{subfigure}
    \caption{
      Benchmark on MVTec AD / Capsule.
      PIMO curves and heatmaps are from SimpleNet WR50.
      132 images (023 normal, 109 anomalous).
    }
    \label{fig:benchmark-002}
\end{figure}

\begin{figure}[ht]
    \centering
    \begin{subfigure}[b]{\linewidth}
      \includegraphics[width=\linewidth,valign=t,keepaspectratio]{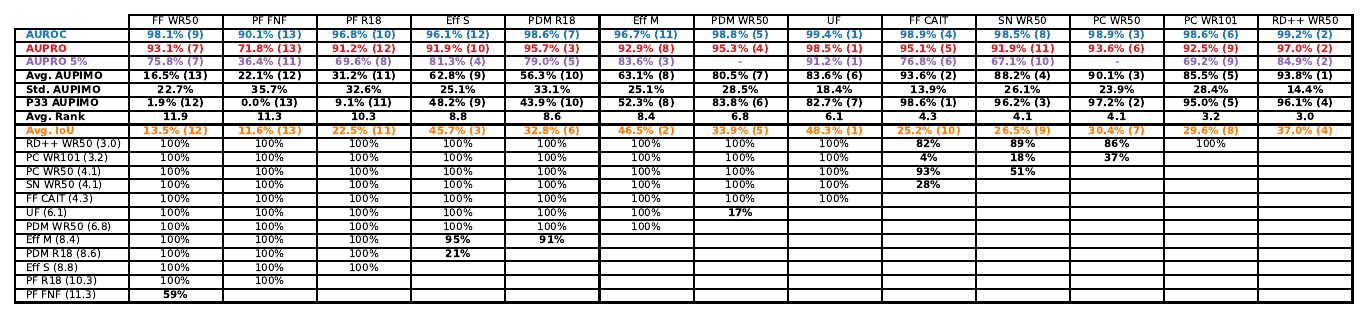}
      \caption{Statistics and pairwise statistical tests.}
      \label{fig:benchmark-003-table}
    \end{subfigure}
    \\ \vspace{2mm}
    \begin{subfigure}[b]{0.5\linewidth}
      \includegraphics[width=\linewidth,valign=t,keepaspectratio]{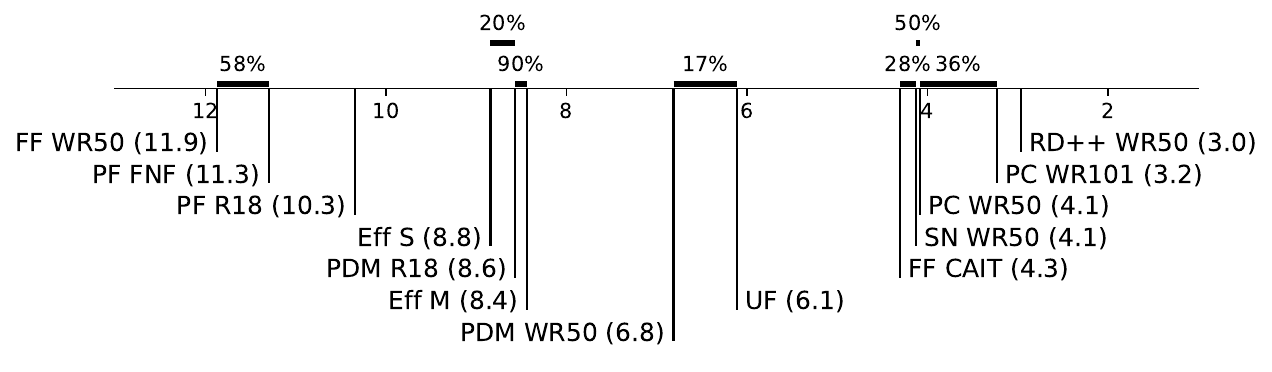}
      \caption{Average rank diagram.}
      \label{fig:benchmark-003-diagram}
    \end{subfigure}
    \\ \vspace{2mm}
    \begin{subfigure}[b]{0.45\linewidth}
      \includegraphics[width=\linewidth,valign=t,keepaspectratio]{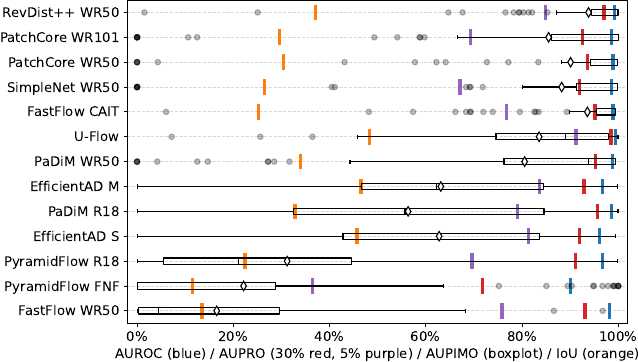}
      \caption{Score distributions.}
      \label{fig:benchmark-003-boxplot}
    \end{subfigure}
    ~
    \begin{subfigure}[b]{0.45\linewidth}
      \includegraphics[width=\linewidth,valign=t,keepaspectratio]{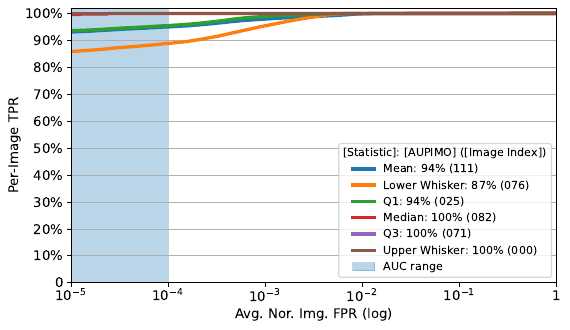}
      \caption{PIMO curves.}
      \label{fig:benchmark-003-pimo-curves}
    \end{subfigure}
    \\  \vspace{2mm}
    \begin{subfigure}[b]{\linewidth}
    
      \begin{minipage}{\linewidth}
        \centering
        \includegraphics[width=.3\linewidth,valign=t,keepaspectratio]{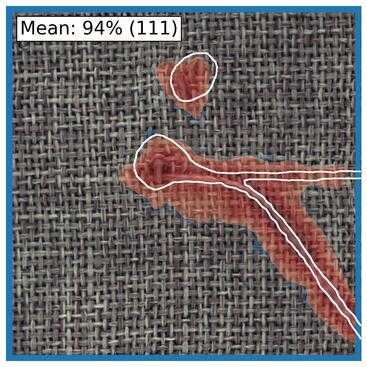}
        \includegraphics[width=.3\linewidth,valign=t,keepaspectratio]{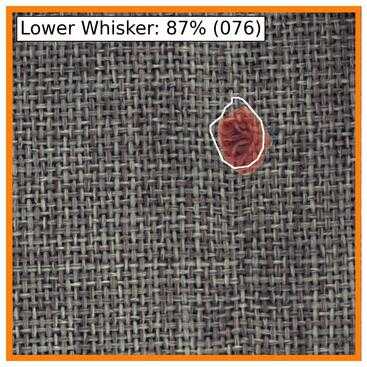}
        \includegraphics[width=.3\linewidth,valign=t,keepaspectratio]{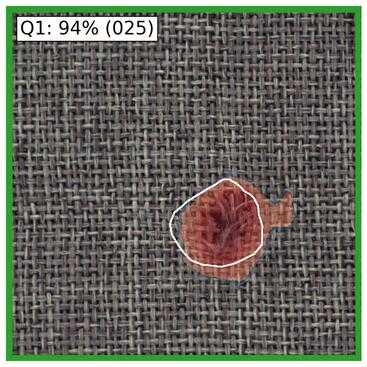}
      \end{minipage}
      \\
      \begin{minipage}{\linewidth}
        \centering
        \includegraphics[width=.3\linewidth,valign=t,keepaspectratio]{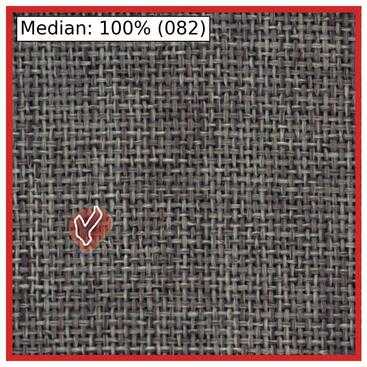}
        \includegraphics[width=.3\linewidth,valign=t,keepaspectratio]{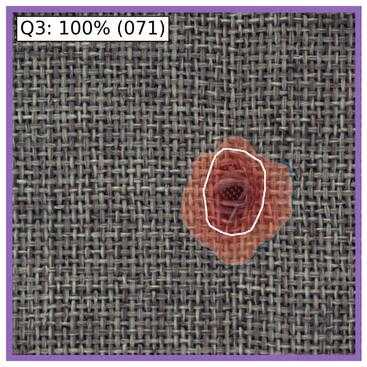}
        \includegraphics[width=.3\linewidth,valign=t,keepaspectratio]{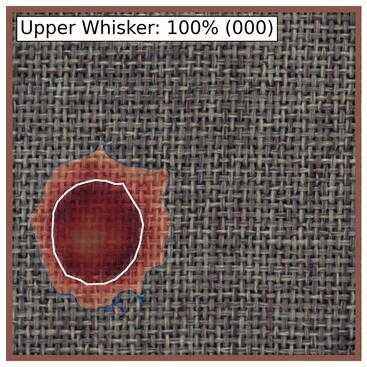}
      \end{minipage}
      \caption{
        Heatmaps.
        Images selected according to AUPIMO's statistics.
        Statistic and image index annotated on upper left corner.
      }
      \label{fig:benchmark-003-heatmap}
    \end{subfigure}
    \caption{
      Benchmark on MVTec AD / Carpet.
      PIMO curves and heatmaps are from RevDist++ WR50.
      117 images (028 normal, 089 anomalous).
    }
    \label{fig:benchmark-003}
\end{figure}

\begin{figure}[ht]
    \centering
    \begin{subfigure}[b]{\linewidth}
      \includegraphics[width=\linewidth,valign=t,keepaspectratio]{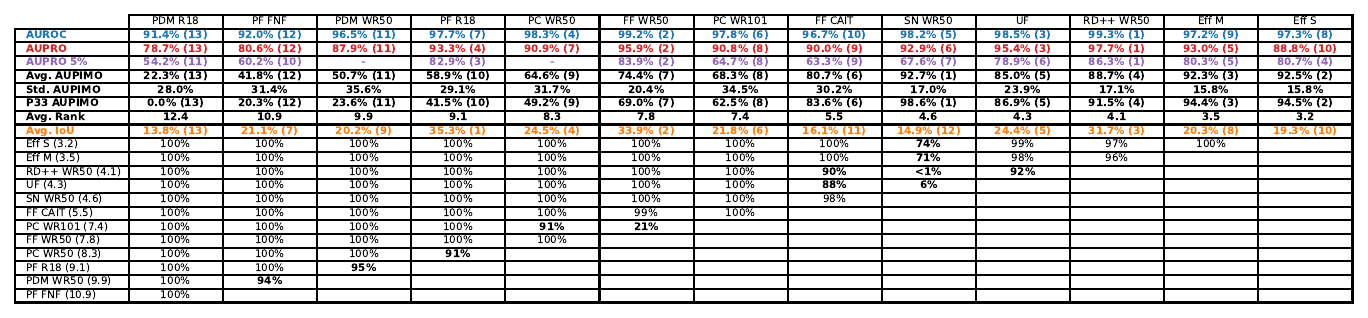}
      \caption{Statistics and pairwise statistical tests.}
      \label{fig:benchmark-004-table}
    \end{subfigure}
    \\ \vspace{2mm}
    \begin{subfigure}[b]{0.5\linewidth}
      \includegraphics[width=\linewidth,valign=t,keepaspectratio]{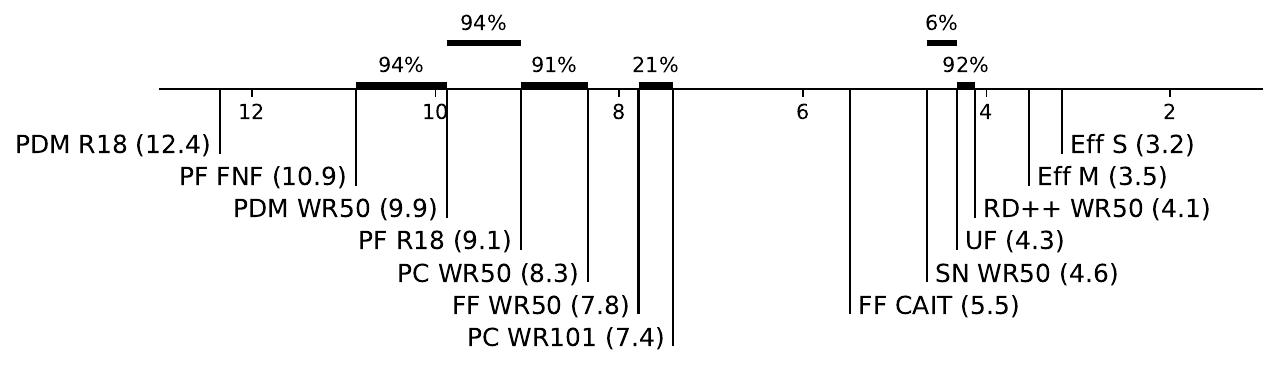}
      \caption{Average rank diagram.}
      \label{fig:benchmark-004-diagram}
    \end{subfigure}
    \\ \vspace{2mm}
    \begin{subfigure}[b]{0.45\linewidth}
      \includegraphics[width=\linewidth,valign=t,keepaspectratio]{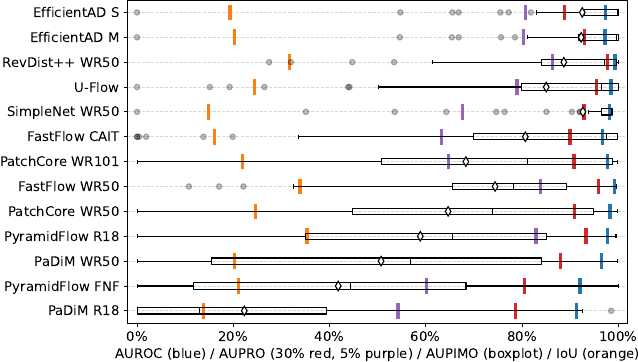}
      \caption{Score distributions.}
      \label{fig:benchmark-004-boxplot}
    \end{subfigure}
    ~
    \begin{subfigure}[b]{0.45\linewidth}
      \includegraphics[width=\linewidth,valign=t,keepaspectratio]{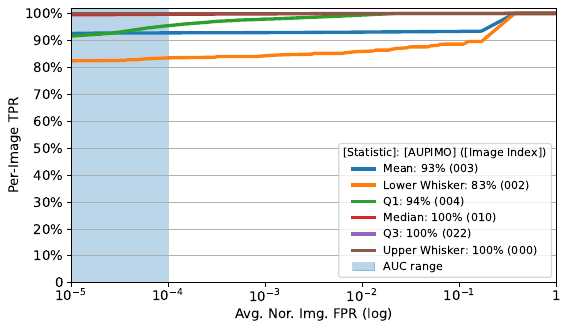}
      \caption{PIMO curves.}
      \label{fig:benchmark-004-pimo-curves}
    \end{subfigure}
    \\  \vspace{2mm}
    \begin{subfigure}[b]{\linewidth}
    
      \begin{minipage}{\linewidth}
        \centering
        \includegraphics[width=.3\linewidth,valign=t,keepaspectratio]{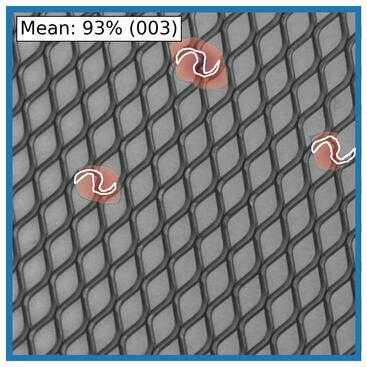}
        \includegraphics[width=.3\linewidth,valign=t,keepaspectratio]{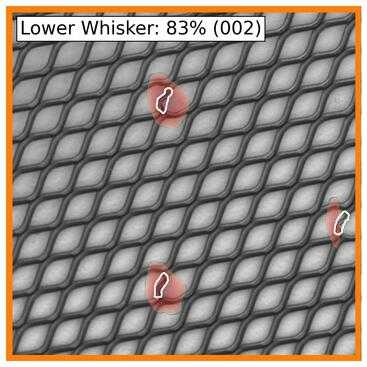}
        \includegraphics[width=.3\linewidth,valign=t,keepaspectratio]{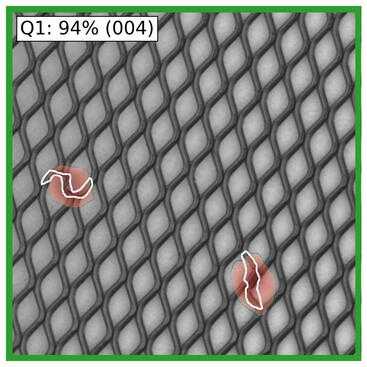}
      \end{minipage}
      \\
      \begin{minipage}{\linewidth}
        \centering
        \includegraphics[width=.3\linewidth,valign=t,keepaspectratio]{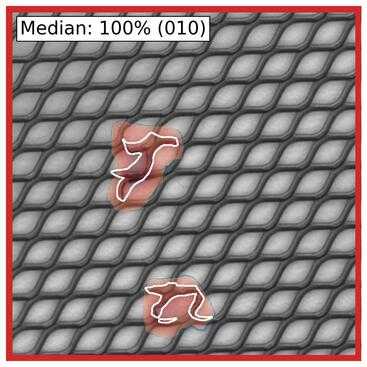}
        \includegraphics[width=.3\linewidth,valign=t,keepaspectratio]{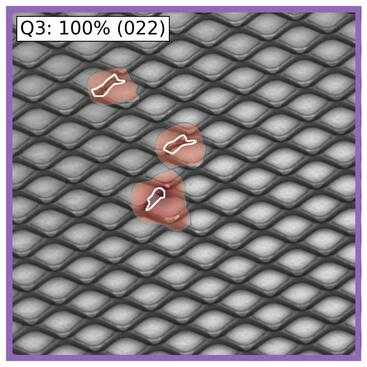}
        \includegraphics[width=.3\linewidth,valign=t,keepaspectratio]{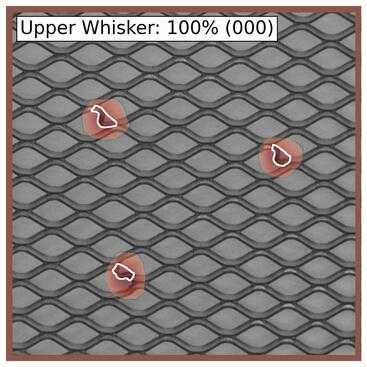}
      \end{minipage}
      \caption{
        Heatmaps.
        Images selected according to AUPIMO's statistics.
        Statistic and image index annotated on upper left corner.
      }
      \label{fig:benchmark-004-heatmap}
    \end{subfigure}
    \caption{
      Benchmark on MVTec AD / Grid.
      PIMO curves and heatmaps are from EfficientAD S.
      078 images (021 normal, 057 anomalous).
    }
    \label{fig:benchmark-004}
\end{figure}

\begin{figure}[ht]
    \centering
    \begin{subfigure}[b]{\linewidth}
      \includegraphics[width=\linewidth,valign=t,keepaspectratio]{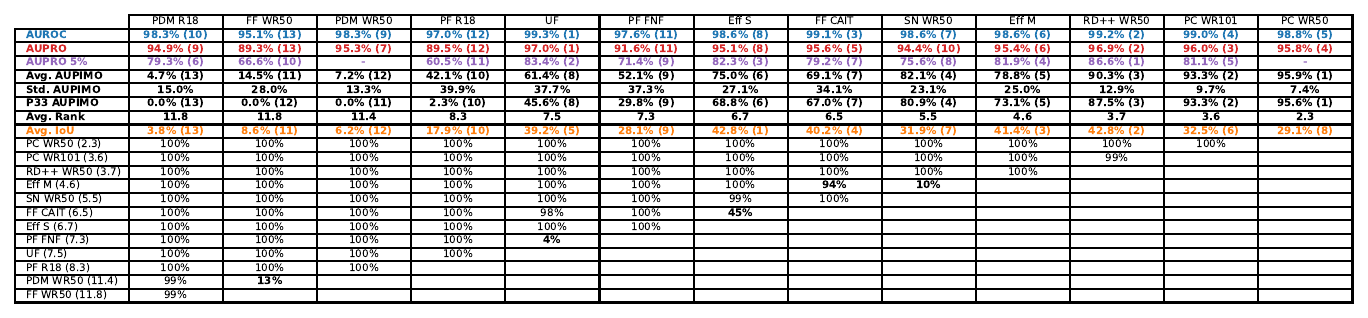}
      \caption{Statistics and pairwise statistical tests.}
      \label{fig:benchmark-005-table}
    \end{subfigure}
    \\ \vspace{2mm}
    \begin{subfigure}[b]{0.5\linewidth}
      \includegraphics[width=\linewidth,valign=t,keepaspectratio]{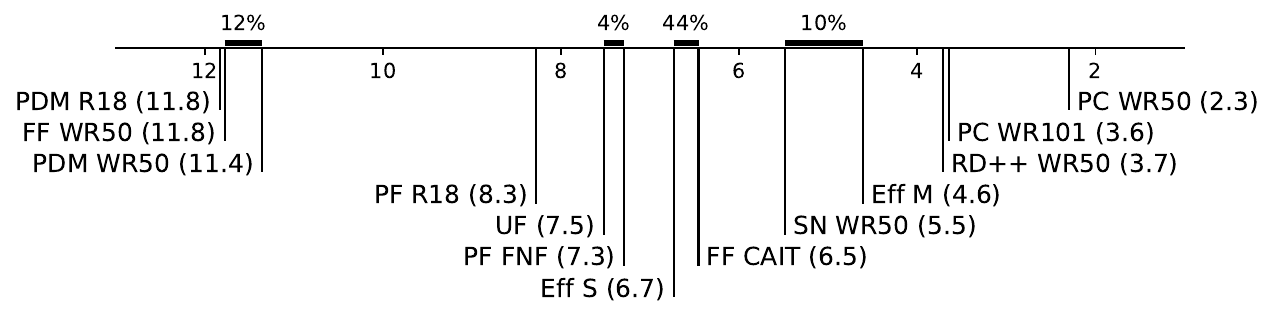}
      \caption{Average rank diagram.}
      \label{fig:benchmark-005-diagram}
    \end{subfigure}
    \\ \vspace{2mm}
    \begin{subfigure}[b]{0.45\linewidth}
      \includegraphics[width=\linewidth,valign=t,keepaspectratio]{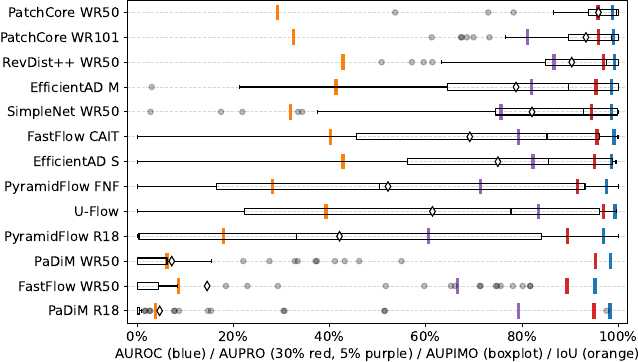}
      \caption{Score distributions.}
      \label{fig:benchmark-005-boxplot}
    \end{subfigure}
    ~
    \begin{subfigure}[b]{0.45\linewidth}
      \includegraphics[width=\linewidth,valign=t,keepaspectratio]{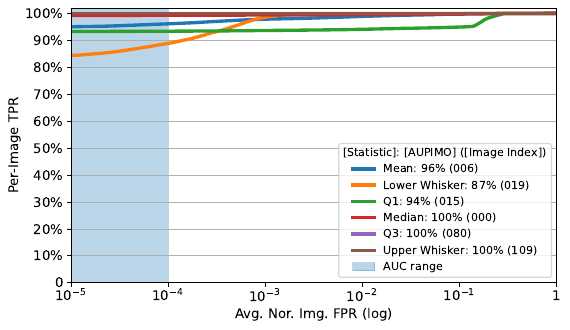}
      \caption{PIMO curves.}
      \label{fig:benchmark-005-pimo-curves}
    \end{subfigure}
    \\  \vspace{2mm}
    \begin{subfigure}[b]{\linewidth}
    
      \begin{minipage}{\linewidth}
        \centering
        \includegraphics[width=.3\linewidth,valign=t,keepaspectratio]{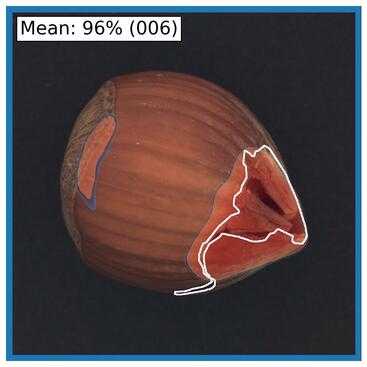}
        \includegraphics[width=.3\linewidth,valign=t,keepaspectratio]{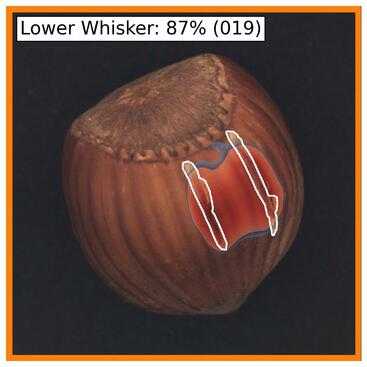}
        \includegraphics[width=.3\linewidth,valign=t,keepaspectratio]{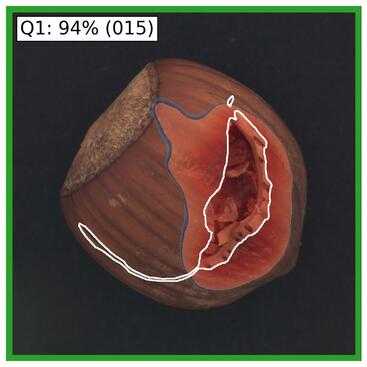}
      \end{minipage}
      \\
      \begin{minipage}{\linewidth}
        \centering
        \includegraphics[width=.3\linewidth,valign=t,keepaspectratio]{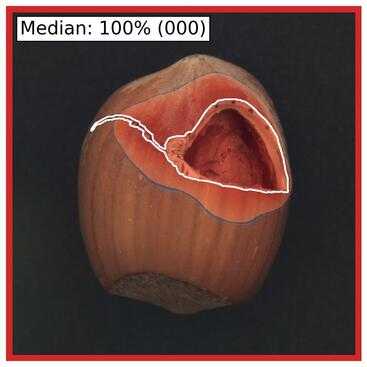}
        \includegraphics[width=.3\linewidth,valign=t,keepaspectratio]{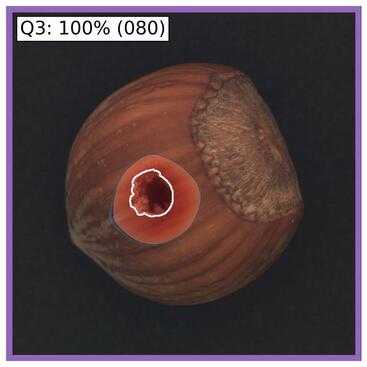}
        \includegraphics[width=.3\linewidth,valign=t,keepaspectratio]{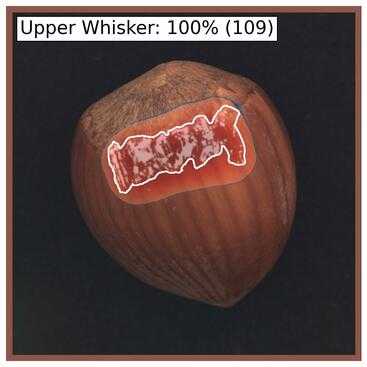}
      \end{minipage}
      \caption{
        Heatmaps.
        Images selected according to AUPIMO's statistics.
        Statistic and image index annotated on upper left corner.
      }
      \label{fig:benchmark-005-heatmap}
    \end{subfigure}
    \caption{
      Benchmark on MVTec AD / Hazelnut.
      PIMO curves and heatmaps are from PatchCore WR50.
      110 images (040 normal, 070 anomalous).
    }
    \label{fig:benchmark-005}
\end{figure}

\begin{figure}[ht]
    \centering
    \begin{subfigure}[b]{\linewidth}
      \includegraphics[width=\linewidth,valign=t,keepaspectratio]{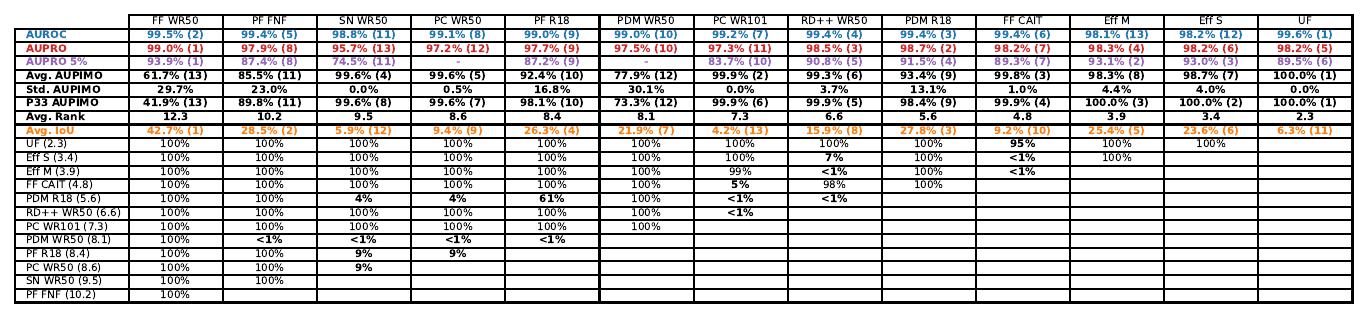}
      \caption{Statistics and pairwise statistical tests.}
      \label{fig:benchmark-006-table}
    \end{subfigure}
    \\ \vspace{2mm}
    \begin{subfigure}[b]{0.5\linewidth}
      \includegraphics[width=\linewidth,valign=t,keepaspectratio]{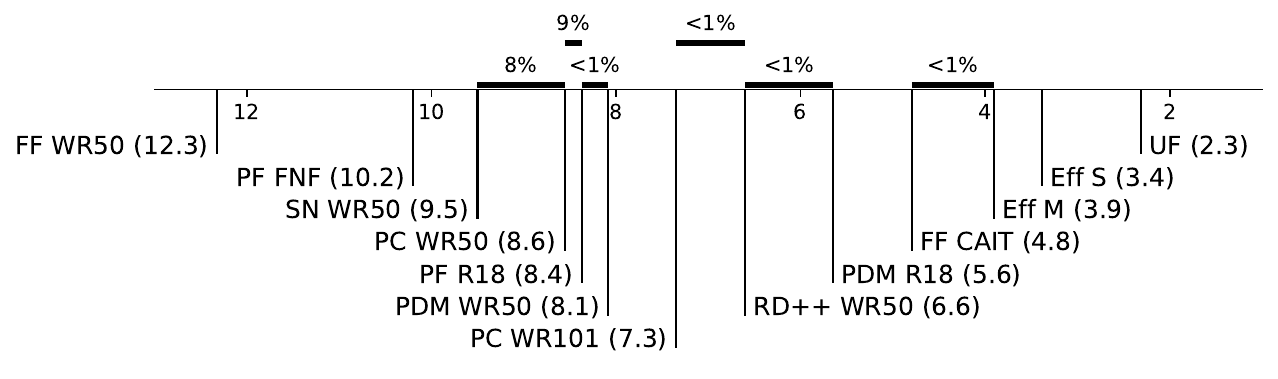}
      \caption{Average rank diagram.}
      \label{fig:benchmark-006-diagram}
    \end{subfigure}
    \\ \vspace{2mm}
    \begin{subfigure}[b]{0.45\linewidth}
      \includegraphics[width=\linewidth,valign=t,keepaspectratio]{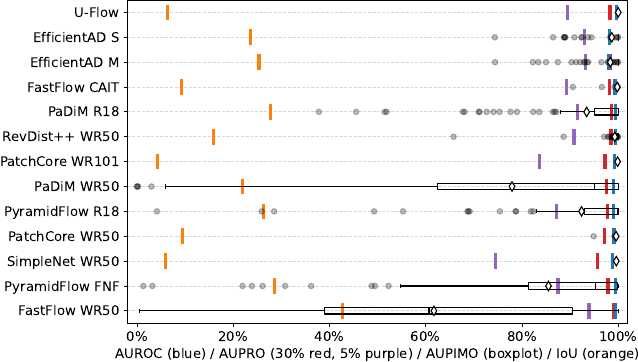}
      \caption{Score distributions.}
      \label{fig:benchmark-006-boxplot}
    \end{subfigure}
    ~
    \begin{subfigure}[b]{0.45\linewidth}
      \includegraphics[width=\linewidth,valign=t,keepaspectratio]{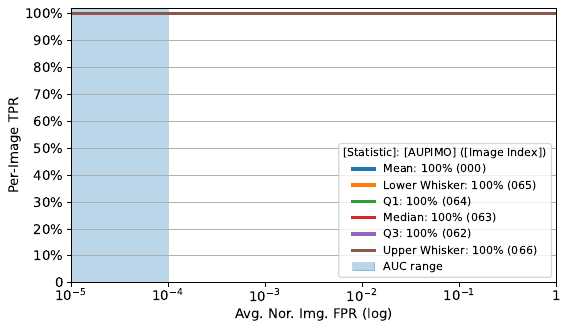}
      \caption{PIMO curves.}
      \label{fig:benchmark-006-pimo-curves}
    \end{subfigure}
    \\  \vspace{2mm}
    \begin{subfigure}[b]{\linewidth}
    
      \begin{minipage}{\linewidth}
        \centering
        \includegraphics[width=.3\linewidth,valign=t,keepaspectratio]{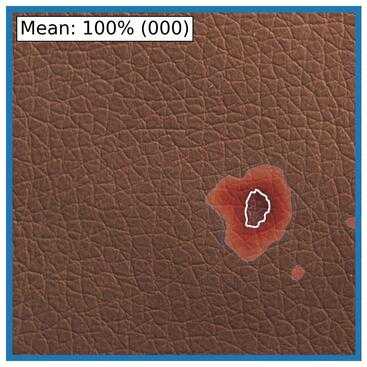}
        \includegraphics[width=.3\linewidth,valign=t,keepaspectratio]{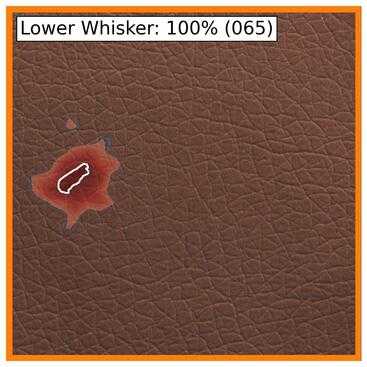}
        \includegraphics[width=.3\linewidth,valign=t,keepaspectratio]{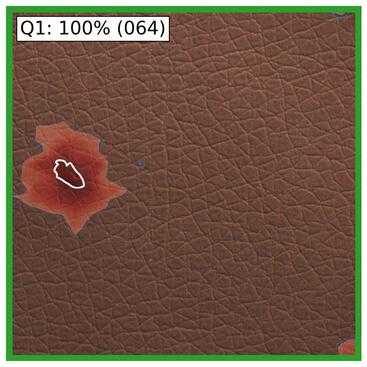}
      \end{minipage}
      \\
      \begin{minipage}{\linewidth}
        \centering
        \includegraphics[width=.3\linewidth,valign=t,keepaspectratio]{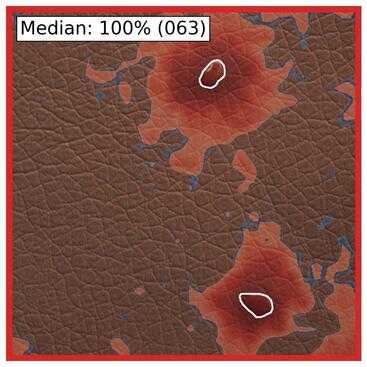}
        \includegraphics[width=.3\linewidth,valign=t,keepaspectratio]{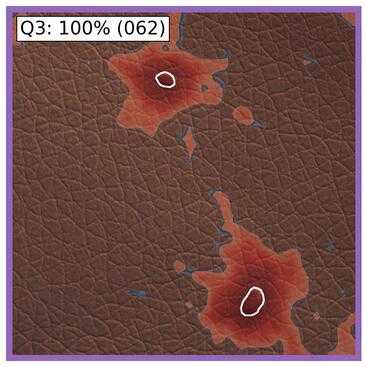}
        \includegraphics[width=.3\linewidth,valign=t,keepaspectratio]{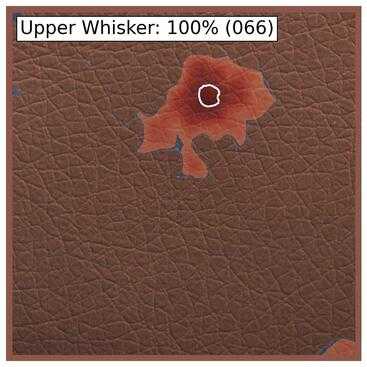}
      \end{minipage}
      \caption{
        Heatmaps.
        Images selected according to AUPIMO's statistics.
        Statistic and image index annotated on upper left corner.
      }
      \label{fig:benchmark-006-heatmap}
    \end{subfigure}
    \caption{
      Benchmark on MVTec AD / Leather.
      PIMO curves and heatmaps are from U-Flow.
      124 images (032 normal, 092 anomalous).
    }
    \label{fig:benchmark-006}
\end{figure}

\begin{figure}[ht]
    \centering
    \begin{subfigure}[b]{\linewidth}
      \includegraphics[width=\linewidth,valign=t,keepaspectratio]{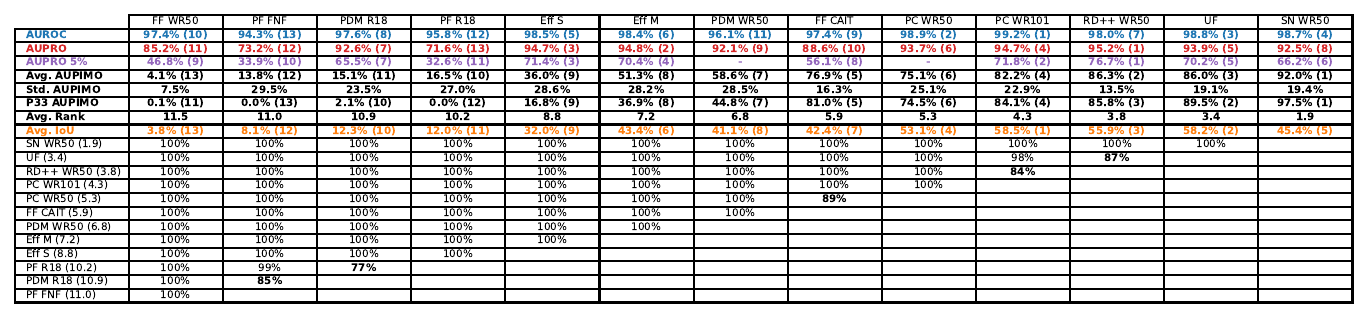}
      \caption{Statistics and pairwise statistical tests.}
      \label{fig:benchmark-007-table}
    \end{subfigure}
    \\ \vspace{2mm}
    \begin{subfigure}[b]{0.5\linewidth}
      \includegraphics[width=\linewidth,valign=t,keepaspectratio]{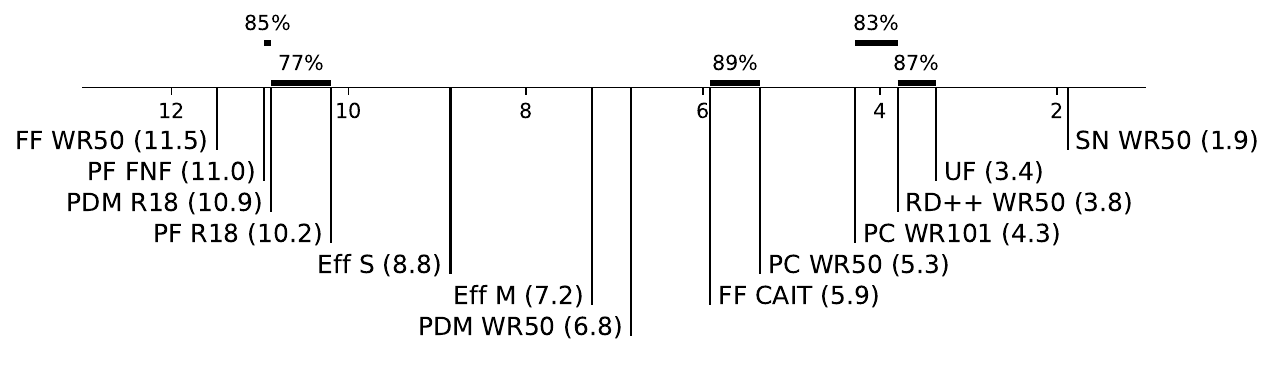}
      \caption{Average rank diagram.}
      \label{fig:benchmark-007-diagram}
    \end{subfigure}
    \\ \vspace{2mm}
    \begin{subfigure}[b]{0.45\linewidth}
      \includegraphics[width=\linewidth,valign=t,keepaspectratio]{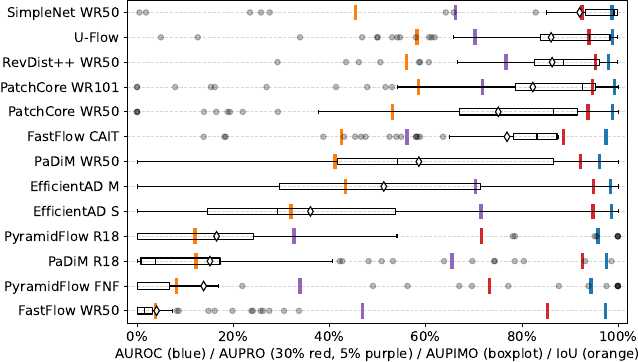}
      \caption{Score distributions.}
      \label{fig:benchmark-007-boxplot}
    \end{subfigure}
    ~
    \begin{subfigure}[b]{0.45\linewidth}
      \includegraphics[width=\linewidth,valign=t,keepaspectratio]{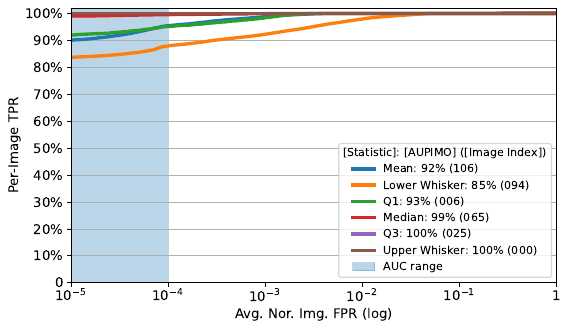}
      \caption{PIMO curves.}
      \label{fig:benchmark-007-pimo-curves}
    \end{subfigure}
    \\  \vspace{2mm}
    \begin{subfigure}[b]{\linewidth}
    
      \begin{minipage}{\linewidth}
        \centering
        \includegraphics[width=.3\linewidth,valign=t,keepaspectratio]{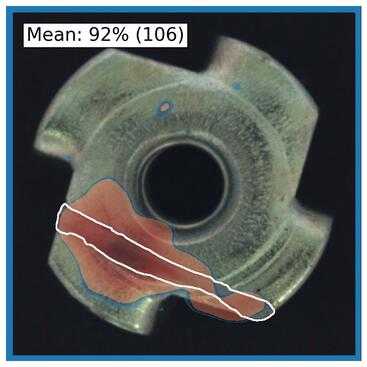}
        \includegraphics[width=.3\linewidth,valign=t,keepaspectratio]{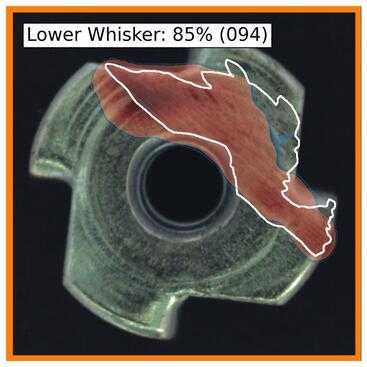}
        \includegraphics[width=.3\linewidth,valign=t,keepaspectratio]{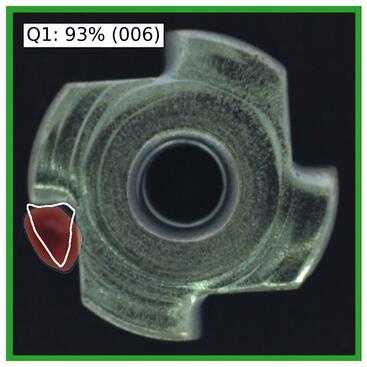}
      \end{minipage}
      \\
      \begin{minipage}{\linewidth}
        \centering
        \includegraphics[width=.3\linewidth,valign=t,keepaspectratio]{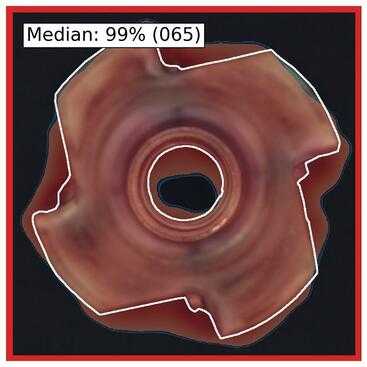}
        \includegraphics[width=.3\linewidth,valign=t,keepaspectratio]{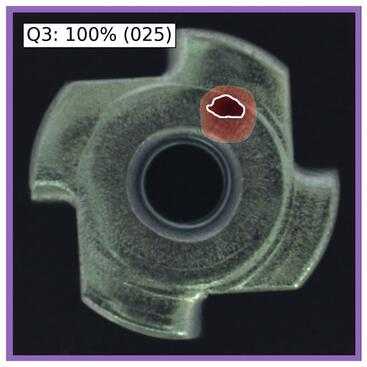}
        \includegraphics[width=.3\linewidth,valign=t,keepaspectratio]{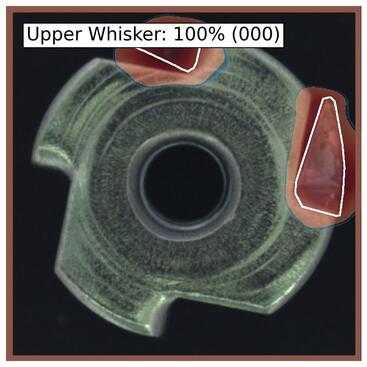}
      \end{minipage}
      \caption{
        Heatmaps.
        Images selected according to AUPIMO's statistics.
        Statistic and image index annotated on upper left corner.
      }
      \label{fig:benchmark-007-heatmap}
    \end{subfigure}
    \caption{
      Benchmark on MVTec AD / Metal Nut.
      PIMO curves and heatmaps are from SimpleNet WR50.
      115 images (022 normal, 093 anomalous).
    }
    \label{fig:benchmark-007}
\end{figure}

\begin{figure}[ht]
    \centering
    \begin{subfigure}[b]{\linewidth}
      \includegraphics[width=\linewidth,valign=t,keepaspectratio]{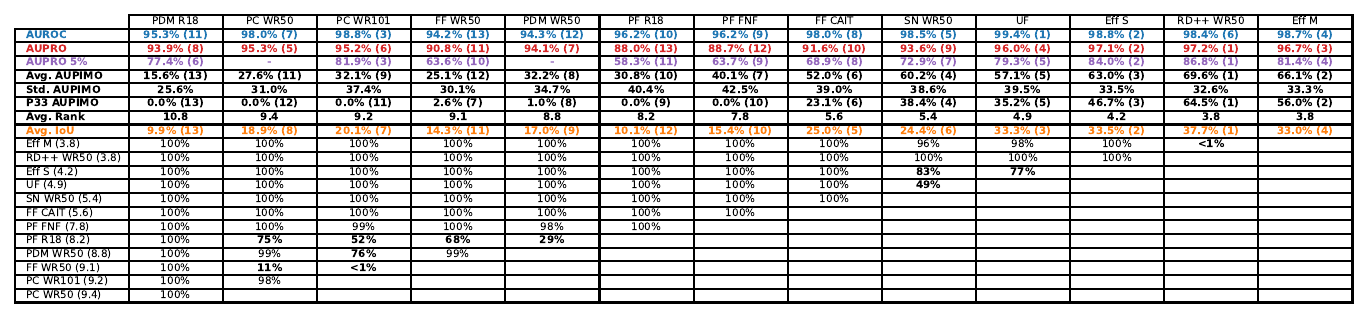}
      \caption{Statistics and pairwise statistical tests.}
      \label{fig:benchmark-008-table}
    \end{subfigure}
    \\ \vspace{2mm}
    \begin{subfigure}[b]{0.5\linewidth}
      \includegraphics[width=\linewidth,valign=t,keepaspectratio]{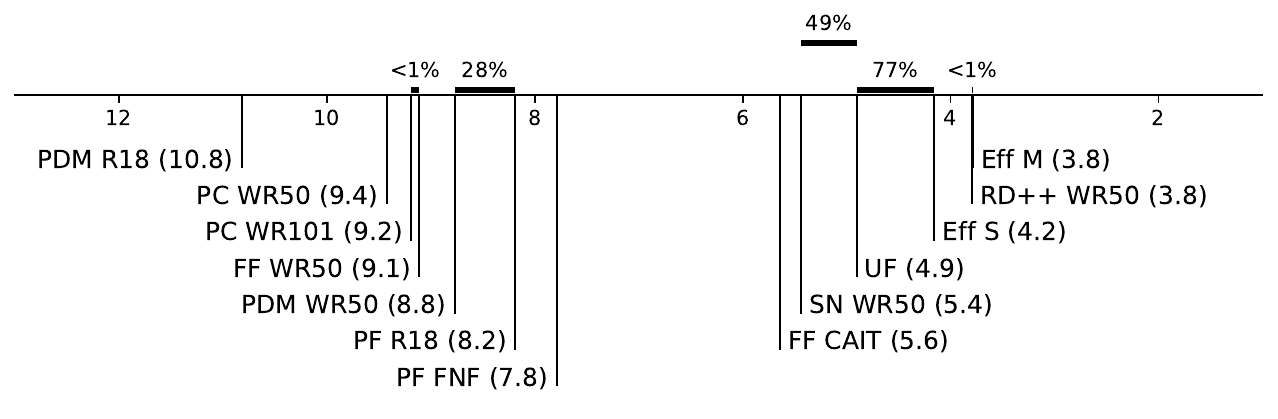}
      \caption{Average rank diagram.}
      \label{fig:benchmark-008-diagram}
    \end{subfigure}
    \\ \vspace{2mm}
    \begin{subfigure}[b]{0.45\linewidth}
      \includegraphics[width=\linewidth,valign=t,keepaspectratio]{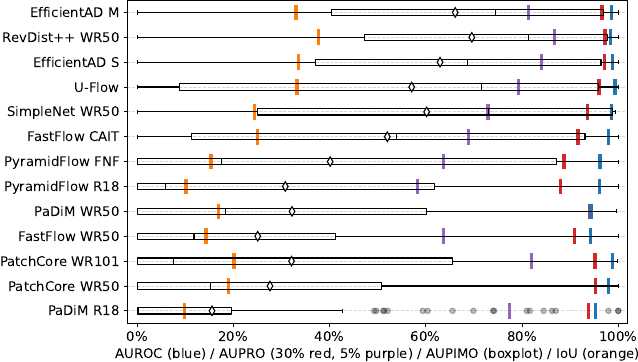}
      \caption{Score distributions.}
      \label{fig:benchmark-008-boxplot}
    \end{subfigure}
    ~
    \begin{subfigure}[b]{0.45\linewidth}
      \includegraphics[width=\linewidth,valign=t,keepaspectratio]{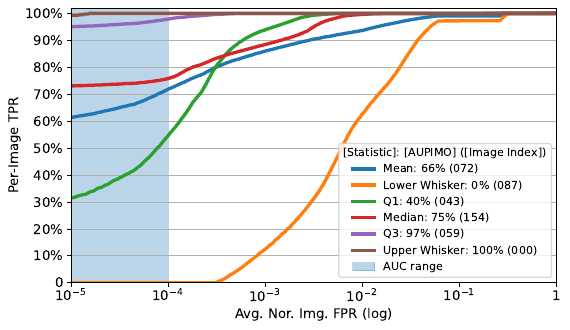}
      \caption{PIMO curves.}
      \label{fig:benchmark-008-pimo-curves}
    \end{subfigure}
    \\  \vspace{2mm}
    \begin{subfigure}[b]{\linewidth}
    
      \begin{minipage}{\linewidth}
        \centering
        \includegraphics[width=.3\linewidth,valign=t,keepaspectratio]{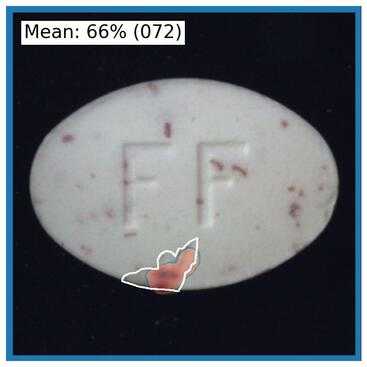}
        \includegraphics[width=.3\linewidth,valign=t,keepaspectratio]{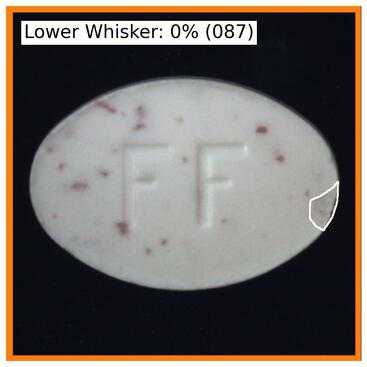}
        \includegraphics[width=.3\linewidth,valign=t,keepaspectratio]{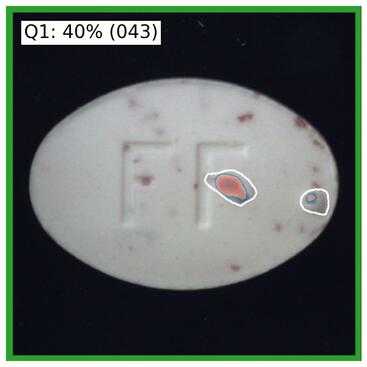}
      \end{minipage}
      \\
      \begin{minipage}{\linewidth}
        \centering
        \includegraphics[width=.3\linewidth,valign=t,keepaspectratio]{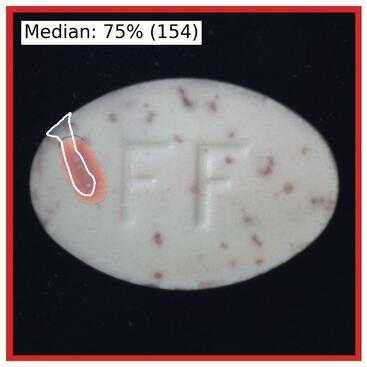}
        \includegraphics[width=.3\linewidth,valign=t,keepaspectratio]{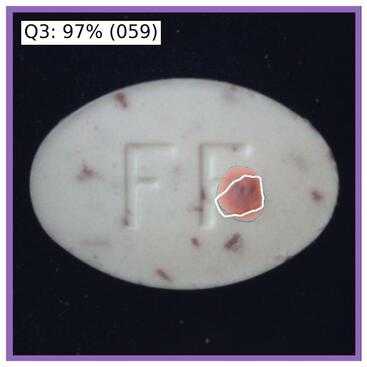}
        \includegraphics[width=.3\linewidth,valign=t,keepaspectratio]{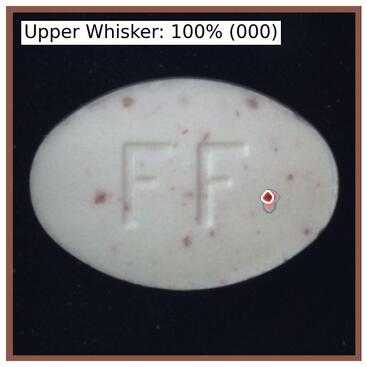}
      \end{minipage}
      \caption{
        Heatmaps.
        Images selected according to AUPIMO's statistics.
        Statistic and image index annotated on upper left corner.
      }
      \label{fig:benchmark-008-heatmap}
    \end{subfigure}
    \caption{
      Benchmark on MVTec AD / Pill.
      PIMO curves and heatmaps are from EfficientAD M.
      167 images (026 normal, 141 anomalous).
    }
    \label{fig:benchmark-008}
\end{figure}

\begin{figure}[ht]
    \centering
    \begin{subfigure}[b]{\linewidth}
      \includegraphics[width=\linewidth,valign=t,keepaspectratio]{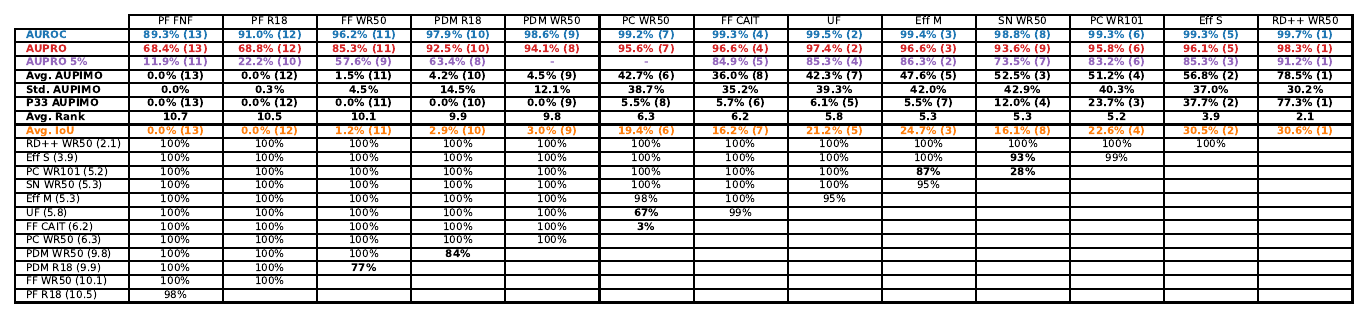}
      \caption{Statistics and pairwise statistical tests.}
      \label{fig:benchmark-009-table}
    \end{subfigure}
    \\ \vspace{2mm}
    \begin{subfigure}[b]{0.5\linewidth}
      \includegraphics[width=\linewidth,valign=t,keepaspectratio]{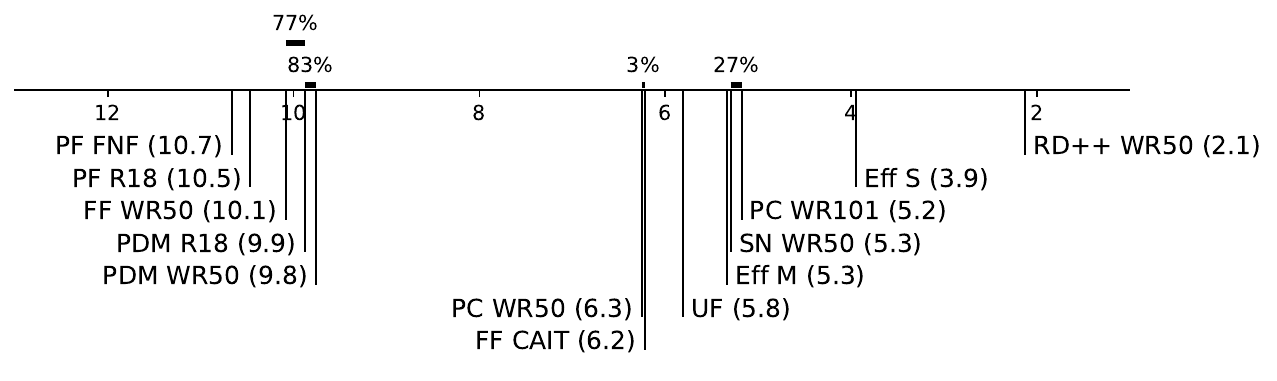}
      \caption{Average rank diagram.}
      \label{fig:benchmark-009-diagram}
    \end{subfigure}
    \\ \vspace{2mm}
    \begin{subfigure}[b]{0.45\linewidth}
      \includegraphics[width=\linewidth,valign=t,keepaspectratio]{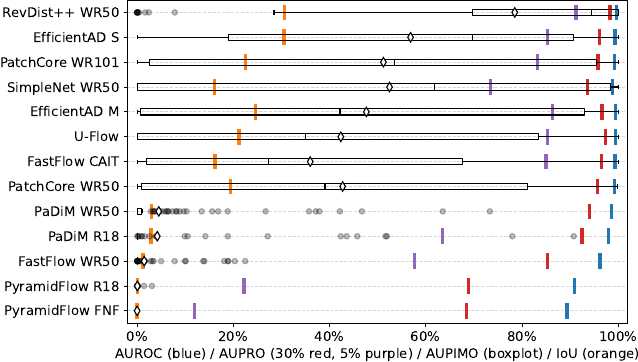}
      \caption{Score distributions.}
      \label{fig:benchmark-009-boxplot}
    \end{subfigure}
    ~
    \begin{subfigure}[b]{0.45\linewidth}
      \includegraphics[width=\linewidth,valign=t,keepaspectratio]{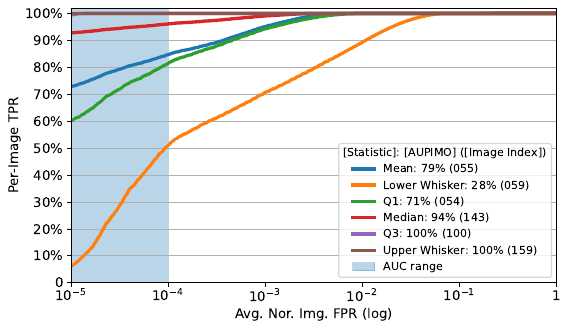}
      \caption{PIMO curves.}
      \label{fig:benchmark-009-pimo-curves}
    \end{subfigure}
    \\  \vspace{2mm}
    \begin{subfigure}[b]{\linewidth}
    
      \begin{minipage}{\linewidth}
        \centering
        \includegraphics[width=.3\linewidth,valign=t,keepaspectratio]{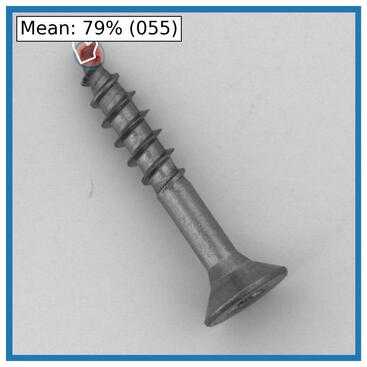}
        \includegraphics[width=.3\linewidth,valign=t,keepaspectratio]{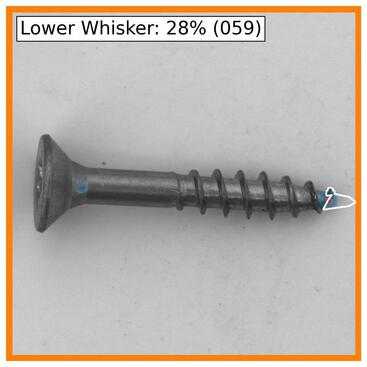}
        \includegraphics[width=.3\linewidth,valign=t,keepaspectratio]{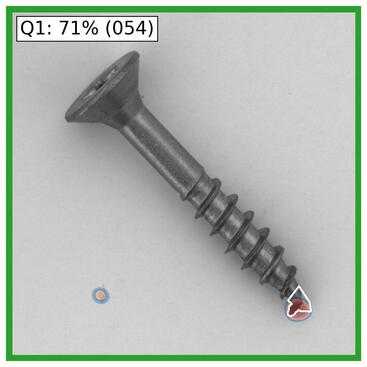}
      \end{minipage}
      \\
      \begin{minipage}{\linewidth}
        \centering
        \includegraphics[width=.3\linewidth,valign=t,keepaspectratio]{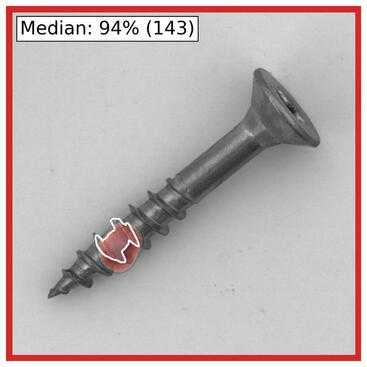}
        \includegraphics[width=.3\linewidth,valign=t,keepaspectratio]{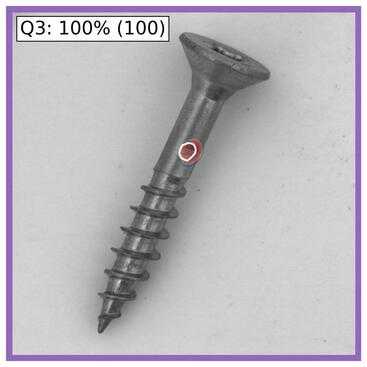}
        \includegraphics[width=.3\linewidth,valign=t,keepaspectratio]{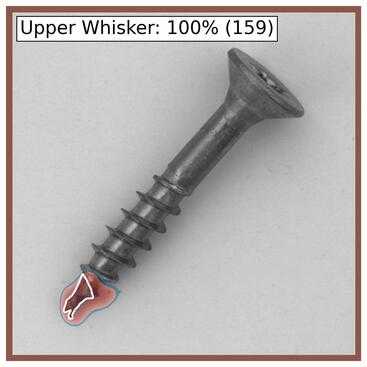}
      \end{minipage}
      \caption{
        Heatmaps.
        Images selected according to AUPIMO's statistics.
        Statistic and image index annotated on upper left corner.
      }
      \label{fig:benchmark-009-heatmap}
    \end{subfigure}
    \caption{
      Benchmark on MVTec AD / Screw.
      PIMO curves and heatmaps are from RevDist++ WR50.
      160 images (041 normal, 119 anomalous).
    }
    \label{fig:benchmark-009}
\end{figure}

\begin{figure}[ht]
    \centering
    \begin{subfigure}[b]{\linewidth}
      \includegraphics[width=\linewidth,valign=t,keepaspectratio]{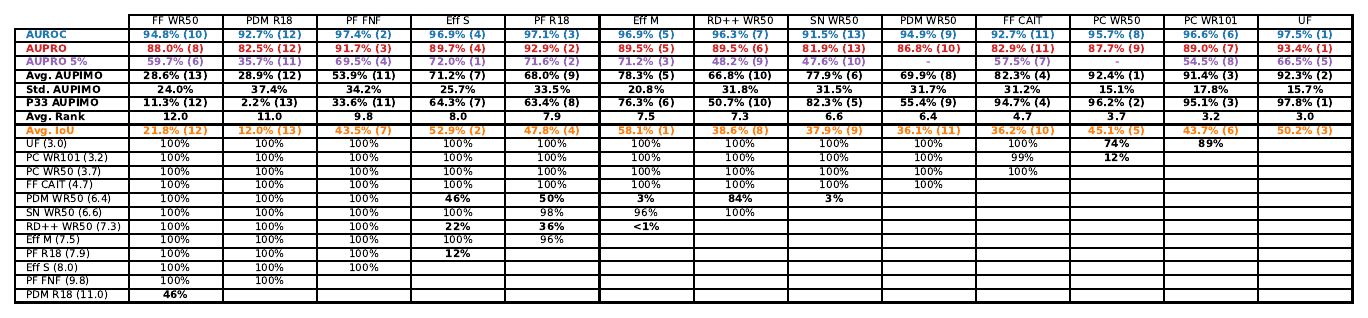}
      \caption{Statistics and pairwise statistical tests.}
      \label{fig:benchmark-010-table}
    \end{subfigure}
    \\ \vspace{2mm}
    \begin{subfigure}[b]{0.5\linewidth}
      \includegraphics[width=\linewidth,valign=t,keepaspectratio]{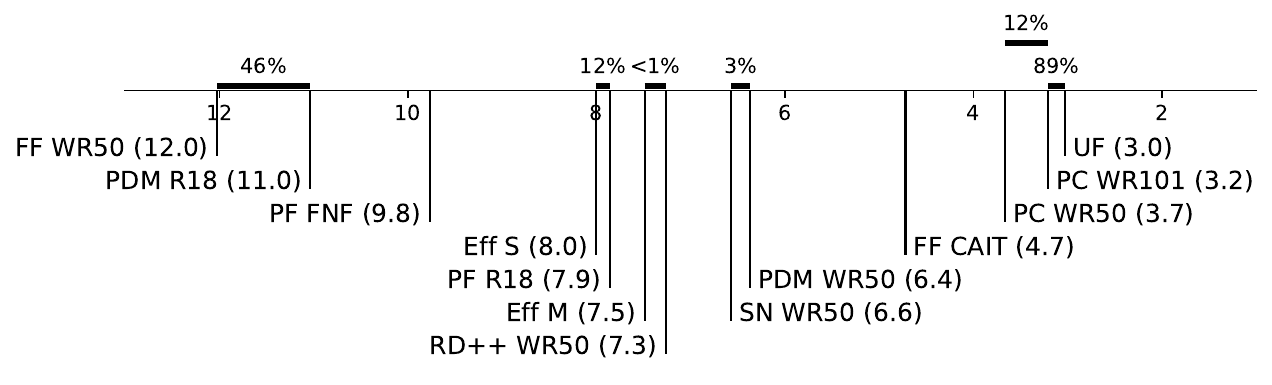}
      \caption{Average rank diagram.}
      \label{fig:benchmark-010-diagram}
    \end{subfigure}
    \\ \vspace{2mm}
    \begin{subfigure}[b]{0.45\linewidth}
      \includegraphics[width=\linewidth,valign=t,keepaspectratio]{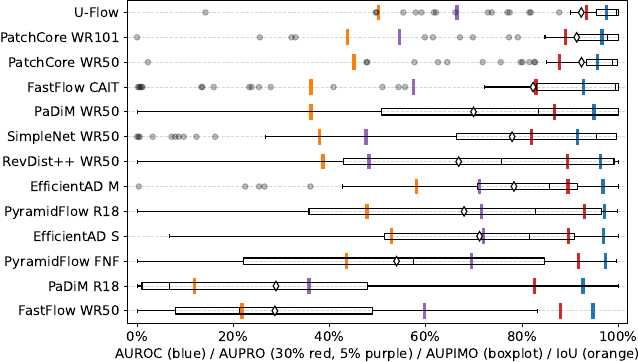}
      \caption{Score distributions.}
      \label{fig:benchmark-010-boxplot}
    \end{subfigure}
    ~
    \begin{subfigure}[b]{0.45\linewidth}
      \includegraphics[width=\linewidth,valign=t,keepaspectratio]{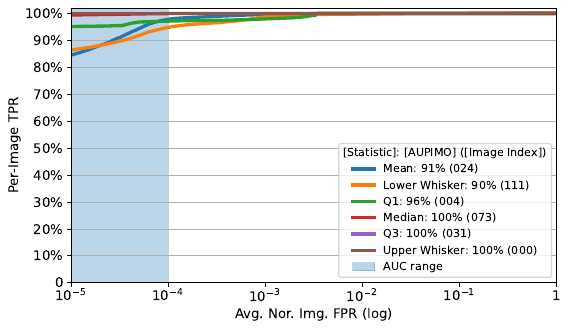}
      \caption{PIMO curves.}
      \label{fig:benchmark-010-pimo-curves}
    \end{subfigure}
    \\  \vspace{2mm}
    \begin{subfigure}[b]{\linewidth}
    
      \begin{minipage}{\linewidth}
        \centering
        \includegraphics[width=.3\linewidth,valign=t,keepaspectratio]{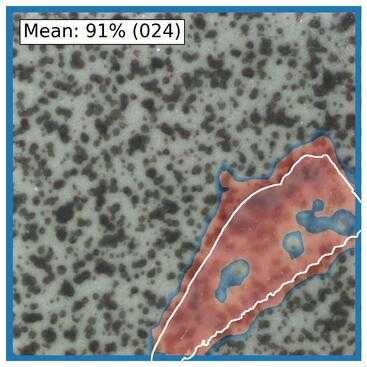}
        \includegraphics[width=.3\linewidth,valign=t,keepaspectratio]{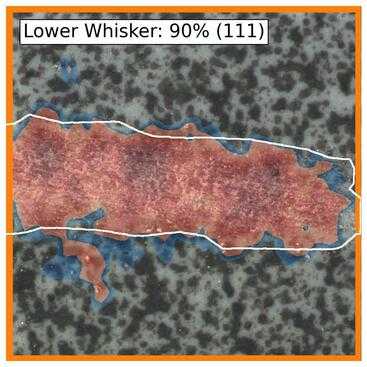}
        \includegraphics[width=.3\linewidth,valign=t,keepaspectratio]{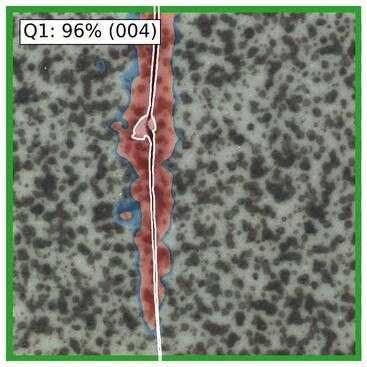}
      \end{minipage}
      \\
      \begin{minipage}{\linewidth}
        \centering
        \includegraphics[width=.3\linewidth,valign=t,keepaspectratio]{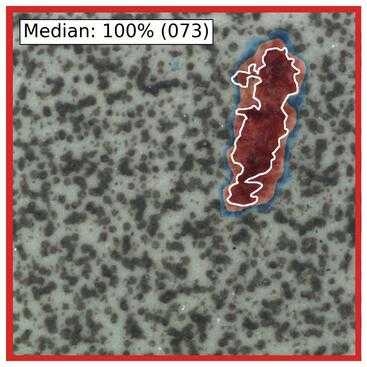}
        \includegraphics[width=.3\linewidth,valign=t,keepaspectratio]{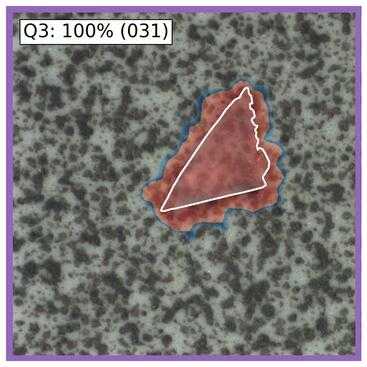}
        \includegraphics[width=.3\linewidth,valign=t,keepaspectratio]{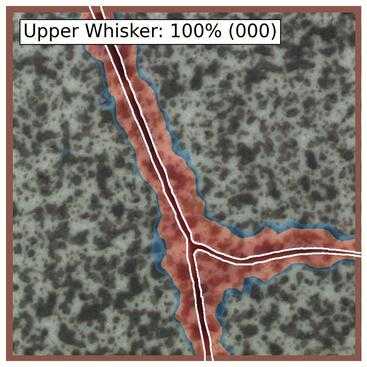}
      \end{minipage}
      \caption{
        Heatmaps.
        Images selected according to AUPIMO's statistics.
        Statistic and image index annotated on upper left corner.
      }
      \label{fig:benchmark-010-heatmap}
    \end{subfigure}
    \caption{
      Benchmark on MVTec AD / Tile.
      PIMO curves and heatmaps are from U-Flow.
      117 images (033 normal, 084 anomalous).
    }
    \label{fig:benchmark-010}
\end{figure}

\begin{figure}[ht]
    \centering
    \begin{subfigure}[b]{\linewidth}
      \includegraphics[width=\linewidth,valign=t,keepaspectratio]{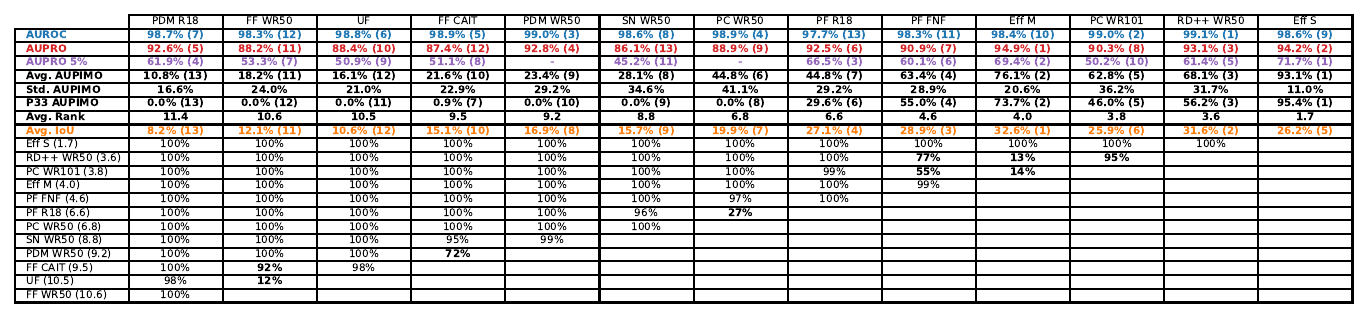}
      \caption{Statistics and pairwise statistical tests.}
      \label{fig:benchmark-011-table}
    \end{subfigure}
    \\ \vspace{2mm}
    \begin{subfigure}[b]{0.5\linewidth}
      \includegraphics[width=\linewidth,valign=t,keepaspectratio]{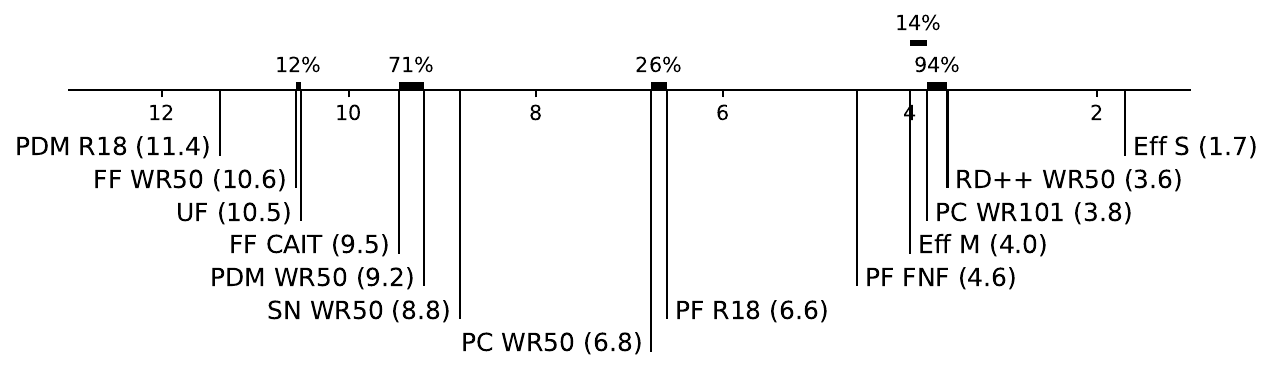}
      \caption{Average rank diagram.}
      \label{fig:benchmark-011-diagram}
    \end{subfigure}
    \\ \vspace{2mm}
    \begin{subfigure}[b]{0.45\linewidth}
      \includegraphics[width=\linewidth,valign=t,keepaspectratio]{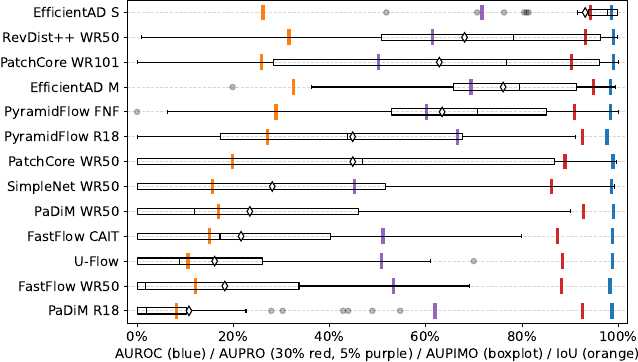}
      \caption{Score distributions.}
      \label{fig:benchmark-011-boxplot}
    \end{subfigure}
    ~
    \begin{subfigure}[b]{0.45\linewidth}
      \includegraphics[width=\linewidth,valign=t,keepaspectratio]{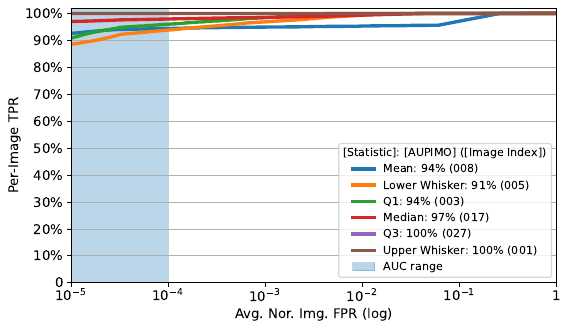}
      \caption{PIMO curves.}
      \label{fig:benchmark-011-pimo-curves}
    \end{subfigure}
    \\  \vspace{2mm}
    \begin{subfigure}[b]{\linewidth}
    
      \begin{minipage}{\linewidth}
        \centering
        \includegraphics[width=.3\linewidth,valign=t,keepaspectratio]{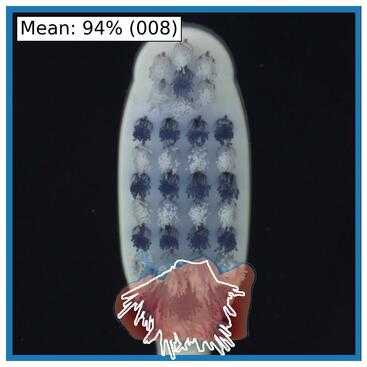}
        \includegraphics[width=.3\linewidth,valign=t,keepaspectratio]{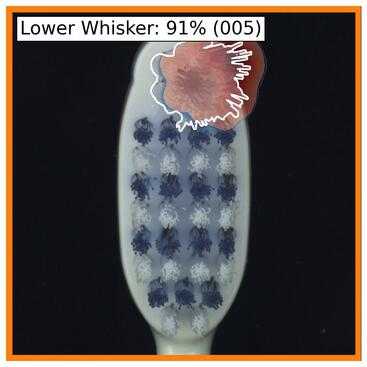}
        \includegraphics[width=.3\linewidth,valign=t,keepaspectratio]{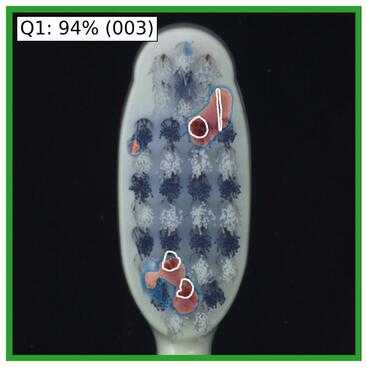}
      \end{minipage}
      \\
      \begin{minipage}{\linewidth}
        \centering
        \includegraphics[width=.3\linewidth,valign=t,keepaspectratio]{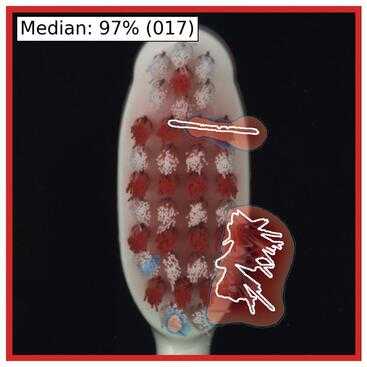}
        \includegraphics[width=.3\linewidth,valign=t,keepaspectratio]{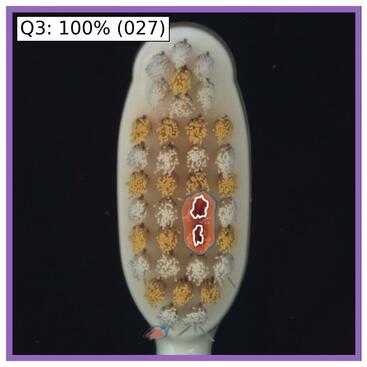}
        \includegraphics[width=.3\linewidth,valign=t,keepaspectratio]{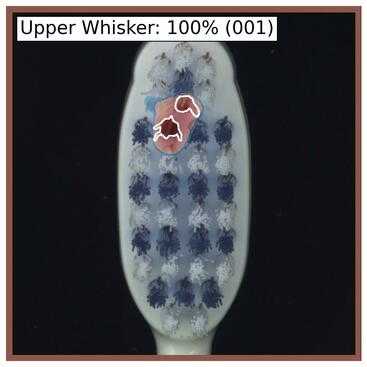}
      \end{minipage}
      \caption{
        Heatmaps.
        Images selected according to AUPIMO's statistics.
        Statistic and image index annotated on upper left corner.
      }
      \label{fig:benchmark-011-heatmap}
    \end{subfigure}
    \caption{
      Benchmark on MVTec AD / Toothbrush.
      PIMO curves and heatmaps are from EfficientAD S.
      042 images (012 normal, 030 anomalous).
    }
    \label{fig:benchmark-011}
\end{figure}

\begin{figure}[ht]
    \centering
    \begin{subfigure}[b]{\linewidth}
      \includegraphics[width=\linewidth,valign=t,keepaspectratio]{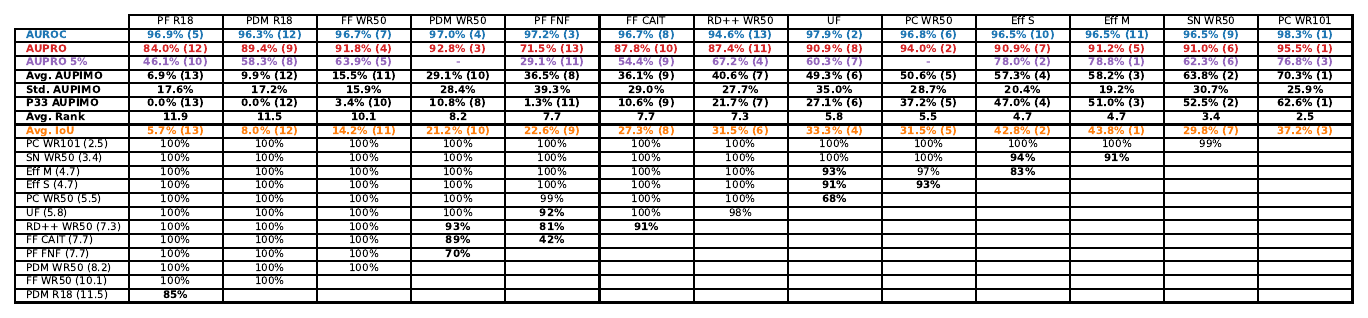}
      \caption{Statistics and pairwise statistical tests.}
      \label{fig:benchmark-012-table}
    \end{subfigure}
    \\ \vspace{2mm}
    \begin{subfigure}[b]{0.5\linewidth}
      \includegraphics[width=\linewidth,valign=t,keepaspectratio]{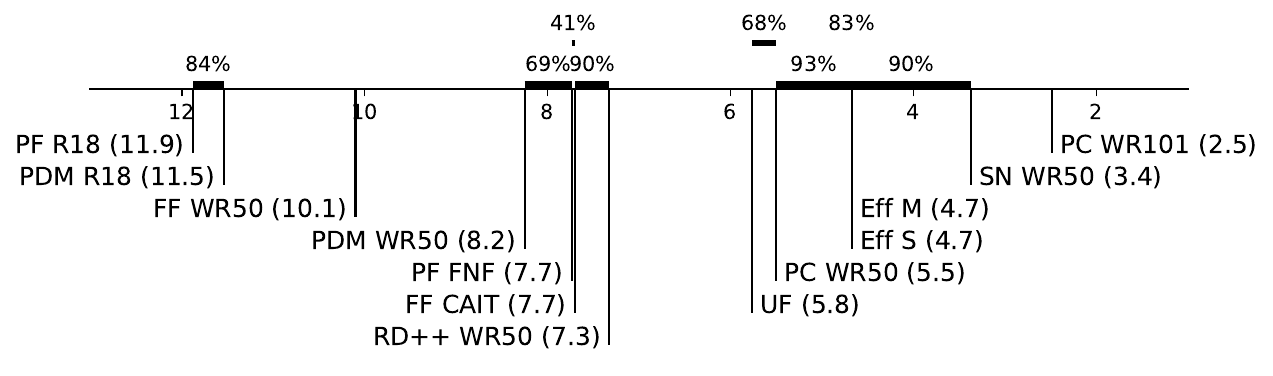}
      \caption{Average rank diagram.}
      \label{fig:benchmark-012-diagram}
    \end{subfigure}
    \\ \vspace{2mm}
    \begin{subfigure}[b]{0.45\linewidth}
      \includegraphics[width=\linewidth,valign=t,keepaspectratio]{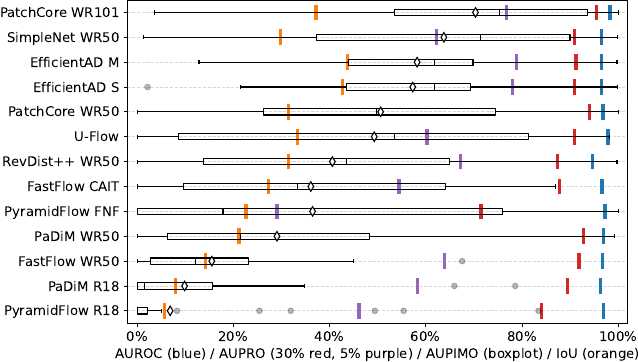}
      \caption{Score distributions.}
      \label{fig:benchmark-012-boxplot}
    \end{subfigure}
    ~
    \begin{subfigure}[b]{0.45\linewidth}
      \includegraphics[width=\linewidth,valign=t,keepaspectratio]{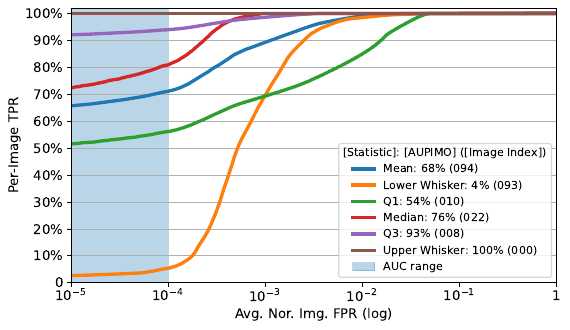}
      \caption{PIMO curves.}
      \label{fig:benchmark-012-pimo-curves}
    \end{subfigure}
    \\  \vspace{2mm}
    \begin{subfigure}[b]{\linewidth}
    
      \begin{minipage}{\linewidth}
        \centering
        \includegraphics[width=.3\linewidth,valign=t,keepaspectratio]{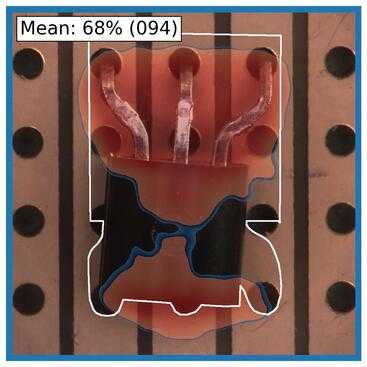}
        \includegraphics[width=.3\linewidth,valign=t,keepaspectratio]{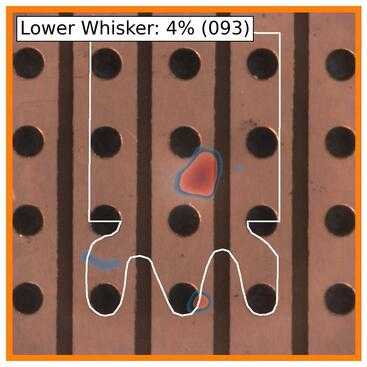}
        \includegraphics[width=.3\linewidth,valign=t,keepaspectratio]{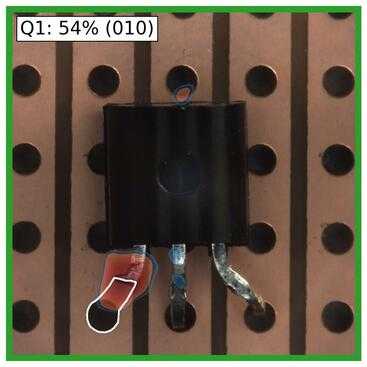}
      \end{minipage}
      \\
      \begin{minipage}{\linewidth}
        \centering
        \includegraphics[width=.3\linewidth,valign=t,keepaspectratio]{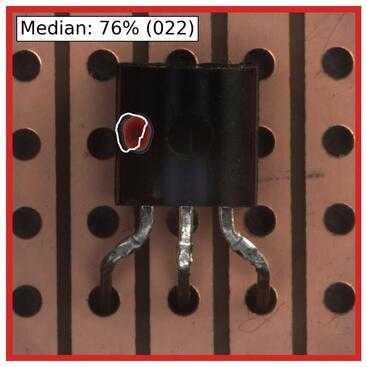}
        \includegraphics[width=.3\linewidth,valign=t,keepaspectratio]{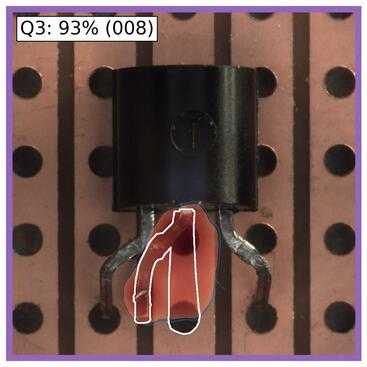}
        \includegraphics[width=.3\linewidth,valign=t,keepaspectratio]{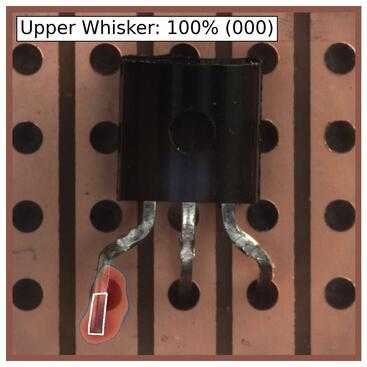}
      \end{minipage}
      \caption{
        Heatmaps.
        Images selected according to AUPIMO's statistics.
        Statistic and image index annotated on upper left corner.
      }
      \label{fig:benchmark-012-heatmap}
    \end{subfigure}
    \caption{
      Benchmark on MVTec AD / Transistor.
      PIMO curves and heatmaps are from PatchCore WR101.
      100 images (060 normal, 040 anomalous).
    }
    \label{fig:benchmark-012}
\end{figure}

\begin{figure}[ht]
    \centering
    \begin{subfigure}[b]{\linewidth}
      \includegraphics[width=\linewidth,valign=t,keepaspectratio]{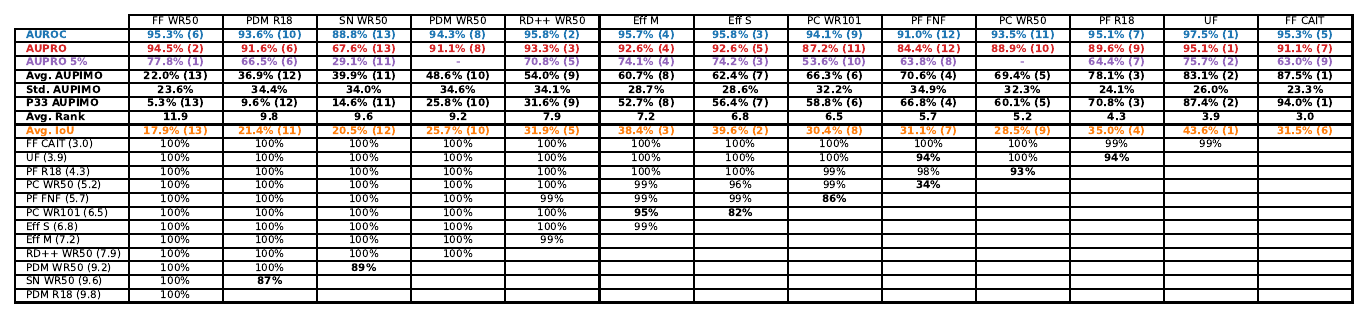}
      \caption{Statistics and pairwise statistical tests.}
      \label{fig:benchmark-013-table}
    \end{subfigure}
    \\ \vspace{2mm}
    \begin{subfigure}[b]{0.5\linewidth}
      \includegraphics[width=\linewidth,valign=t,keepaspectratio]{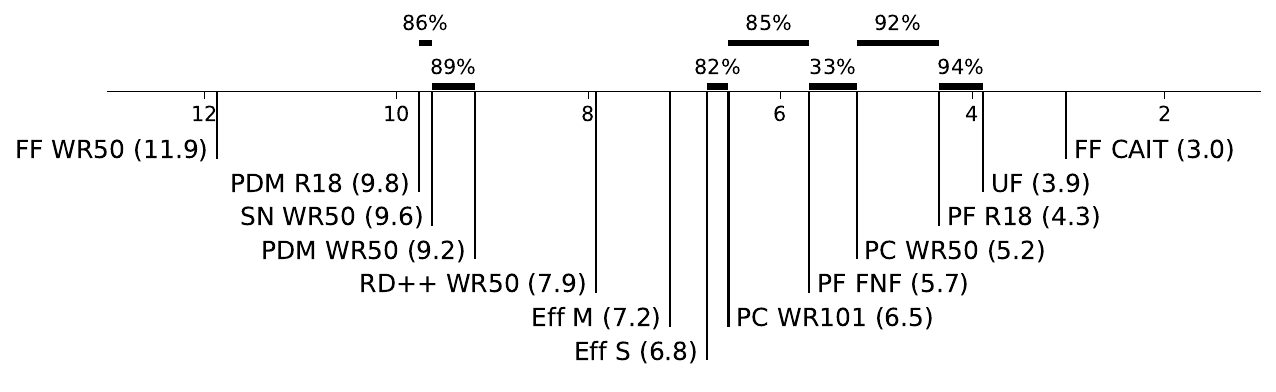}
      \caption{Average rank diagram.}
      \label{fig:benchmark-013-diagram}
    \end{subfigure}
    \\ \vspace{2mm}
    \begin{subfigure}[b]{0.45\linewidth}
      \includegraphics[width=\linewidth,valign=t,keepaspectratio]{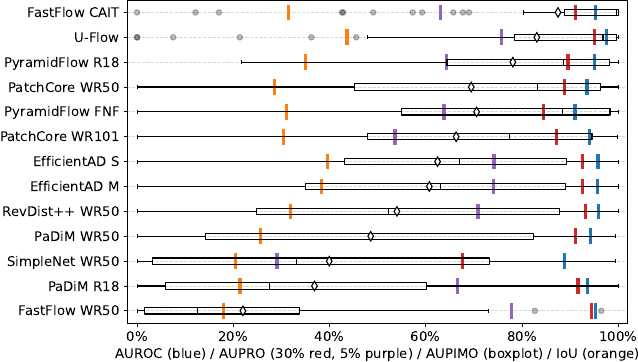}
      \caption{Score distributions.}
      \label{fig:benchmark-013-boxplot}
    \end{subfigure}
    ~
    \begin{subfigure}[b]{0.45\linewidth}
      \includegraphics[width=\linewidth,valign=t,keepaspectratio]{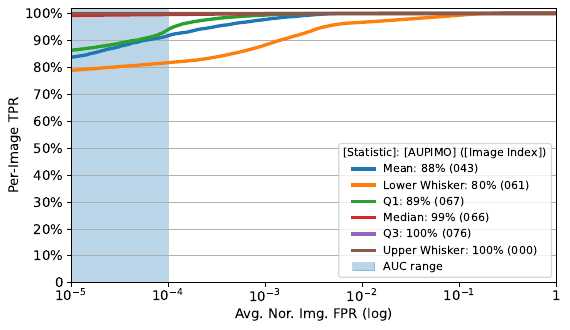}
      \caption{PIMO curves.}
      \label{fig:benchmark-013-pimo-curves}
    \end{subfigure}
    \\  \vspace{2mm}
    \begin{subfigure}[b]{\linewidth}
    
      \begin{minipage}{\linewidth}
        \centering
        \includegraphics[width=.3\linewidth,valign=t,keepaspectratio]{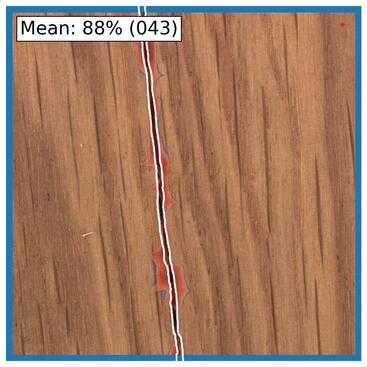}
        \includegraphics[width=.3\linewidth,valign=t,keepaspectratio]{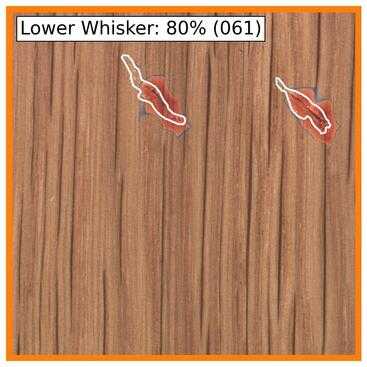}
        \includegraphics[width=.3\linewidth,valign=t,keepaspectratio]{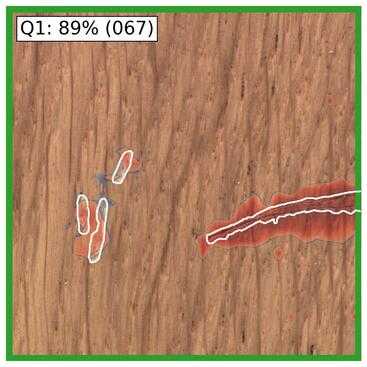}
      \end{minipage}
      \\
      \begin{minipage}{\linewidth}
        \centering
        \includegraphics[width=.3\linewidth,valign=t,keepaspectratio]{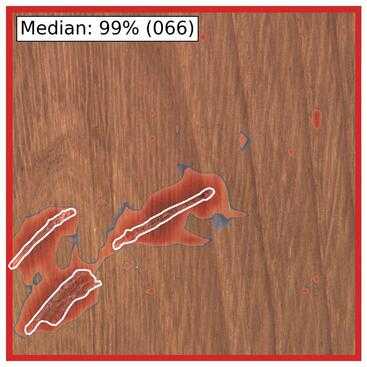}
        \includegraphics[width=.3\linewidth,valign=t,keepaspectratio]{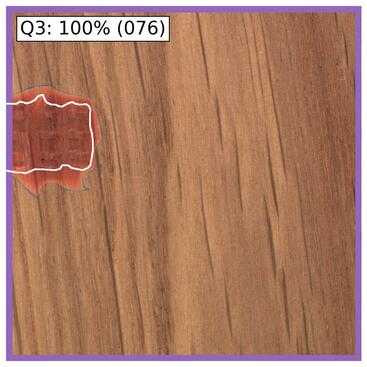}
        \includegraphics[width=.3\linewidth,valign=t,keepaspectratio]{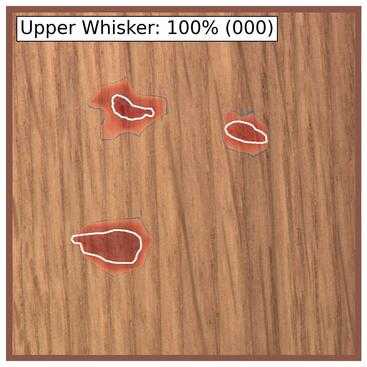}
      \end{minipage}
      \caption{
        Heatmaps.
        Images selected according to AUPIMO's statistics.
        Statistic and image index annotated on upper left corner.
      }
      \label{fig:benchmark-013-heatmap}
    \end{subfigure}
    \caption{
      Benchmark on MVTec AD / Wood.
      PIMO curves and heatmaps are from FastFlow CAIT.
      079 images (019 normal, 060 anomalous).
    }
    \label{fig:benchmark-013}
\end{figure}

\begin{figure}[ht]
    \centering
    \begin{subfigure}[b]{\linewidth}
      \includegraphics[width=\linewidth,valign=t,keepaspectratio]{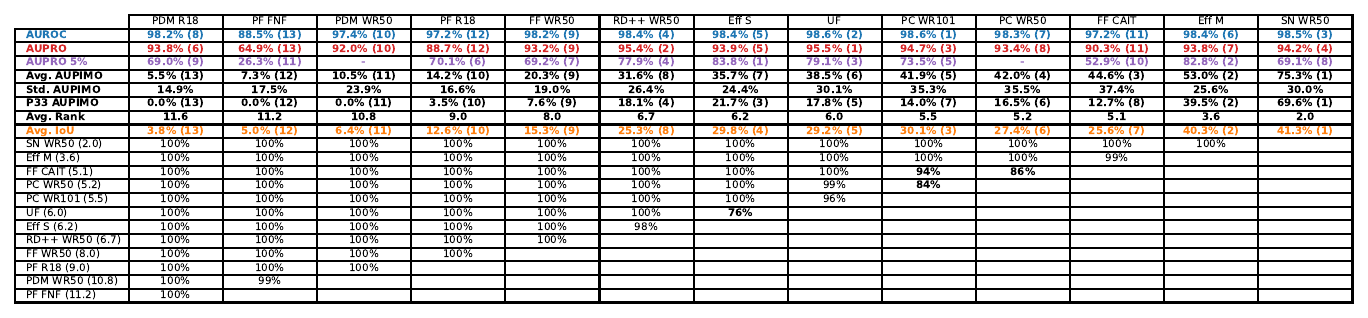}
      \caption{Statistics and pairwise statistical tests.}
      \label{fig:benchmark-014-table}
    \end{subfigure}
    \\ \vspace{2mm}
    \begin{subfigure}[b]{0.5\linewidth}
      \includegraphics[width=\linewidth,valign=t,keepaspectratio]{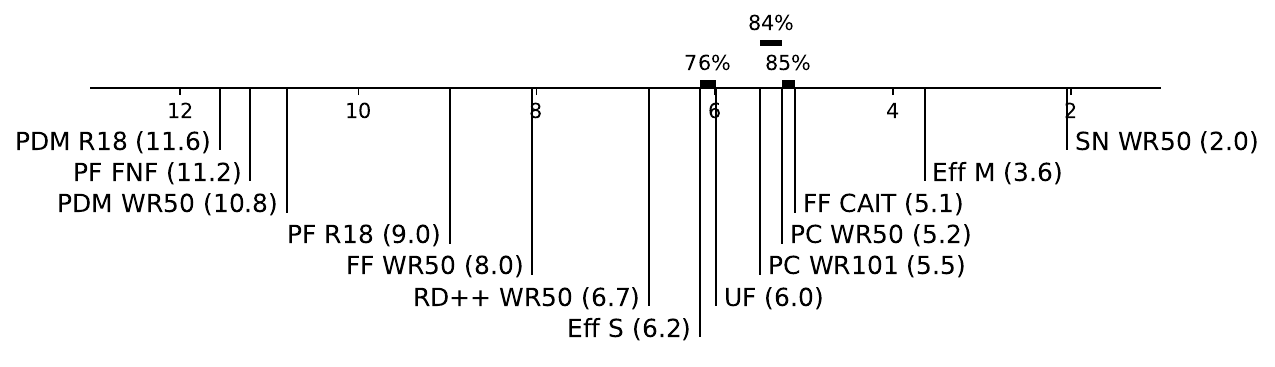}
      \caption{Average rank diagram.}
      \label{fig:benchmark-014-diagram}
    \end{subfigure}
    \\ \vspace{2mm}
    \begin{subfigure}[b]{0.45\linewidth}
      \includegraphics[width=\linewidth,valign=t,keepaspectratio]{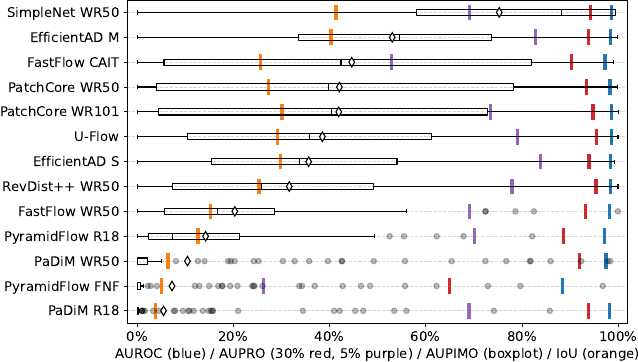}
      \caption{Score distributions.}
      \label{fig:benchmark-014-boxplot}
    \end{subfigure}
    ~
    \begin{subfigure}[b]{0.45\linewidth}
      \includegraphics[width=\linewidth,valign=t,keepaspectratio]{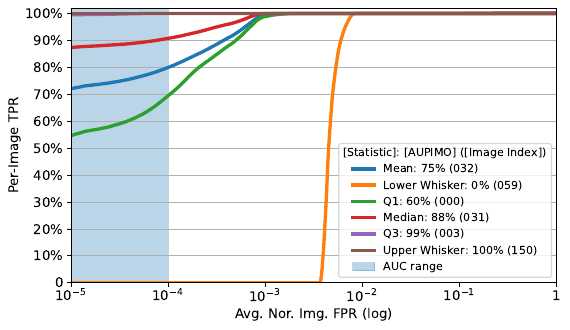}
      \caption{PIMO curves.}
      \label{fig:benchmark-014-pimo-curves}
    \end{subfigure}
    \\  \vspace{2mm}
    \begin{subfigure}[b]{\linewidth}
    
      \begin{minipage}{\linewidth}
        \centering
        \includegraphics[width=.3\linewidth,valign=t,keepaspectratio]{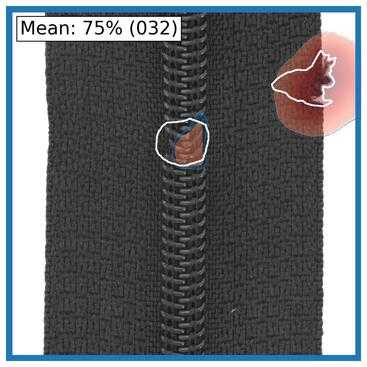}
        \includegraphics[width=.3\linewidth,valign=t,keepaspectratio]{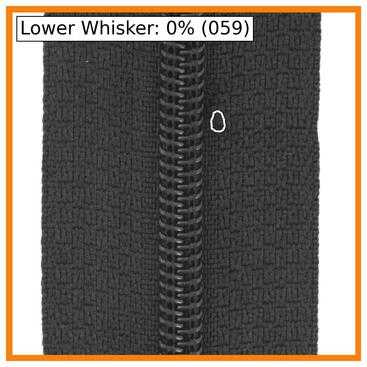}
        \includegraphics[width=.3\linewidth,valign=t,keepaspectratio]{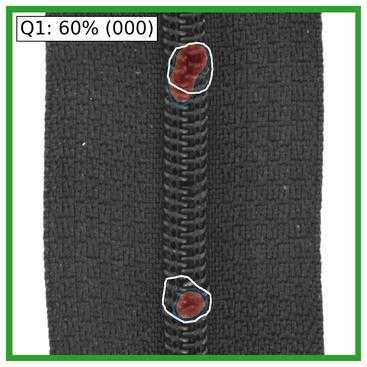}
      \end{minipage}
      \\
      \begin{minipage}{\linewidth}
        \centering
        \includegraphics[width=.3\linewidth,valign=t,keepaspectratio]{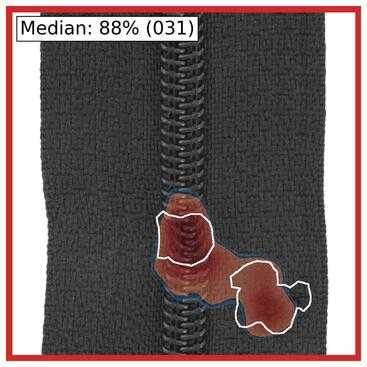}
        \includegraphics[width=.3\linewidth,valign=t,keepaspectratio]{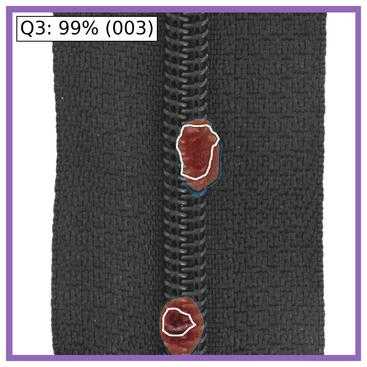}
        \includegraphics[width=.3\linewidth,valign=t,keepaspectratio]{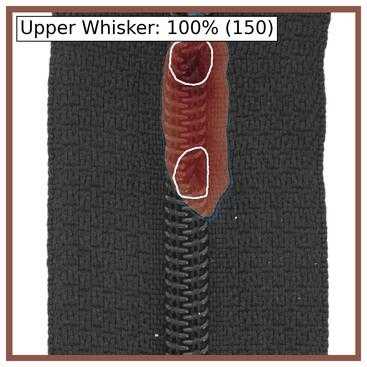}
      \end{minipage}
      \caption{
        Heatmaps.
        Images selected according to AUPIMO's statistics.
        Statistic and image index annotated on upper left corner.
      }
      \label{fig:benchmark-014-heatmap}
    \end{subfigure}
    \caption{
      Benchmark on MVTec AD / Zipper.
      PIMO curves and heatmaps are from SimpleNet WR50.
      151 images (032 normal, 119 anomalous).
    }
    \label{fig:benchmark-014}
\end{figure}

\begin{figure}[ht]
    \centering
    \begin{subfigure}[b]{\linewidth}
      \includegraphics[width=\linewidth,valign=t,keepaspectratio]{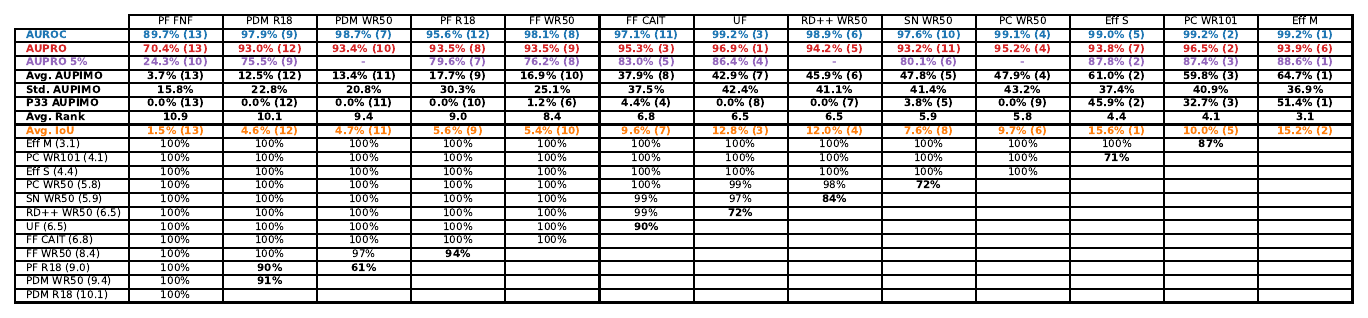}
      \caption{Statistics and pairwise statistical tests.}
      \label{fig:benchmark-015-table}
    \end{subfigure}
    \\ \vspace{2mm}
    \begin{subfigure}[b]{0.5\linewidth}
      \includegraphics[width=\linewidth,valign=t,keepaspectratio]{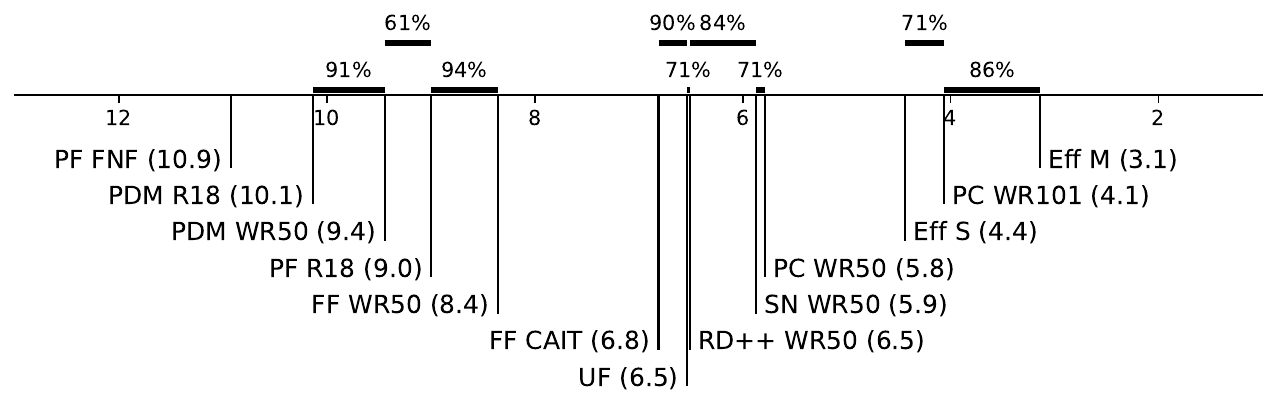}
      \caption{Average rank diagram.}
      \label{fig:benchmark-015-diagram}
    \end{subfigure}
    \\ \vspace{2mm}
    \begin{subfigure}[b]{0.45\linewidth}
      \includegraphics[width=\linewidth,valign=t,keepaspectratio]{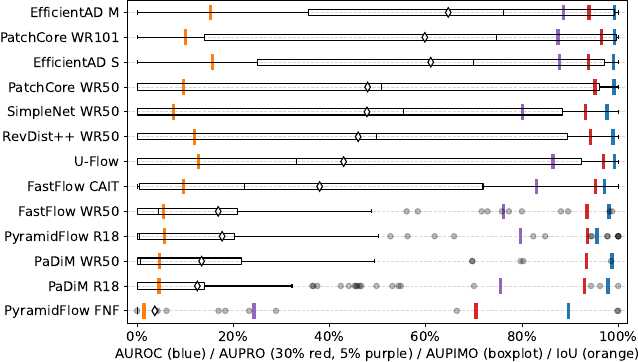}
      \caption{Score distributions.}
      \label{fig:benchmark-015-boxplot}
    \end{subfigure}
    ~
    \begin{subfigure}[b]{0.45\linewidth}
      \includegraphics[width=\linewidth,valign=t,keepaspectratio]{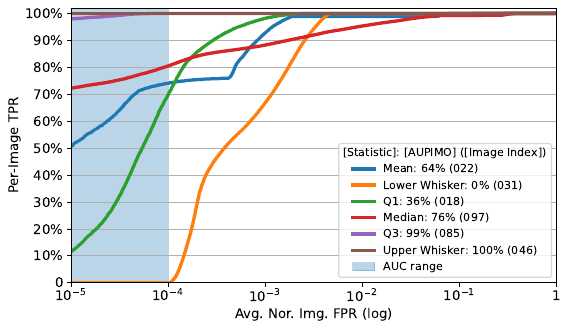}
      \caption{PIMO curves.}
      \label{fig:benchmark-015-pimo-curves}
    \end{subfigure}
    \\  \vspace{2mm}
    \begin{subfigure}[b]{\linewidth}
    
      \begin{minipage}{\linewidth}
        \centering
        \includegraphics[width=.3\linewidth,valign=t,keepaspectratio]{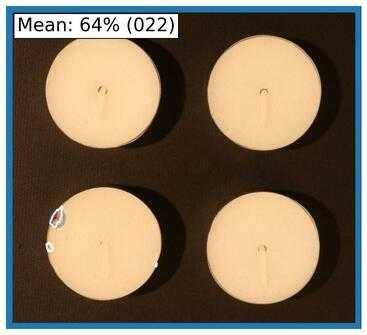}
        \includegraphics[width=.3\linewidth,valign=t,keepaspectratio]{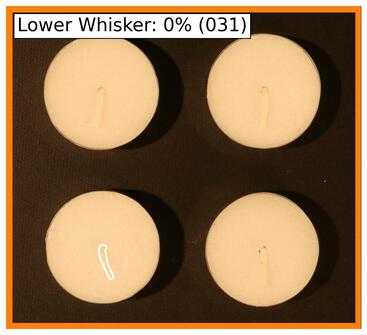}
        \includegraphics[width=.3\linewidth,valign=t,keepaspectratio]{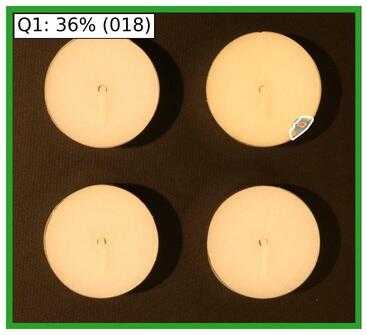}
      \end{minipage}
      \\
      \begin{minipage}{\linewidth}
        \centering
        \includegraphics[width=.3\linewidth,valign=t,keepaspectratio]{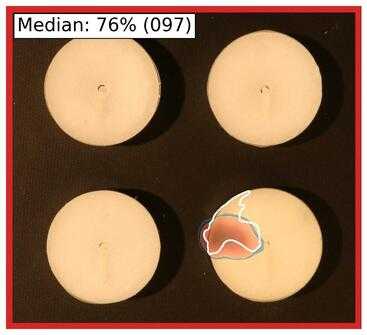}
        \includegraphics[width=.3\linewidth,valign=t,keepaspectratio]{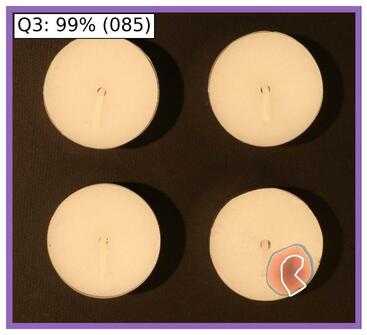}
        \includegraphics[width=.3\linewidth,valign=t,keepaspectratio]{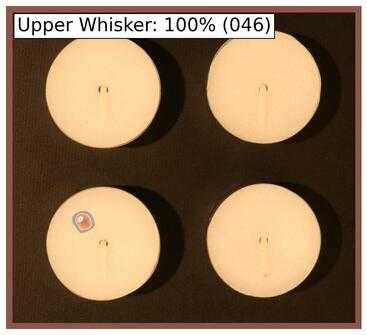}
      \end{minipage}
      \caption{
        Heatmaps.
        Images selected according to AUPIMO's statistics.
        Statistic and image index annotated on upper left corner.
      }
      \label{fig:benchmark-015-heatmap}
    \end{subfigure}
    \caption{
      Benchmark on VisA / Candle.
      PIMO curves and heatmaps are from EfficientAD M.
      200 images (100 normal, 100 anomalous).
    }
    \label{fig:benchmark-015}
\end{figure}

\begin{figure}[ht]
    \centering
    \begin{subfigure}[b]{\linewidth}
      \includegraphics[width=\linewidth,valign=t,keepaspectratio]{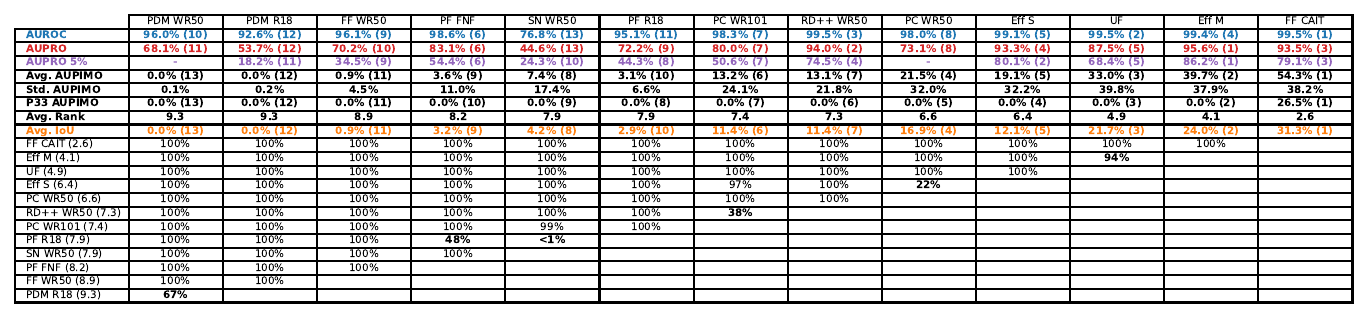}
      \caption{Statistics and pairwise statistical tests.}
      \label{fig:benchmark-016-table}
    \end{subfigure}
    \\ \vspace{2mm}
    \begin{subfigure}[b]{0.5\linewidth}
      \includegraphics[width=\linewidth,valign=t,keepaspectratio]{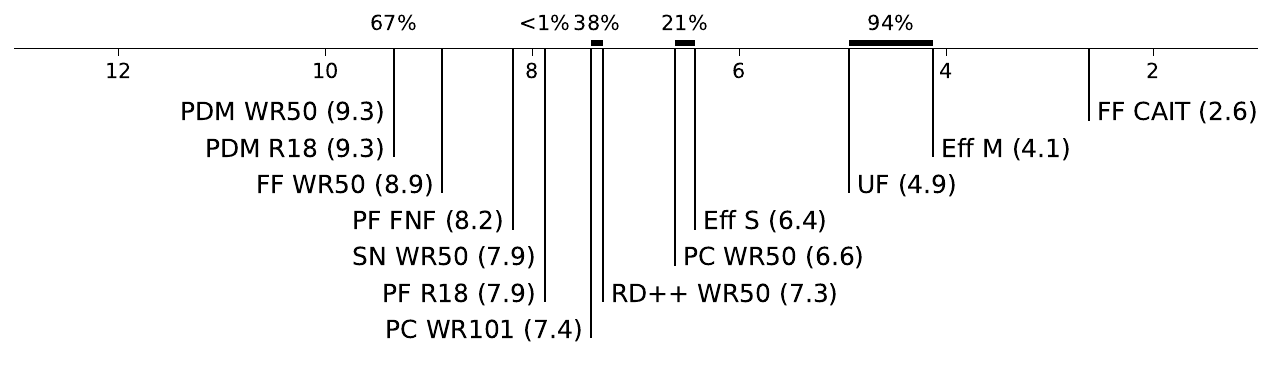}
      \caption{Average rank diagram.}
      \label{fig:benchmark-016-diagram}
    \end{subfigure}
    \\ \vspace{2mm}
    \begin{subfigure}[b]{0.45\linewidth}
      \includegraphics[width=\linewidth,valign=t,keepaspectratio]{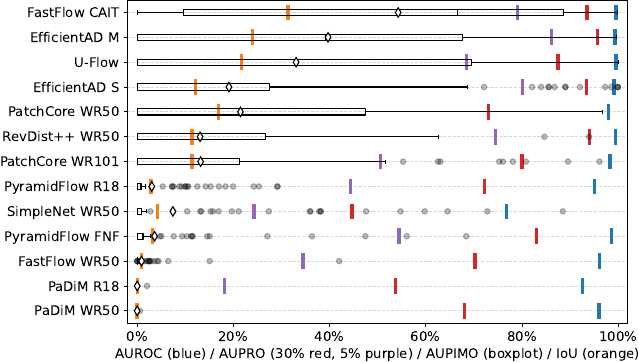}
      \caption{Score distributions.}
      \label{fig:benchmark-016-boxplot}
    \end{subfigure}
    ~
    \begin{subfigure}[b]{0.45\linewidth}
      \includegraphics[width=\linewidth,valign=t,keepaspectratio]{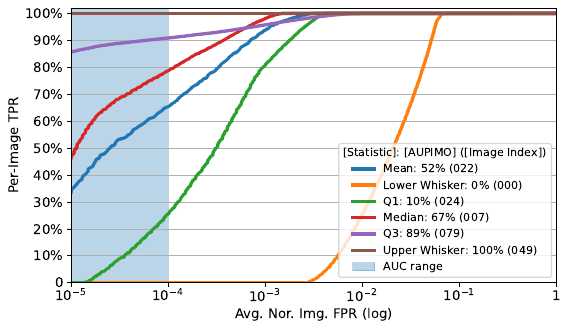}
      \caption{PIMO curves.}
      \label{fig:benchmark-016-pimo-curves}
    \end{subfigure}
    \\  \vspace{2mm}
    \begin{subfigure}[b]{\linewidth}
    
      \begin{minipage}{\linewidth}
        \centering
        \includegraphics[width=.3\linewidth,valign=t,keepaspectratio]{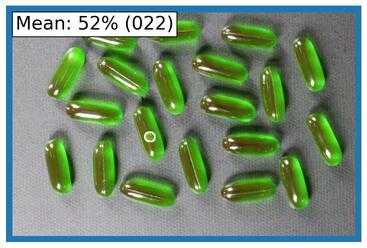}
        \includegraphics[width=.3\linewidth,valign=t,keepaspectratio]{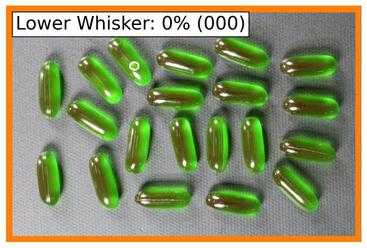}
        \includegraphics[width=.3\linewidth,valign=t,keepaspectratio]{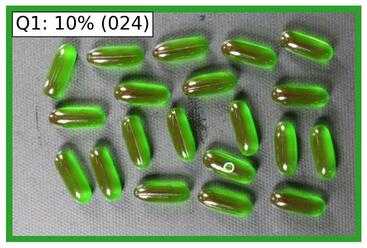}
      \end{minipage}
      \\
      \begin{minipage}{\linewidth}
        \centering
        \includegraphics[width=.3\linewidth,valign=t,keepaspectratio]{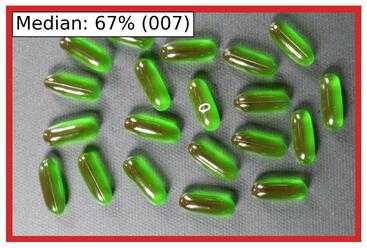}
        \includegraphics[width=.3\linewidth,valign=t,keepaspectratio]{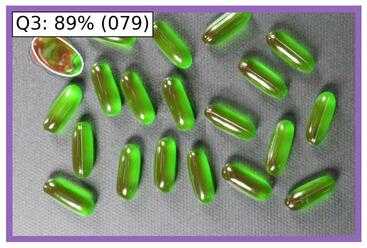}
        \includegraphics[width=.3\linewidth,valign=t,keepaspectratio]{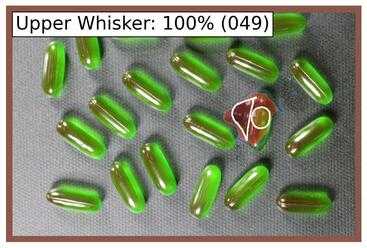}
      \end{minipage}
      \caption{
        Heatmaps.
        Images selected according to AUPIMO's statistics.
        Statistic and image index annotated on upper left corner.
      }
      \label{fig:benchmark-016-heatmap}
    \end{subfigure}
    \caption{
      Benchmark on VisA / Capsules.
      PIMO curves and heatmaps are from FastFlow CAIT.
      160 images (060 normal, 100 anomalous).
    }
    \label{fig:benchmark-016}
\end{figure}

\begin{figure}[ht]
    \centering
    \begin{subfigure}[b]{\linewidth}
      \includegraphics[width=\linewidth,valign=t,keepaspectratio]{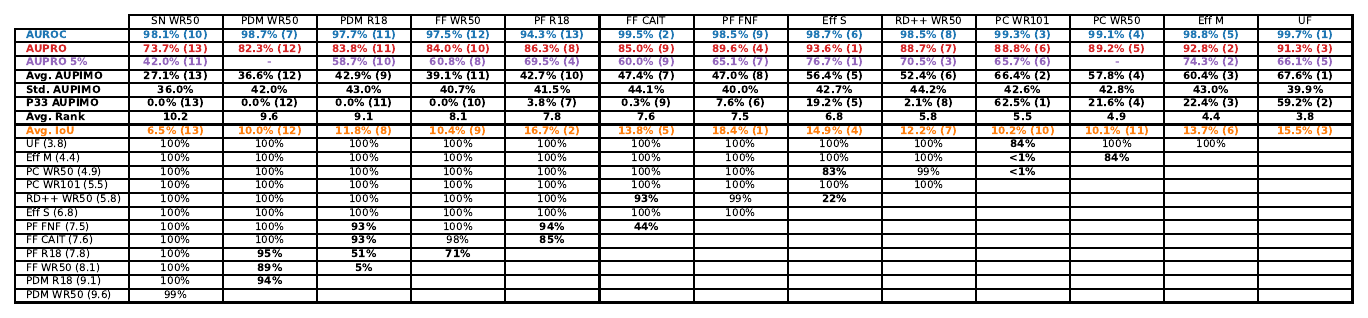}
      \caption{Statistics and pairwise statistical tests.}
      \label{fig:benchmark-017-table}
    \end{subfigure}
    \\ \vspace{2mm}
    \begin{subfigure}[b]{0.5\linewidth}
      \includegraphics[width=\linewidth,valign=t,keepaspectratio]{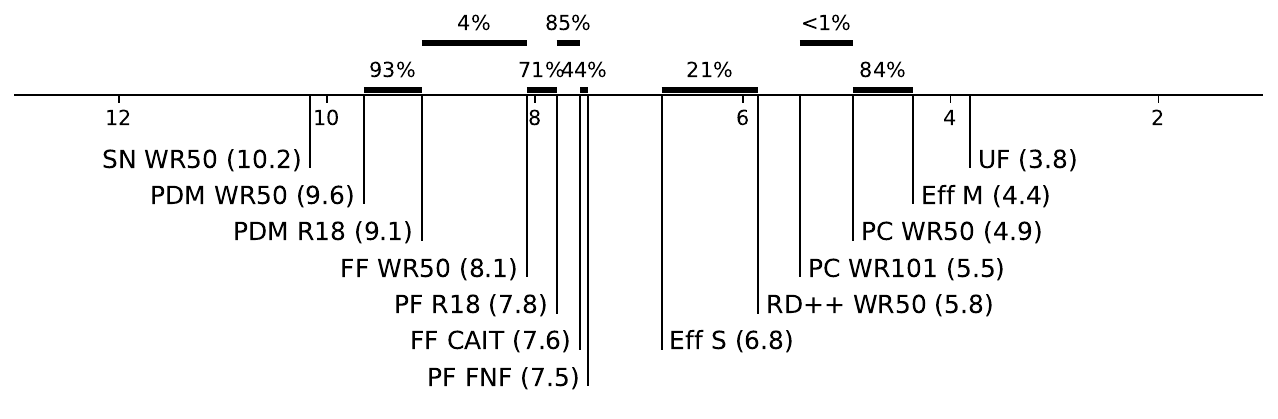}
      \caption{Average rank diagram.}
      \label{fig:benchmark-017-diagram}
    \end{subfigure}
    \\ \vspace{2mm}
    \begin{subfigure}[b]{0.45\linewidth}
      \includegraphics[width=\linewidth,valign=t,keepaspectratio]{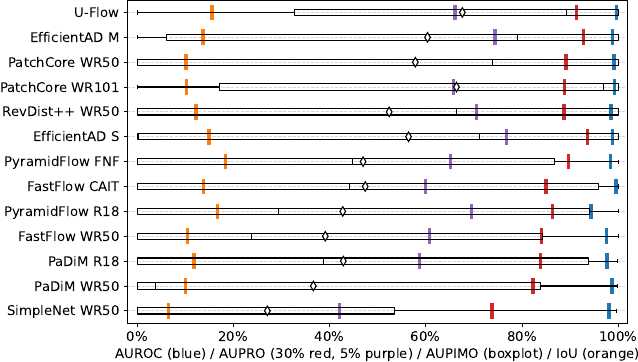}
      \caption{Score distributions.}
      \label{fig:benchmark-017-boxplot}
    \end{subfigure}
    ~
    \begin{subfigure}[b]{0.45\linewidth}
      \includegraphics[width=\linewidth,valign=t,keepaspectratio]{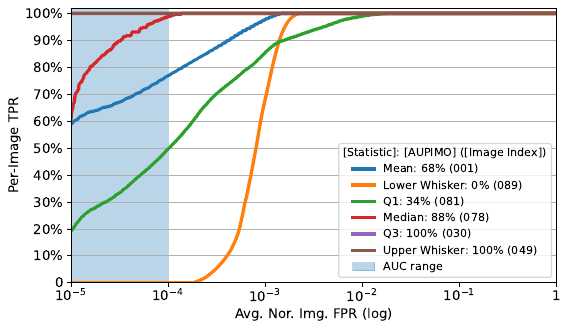}
      \caption{PIMO curves.}
      \label{fig:benchmark-017-pimo-curves}
    \end{subfigure}
    \\  \vspace{2mm}
    \begin{subfigure}[b]{\linewidth}
    
      \begin{minipage}{\linewidth}
        \centering
        \includegraphics[width=.3\linewidth,valign=t,keepaspectratio]{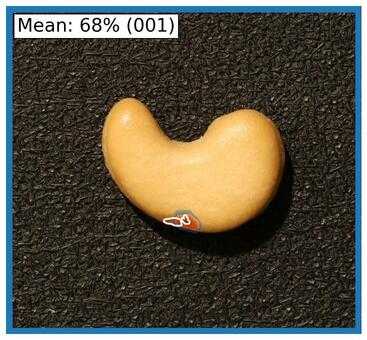}
        \includegraphics[width=.3\linewidth,valign=t,keepaspectratio]{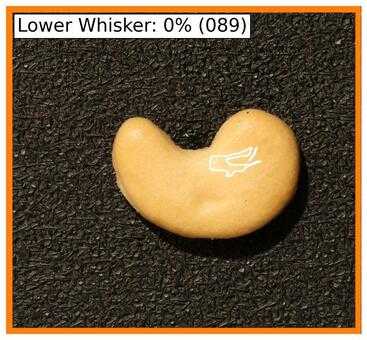}
        \includegraphics[width=.3\linewidth,valign=t,keepaspectratio]{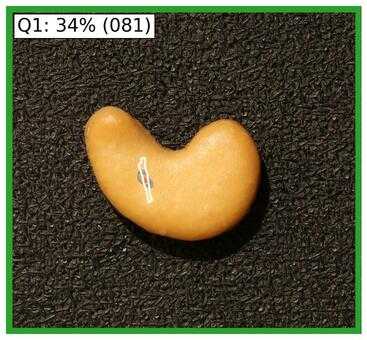}
      \end{minipage}
      \\
      \begin{minipage}{\linewidth}
        \centering
        \includegraphics[width=.3\linewidth,valign=t,keepaspectratio]{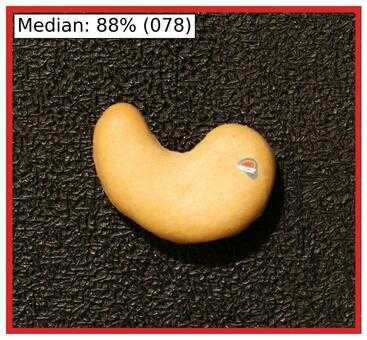}
        \includegraphics[width=.3\linewidth,valign=t,keepaspectratio]{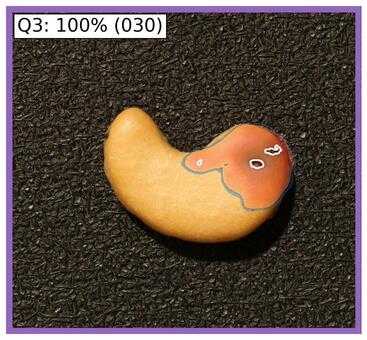}
        \includegraphics[width=.3\linewidth,valign=t,keepaspectratio]{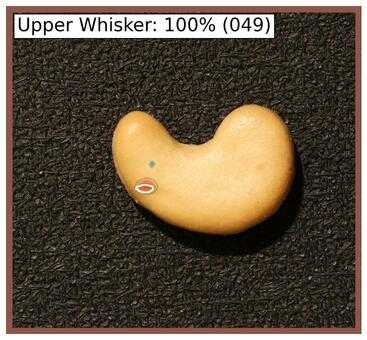}
      \end{minipage}
      \caption{
        Heatmaps.
        Images selected according to AUPIMO's statistics.
        Statistic and image index annotated on upper left corner.
      }
      \label{fig:benchmark-017-heatmap}
    \end{subfigure}
    \caption{
      Benchmark on VisA / Cashew.
      PIMO curves and heatmaps are from U-Flow.
      150 images (050 normal, 100 anomalous).
    }
    \label{fig:benchmark-017}
\end{figure}

\begin{figure}[ht]
    \centering
    \begin{subfigure}[b]{\linewidth}
      \includegraphics[width=\linewidth,valign=t,keepaspectratio]{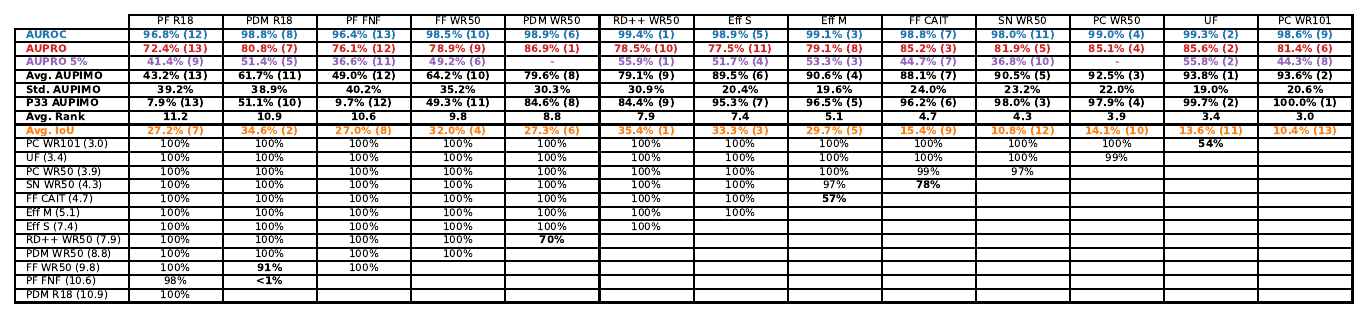}
      \caption{Statistics and pairwise statistical tests.}
      \label{fig:benchmark-018-table}
    \end{subfigure}
    \\ \vspace{2mm}
    \begin{subfigure}[b]{0.5\linewidth}
      \includegraphics[width=\linewidth,valign=t,keepaspectratio]{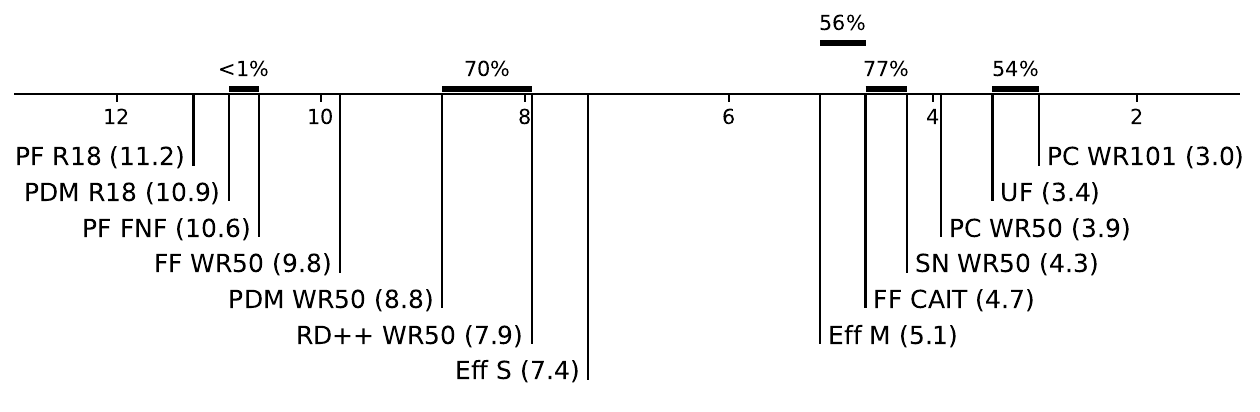}
      \caption{Average rank diagram.}
      \label{fig:benchmark-018-diagram}
    \end{subfigure}
    \\ \vspace{2mm}
    \begin{subfigure}[b]{0.45\linewidth}
      \includegraphics[width=\linewidth,valign=t,keepaspectratio]{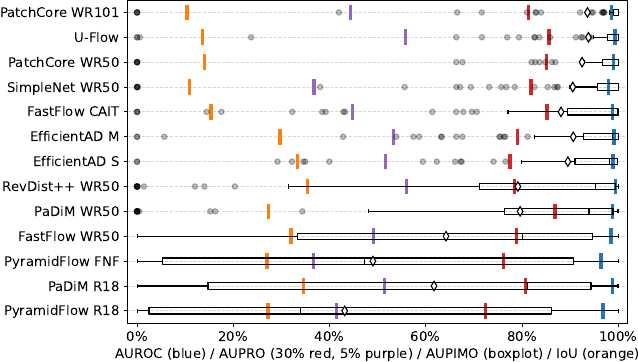}
      \caption{Score distributions.}
      \label{fig:benchmark-018-boxplot}
    \end{subfigure}
    ~
    \begin{subfigure}[b]{0.45\linewidth}
      \includegraphics[width=\linewidth,valign=t,keepaspectratio]{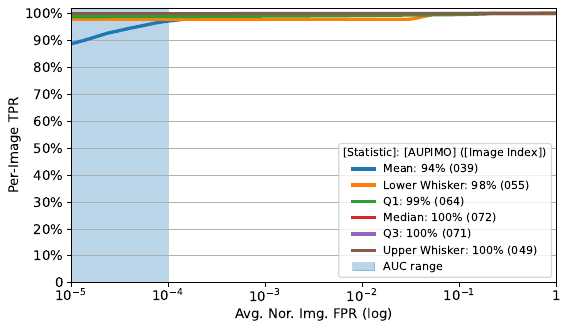}
      \caption{PIMO curves.}
      \label{fig:benchmark-018-pimo-curves}
    \end{subfigure}
    \\  \vspace{2mm}
    \begin{subfigure}[b]{\linewidth}
    
      \begin{minipage}{\linewidth}
        \centering
        \includegraphics[width=.3\linewidth,valign=t,keepaspectratio]{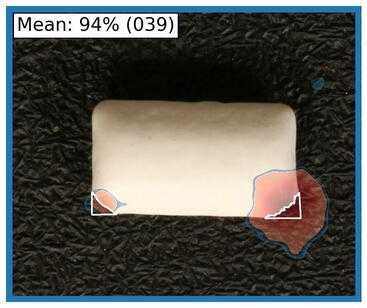}
        \includegraphics[width=.3\linewidth,valign=t,keepaspectratio]{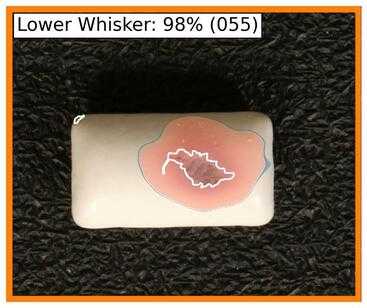}
        \includegraphics[width=.3\linewidth,valign=t,keepaspectratio]{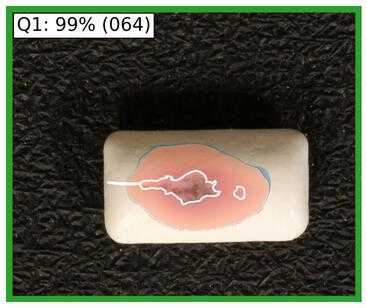}
      \end{minipage}
      \\
      \begin{minipage}{\linewidth}
        \centering
        \includegraphics[width=.3\linewidth,valign=t,keepaspectratio]{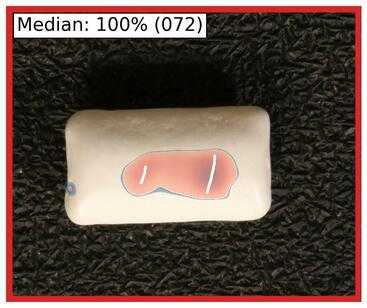}
        \includegraphics[width=.3\linewidth,valign=t,keepaspectratio]{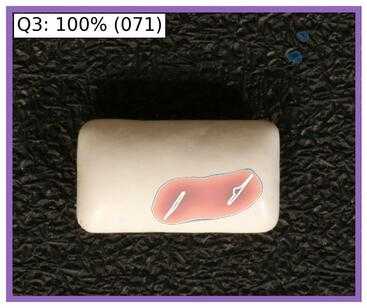}
        \includegraphics[width=.3\linewidth,valign=t,keepaspectratio]{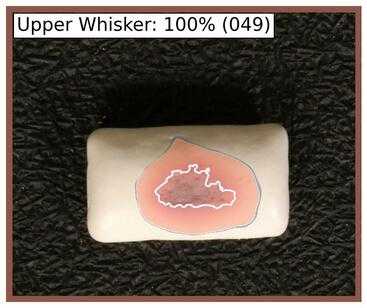}
      \end{minipage}
      \caption{
        Heatmaps.
        Images selected according to AUPIMO's statistics.
        Statistic and image index annotated on upper left corner.
      }
      \label{fig:benchmark-018-heatmap}
    \end{subfigure}
    \caption{
      Benchmark on VisA / Chewing Gum.
      PIMO curves and heatmaps are from PatchCore WR101.
      150 images (050 normal, 100 anomalous).
    }
    \label{fig:benchmark-018}
\end{figure}

\begin{figure}[ht]
    \centering
    \begin{subfigure}[b]{\linewidth}
      \includegraphics[width=\linewidth,valign=t,keepaspectratio]{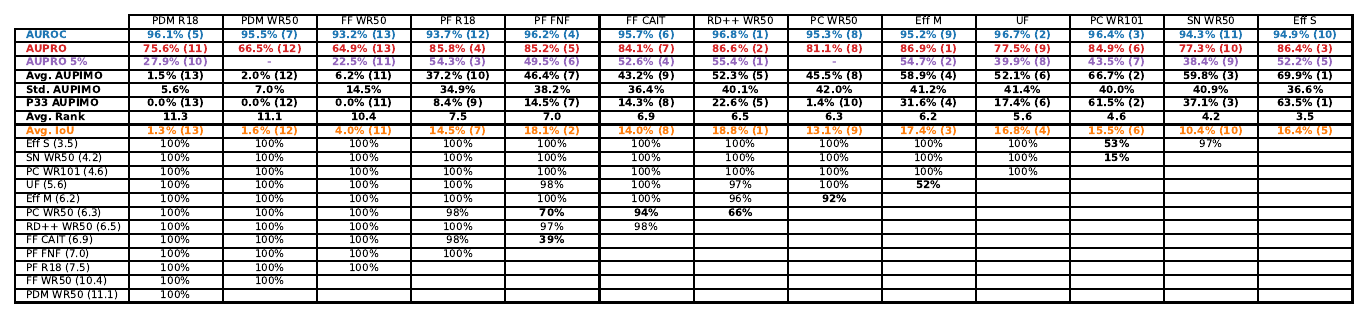}
      \caption{Statistics and pairwise statistical tests.}
      \label{fig:benchmark-019-table}
    \end{subfigure}
    \\ \vspace{2mm}
    \begin{subfigure}[b]{0.5\linewidth}
      \includegraphics[width=\linewidth,valign=t,keepaspectratio]{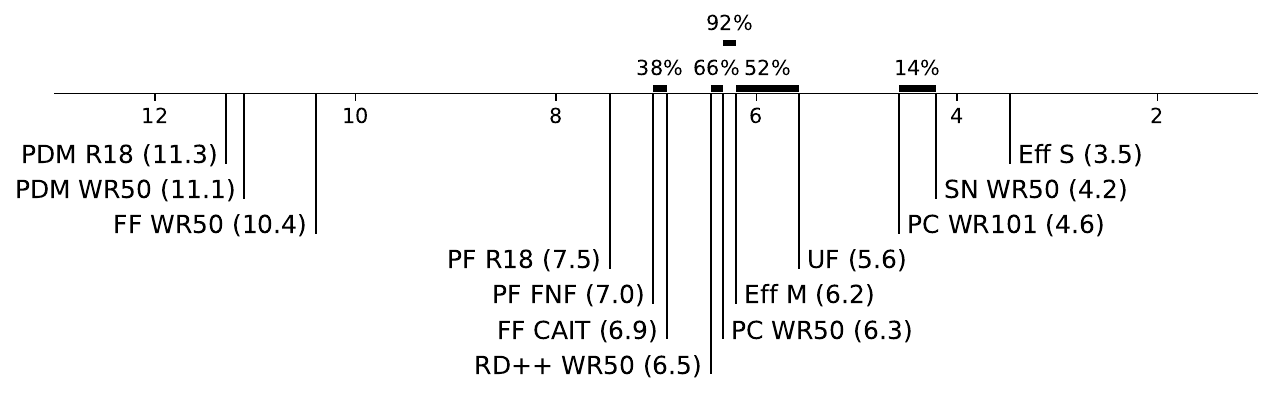}
      \caption{Average rank diagram.}
      \label{fig:benchmark-019-diagram}
    \end{subfigure}
    \\ \vspace{2mm}
    \begin{subfigure}[b]{0.45\linewidth}
      \includegraphics[width=\linewidth,valign=t,keepaspectratio]{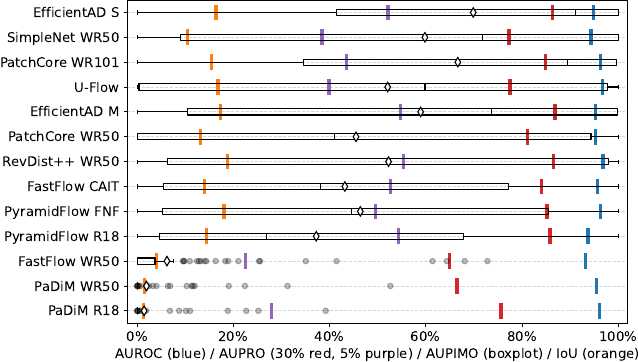}
      \caption{Score distributions.}
      \label{fig:benchmark-019-boxplot}
    \end{subfigure}
    ~
    \begin{subfigure}[b]{0.45\linewidth}
      \includegraphics[width=\linewidth,valign=t,keepaspectratio]{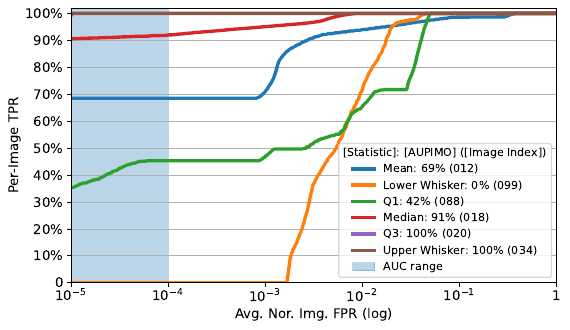}
      \caption{PIMO curves.}
      \label{fig:benchmark-019-pimo-curves}
    \end{subfigure}
    \\  \vspace{2mm}
    \begin{subfigure}[b]{\linewidth}
    
      \begin{minipage}{\linewidth}
        \centering
        \includegraphics[width=.3\linewidth,valign=t,keepaspectratio]{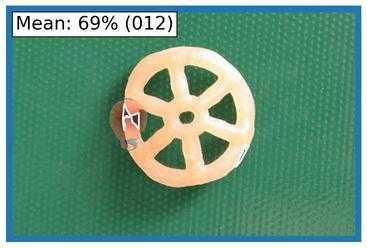}
        \includegraphics[width=.3\linewidth,valign=t,keepaspectratio]{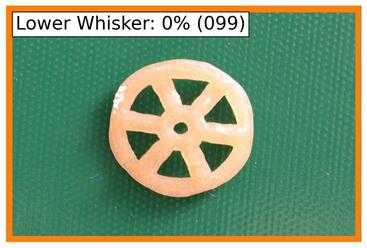}
        \includegraphics[width=.3\linewidth,valign=t,keepaspectratio]{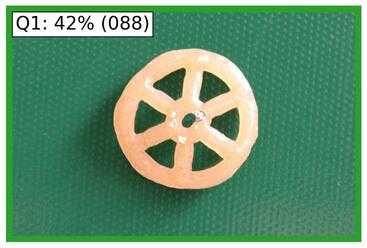}
      \end{minipage}
      \\
      \begin{minipage}{\linewidth}
        \centering
        \includegraphics[width=.3\linewidth,valign=t,keepaspectratio]{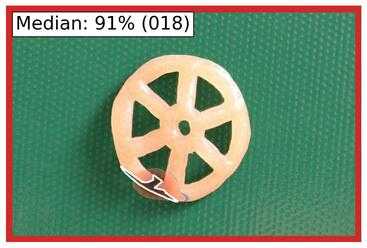}
        \includegraphics[width=.3\linewidth,valign=t,keepaspectratio]{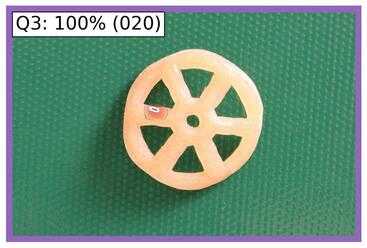}
        \includegraphics[width=.3\linewidth,valign=t,keepaspectratio]{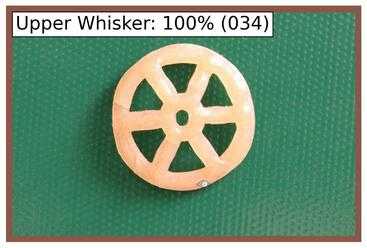}
      \end{minipage}
      \caption{
        Heatmaps.
        Images selected according to AUPIMO's statistics.
        Statistic and image index annotated on upper left corner.
      }
      \label{fig:benchmark-019-heatmap}
    \end{subfigure}
    \caption{
      Benchmark on VisA / Fryum.
      PIMO curves and heatmaps are from EfficientAD S.
      150 images (050 normal, 100 anomalous).
    }
    \label{fig:benchmark-019}
\end{figure}

\begin{figure}[ht]
    \centering
    \begin{subfigure}[b]{\linewidth}
      \includegraphics[width=\linewidth,valign=t,keepaspectratio]{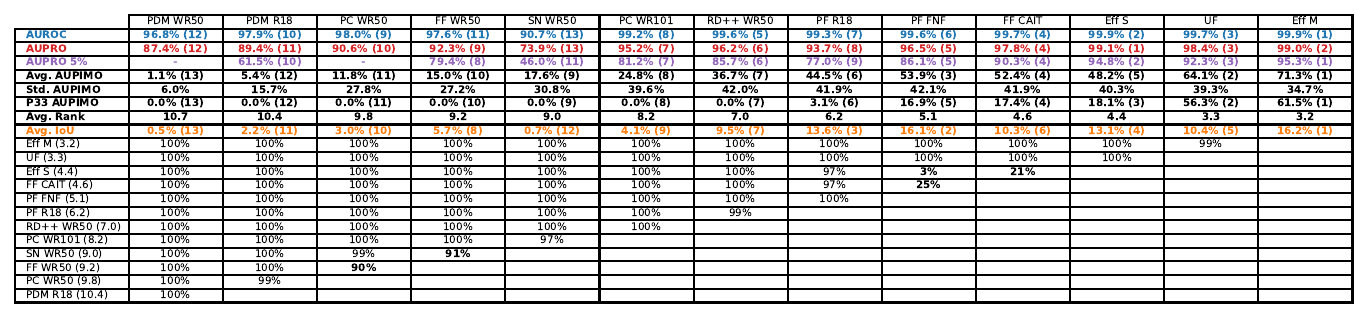}
      \caption{Statistics and pairwise statistical tests.}
      \label{fig:benchmark-020-table}
    \end{subfigure}
    \\ \vspace{2mm}
    \begin{subfigure}[b]{0.5\linewidth}
      \includegraphics[width=\linewidth,valign=t,keepaspectratio]{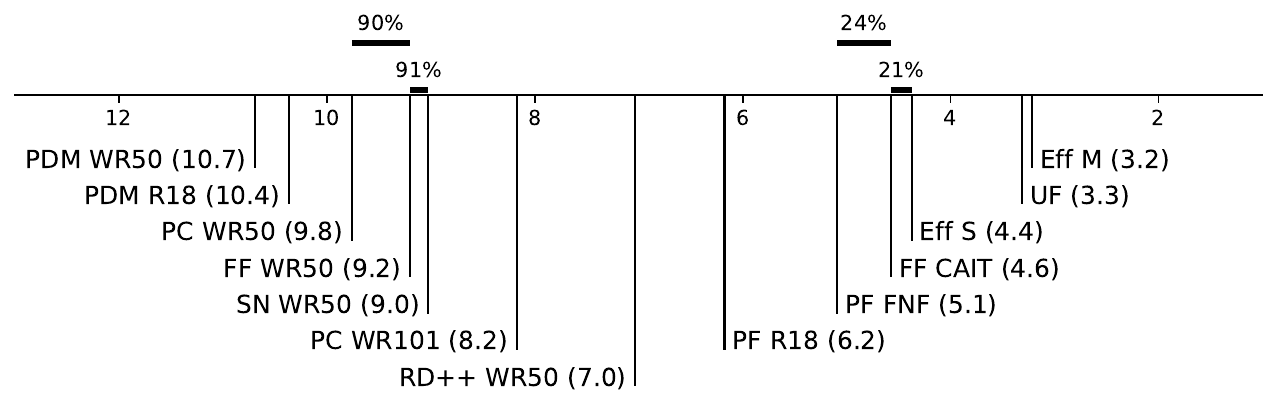}
      \caption{Average rank diagram.}
      \label{fig:benchmark-020-diagram}
    \end{subfigure}
    \\ \vspace{2mm}
    \begin{subfigure}[b]{0.45\linewidth}
      \includegraphics[width=\linewidth,valign=t,keepaspectratio]{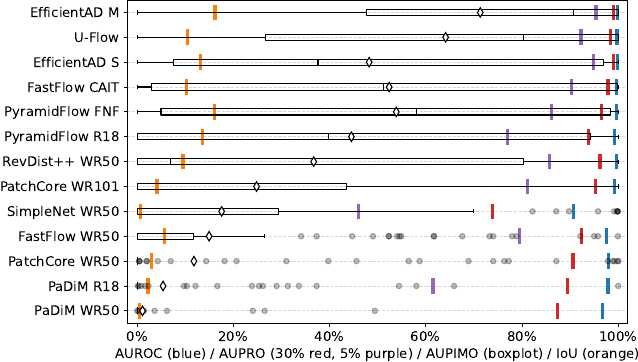}
      \caption{Score distributions.}
      \label{fig:benchmark-020-boxplot}
    \end{subfigure}
    ~
    \begin{subfigure}[b]{0.45\linewidth}
      \includegraphics[width=\linewidth,valign=t,keepaspectratio]{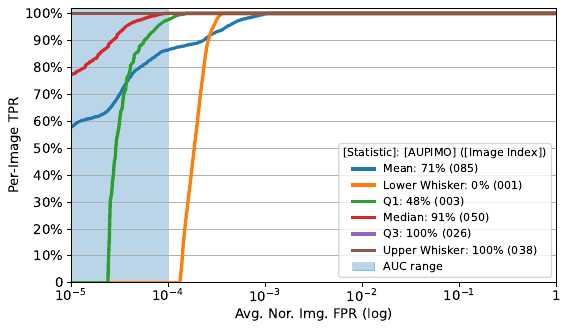}
      \caption{PIMO curves.}
      \label{fig:benchmark-020-pimo-curves}
    \end{subfigure}
    \\  \vspace{2mm}
    \begin{subfigure}[b]{\linewidth}
    
      \begin{minipage}{\linewidth}
        \centering
        \includegraphics[width=.3\linewidth,valign=t,keepaspectratio]{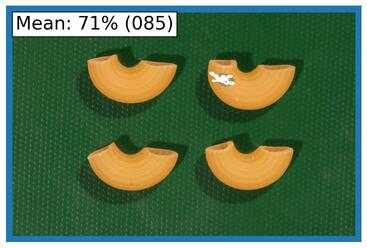}
        \includegraphics[width=.3\linewidth,valign=t,keepaspectratio]{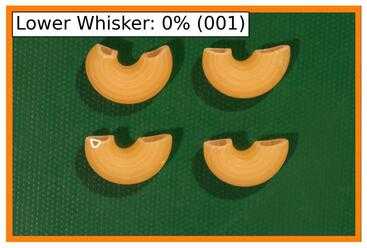}
        \includegraphics[width=.3\linewidth,valign=t,keepaspectratio]{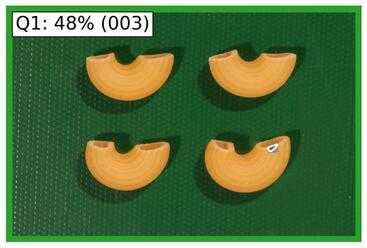}
      \end{minipage}
      \\
      \begin{minipage}{\linewidth}
        \centering
        \includegraphics[width=.3\linewidth,valign=t,keepaspectratio]{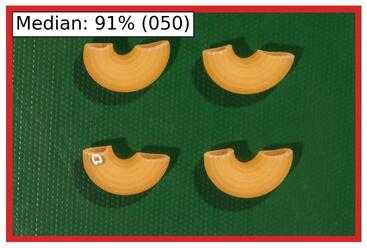}
        \includegraphics[width=.3\linewidth,valign=t,keepaspectratio]{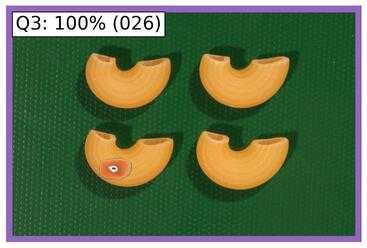}
        \includegraphics[width=.3\linewidth,valign=t,keepaspectratio]{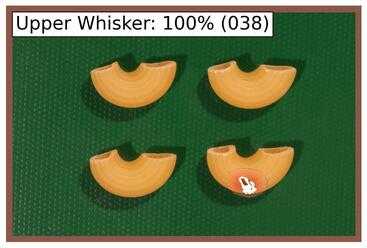}
      \end{minipage}
      \caption{
        Heatmaps.
        Images selected according to AUPIMO's statistics.
        Statistic and image index annotated on upper left corner.
      }
      \label{fig:benchmark-020-heatmap}
    \end{subfigure}
    \caption{
      Benchmark on VisA / Macaroni 1.
      PIMO curves and heatmaps are from EfficientAD M.
      200 images (100 normal, 100 anomalous).
    }
    \label{fig:benchmark-020}
\end{figure}

\begin{figure}[ht]
    \centering
    \begin{subfigure}[b]{\linewidth}
      \includegraphics[width=\linewidth,valign=t,keepaspectratio]{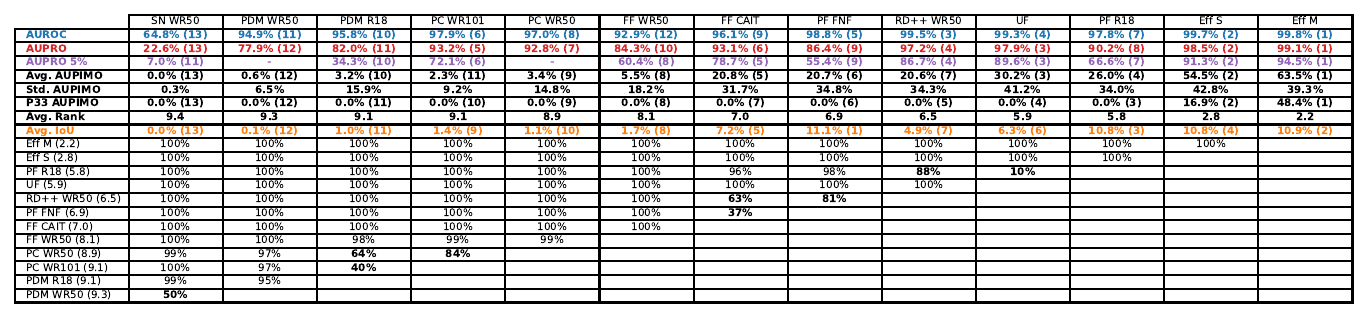}
      \caption{Statistics and pairwise statistical tests.}
      \label{fig:benchmark-021-table}
    \end{subfigure}
    \\ \vspace{2mm}
    \begin{subfigure}[b]{0.5\linewidth}
      \includegraphics[width=\linewidth,valign=t,keepaspectratio]{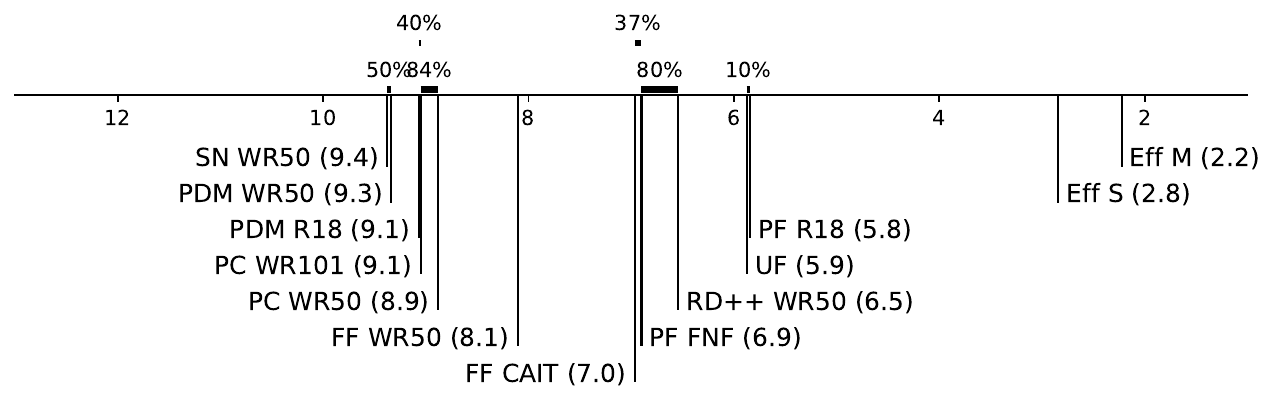}
      \caption{Average rank diagram.}
      \label{fig:benchmark-021-diagram}
    \end{subfigure}
    \\ \vspace{2mm}
    \begin{subfigure}[b]{0.45\linewidth}
      \includegraphics[width=\linewidth,valign=t,keepaspectratio]{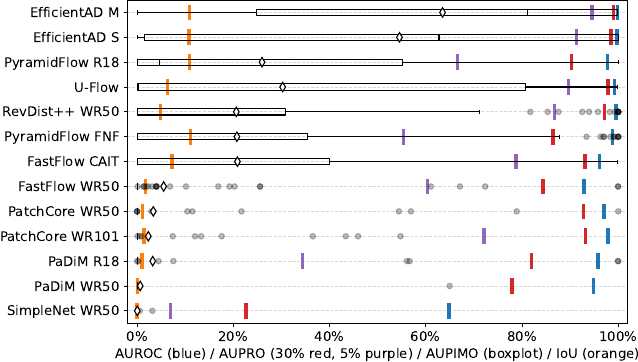}
      \caption{Score distributions.}
      \label{fig:benchmark-021-boxplot}
    \end{subfigure}
    ~
    \begin{subfigure}[b]{0.45\linewidth}
      \includegraphics[width=\linewidth,valign=t,keepaspectratio]{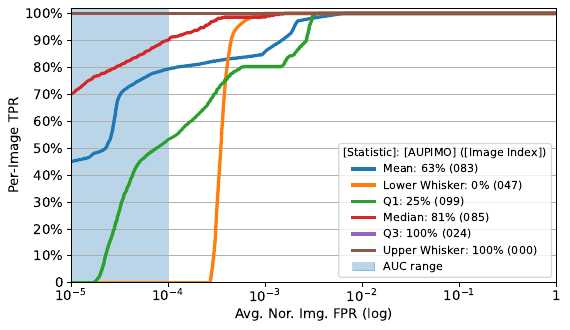}
      \caption{PIMO curves.}
      \label{fig:benchmark-021-pimo-curves}
    \end{subfigure}
    \\  \vspace{2mm}
    \begin{subfigure}[b]{\linewidth}
    
      \begin{minipage}{\linewidth}
        \centering
        \includegraphics[width=.3\linewidth,valign=t,keepaspectratio]{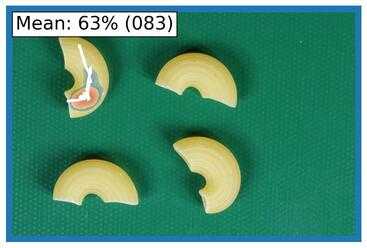}
        \includegraphics[width=.3\linewidth,valign=t,keepaspectratio]{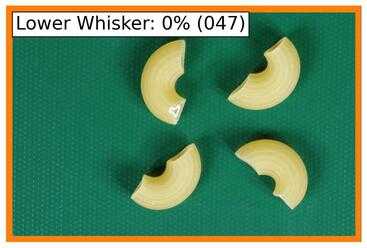}
        \includegraphics[width=.3\linewidth,valign=t,keepaspectratio]{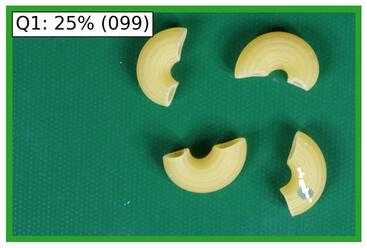}
      \end{minipage}
      \\
      \begin{minipage}{\linewidth}
        \centering
        \includegraphics[width=.3\linewidth,valign=t,keepaspectratio]{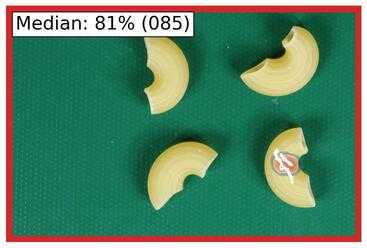}
        \includegraphics[width=.3\linewidth,valign=t,keepaspectratio]{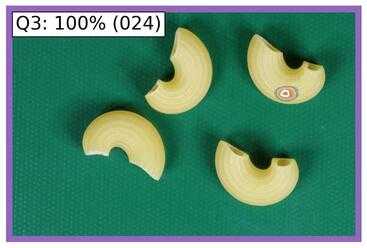}
        \includegraphics[width=.3\linewidth,valign=t,keepaspectratio]{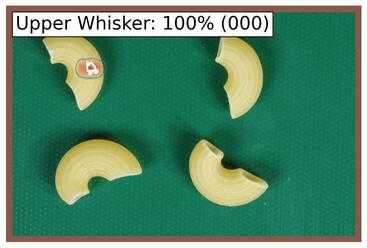}
      \end{minipage}
      \caption{
        Heatmaps.
        Images selected according to AUPIMO's statistics.
        Statistic and image index annotated on upper left corner.
      }
      \label{fig:benchmark-021-heatmap}
    \end{subfigure}
    \caption{
      Benchmark on VisA / Macaroni 2.
      PIMO curves and heatmaps are from EfficientAD M.
      200 images (100 normal, 100 anomalous).
    }
    \label{fig:benchmark-021}
\end{figure}

\begin{figure}[ht]
    \centering
    \begin{subfigure}[b]{\linewidth}
      \includegraphics[width=\linewidth,valign=t,keepaspectratio]{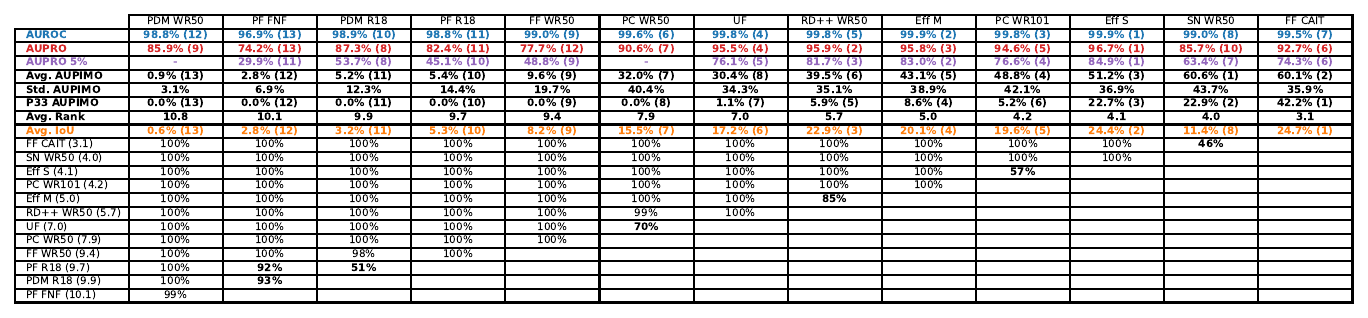}
      \caption{Statistics and pairwise statistical tests.}
      \label{fig:benchmark-022-table}
    \end{subfigure}
    \\ \vspace{2mm}
    \begin{subfigure}[b]{0.5\linewidth}
      \includegraphics[width=\linewidth,valign=t,keepaspectratio]{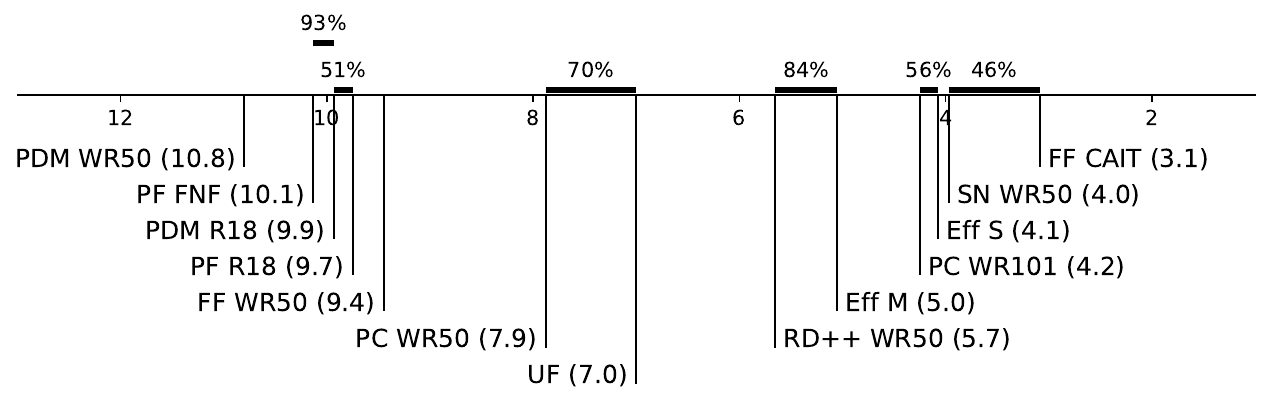}
      \caption{Average rank diagram.}
      \label{fig:benchmark-022-diagram}
    \end{subfigure}
    \\ \vspace{2mm}
    \begin{subfigure}[b]{0.45\linewidth}
      \includegraphics[width=\linewidth,valign=t,keepaspectratio]{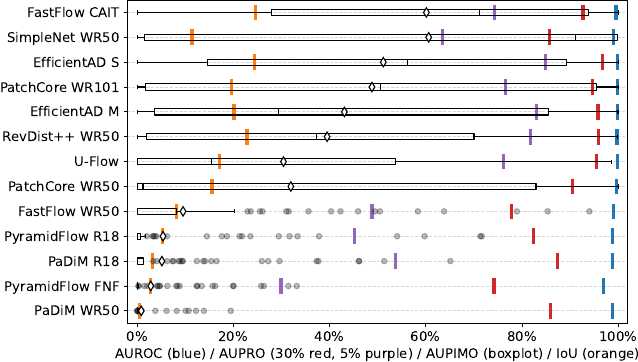}
      \caption{Score distributions.}
      \label{fig:benchmark-022-boxplot}
    \end{subfigure}
    ~
    \begin{subfigure}[b]{0.45\linewidth}
      \includegraphics[width=\linewidth,valign=t,keepaspectratio]{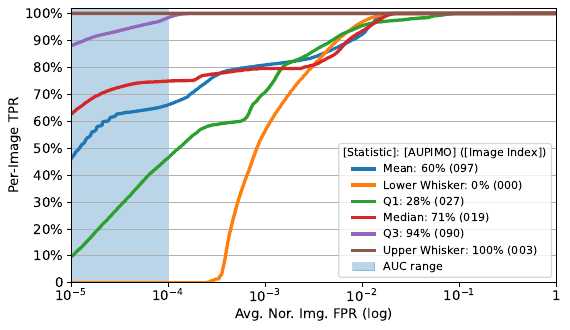}
      \caption{PIMO curves.}
      \label{fig:benchmark-022-pimo-curves}
    \end{subfigure}
    \\  \vspace{2mm}
    \begin{subfigure}[b]{\linewidth}
    
      \begin{minipage}{\linewidth}
        \centering
        \includegraphics[width=.3\linewidth,valign=t,keepaspectratio]{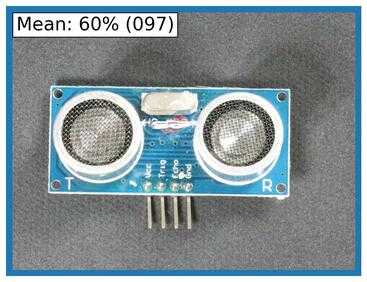}
        \includegraphics[width=.3\linewidth,valign=t,keepaspectratio]{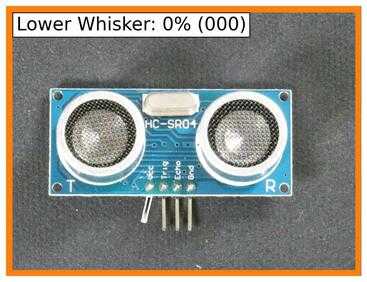}
        \includegraphics[width=.3\linewidth,valign=t,keepaspectratio]{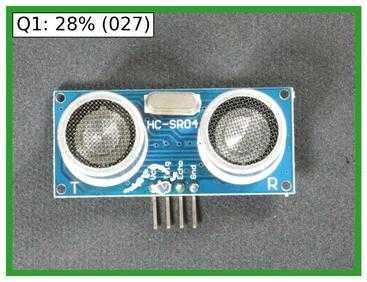}
      \end{minipage}
      \\
      \begin{minipage}{\linewidth}
        \centering
        \includegraphics[width=.3\linewidth,valign=t,keepaspectratio]{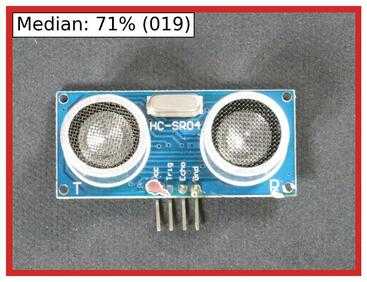}
        \includegraphics[width=.3\linewidth,valign=t,keepaspectratio]{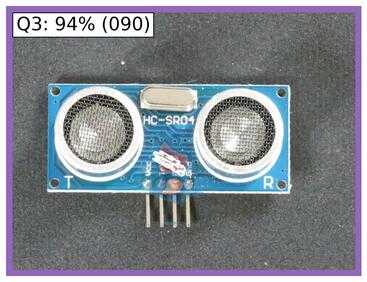}
        \includegraphics[width=.3\linewidth,valign=t,keepaspectratio]{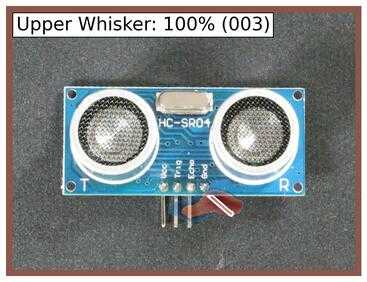}
      \end{minipage}
      \caption{
        Heatmaps.
        Images selected according to AUPIMO's statistics.
        Statistic and image index annotated on upper left corner.
      }
      \label{fig:benchmark-022-heatmap}
    \end{subfigure}
    \caption{
      Benchmark on VisA / PCB 1.
      PIMO curves and heatmaps are from FastFlow CAIT.
      200 images (100 normal, 100 anomalous).
    }
    \label{fig:benchmark-022}
\end{figure}

\begin{figure}[ht]
    \centering
    \begin{subfigure}[b]{\linewidth}
      \includegraphics[width=\linewidth,valign=t,keepaspectratio]{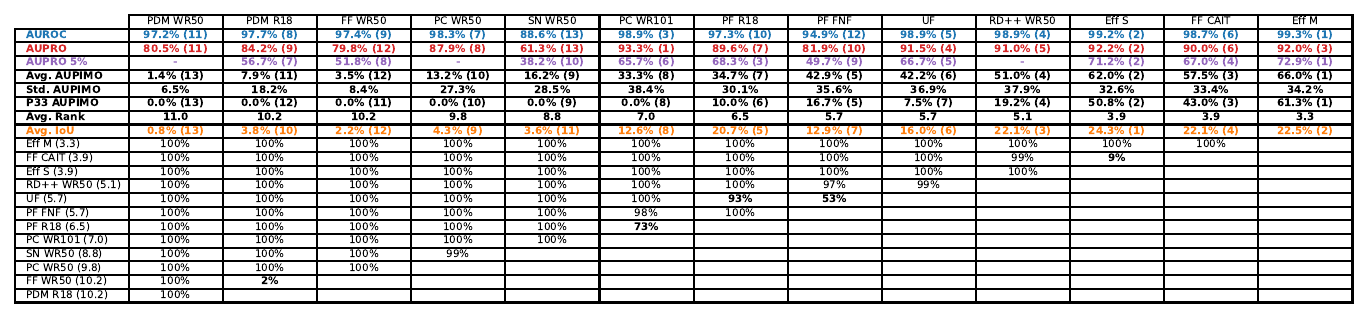}
      \caption{Statistics and pairwise statistical tests.}
      \label{fig:benchmark-023-table}
    \end{subfigure}
    \\ \vspace{2mm}
    \begin{subfigure}[b]{0.5\linewidth}
      \includegraphics[width=\linewidth,valign=t,keepaspectratio]{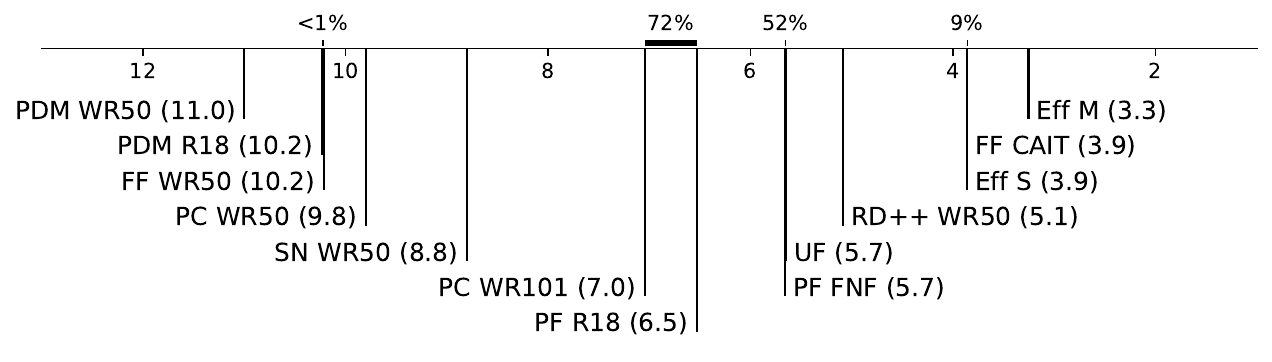}
      \caption{Average rank diagram.}
      \label{fig:benchmark-023-diagram}
    \end{subfigure}
    \\ \vspace{2mm}
    \begin{subfigure}[b]{0.45\linewidth}
      \includegraphics[width=\linewidth,valign=t,keepaspectratio]{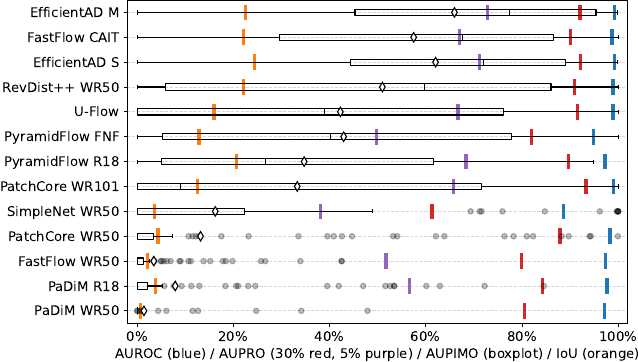}
      \caption{Score distributions.}
      \label{fig:benchmark-023-boxplot}
    \end{subfigure}
    ~
    \begin{subfigure}[b]{0.45\linewidth}
      \includegraphics[width=\linewidth,valign=t,keepaspectratio]{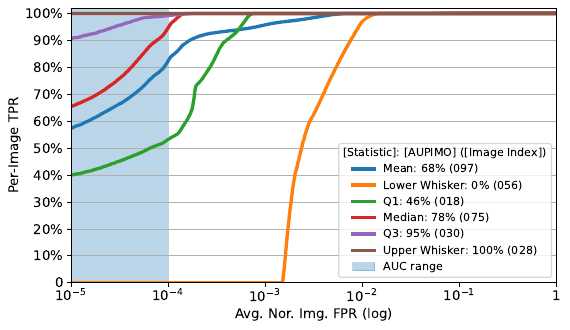}
      \caption{PIMO curves.}
      \label{fig:benchmark-023-pimo-curves}
    \end{subfigure}
    \\  \vspace{2mm}
    \begin{subfigure}[b]{\linewidth}
    
      \begin{minipage}{\linewidth}
        \centering
        \includegraphics[width=.3\linewidth,valign=t,keepaspectratio]{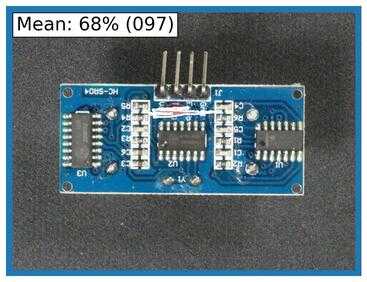}
        \includegraphics[width=.3\linewidth,valign=t,keepaspectratio]{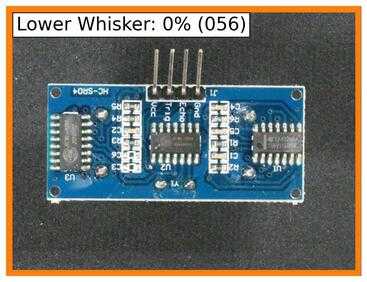}
        \includegraphics[width=.3\linewidth,valign=t,keepaspectratio]{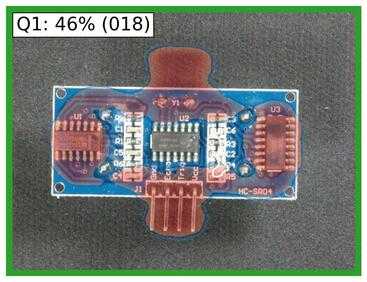}
      \end{minipage}
      \\
      \begin{minipage}{\linewidth}
        \centering
        \includegraphics[width=.3\linewidth,valign=t,keepaspectratio]{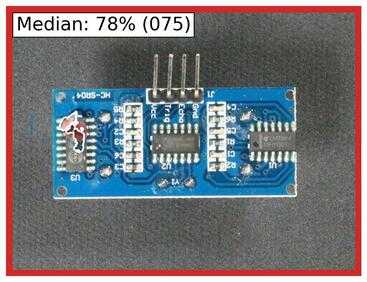}
        \includegraphics[width=.3\linewidth,valign=t,keepaspectratio]{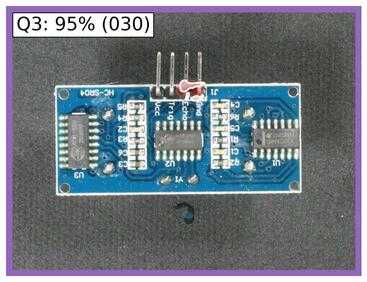}
        \includegraphics[width=.3\linewidth,valign=t,keepaspectratio]{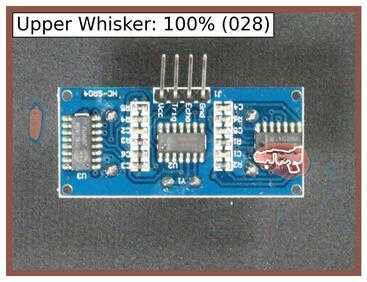}
      \end{minipage}
      \caption{
        Heatmaps.
        Images selected according to AUPIMO's statistics.
        Statistic and image index annotated on upper left corner.
      }
      \label{fig:benchmark-023-heatmap}
    \end{subfigure}
    \caption{
      Benchmark on VisA / PCB 2.
      PIMO curves and heatmaps are from EfficientAD M.
      200 images (100 normal, 100 anomalous).
    }
    \label{fig:benchmark-023}
\end{figure}

\begin{figure}[ht]
    \centering
    \begin{subfigure}[b]{\linewidth}
      \includegraphics[width=\linewidth,valign=t,keepaspectratio]{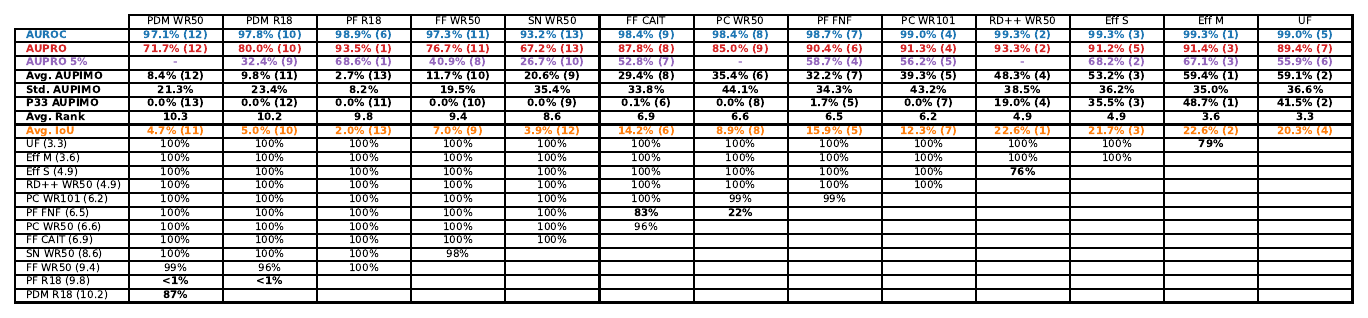}
      \caption{Statistics and pairwise statistical tests.}
      \label{fig:benchmark-024-table}
    \end{subfigure}
    \\ \vspace{2mm}
    \begin{subfigure}[b]{0.5\linewidth}
      \includegraphics[width=\linewidth,valign=t,keepaspectratio]{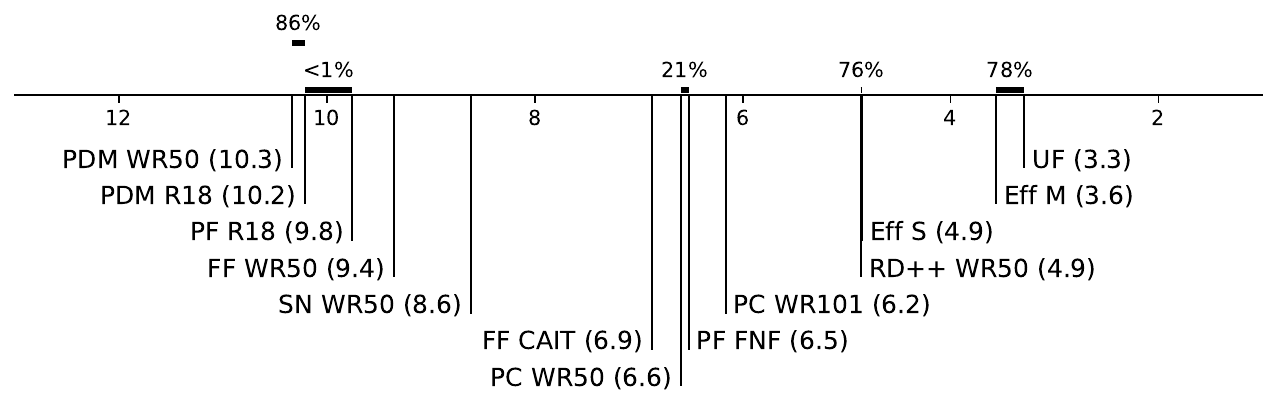}
      \caption{Average rank diagram.}
      \label{fig:benchmark-024-diagram}
    \end{subfigure}
    \\ \vspace{2mm}
    \begin{subfigure}[b]{0.45\linewidth}
      \includegraphics[width=\linewidth,valign=t,keepaspectratio]{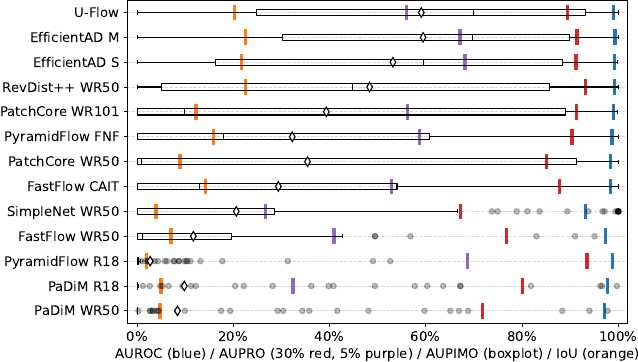}
      \caption{Score distributions.}
      \label{fig:benchmark-024-boxplot}
    \end{subfigure}
    ~
    \begin{subfigure}[b]{0.45\linewidth}
      \includegraphics[width=\linewidth,valign=t,keepaspectratio]{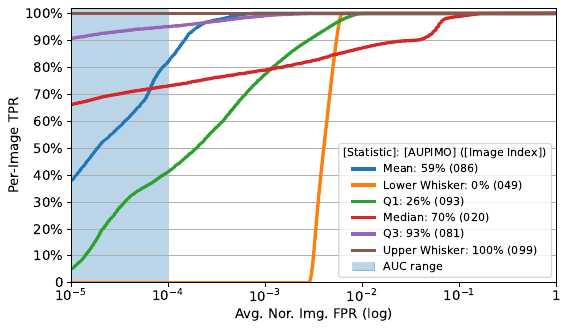}
      \caption{PIMO curves.}
      \label{fig:benchmark-024-pimo-curves}
    \end{subfigure}
    \\  \vspace{2mm}
    \begin{subfigure}[b]{\linewidth}
    
      \begin{minipage}{\linewidth}
        \centering
        \includegraphics[width=.3\linewidth,valign=t,keepaspectratio]{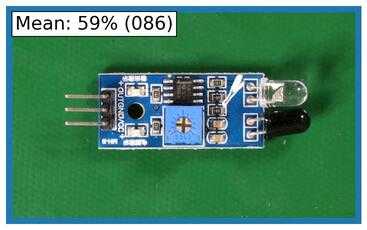}
        \includegraphics[width=.3\linewidth,valign=t,keepaspectratio]{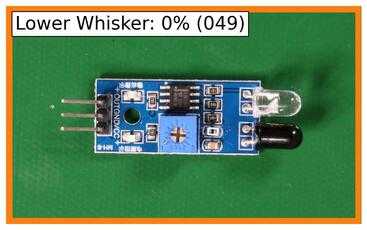}
        \includegraphics[width=.3\linewidth,valign=t,keepaspectratio]{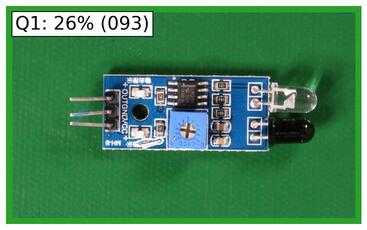}
      \end{minipage}
      \\
      \begin{minipage}{\linewidth}
        \centering
        \includegraphics[width=.3\linewidth,valign=t,keepaspectratio]{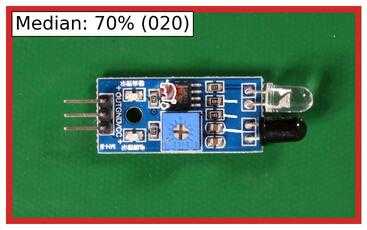}
        \includegraphics[width=.3\linewidth,valign=t,keepaspectratio]{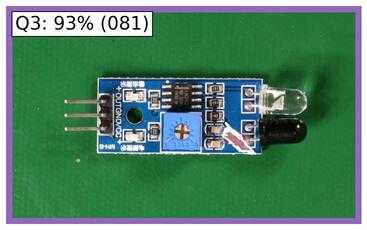}
        \includegraphics[width=.3\linewidth,valign=t,keepaspectratio]{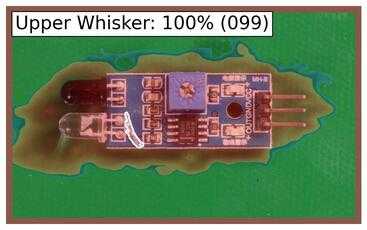}
      \end{minipage}
      \caption{
        Heatmaps.
        Images selected according to AUPIMO's statistics.
        Statistic and image index annotated on upper left corner.
      }
      \label{fig:benchmark-024-heatmap}
    \end{subfigure}
    \caption{
      Benchmark on VisA / PCB 3.
      PIMO curves and heatmaps are from U-Flow.
      201 images (101 normal, 100 anomalous).
    }
    \label{fig:benchmark-024}
\end{figure}

\begin{figure}[ht]
    \centering
    \begin{subfigure}[b]{\linewidth}
      \includegraphics[width=\linewidth,valign=t,keepaspectratio]{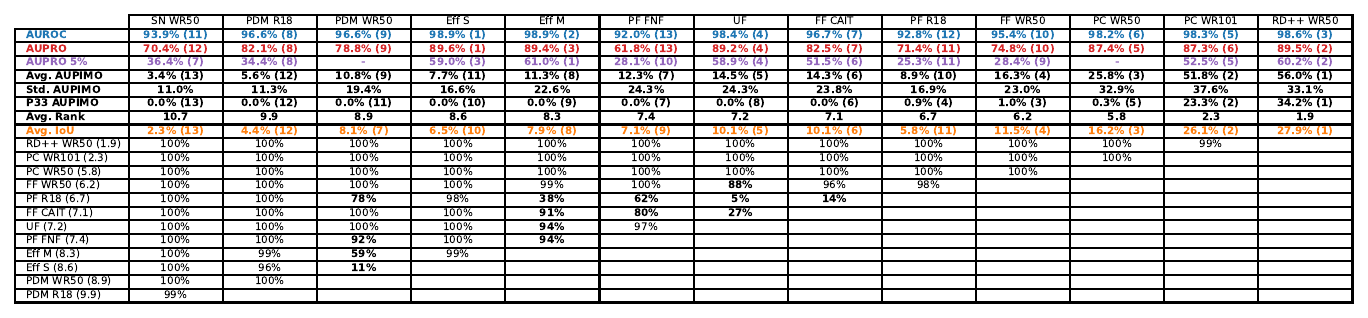}
      \caption{Statistics and pairwise statistical tests.}
      \label{fig:benchmark-025-table}
    \end{subfigure}
    \\ \vspace{2mm}
    \begin{subfigure}[b]{0.5\linewidth}
      \includegraphics[width=\linewidth,valign=t,keepaspectratio]{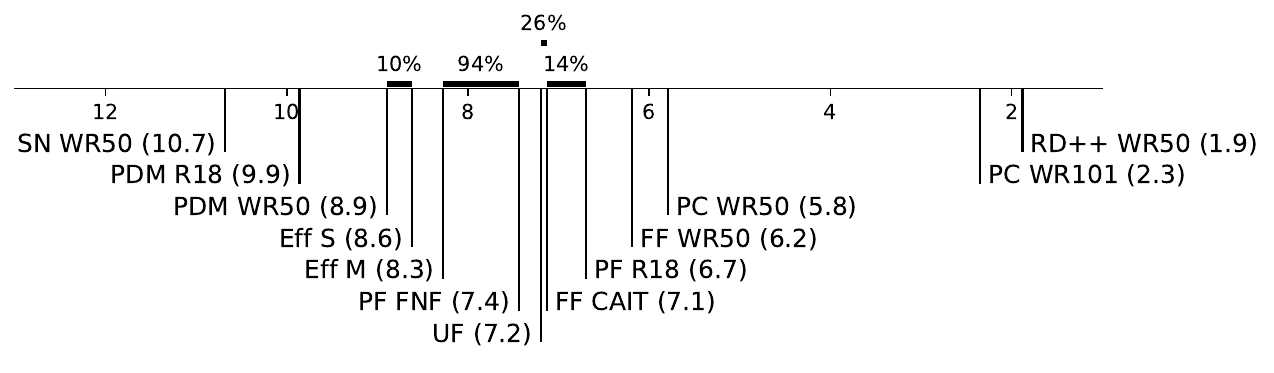}
      \caption{Average rank diagram.}
      \label{fig:benchmark-025-diagram}
    \end{subfigure}
    \\ \vspace{2mm}
    \begin{subfigure}[b]{0.45\linewidth}
      \includegraphics[width=\linewidth,valign=t,keepaspectratio]{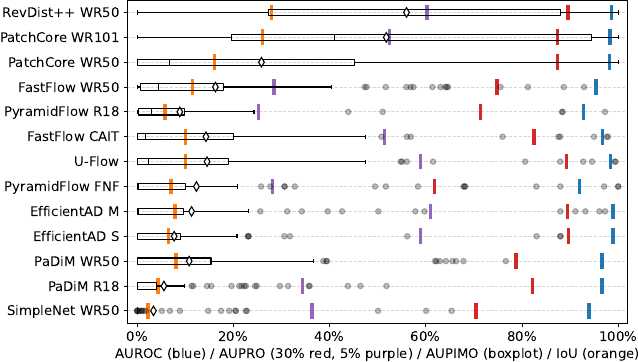}
      \caption{Score distributions.}
      \label{fig:benchmark-025-boxplot}
    \end{subfigure}
    ~
    \begin{subfigure}[b]{0.45\linewidth}
      \includegraphics[width=\linewidth,valign=t,keepaspectratio]{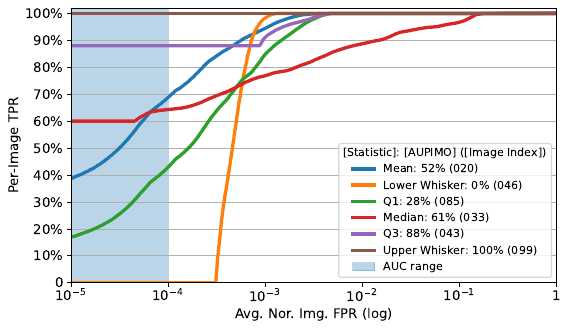}
      \caption{PIMO curves.}
      \label{fig:benchmark-025-pimo-curves}
    \end{subfigure}
    \\  \vspace{2mm}
    \begin{subfigure}[b]{\linewidth}
    
      \begin{minipage}{\linewidth}
        \centering
        \includegraphics[width=.3\linewidth,valign=t,keepaspectratio]{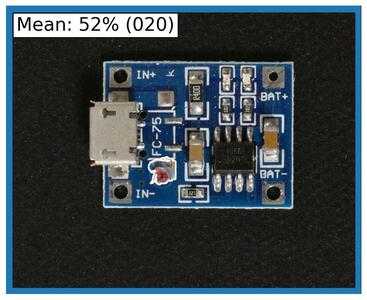}
        \includegraphics[width=.3\linewidth,valign=t,keepaspectratio]{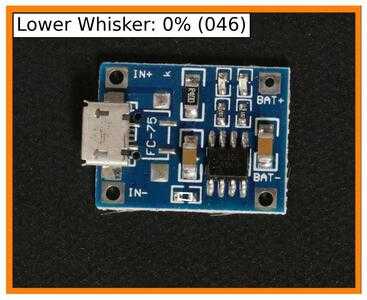}
        \includegraphics[width=.3\linewidth,valign=t,keepaspectratio]{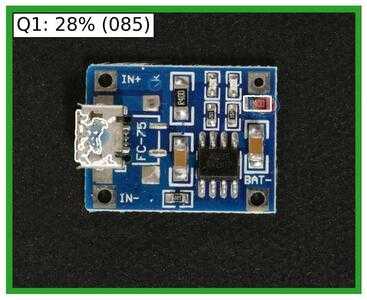}
      \end{minipage}
      \\
      \begin{minipage}{\linewidth}
        \centering
        \includegraphics[width=.3\linewidth,valign=t,keepaspectratio]{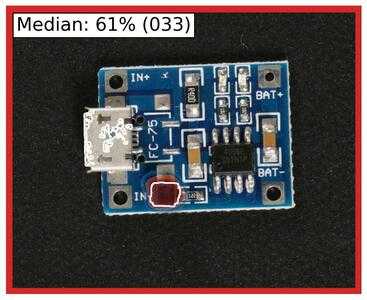}
        \includegraphics[width=.3\linewidth,valign=t,keepaspectratio]{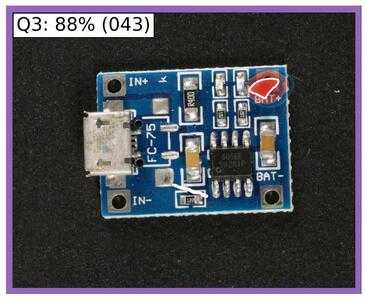}
        \includegraphics[width=.3\linewidth,valign=t,keepaspectratio]{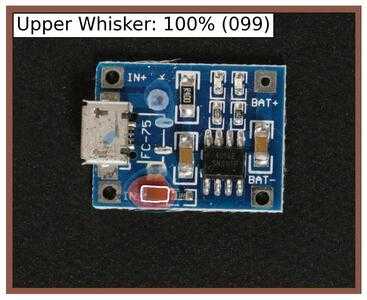}
      \end{minipage}
      \caption{
        Heatmaps.
        Images selected according to AUPIMO's statistics.
        Statistic and image index annotated on upper left corner.
      }
      \label{fig:benchmark-025-heatmap}
    \end{subfigure}
    \caption{
      Benchmark on VisA / PCB 4.
      PIMO curves and heatmaps are from RevDist++ WR50.
      201 images (101 normal, 100 anomalous).
    }
    \label{fig:benchmark-025}
\end{figure}

\begin{figure}[ht]
    \centering
    \begin{subfigure}[b]{\linewidth}
      \includegraphics[width=\linewidth,valign=t,keepaspectratio]{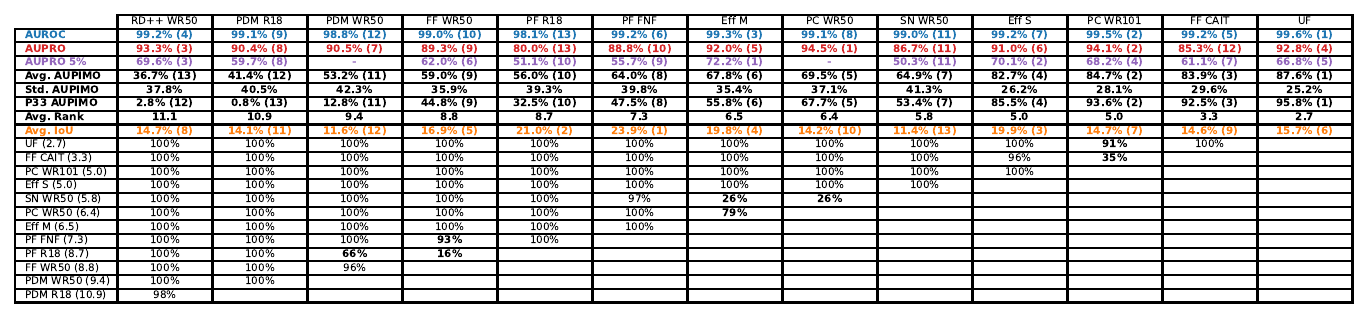}
      \caption{Statistics and pairwise statistical tests.}
      \label{fig:benchmark-026-table}
    \end{subfigure}
    \\ \vspace{2mm}
    \begin{subfigure}[b]{0.5\linewidth}
      \includegraphics[width=\linewidth,valign=t,keepaspectratio]{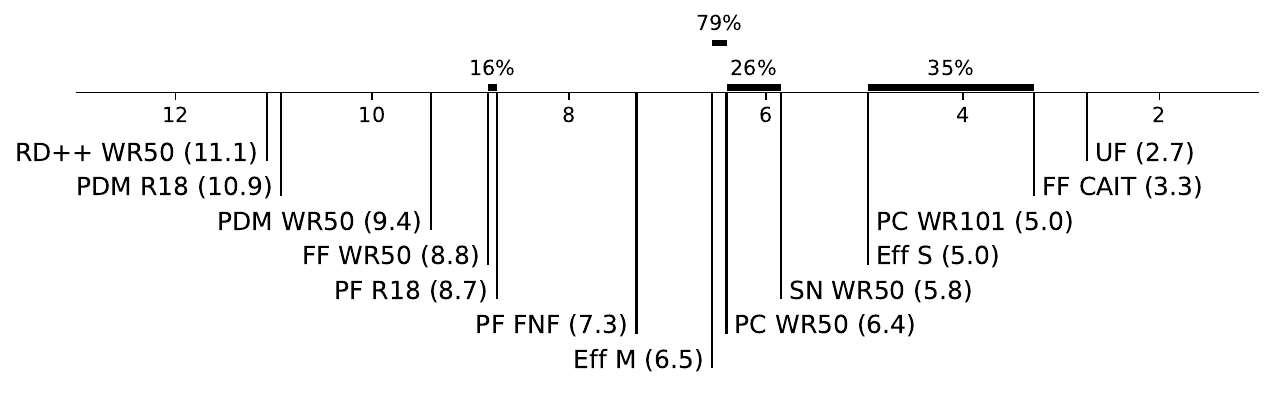}
      \caption{Average rank diagram.}
      \label{fig:benchmark-026-diagram}
    \end{subfigure}
    \\ \vspace{2mm}
    \begin{subfigure}[b]{0.45\linewidth}
      \includegraphics[width=\linewidth,valign=t,keepaspectratio]{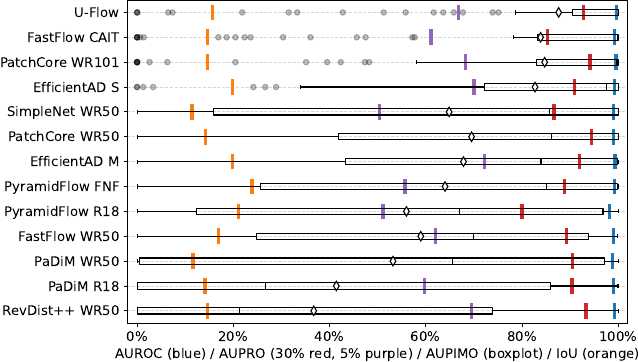}
      \caption{Score distributions.}
      \label{fig:benchmark-026-boxplot}
    \end{subfigure}
    ~
    \begin{subfigure}[b]{0.45\linewidth}
      \includegraphics[width=\linewidth,valign=t,keepaspectratio]{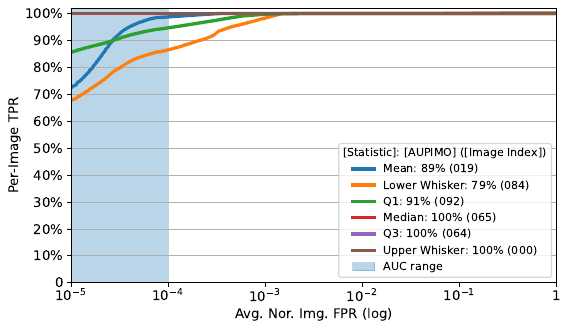}
      \caption{PIMO curves.}
      \label{fig:benchmark-026-pimo-curves}
    \end{subfigure}
    \\  \vspace{2mm}
    \begin{subfigure}[b]{\linewidth}
    
      \begin{minipage}{\linewidth}
        \centering
        \includegraphics[width=.3\linewidth,valign=t,keepaspectratio]{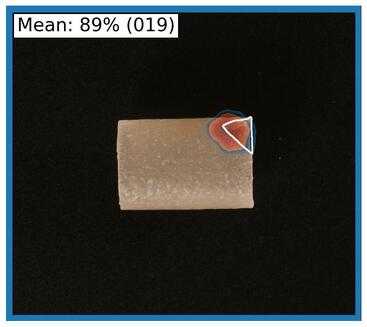}
        \includegraphics[width=.3\linewidth,valign=t,keepaspectratio]{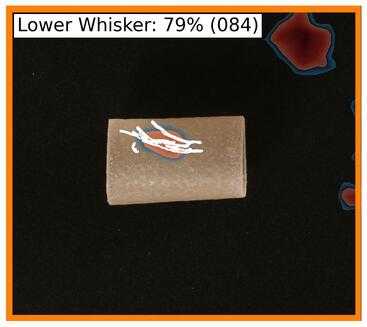}
        \includegraphics[width=.3\linewidth,valign=t,keepaspectratio]{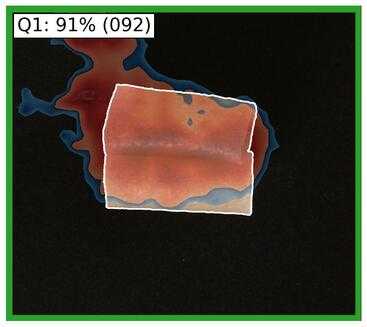}
      \end{minipage}
      \\
      \begin{minipage}{\linewidth}
        \centering
        \includegraphics[width=.3\linewidth,valign=t,keepaspectratio]{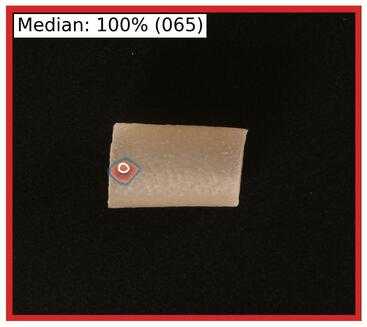}
        \includegraphics[width=.3\linewidth,valign=t,keepaspectratio]{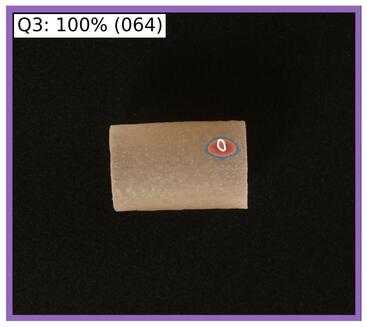}
        \includegraphics[width=.3\linewidth,valign=t,keepaspectratio]{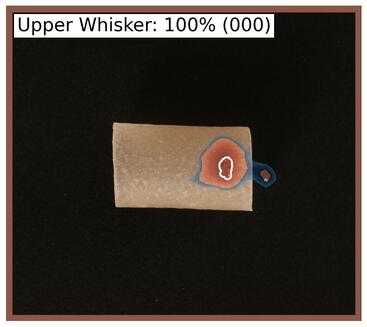}
      \end{minipage}
      \caption{
        Heatmaps.
        Images selected according to AUPIMO's statistics.
        Statistic and image index annotated on upper left corner.
      }
      \label{fig:benchmark-026-heatmap}
    \end{subfigure}
    \caption{
      Benchmark on VisA / Pipe Fryum.
      PIMO curves and heatmaps are from U-Flow.
      150 images (050 normal, 100 anomalous).
    }
    \label{fig:benchmark-026}
\end{figure}

\begin{multicols}{2}
    \begin{enumerate}
        \small
        \item \cref{fig:benchmark-000}: MVTec AD / Bottle
        \item \cref{fig:benchmark-001}: MVTec AD / Cable
        \item \cref{fig:benchmark-002}: MVTec AD / Capsule
        \item \cref{fig:benchmark-003}: MVTec AD / Carpet
        \item \cref{fig:benchmark-004}: MVTec AD / Grid
        \item \cref{fig:benchmark-005}: MVTec AD / Hazelnut
        \item \cref{fig:benchmark-006}: MVTec AD / Leather
        \item \cref{fig:benchmark-007}: MVTec AD / Metal Nut
        \item \cref{fig:benchmark-008}: MVTec AD / Pill
        \item \cref{fig:benchmark-009}: MVTec AD / Screw
        \item \cref{fig:benchmark-010}: MVTec AD / Tile
        \item \cref{fig:benchmark-011}: MVTec AD / Toothbrush
        \item \cref{fig:benchmark-012}: MVTec AD / Transistor
        \item \cref{fig:benchmark-013}: MVTec AD / Wood
        \item \cref{fig:benchmark-014}: MVTec AD / Zipper
        \item \cref{fig:benchmark-015}: VisA / Candle
        \item \cref{fig:benchmark-016}: VisA / Capsules
        \item \cref{fig:benchmark-017}: VisA / Cashew
        \item \cref{fig:benchmark-018}: VisA / Chewing Gum
        \item \cref{fig:benchmark-019}: VisA / Fryum
        \item \cref{fig:benchmark-020}: VisA / Macaroni 1
        \item \cref{fig:benchmark-021}: VisA / Macaroni 2
        \item \cref{fig:benchmark-022}: VisA / PCB 1
        \item \cref{fig:benchmark-023}: VisA / PCB 2
        \item \cref{fig:benchmark-024}: VisA / PCB 3
        \item \cref{fig:benchmark-025}: VisA / PCB 4
        \item \cref{fig:benchmark-026}: VisA / Pipe Fryum
    \end{enumerate}
\end{multicols}

\noindent    
Each figure contains the following elements:
\begin{enumerate}

    \item a plot with one model per row containing:
    \begin{enumerate}
        \item the \gls{auroc} score as a blue vertical line;
        \item the \gls{aupro} score as a red vertical line; 
        \item a boxplot of \gls{aupimo} scores; 
        \begin{enumerate}
            \item lower and upper whiskers set with maximum size to $1.5$ inter-quartile range (IQR);
            \item the mean is displayed as a white diamond;
            \item fliers are displayed as gray dots;
        \end{enumerate}
    \end{enumerate}

    \item a diagram of (image-wise) average rank according to \gls{aupimo} scores; 
    lower is better;
    $1$ means that the model has the best \gls{aupimo} score at all images;

    \item a table comprising two parts:
    \begin{enumerate}

        \item the upper part, in bold, comprises:
        \begin{enumerate}
            \item the \gls{auroc} scores (in blue);
            \item the \gls{aupro} scores (in red);
            \item the average and standard deviation \gls{aupimo} score (in black);
            \item the \nth{33} percentile \gls{aupimo} score (in black);
            \item the values in parentheses are the ranks of the models according to the respective score metric in each row;
        \end{enumerate}

        \item the lower part shows the results of pairwise Wilcoxon signed rank tests using \gls{aupimo} scores; 
        each cell shows the confidence to reject the null hypothesis $\confidence = 1 - p$ (where $p$ is the p-value) assuming that the row model is better than the column model as alternative hypothesis; 
        confidence values below $95\%$ (\ie \say{low confidence}) are highlighted in bold;

    \end{enumerate}

    \item \gls{pimo} curves and heatmap samples from the model with best average \gls{aupimo} rank;
    \begin{enumerate}
        \item samples are selected according to the recommendations from the paragraph \say{Sample selection};
        \item the (2-pixel wide, outter) countour of the ground truth mask is shown in white.
        \item heatmaps are colored according to the color scheme described below;
    \end{enumerate}
    
\end{enumerate}

\paragraph{Heatmaps coloring scheme}
The input images are superimposed by their respective anomaly score map $\amap$.
Coloring rules are linked to the thresholds in \gls{aupimo}'s integration bounds: transparent is for scores below the lowest threshold, blues are for scores between the lowest and the highest thresholds, and reds are for scores above the highest threshold.
Darker blue/red tones mean higher scores.
The coloring strategy links the heatmaps to the validation-evaluation framework employed in \gls{aupimo}.
Transparent heatmap zones are never accounted in the metric because the validation requirement is not respected.
Blue zones visually express the average recall measured by the integration in \gls{aupimo}.
Additionally, red zones show the model's local behavior (per-image normalization) within the \emph{valid} score range (\ie scores above the threshold given by the \gls{shfpr} lower bound).

\clearpage

\clearpage

\clearpage

\clearpage

\clearpage

\clearpage

\clearpage

\clearpage

\clearpage

\clearpage

\clearpage

\clearpage

\clearpage

\clearpage

\clearpage

\clearpage

\clearpage

\clearpage

\clearpage

\clearpage

\clearpage

\clearpage

\clearpage

\clearpage

\clearpage

\clearpage

\clearpage

\end{document}